\documentclass[11pt,a4paper]{article}
\usepackage[margin=25mm,headheight=14pt,headsep=14pt]{geometry}
\usepackage[T1]{fontenc}
\usepackage[no-math]{fontspec}
\usepackage{amsmath,amssymb}
\usepackage{XCharter}
\usepackage[xcharter]{newtxmath}
\SetSymbolFont{operators}{normal}{OT1}{XCharter-TLF}{m}{n}
\SetSymbolFont{operators}{bold}{OT1}{XCharter-TLF}{b}{n}
\SetMathAlphabet{\mathit}{normal}{OT1}{XCharter-TLF}{m}{it}
\SetMathAlphabet{\mathit}{bold}{OT1}{XCharter-TLF}{b}{it}
\SetMathAlphabet{\mathbf}{normal}{OT1}{XCharter-TLF}{b}{n}
\usepackage[semibold]{plex-sans}
\usepackage[scale=0.95]{plex-mono}
\usepackage{microtype}
\usepackage{graphicx,booktabs,longtable}
\usepackage{pdflscape}
\usepackage{parskip}
\usepackage{xcolor,titlesec,caption,fancyhdr}
\usepackage{tcolorbox}
\tcbuselibrary{breakable}
\usepackage[numbers,sort&compress]{natbib}
\usepackage{bibunits}
\usepackage{mathtools,array,tabularx,placeins}
\usepackage{needspace}
\usepackage{float,listings}
\usepackage{algorithm}
\usepackage{algorithmicx,algpseudocode}

\usepackage[hidelinks,unicode]{hyperref}

\definecolor{reportgray}{gray}{0.45}
\titleformat{\section}{\sffamily\bfseries\large}{\thesection.}{0.6em}{}
\titleformat{\subsection}{\sffamily\bfseries\normalsize}{\thesubsection}{0.6em}{}
\titleformat{\subsubsection}{\sffamily\bfseries\normalsize}{\thesubsubsection}{0.6em}{}
\titleformat{\paragraph}[runin]{\sffamily\bfseries}{\theparagraph}{0.6em}{}[.]
\titleformat{\subparagraph}[runin]{\sffamily\bfseries}{\thesubparagraph}{0.6em}{}[.]
\titlespacing*{\section}{0pt}{1.6\baselineskip}{0.6\baselineskip}
\titlespacing*{\subsection}{0pt}{1.1\baselineskip}{0.4\baselineskip}
\titlespacing*{\subsubsection}{0pt}{0.8\baselineskip}{0.3\baselineskip}
\DeclareCaptionLabelFormat{continuedfigure}{#1~#2 (continued)}

\definecolor{codebg}{RGB}{248,248,248}
\definecolor{codecomment}{RGB}{90,90,90}
\newcommand{\reportheader}{%
  \raisebox{-2.3pt}{\includegraphics[height=12pt]{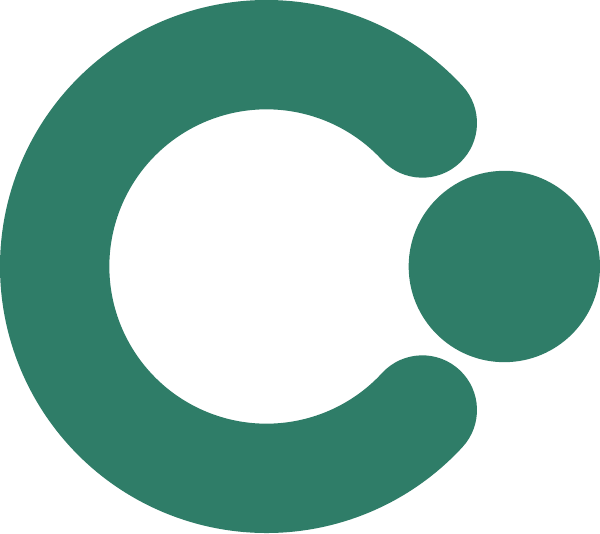}}%
  \hspace{0.5em}{\sffamily\footnotesize\color{reportgray}Cyto.CoScientist}}
\fancypagestyle{plain}{%
  \fancyhf{}
  \fancyhead[L]{\reportheader}
  \fancyfoot[C]{\sffamily\footnotesize\color{reportgray}\thepage}
  }

\newif\ifreportauthorinfo
\reportauthorinfotrue
\makeatletter
\newcommand{\report@affiliations}{}
\newcommand{\report@correspondence}{}
\def\report@keywords{}
\newcommand{\affiliations}[1]{\gdef\report@affiliations{#1}}
\newcommand{\correspondence}[1]{\gdef\report@correspondence{#1}}
\newcommand{\keywords}[1]{\gdef\report@keywords{#1}}
\renewcommand{\maketitle}{%
  \thispagestyle{plain}
  \begin{center}
    {\sffamily\bfseries\fontsize{22}{27}\selectfont\@title\par}
    \ifreportauthorinfo
      \ifx\@author\@empty\else\vspace{1em}{\sffamily\@author\par}\fi
      \ifx\report@affiliations\@empty\else
        \vspace{0.5em}{\sffamily\small\color{reportgray}\report@affiliations\par}\fi
      \ifx\report@correspondence\@empty\else
        \vspace{0.5em}{\small Correspondence: \report@correspondence\par}\fi
    \fi
    \ifx\@date\@empty\else\vspace{0.5em}{\small\@date\par}\fi
  \end{center}
  \vspace{1em}}
\newcommand{\reportkeywords}{%
  \ifx\report@keywords\@empty\else
    \par\noindent{\small\textbf{Keywords:} \report@keywords\par}\medskip
  \fi}
\makeatother

\newlength{\reporttocnumberwidth}
\makeatletter
\newcommand{\reporttocsection}[2]{%
  \addpenalty{-\@highpenalty}\addvspace{7pt}%
  \begingroup\bfseries
    \def\@dotsep{10000}%
    \@dottedtocline{1}{0em}{\reporttocnumberwidth}{#1}{#2}%
  \endgroup\nobreak}
\newcommand{\reporttocsubsection}[2]{%
  \addvspace{1pt}%
  \@dottedtocline{2}{0em}{\reporttocnumberwidth}{#1}{#2}}
\newcommand{\reportcontents}{%
  \begingroup
    \setlength{\parskip}{0pt}%
    \renewcommand{\@pnumwidth}{2.2em}%
    \renewcommand{\@tocrmarg}{3em}%
    \let\l@section\reporttocsection
    \let\l@subsection\reporttocsubsection
    \tableofcontents
  \endgroup}
\makeatother

\newtcolorbox{reportnote}[1]{breakable,
  colback=white,colframe=reportgray,colbacktitle=white,coltitle=black,
  boxrule=0.4pt,arc=0pt,left=3mm,right=3mm,top=2mm,bottom=2mm,
  fonttitle=\sffamily\bfseries,title={#1},
  before skip=0.7\baselineskip,after skip=0.7\baselineskip}

\newtcolorbox{caseprompt}[1]{breakable,
  colback=white,colframe=teal!60!black,colbacktitle=teal!60!black,coltitle=white,
  boxrule=0.6pt,arc=0pt,left=4mm,right=4mm,top=3mm,bottom=3mm,
  fonttitle=\sffamily\bfseries,title={#1},
  before skip=0.8\baselineskip,after skip=0.9\baselineskip}

\let\reportoriginalappendix\appendix
\renewcommand{\appendix}{%
  \clearpage\reportoriginalappendix
  \setcounter{figure}{0}\setcounter{table}{0}\setcounter{equation}{0}%
  \renewcommand{\thesection}{S\arabic{section}}%
  \renewcommand{\thefigure}{S\arabic{figure}}%
  \renewcommand{\thetable}{S\arabic{table}}%
  \renewcommand{\theequation}{S\arabic{equation}}%
  \renewcommand{\theHsection}{supp.\arabic{section}}%
  \renewcommand{\theHfigure}{supp.\arabic{figure}}%
  \renewcommand{\theHtable}{supp.\arabic{table}}%
  \renewcommand{\theHequation}{supp.\arabic{equation}}%
}

\title{CytoCoScientist:\\Within-Task Recursive Self-Improvement\\for Autonomous Biological Research}
\author{Jun~Huo\textsuperscript{1,*}, Junjie~Pang\textsuperscript{1,2,*}, Kai~Liao\textsuperscript{1,2,*}, Ruofan~Jin\textsuperscript{3,*}, Chuan~Wang\textsuperscript{4,5,*}, Yue~Wang\textsuperscript{1,*}, Xiaofeng~Shi\textsuperscript{1}, Di~Wang\textsuperscript{1}, Qiuyi~Li\textsuperscript{3}, Yuan~He\textsuperscript{3}, Yan~Gao\textsuperscript{3}, Xiaorong~Zhang\textsuperscript{1,2}, Meng~Sun\textsuperscript{1,2}, Qian~Zhang\textsuperscript{1,2}, Yong~Zhang\textsuperscript{1}, Ziqing~Deng\textsuperscript{1}, Yuxiang~Li\textsuperscript{1}, Tao~Qin\textsuperscript{3}, Xun~Xu\textsuperscript{1,\dag}, Haiguang~Liu\textsuperscript{3,\dag}, Qijin~Yin\textsuperscript{3,\dag}, Shaoshuai~Li\textsuperscript{1,\dag}}
\affiliations{
  \textsuperscript{1}BGI Research\\
  \textsuperscript{2}University of the Chinese Academy of Sciences\\
  \textsuperscript{3}Zhongguancun Academy\\
  \textsuperscript{4}Department of Cardiac Surgery, Beijing Anzhen Hospital, Capital Medical University\\
  \textsuperscript{5}Beijing Institute of Brain Disorders, Capital Medical University\\
  [3pt]
  \textsuperscript{*}Equal contribution. \quad \textsuperscript{\dag}Corresponding authors.
}
\correspondence{lishaoshuai@genomics.cn, yinqijin@bza.edu.cn, liuhaiguang@bza.edu.cn, xuxun@genomics.cn}
\date{}

\renewcommand{\refname}{References}
\renewcommand{\tablename}{Table}

\begin{document}
\maketitle
\begin{bibunit}[plainnat]
\makeatletter
\def\@extra@b@citeb{.main}
\def\@extra@binfo{.main}
\makeatother
We present CytoCoScientist, an autonomous biological research system built for long-horizon scientific exploration that realizes \emph{within-task recursive self-improvement} (RSI) under fixed model parameters across bioinformatics algorithm development and open-ended bioinformatics inquiry. During a multi-stage investigation, retrieved literature and empirical outcomes continually reshape how hypotheses, computational methods, and analytical workflows are judged and revised, separating whether a result was produced from whether it is admissible to support a conclusion. A central main agent sustains this self-correcting loop over a persistent research state through three coupled mechanisms: multi-branch evolution in explicit candidate search spaces that decouples branch selection from critique-guided revision; evidence-based control spanning pre-execution checks, layered post-execution review, completion gating, and independent outer-loop meta-review; and isolated subagent execution paired with a revisable timeline, hierarchical provenance-linked memory, and on-demand domain skills that update superseded interpretations while accumulating reusable experience. We evaluate CytoCoScientist across five long-horizon biological studies. In algorithm development, the evolved estimator reduces three-tissue mean squared error by $4.7\%$ over the top baseline MAGIC on the OpenProblems single-cell denoising benchmark, and a review-guided late-fusion survival pipeline improves C-index by $0.0508$ over matched DeepProg on the fixed TCGA-LUAD evaluation split. In open-ended bioinformatics inquiry, CytoCoScientist autonomously pivots from single-draw sepsis proteomics to within-patient longitudinal trajectories, decouples intrinsic memory B-cell \textit{TOP2A} elevation from compositional shifts while auditing \textup{CXCL13} cellular origin in paired esophageal squamous cell carcinoma (ESCC) lymph nodes, and aligns five blood-derived modalities in acute aortic dissection to identify a cross-material \textit{SMAD3} association while retracting an unsupported chromatin-contact claim. By coupling open-ended exploration with empirical admissibility control, CytoCoScientist enables autonomous systems to refine not only computational implementations, but also the underlying biological hypotheses, analytical designs, and scientific conclusions.

\reportkeywords\clearpage\reportcontents\clearpage
\section{Introduction}\label{sec:intro-need}

Modern biological research increasingly integrates diverse knowledge, data, and analytical methods \citep{Kitano2002SystemsBiology,Huang2025Biomni,Mitchener2025Kosmos}. Investigating disease mechanisms, cellular functions, and potential interventions often requires connecting prior literature with molecular measurements, cellular profiles, and clinical observations. High-throughput assays and public datasets expand the scope of tractable questions, but they also place greater demands on domain expertise and research coordination \citep{Luecken2025OpenProblems,Huang2025Biomni}. Starting from an open question, researchers must synthesize existing knowledge, formulate testable hypotheses, and select or develop appropriate methods \citep{Gottweis2026AcceleratingScientific3906c8d9a2f0,Mitchener2025Kosmos,Swanson2025VirtualLab,Ghareeb2025Robin}. Emerging evidence can either support the current approach or overturn earlier judgments, prompting changes to analytical and validation plans. Real-world biological discovery is therefore inherently a long-horizon task, requiring autonomous research systems that can coordinate dozens of interdependent decisions, retain useful experience, and continually revise prior interpretations as new evidence emerges \citep{Kitano2021NobelTuring56b31b936a7c,Shinn2023ReflexionLanguageb7573199fe22,Zhao2023ExpeL}.

Recent advances in large language models (LLMs) have brought autonomous scientific agents closer to this goal by combining literature synthesis, reasoning, and code generation. Early biological agents focus on workflow orchestration and tool integration, coordinating bioinformatics pipelines and composing specialized software and databases \citep{Xiao2024CellAgent,Su2026BioMaster,Huang2025Biomni,Jin2026StellaTowards022cedda42b0}. More recent systems close narrower research loops: CellVoyager, BioDiscoveryAgent, and Robin feed analytical or experimental outcomes back into subsequent questions and designs \citep{Alber2026CellVoyager,Roohani2024BioDiscoveryAgent,Ghareeb2025Robin}, while Google AI co-scientist and multi-agent frameworks such as Virtual Lab and BioLab evolve biomedical hypotheses and biomolecular designs through iterative critique \citep{Gottweis2026AcceleratingScientific3906c8d9a2f0,Swanson2025VirtualLab,Jin2025BioLab}.

Sustaining multi-round inquiry beyond single-pass workflows further requires searching across competing candidates while maintaining persistent records of prior results \citep{Yao2023TreeThoughtsba7b3b20b200,Yamada2025AIScientistV2,Shinn2023ReflexionLanguageb7573199fe22,MemGPT2023,Xu2025AMEM}. Program-evolution systems such as FunSearch, AlphaEvolve, and ERA pair execution-based evaluation with structured search to discover high-performing algorithms, including single-cell batch integration methods enriched with literature and external research ideas \citep{RomeraParedes2024FunSearch,Novikov2025AlphaEvolve,Aygun2026ERA}. End-to-end discovery frameworks connect ideation, computational experimentation, and reporting \citep{Yamada2025AIScientistV2,Tang2025AiresearcherAutonomous8d190a9211c3,Schmidgall2025AgentLaboratory,Jansen2025CodeScientist,VillaescusaNavarro2025Denario}; in biology, Kosmos coordinates literature and multi-omics analysis through a shared research state \citep{Mitchener2025Kosmos}, while Scientist-One and xScientist strengthen traceability through evidence-linked claims and versioned records \citep{Meng2026ScientistOne,Luo2026XscientistGitlikec9a109c50b3c}. Yet these approaches largely address candidate search, workflow coordination, or traceable recordkeeping in isolation, leaving open how execution outcomes and review feedback can jointly update candidate rankings, revise active interpretations, and steer subsequent decisions over a long-horizon study.

In bioinformatics, this challenge spans two tightly coupled regimes: bioinformatics algorithm development and open-ended bioinformatics inquiry. Algorithm development seeks computational formulations tailored to data characteristics and modeling objectives \citep{Luecken2025OpenProblems,Poirion2021DeepProg,Aygun2026ERA}, whereas open-ended inquiry organizes analytical workflows around biological questions, tests competing explanations, and builds mechanistic interpretations \citep{Huang2025Biomni,Alber2026CellVoyager,Mitchener2025Kosmos}. In practice, the two regimes interact repeatedly within a single investigation. For example, sparse counts or scale disparities across omics modalities often force researchers to revise an estimator, model architecture, or integration strategy \citep{Luecken2025OpenProblems,Poirion2021DeepProg}. Similarly, a cohort analysis may yield only partial support for an initial hypothesis, prompting follow-up work to separate competing explanations, adjust statistical designs, or test alternative candidates \citep{Mitchener2025Kosmos,Roohani2024BioDiscoveryAgent,Ghareeb2025Robin}. Supporting both regimes within one autonomous system therefore requires comparing competing directions and using empirical feedback to decide which lines of inquiry merit continuation \citep{Gottweis2026AcceleratingScientific3906c8d9a2f0,Aygun2026ERA,Yamada2025AIScientistV2}.

Turning iterative exploration into cumulative self-improvement over long horizons faces two coupled failure modes. First, explicit search amplifies whatever evaluation signals it receives, yet successful code execution or higher metrics alone do not establish scientific validity \citep{Son2025SPOT,Trehan2026ResearchFailures,Huang2025Popper}. Data leakage can inflate predictive performance \citep{Kapoor2023Leakage}, and shifts in cell-type composition can create disease associations that appear to reflect cell-intrinsic changes \citep{Buttner2021scCODA,Farahbod2020CellularComposition}; when subsequent sensitivity analyses or background calibrations overturn earlier conclusions, candidate rankings and follow-up plans must be updated accordingly \citep{Huang2025Popper,Mitchener2025Kosmos}. Second, as code, analytical artifacts, and review records accumulate across rounds, the full research history quickly exceeds an LLM's effective context window \citep{LostMiddle2024,LongLLMLingua2024,Kang2025ACON,MemGPT2023}. If compressed summaries retain superseded interpretations, later actions risk relying on outdated judgments \citep{Xu2025AMEM,Chhikara2025Mem0}. Achieving reliable self-improvement over long horizons therefore requires synchronizing result auditing, candidate evaluation, and revisable research-state management so that methodological repairs, valid negative findings, and updated interpretations continually improve subsequent exploration.

To address these challenges, we introduce CytoCoScientist, an autonomous biological research system built for long-horizon scientific exploration. CytoCoScientist operationalizes \emph{within-task recursive self-improvement} (RSI) under fixed model parameters across both bioinformatics algorithm development and open-ended bioinformatics inquiry. Rather than following a fixed stage pipeline, a central main agent dynamically coordinates literature retrieval, computational experimentation, and scientific writing through a persistent research state, using the empirical evidence and review feedback gathered in each round to revise hypotheses, executable programs, and analytical plans and to compile methods and findings into manuscript drafts. Each reviewed cycle improves not only the research artifacts themselves, but also the candidate rankings, progress records, and distilled experience that guide subsequent decisions.

CytoCoScientist makes within-task RSI reliable over long-horizon autonomous runs through three interconnected mechanisms governing how candidate solutions are explored, how evidence is audited and admitted, and how research state is maintained. \emph{First}, multi-branch exploration tracks candidates, derivation lineages, and evaluation records in explicit candidate search spaces, decoupling branch selection from content revision. Search strategies determine which branch to expand, while review feedback and prior experience guide targeted modifications; new versions enter the search space, and newly acquired evidence can trigger re-evaluation of unchanged historical candidates. \emph{Second}, evidence-based control combines pre-execution checks, post-execution review, completion gating, and an outer-loop meta-review. Step-level safeguards verify prerequisites, determine which outputs may update candidate evaluations or support study conclusions, return diagnostic feedback on flawed runs while retaining sound negative findings to guide exploration, and verify objective coverage before termination. Complementing these step-level checks, an outer-loop meta-review periodically audits the broader trajectory from an independent context to catch overlooked flaws and redirect strategic decisions. \emph{Third}, isolated subagent execution, a revisable timeline, and provenance-linked hierarchical memory maintain research state and distill reusable experience across rounds. Detailed activity execution runs in isolated subagent contexts that return concise summaries to the main agent, while the revisable timeline and hierarchical memory keep the bounded working context current by updating superseded interpretations, retaining applicability conditions and links to primary artifacts, and supporting on-demand reuse of domain methodological skills. Admitted evaluations thus update candidate rankings for branch selection, and the revised research state directs subsequent analysis and generation.

We evaluate CytoCoScientist across five long-horizon biological studies spanning bioinformatics algorithm development and open-ended bioinformatics inquiry. Within algorithm development under sealed quantitative evaluators, single-cell RNA-seq denoising evolves a support-restricted graph propagation operator that reduces three-tissue mean squared error by $4.7\%$ over the top baseline MAGIC, and TCGA-LUAD multi-omics survival prediction rejects a higher-scoring but uncalibrated early-concatenation baseline to build a leak-free late-fusion pipeline that improves C-index by $0.0508$ over matched DeepProg. Within open-ended bioinformatics inquiry without a fixed scalar objective, three studies progress from broad clinical discovery to five-omics integration: autonomously sourcing public cohorts and pivoting from cross-sectional plasma proteomics to within-patient longitudinal trajectories in sepsis mortality; decoupling rising cell-intrinsic \textit{TOP2A} expression from declining memory B-cell proportions while auditing \textup{CXCL13} cellular origin in paired esophageal squamous cell carcinoma (ESCC) lymph nodes; and calibrating group-level shifts against participant-level concordance across five blood-derived modalities in acute aortic dissection to identify a cross-material \textit{SMAD3} association while retracting an unsupported chromatin-contact claim. Across both regimes, these results demonstrate that coupling multi-branch exploration with evidence-based control and revisable research state enables autonomous systems to refine not only computational implementations, but also the underlying biological hypotheses, analytical designs, and scientific conclusions over long horizons.

\section{Related Work}\label{sec:related-work}

Recursive self-improvement (RSI) and related self-improving methods use execution experience and evaluation feedback to refine subsequent problem solving. Prior approaches vary along two orthogonal dimensions: \emph{what is updated} (model weights, the agent's own harness, code, or control policies, or the candidates and research state of an ongoing study) and \emph{whether updates occur across tasks or within a single investigation}. At the parameter level, SEAL updates model weights from self-generated fine-tuning edits across adaptation cycles, whereas MELD jointly updates coordinator and researcher policies within a single scientific discovery process \citep{Zweiger2025SEAL,MELD2026}. Beyond model weights, a large body of work evolves the agent itself across tasks: GEPA, ADAS, and AFlow optimize prompts, agent designs, and workflows; SICA, the Darwin G\"odel Machine, and the Mendel G\"odel Machine iteratively rewrite agent code; and Meta-Harness, HarnessFlow, and EvoSkill v2 automate harness and skill optimization \citep{Agrawal2025GEPA,Hu2024AutomatedDesignAgenticSystems,Zhang2024AFlowWorkflowGeneration,Robeyns2025SelfImprovingCodingAgent,Zhang2025DarwinGodelMachine,Liu2026MendelGodelMachine,MetaHarness2026,HarnessFlow2026,EvoSkillV22026}. Within a single study, Organon and SciMARS likewise adjust search strategies, role collaboration, or memory policies during the investigation \citep{Organon2026,SciMARS2026}.

Closer to our setting, a third line of work focuses directly on evolving the candidates and research state of the current investigation. SwarmRSI revises hypotheses, programs, and shared research state from experimental feedback, and EvoSCM continually tests and revises candidate causal models through experimentation \citep{SwarmRSI2026,EvoSCM2026}; hybrid systems such as ScienceBuddy further combine harness evolution with model learning \citep{ScienceBuddy2026}. CytoCoScientist operates in this third regime, applying within-task RSI to continually update candidate solutions, candidate evaluations, research state, and domain methodological skills over a single long-horizon biological study. Unlike self-refinement loops confined to a single stage such as code optimization or hypothesis ranking, autonomous biological research must coordinate multiple interdependent activities without constant human oversight. Below, we review prior work across the three mechanisms central to sustaining this autonomous loop: search spaces and candidate evolution, evidence-based verification and execution control, and long-horizon research state and experience reuse.

\paragraph{Search Spaces and Candidate Evolution}
Sustained exploration requires retaining competing directions in an explicit candidate search space. The original ReAct framework advances tasks by alternating reasoning and acting, but does not maintain a persistent candidate pool, derivation lineages, or search statistics \citep{ReAct2023}. Program-evolution systems such as FunSearch and AlphaEvolve maintain populations of executable programs and evolve computational solutions through generation, execution, and evaluation, while ShinkaEvolve, PACEvolve, and ERA further investigate candidate scheduling, search-history management, and scientific program evolution \citep{RomeraParedes2024FunSearch,Novikov2025AlphaEvolve,Lange2025ShinkaEvolve,Yan2026PACEvolve,Aygun2026ERA}. Complementing branch selection over candidate pools, Self-Refine, Self-Debugging, and TextGrad use self-feedback, execution diagnostics, and textual critiques to guide targeted revisions of candidate content \citep{SelfRefine2023,Chen2024SelfDebugging,Yuksekgonul2024TextGrad}.

Beyond program optimization under fixed scalar benchmarks, open-ended scientific search operates over hypotheses, analyses, and models whose evidential support shifts as new findings arrive. HypoGeniC iteratively updates candidate hypotheses from sample feedback while balancing exploration and exploitation, and Google AI co-scientist evolves biomedical hypotheses through generation, critique, and tournament comparison \citep{Zhou2024HypoGeniC,Gottweis2026AcceleratingScientific3906c8d9a2f0}. HEP formalizes hypothesis generation, testing, evidence linking, and evidence-constrained updating as explicit operations with revision and merge lineages \citep{Takahara2026HEP}, while EvoSCM and SwarmRSI link experimental feedback to cross-branch model revision and recombination \citep{EvoSCM2026,SwarmRSI2026}. However, higher-ranked ideas do not reliably translate into stronger empirical outcomes once implemented \citep{Si2025IdeationExecutionGap}. Sustaining autonomous within-task RSI therefore requires decoupling parent selection from critique-guided content revision and coupling explicit candidate search spaces directly to execution-grounded evidence and cross-round re-evaluation.

\paragraph{Evidence-Based Verification and Execution Control}
Scientific feedback must both guide candidate edits and determine whether empirical results are admissible to support a claim. Popper evaluates hypotheses through sequential falsification tests, Aletheia couples solution generation with verification, and CycleResearcher and ScholarPeer study automated peer review and structured critique \citep{Huang2025Popper,Feng2026Aletheia,Weng2024CycleResearcher,Goyal2026ScholarPeer}. More recently, Triage-PiEvo separates anomalous observations that guide exploration from verified evidence permitted to modify persistent principles, retracting principles contradicted by counter-evidence \citep{TriagePiEvo2026}, while Verify by Doing inspects artifacts and environment state to jointly evolve execution and evaluation harnesses within a task and re-evaluate existing candidates \citep{VerifyByDoing2026}.

Verification also governs when an autonomous agent may advance a study or stop. SmartSnap actively acquires task-completion evidence to verify outcomes, and Evidence-Carrying Termination binds termination claims to verifiable execution evidence \citep{Cai2025SmartSnap,Liu2026ECT}. Yet most prior systems address step-level auditing or stopping criteria in isolation. In long-horizon biological research, preventing recursive error amplification and gradual strategic drift requires integrating operation preconditions, layered review, completion gating, and trajectory-level outer-loop meta-review so that flawed runs cannot distort candidate rankings and local execution success is not mistaken for achieving the research objective.

\paragraph{Long-Horizon Research State and Experience Reuse}
Long-horizon investigations quickly exceed an agent's effective context window. Multi-agent context engineering isolates verbose step execution within dedicated subagent contexts so only concise summaries return to the main agent \citep{Anthropic2025MultiAgentResearch,Anthropic2025ContextEngineering}. Across rounds, MemGPT and ACON address context limits through hierarchical memory management and trajectory compression \citep{MemGPT2023,Kang2025ACON}, while SKILL.state replaces an ever-growing dialogue history with an explicit, mutable execution state updated after validated steps \citep{Badhe2026SkillState}. Together, these approaches sustain long-horizon execution through context isolation, compression, and explicit runtime state.

Beyond maintaining working context, agents distill past trials into reusable experience along three complementary lines: extracting episodic reflections and reasoning lessons (Reflexion, ExpeL, and ReasoningBank) \citep{Shinn2023ReflexionLanguageb7573199fe22,Zhao2023ExpeL,Ouyang2025ReasoningBank}; compiling workflows, skills, and evolving contextual playbooks (Voyager, Agent Workflow Memory, ACE, Evo-Harness, and Recuris) \citep{Wang2023Voyager,Wang2024AgentWorkflowMemory,Zhang2025ACE,Wei2026EvoHarness,Yu2026Recuris}; and organizing long-term knowledge through hierarchical retrieval and mutable structured memory (RAPTOR, A-MEM, and Mem0) \citep{Sarthi2024RAPTOR,Xu2025AMEM,Chhikara2025Mem0}.

In scientific discovery, Scientist-One and xScientist adapt these ideas through evidence-linked claims and versioned research records, while JANUS and SciMARS use feedback to refine world models, skills, or memory policies \citep{Meng2026ScientistOne,Luo2026XscientistGitlikec9a109c50b3c,JANUS2026,SciMARS2026}. CytoCoScientist builds directly on this foundation by synchronizing isolated subagent execution, candidate evaluations, review verdicts, and revisable research state so that each verified revision compounds into better subsequent decisions over long-horizon studies. Representative benchmarks for scientific agents spanning computational execution, method development, study reproduction, and bioinformatics analysis are summarized in SI Section~\ref{sec:si-related-evaluation} and SI Table~\ref{tab:related-benchmarks} \citep{Chen2024ScienceAgentBench,Chan2024MLEBench,Starace2025PaperBench,Mitchener2025BixBench,Diks2026ScBenchLong}.

\section{The CytoCoScientist Framework}\label{sec:method}

CytoCoScientist operationalizes within-task recursive self-improvement (RSI) for long-horizon autonomous biological research through repeated cycles of activity selection, execution, review, and research-state updates (Figure~\ref{fig:framework-overview}). Rather than following a fixed stage pipeline, a central main agent autonomously coordinates evidence acquisition, hypothesis exploration, program evolution, bioinformatics analysis, and scientific writing through a persistent research state (Section~\ref{sec:method-overview}). Reviewed outcomes recursively update this state and refine the decision basis for subsequent rounds, allowing research artifacts, candidate evaluations, and analytical strategies to improve within a single study under fixed model parameters. This unified loop supports both quantitative bioinformatics algorithm development under defined evaluation benchmarks and open-ended bioinformatics inquiry. In open-ended studies without a fixed scalar objective, it coordinates literature synthesis, bioinformatics analysis, and discriminative experimental design.

Explicit candidate search spaces separate branch selection from content revision and retain candidates for later re-evaluation (Section~\ref{sec:method-evolution}). Evidence-based control verifies action prerequisites, audits outputs, and gates completion, complemented by an independent outer-loop meta-review of decisions across the investigation (Section~\ref{sec:method-control}). Local work proceeds in isolated subagent contexts, while a revisable timeline and hierarchical memory preserve progress and reusable experience across rounds. Domain methodological skills are discovered, composed, and refined on demand as the investigation proceeds (Section~\ref{sec:method-long-horizon}). Together, these mechanisms govern candidate exploration and revision across the core research activities (Section~\ref{sec:domain-instantiations-v2}).

\begin{figure}[!htbp]
\centering
\includegraphics[width=\linewidth]{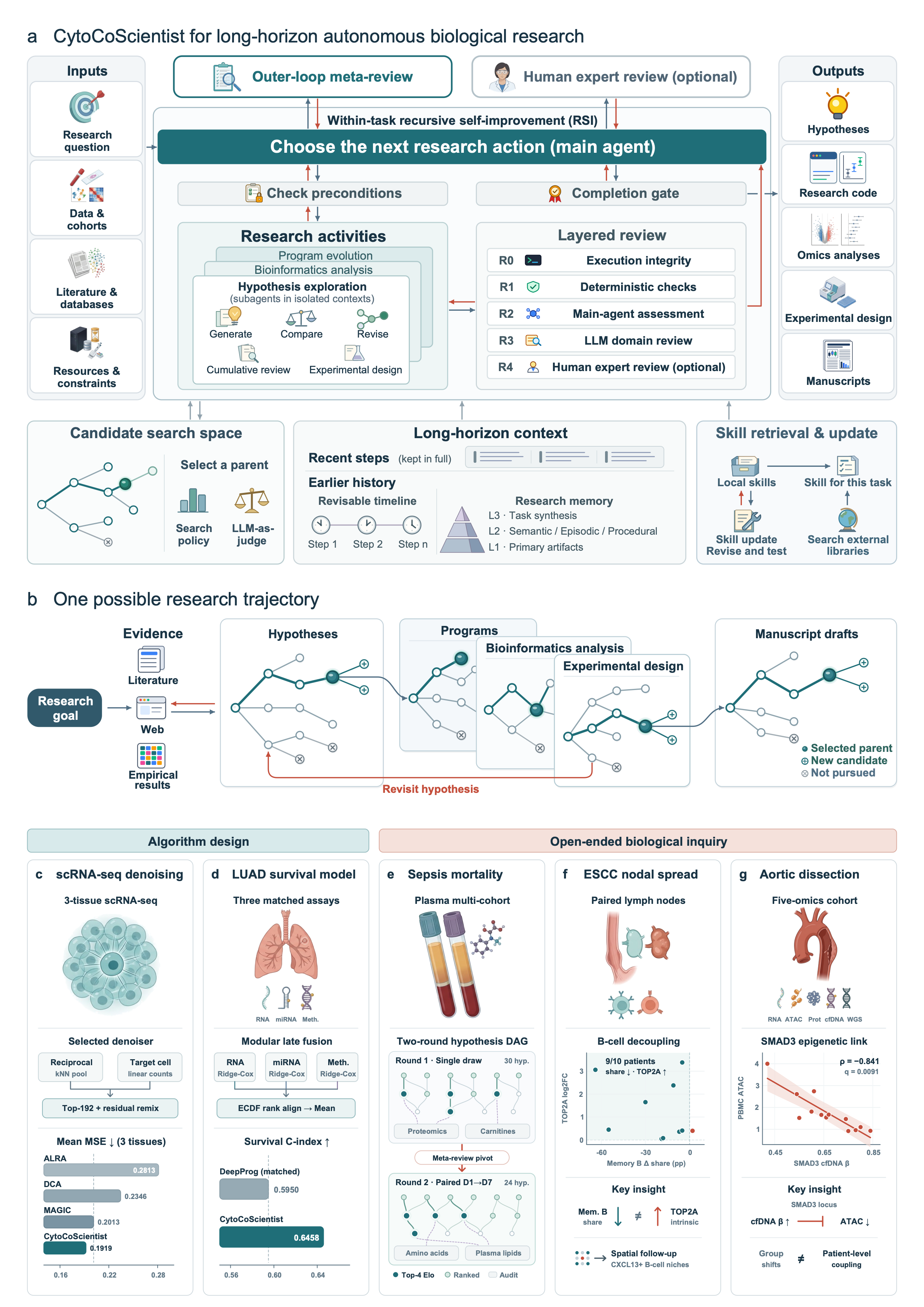}
\caption[Overview of the CytoCoScientist framework]{\textbf{How CytoCoScientist selects and revises research actions over a long-horizon task.}}
\label{fig:framework-overview}
\end{figure}
\clearpage

\begin{figure}[!htbp]
\ContinuedFloat
\captionsetup{labelformat=continuedfigure}
\caption[]{\textbf{a,} Architecture of CytoCoScientist. Given inputs on the left (research question, data and cohorts, literature and databases, and resources and constraints), the central main agent iteratively selects the next research action. Precondition checks gate actions with missing inputs or unresolved reviews before delegating research activities (stacked cards, including hypothesis exploration, bioinformatics analysis, and program evolution) to subagents running in isolated contexts. Directly coupled with this execution loop, an explicit candidate search space tracks candidate lineages and evaluations, guiding branch selection via search policies or LLM-as-judge and updating rankings from admitted outcomes. Layered review (R0--R4) audits execution integrity, deterministic checks, main-agent assessment, LLM domain review, and optional human expert review to govern which results inform subsequent rounds. Operating outside the step-level loop in fully isolated contexts, outer-loop meta-review and optional human expert review independently audit the broader trajectory and inject strategic feedback. Separately, a long-horizon context module sustains state across rounds by pairing full recent steps with a revisable timeline and three-level research memory (L1--L3), while an independent skill retrieval and update module discovers local or external domain methodological skills and revises and tests them for the current task. When the main agent requests termination, a completion gate verifies objective coverage before releasing the final outputs on the right (hypotheses, research code, omics analyses, experimental designs, and manuscripts).
\textbf{b,} Example research trajectory illustrating branching lineages across hypotheses, programs, bioinformatics analyses, experimental designs, and manuscript drafts, with red return arrows marking where experimental design prompts revisiting hypotheses and acquiring additional evidence.
\textbf{c,d,} Algorithm design under sealed quantitative evaluators. (\textbf{c}) Three-tissue scRNA-seq denoising: CytoCoScientist evolves a denoiser combining reciprocal $k$-nearest-neighbor pooling with target-cell linear counts (top-192 residual remix), reducing mean MSE to $0.1919$ vs.\ MAGIC ($0.2013$), DCA ($0.2346$), and ALRA ($0.2813$). (\textbf{d}) TCGA-LUAD survival prediction across three matched assays (RNA, miRNA, and DNA methylation): a modular late-fusion pipeline combining per-assay ridge-Cox models via ECDF rank alignment raises the survival C-index to $0.6458$ vs.\ matched DeepProg ($0.5950$).
\textbf{e--g,} Open-ended biological inquiry without a fixed scalar objective. (\textbf{e}) Cross-cohort hypothesis exploration in plasma sepsis mortality: a two-round hypothesis evolution and evaluation DAG tracks Elo ranking and auditing across 30 single-draw proteomics and carnitine hypotheses in Round 1, after which an outer-loop meta-review triggers a pivot to 24 paired day-1 to day-7 amino-acid and plasma-lipid hypotheses in Round 2. (\textbf{f}) Paired esophageal squamous cell carcinoma (ESCC) lymph nodes: B-cell decoupling separates declining memory B-cell proportion from rising cell-intrinsic \textit{TOP2A} expression (jointly observed in 9 of 10 evaluable patients) and motivates spatial follow-up of \textup{CXCL13}$^+$ B-cell niches. (\textbf{g}) Acute aortic dissection across five blood-derived modalities (RNA, ATAC, proteomics, cfDNA methylation, and WGS): cross-layer calibration links higher regional \textit{SMAD3} cfDNA methylation to lower PBMC ATAC accessibility ($\rho = -0.841$, $q = 0.0091$), distinguishing group-level shifts from patient-level cross-omics coupling.}
\end{figure}

\subsection{Research State and Agent Architecture}\label{sec:method-overview}

In long-horizon biological research, emerging findings frequently require backtracking across stages. CytoCoScientist therefore enables autonomous end-to-end execution through a central main agent that schedules research activities adaptively. At each step, the main agent inspects current progress, accumulated evidence, and unresolved bottlenecks to invoke evidence acquisition or evolve persistent \emph{research objects}, including scientific hypotheses, executable programs, bioinformatics analyses, and manuscripts.

Local generation, critique, and domain review run in isolated subagent contexts, returning concise summaries to the main agent to guide subsequent decisions (Section~\ref{sec:method-long-horizon}). Operating outside the step-level loop in an independent context, outer-loop meta-review periodically audits execution records and intermediate results, identifying overlooked issues and returning strategic feedback to the main agent (Section~\ref{sec:method-meta-review}).

To decouple long-horizon autonomous exploration from linear conversation histories, the system maintains a persistent research state $S_t = (\mathcal X_t,\mathcal E_t,\mathcal H_t)$ and drives within-task RSI through a closed decision, execution, and state-update loop:
\[
S_t \xrightarrow{\text{assemble context}} C_t
\xrightarrow{\text{select and execute }(g_t,a_t)\in\Omega_t} o_t
\xrightarrow{\text{review and update state}} S_{t+1}.
\]
Within $S_t$, three state components support the framework's core mechanisms:
\begin{itemize}
\item \textbf{Candidate space $\mathcal X_t$} (Section~\ref{sec:method-evolution}): Stores candidate solutions across research objects alongside their derivation lineages and dependency links. For each research goal or subproblem $g$, a dedicated search state $Q_t^g$ aggregates candidate nodes, admitted evaluations, and exploration statistics.
\item \textbf{Evidence and review store $\mathcal E_t$} (Section~\ref{sec:method-control}; SI Section~\ref{sec:si-domain-instantiations}): Records retrieved literature and web evidence, empirical outputs from code and data analyses, and review verdicts under their evaluation conditions.
\item \textbf{History and hierarchical memory $\mathcal H_t$} (Section~\ref{sec:method-long-horizon}): Maintains the full interaction log, the revisable timeline, and distilled research experience, while primary code, data matrices, and figures remain stored on disk.
\end{itemize}
Over long horizons, the full state $S_t$ exceeds the model's context window. The main agent therefore assembles a bounded working context $C_t$ from task instructions, recent interactions, and the revisable timeline, retrieving relevant evidence, historical lessons, and domain methodological skills on demand (Section~\ref{sec:method-long-horizon}).

Within the decision loop, operation preconditions determine the eligible action space $\Omega_t$ by verifying that required inputs and prior reviews are settled. Given $C_t$, the main agent selects a target goal $g_t$ and a bounded activity $a_t$. Typical activities include retrieving literature, evolving hypotheses or executable programs, running a bioinformatics analysis, and revising a manuscript section. Each operation produces an observation $o_t$ containing its results, diagnostic feedback, and artifact references, and then returns control to the main agent.

After execution, layered review determines which results in $o_t$ can update candidate evaluations or support study conclusions (Section~\ref{sec:method-control}). Admitted evaluations update candidate rankings in $\mathcal X_t$, and new evidence added to $\mathcal E_t$ can trigger re-evaluation of existing candidates, whereas failed runs retain diagnostic feedback for repair. In parallel, the revisable timeline and hierarchical memory carry revised interpretations and accumulated experience forward in $\mathcal H_t$, while execution and review feedback also guides within-task refinement of domain methodological skills. Together, these updates form the next state $S_{t+1}$, closing the within-task RSI loop so that subsequent decisions draw on updated rankings, verified evidence, distilled lessons, and refined skills. When the main agent requests termination, completion gating verifies the accumulated evidence and deliverables against the research goal.

\subsection{Explicit Search Spaces and Candidate Evolution}\label{sec:method-evolution}

Tracking only the latest attempt makes it difficult to compare competing research directions across rounds. CytoCoScientist therefore couples the main agent's activity scheduling with persistent search states that retain candidate lineages, evaluations, and exploration statistics. Maintaining these records outside the conversation context allows the autonomous main agent to backtrack and revisit earlier branches across long-horizon runs without reconstructing the candidate pool from the interaction history.

As shown in Algorithm~\ref{alg:react-evolution-v2} (where $\mathcal Q = \{Q_t^g\}$ collects the per-goal search states within $\mathcal X_t$), the system embeds a ``select--expand--evaluate--update'' cycle within the main agent's decision loop. A candidate-expanding activity queries $Q_t^g$ for parent versions and draws on review critiques and memory lessons to generate revised or combined offspring, whereas other activities acquire new evidence or audit existing candidates without creating a new version. Although all execution outcomes return to the main agent, only admitted evaluations update candidate utilities and candidate rankings.

\begin{algorithm}[tbp]
\caption{\textbf{Main-agent decision loop with persistent candidate search and evidence-based control.} Given the initial context and research repositories, the loop iteratively filters eligible actions ($\Omega$), couples main-agent decision steps with parent selection (\textsc{SelectProposal}) and review-gated search updates (\textsc{UpdateSearch}), and verifies termination requests through \textsc{CompletionGate}. Outer-loop meta-review is described separately in Section~\ref{sec:method-meta-review}.}
\label{alg:react-evolution-v2}
\begingroup
\algrenewcommand{\algorithmicrequire}{\textbf{Input:}}
\algrenewcommand{\algorithmicensure}{\textbf{Output:}}
\algrenewcommand{\algorithmicwhile}{\textbf{while}}
\algrenewcommand{\algorithmicdo}{\textbf{do}}
\algrenewcommand{\algorithmicif}{\textbf{if}}
\algrenewcommand{\algorithmicthen}{\textbf{then}}
\algrenewcommand{\algorithmicend}{\textbf{end}}
\algrenewcommand{\algorithmicreturn}{\textbf{return}}
\begin{algorithmic}[1]
\Require Initial context $C_0$, domain search states $\mathcal Q_0$, evidence store $\mathcal E_0$
\Ensure Final context $C$, domain search states $\mathcal Q$, and evidence store $\mathcal E$
\State $C \gets C_0$, $\mathcal Q \gets \mathcal Q_0$, $\mathcal E \gets \mathcal E_0$
\While{\textbf{true}}
    \If{$\Call{RequestsTermination}{C}$}
        \If{$\Call{CompletionGate}{C,\mathcal Q,\mathcal E}$} \Comment{Gate 3: verify completion readiness}
            \State \Return $(C,\mathcal Q,\mathcal E)$
        \EndIf
        \State $C \gets \Call{AppendCompletionGaps}{C}$
    \EndIf
    \State $\Omega \gets \Call{EligibleActions}{C,\mathcal Q,\mathcal E}$ \Comment{Gate 1: operation preconditions}
    \State $(g,a) \gets \Call{Think}{C,\Omega}$ \Comment{Select research goal $g$ and activity $a$}
    \State $\eta \gets \varnothing$
    \If{$\Call{ProposesCandidate}{a}$}
        \State $\eta \gets \Call{SelectProposal}{\mathcal Q[g],a}$ \Comment{Select parent candidates from $Q_t^g$}
    \EndIf
    \State $(o,\mathcal E) \gets \Call{Act}{g,a,\eta,C,\mathcal E}$ \Comment{Generate, execute, or review candidates}
    \If{$\Call{HasSearchRecords}{g,a,\eta,o}$ \textbf{and} $\Call{Admissible}{o}$} \Comment{Gate 2: R0--R4 layered review}
        \State $\mathcal Q[g] \gets \Call{UpdateSearch}{\mathcal Q[g],a,\eta,o}$ \Comment{Update $Q_t^g$ with admitted evaluations}
    \EndIf
    \State $C \gets \Call{Observe}{C,a,o}$ \Comment{Incorporate outcomes and critiques into $C_t$}
\EndWhile
\end{algorithmic}
\endgroup
\end{algorithm}

\subsubsection{Search Space Representation and Candidate Lineage}\label{sec:evolution-representation-v2}

For each research goal $g$, the system maintains an explicit candidate search space $Q_t^g$ and organizes candidate evolution as a directed acyclic lineage graph $G_t^g = (V_t^g, E_t^g)$. Each node $v \in V_t^g$ links a candidate artifact to its evaluation history, while directed edges $(u \to v) \in E_t^g$ record whether a candidate derives from modifying a single parent or combining multiple parents. Depending on the research activity, each node represents a hypothesis, an executable program, a bioinformatics analysis step, or a manuscript version, with activity-specific artifacts summarized in Table~\ref{tab:domain-instantiations-v2}.

Because biological studies span both metric-driven algorithm development and open-ended bioinformatics inquiry without a fixed scalar objective, this representation supports goal-specific evaluation criteria (Section~\ref{sec:domain-instantiations-v2}; SI Section~\ref{sec:si-domain-instantiations}). Across all four research objects, decoupling candidate content from its supporting evidence and evaluation records keeps historical branches traceable and allows them to be re-evaluated as new evidence accumulates.

\subsubsection{Candidate Selection, Feedback-Driven Revision, and State Update}\label{sec:evolution-selection-v2}\label{sec:evolution-evaluation-v2}

CytoCoScientist separates \emph{which candidate to explore next} from \emph{how to revise it}: search strategies select parent candidates, while review feedback and prior experience guide changes to their content.

For candidate selection, the system supports both pluggable search algorithms and direct LLM-as-judge selection according to the evaluation structure of the research object. Under score-based policies such as Upper Confidence Bound (UCB) and rank-prior tree search (Flat Rank PUCT), $Q_t^g$ tracks candidate utility and policy-specific exploration statistics to balance exploitation of high-scoring candidates with exploration of less-visited branches \citep{UCB2002,Browne2012SurveyMonte5004524835d3,Silver2017MasteringChess8aa374b3b78c}. Through the same selection interface, Bayesian optimization (BO), Island MAP-Elites, and multi-objective Pareto selection (NSGA-II) respectively account for predictive uncertainty, diversity across behavioral descriptors, and trade-offs among objectives \citep{Snoek2012PracticalBayesian,Bischl2023HyperparameterOptimizationb308b5f85fc5,MAPElites2015,Pugh2016QualityDiversity6585d7b4eb89,NSGA2002}. When parent selection instead requires qualitative domain judgment, as in bioinformatics analysis and manuscript revision, the system invokes an LLM-as-judge over candidate content, current evidence, and historical evaluations.

Once parent nodes are selected, the generator draws on accumulated review critiques and lessons from research memory to revise a single parent or recombine complementary elements across parents. Conditioning offspring on review diagnostics rather than unguided perturbations allows candidate quality to improve across rounds. New candidates and their derivation edges are registered in $G_t^g$, while execution and review yield two complementary signals \citep{Zheng2023JudgingLLM,Weng2024CycleResearcher,Goyal2026ScholarPeer}: admitted numerical scores or comparative ratings update candidate rankings, and diagnostic critiques guide subsequent revisions.

These signals drive two distinct state-update paths. \emph{Modifying candidate content} creates a new node whose eligibility for further expansion depends on layered review (Section~\ref{sec:method-evaluation}). \emph{Expanding the evidence} in $\mathcal E_t$ can instead update the evaluation of an existing candidate without altering its historical artifact \citep{Takahara2026HEP}. Revised evaluations then redirect autonomous branch selection and update the interpretations carried forward in research memory.

\subsection{Evidence-Based Execution Control and Research Supervision}\label{sec:method-control}

Because candidate search amplifies whatever evaluation signals enter $Q_t^g$, successful code execution or a higher scalar metric alone cannot establish methodological soundness or resolve the overarching research question \citep{Trehan2026ResearchFailures,Son2025SPOT}. In biological datasets, measurement noise, patient-level leakage, sample confounding, and cell-level pseudoreplication frequently produce spurious gains. Admitting such flawed evaluations into $Q_t^g$ amplifies errors across RSI rounds, distorts candidate rankings, and can trigger premature termination. CytoCoScientist therefore stabilizes long-horizon autonomous execution across three stages of the action lifecycle: operation preconditions verify required inputs before execution, layered review determines whether outputs can inform candidate evaluations or scientific claims, and completion gating checks whether termination requirements are met. Operating outside the step-level loop in an independent context, outer-loop meta-review further audits the broader execution trajectory periodically, returning strategic feedback on issues that step-level review may miss.

\subsubsection{Operation Preconditions and Local Constraints}\label{sec:method-operation-preconditions-v2}

Although the main agent can flexibly switch among research activities as evidence evolves, each operation depends on its required inputs and prior reviews. We formalize these local constraints as \emph{operation preconditions}. For example, launching a new expansion batch in hypothesis or program search requires the preceding batch to be evaluated, reviewed, and incorporated into $Q_t^g$, preventing unsettled evaluations from biasing subsequent parent selection. Similarly, downstream analyses require validated upstream artifacts, and manuscript operations depend on the corresponding outline, section, or figure versions. These dependencies safeguard individual actions without imposing a rigid stage sequence on the investigation. When a precondition is unmet, the system blocks the call and reports the missing prerequisite so the main agent can resolve it first.

\subsubsection{Layered Review}\label{sec:method-evaluation}

After an operation executes, the system assesses whether its output is computationally and scientifically reliable across five complementary review levels (R0--R4; Table~\ref{tab:review-levels}), invoked on demand according to the operation type. Deterministic checks establish execution integrity (R0) and validate schemas, numerical bounds, sample identifiers, and evaluation protocols (R1). Within the \textsc{Observe}--\textsc{Think} loop, the main agent inspects outputs alongside R0--R1 diagnostics (R2) to route runtime repairs, supplementary analyses, or formal domain review. For scientific scrutiny, an isolated LLM-as-judge reviewer (R3) audits methodological soundness, evidential support, and research integrity \citep{Zheng2023JudgingLLM,Wang2024GeneAgent,Weng2024CycleResearcher,Goyal2026ScholarPeer,Huang2025Popper}, returning admissibility verdicts and revision critiques. Whereas R0--R3 support fully autonomous end-to-end execution, optional human-in-the-loop review (R4) allows domain experts to inspect key hypotheses, experimental designs, or stage conclusions when requested.

\begin{table}[!htbp]
\centering\small
\caption{\textbf{Layered review hierarchy (R0--R4) in CytoCoScientist.} Each level targets a distinct verification scope and produces structured feedback to govern state admission and candidate revision.}
\label{tab:review-levels}
\begin{tabularx}{\linewidth}{@{}>{\raggedright\arraybackslash}p{0.10\linewidth}>{\raggedright\arraybackslash}p{0.28\linewidth}>{\raggedright\arraybackslash}X@{}}
\toprule
\textbf{Level} & \textbf{Review role} & \textbf{Checks and feedback} \\
\midrule
R0 & Execution layer & Process status, output completeness, and runtime failure diagnostics \\
\addlinespace[0.5em]
R1 & Deterministic validator & Schema, numerical, identifier, and protocol checks, plus validation status \\
\addlinespace[0.5em]
R2 & Main agent (in-loop assessment) & Contextual assessment against current goal; repair or next-action routing \\
\addlinespace[0.5em]
R3 & LLM domain reviewer & Methodological soundness, evidential support, and integrity, plus admissibility and revision critiques \\
\addlinespace[0.5em]
R4 & Human expert review (optional) & Expert approval, scientific corrections, or updated task constraints \\
\bottomrule
\end{tabularx}
\end{table}

When integrating verdicts, the system enforces deterministic validation first and separates methodological validity from positive effect size. R0--R1 execution or schema failures cannot be overridden by favorable semantic judgments, though their diagnostic logs remain available for repair. Accordingly, a methodologically sound analysis that yields a low score or null finding is retained as valid evidence, whereas a high-scoring run compromised by data leakage, pseudoreplication, or confounding is rejected so it cannot distort the candidate search space. Review critiques and any unresolved qualifications remain attached to the output to guide subsequent revisions.

\subsubsection{Completion Gating}\label{sec:method-completion-v2}

To prevent local progress from being mistaken for a completed study, CytoCoScientist applies \emph{completion gating} whenever the main agent requests termination \citep{Cai2025SmartSnap,Liu2026ECT,Ma2026LongHorizonHarness}. Programmatic checks and an LLM audit compare the accumulated work against the approved research goal \citep{Zheng2023JudgingLLM}, verifying that adopted conclusions have passed their required validations and reviews, each sub-objective has been addressed with the evidence required by the goal, and all required outputs have been produced in their latest reviewed versions.

When required analyses, reviews, or outputs remain outstanding, completion gating rejects the termination request and writes the specific gaps back into $C_t$, prompting the main agent to perform supplementary work or finalize missing outputs. Neither a favorable step-level review nor outer-loop meta-review substitutes for this goal-level check.

\subsubsection{Outer-Loop Meta-Review}\label{sec:method-meta-review}

Over long autonomous horizons, step-level checks alone may miss gradual strategic drift or repeated ineffective loops. Operating outside the step-level execution loop in an independent context, outer-loop meta-review periodically reassesses the broader investigation. Drawing on trace-based diagnosis and reflective feedback \citep{MetaHarness2026,Agrawal2025GEPA}, it examines historical decisions, execution records, and supporting artifacts against the overarching research goal. The review identifies repeated ineffective actions, overlooked evidence, and assumptions that warrant reconsideration, and returns targeted strategic feedback to the main agent. This feedback can trigger a revised analysis, additional validation, or a shift in research direction, while the resulting outputs remain subject to layered review and completion gating.

\subsection{Long-Horizon Research State and Experience Reuse}\label{sec:method-long-horizon}

\begin{figure}[H]
\centering
\includegraphics[width=\linewidth]{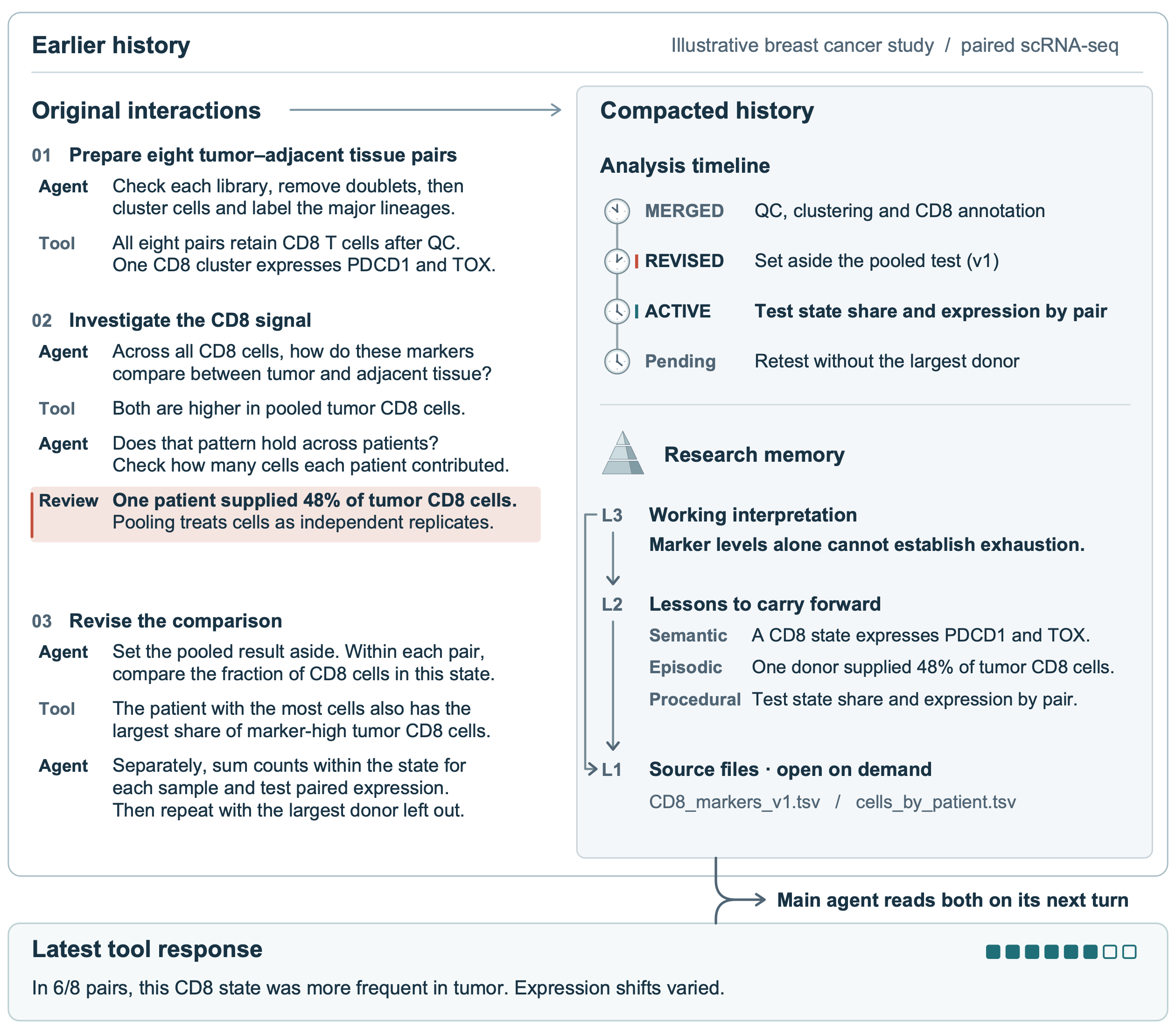}
\caption{\textbf{How earlier interactions and the current exchange form the next-turn context.} In this illustrative paired breast cancer scRNA-seq analysis, one patient contributed 48\% of the tumor CD8 cells. The timeline records why the pooled comparison was set aside, which paired analysis replaced it, and which sensitivity check remains pending. Research memory retains what was learned: arrows trace the working interpretation (L3) to the relevant lessons (L2) or directly to source files (L1), and link L2 lessons to their sources. The current tool response is kept verbatim; the agent receives it alongside the compacted history, task instructions, and relevant retrieved material.}
\label{fig:context-compression-memory}
\end{figure}

Sustaining within-task RSI over long-horizon autonomous studies requires keeping decision-relevant context accurate as tool executions, candidate revisions, literature searches, and review feedback quickly exceed the model's effective context window \citep{LostMiddle2024,Kang2025ACON}. CytoCoScientist therefore isolates detailed step execution in isolated subagent contexts and returns concise summaries to the main agent. Across rounds, a revisable timeline records completed steps, shifts in direction, and pending work, while hierarchical memory distills research findings and lessons from prior attempts (Figure~\ref{fig:context-compression-memory}). Complementing these state representations, domain methodological skills are discovered, loaded, and refined on demand to guide analytical workflows for the current task.

\subsubsection{Context Isolation across Research Activities}\label{sec:method-context-isolation}

While the main agent coordinates the overall investigation, the internal execution of each research activity proceeds in an isolated subagent context \citep{Anthropic2025MultiAgentResearch,Anthropic2025ContextEngineering}. Individual activities often span multiple roles and steps, such as hypothesis generation, comparison, revision, and experimental design (Section~\ref{sec:method-hypothesis-evolution}), or the Strategist--Skeptic--Planner collaboration in program evolution (SI Section~\ref{sec:si-program-planning}). Distilled findings, review critiques, unresolved issues, and artifact references return to the main agent to guide subsequent decisions, whereas verbose intermediate exchanges remain outside the main conversation with full records preserved on disk for inspection when needed.

\subsubsection{Preserving Research Progress through Compaction}\label{sec:method-timeline-compaction}

Across studies spanning many operations, an agent must track which steps have been executed and revise earlier interpretations when subsequent evidence alters their support \citep{Badhe2026SkillState}. Building on cumulative compaction \citep{Pi2026Compaction}, which updates an existing summary with newly elapsed interactions, CytoCoScientist maintains a revisable timeline $T_k$ through \emph{pipeline timeline compaction}. At compaction step $k$, this timeline records key steps and their current status, empirical observations, decision rationales, and pending tasks, while retaining artifact paths and candidate version identifiers so the main agent can inspect primary records whenever a later decision requires finer detail.

When the active prompt approaches the input token budget (reserving space for model generation and a safety margin), the system keeps recent interactions intact and folds an earlier history prefix into $T_k$ to produce $T_{k+1}$. Timeline updates perform three state-maintenance operations (Figure~\ref{fig:context-compression-memory}):
\begin{itemize}
\item \textbf{Merging routine completed steps} (\texttt{MERGED}): The system consolidates completed prerequisite steps (such as data import and routine quality control) into a single milestone with summary statistics and output paths, whereas steps involving unresolved errors or key sample exclusions retain detailed records.
\item \textbf{Revising earlier step conclusions} (\texttt{REVISED}): When subsequent validation or R3 domain review overturns an earlier verdict (for example, identifying that a pooled comparison treats cells from the same patient as independent and requiring a patient-level comparison), the timeline preserves the historical execution record but updates its status to \texttt{REVISED} alongside the rationale for the change.
\item \textbf{Registering newly initiated work} (\texttt{ACTIVE}): Newly launched analyses, candidate expansions, or pending reviews are registered on the timeline as \texttt{ACTIVE} entries with their live progress, clearly separating settled empirical results from ongoing or planned work.
\end{itemize}

\subsubsection{Hierarchical Research Memory}\label{sec:method-hierarchical-memory}

The revisable timeline tracks where the study stands, and hierarchical memory distills what has been learned across attempts. It records biological findings, cohort- or assay-specific applicability conditions, and failure modes so later branches avoid overgeneralizing across datasets. Drawing on multilevel memory architectures \citep{GenerativeAgents2023,MemGPT2023,Sarthi2024RAPTOR,Sumers2023CognitiveArchitectures5dff93153fff}, the system organizes research memory into three interconnected tiers (L1--L3 in Figure~\ref{fig:context-compression-memory}):
\begin{itemize}
\item \textbf{L1 (Primary artifacts):} Tier L1 stores code versions, data matrices, analytical outputs, literature passages, and review records on disk, with file paths and version identifiers indexed in upper-tier summaries so the main agent can inspect primary evidence on demand without inflating the context window \citep{MemGPT2023,Ji2025ManusContextEngineering,Anthropic2025ContextEngineering}.
\item \textbf{L2 (Attempt- and object-level experiential knowledge):} Tier L2 distills L1 artifacts and stage summaries into three forms of knowledge \citep{Sumers2023CognitiveArchitectures5dff93153fff}: \emph{semantic} knowledge records findings and object-level summaries; \emph{episodic} knowledge captures individual attempts and outcomes; and \emph{procedural} knowledge describes reusable methods, validation checks, and tool-use practices.
\item \textbf{L3 (Task-level knowledge synthesis):} Tier L3 maintains a global synthesis centered on the overarching research question, integrating active L2 records across research objects and activities while preserving applicability conditions, cross-source disagreements, and provenance links back to L2 and L1.
\end{itemize}

Routine outputs append new experience records to L2, whereas comprehensive reviews (such as hypothesis tournament reviews and program retrospectives) revise the accumulated lessons for the corresponding research object \citep{Xu2025AMEM,Chhikara2025Mem0}. Each updated summary replaces its predecessor in active memory, avoiding redundant retrieval while preserving the underlying history. L3 then integrates new L2 records and updated object-level summaries into the task-level synthesis, revising interpretations as evidence shifts while retaining supported prior knowledge. At each autonomous decision step, the working context combines the latest L3 synthesis, recent unsynthesized L2 entries, and relevant historical L2 records, allowing subsequent RSI rounds to build on corrected findings.

\subsubsection{Research Skill Discovery and Refinement}\label{sec:method-skill-loading}

To support long-horizon autonomous execution across diverse biological domains without modifying core control logic, CytoCoScientist organizes domain methodological knowledge and analytical workflows into modular skills \citep{Li2026SkillsBench,Liu2026SkillsInTheWild}. Each skill pairs an execution procedure with its applicability conditions and validation guidance, allowing the main agent to retrieve and load relevant instructions and supporting materials on demand \citep{Anthropic2026ClaudeCodeSkills,Anthropic2025ContextEngineering}.

The local skill library spans genomic variant analysis, bulk, single-cell, and spatial transcriptomics, chromatin accessibility, DNA methylation, quantitative proteomics, and multi-omics integration, covering procedures from cell-state annotation and donor-level differential comparisons to spatial domain analysis. The five-omics aortic dissection study illustrates why these skills must sometimes be composed across modalities: RNA, chromatin accessibility, and plasma protein measurements were compared within participants, whereas comparisons with cell-free DNA methylation also required alignment by genomic region.

When a study encounters an analytical requirement not covered by existing local skills, the main agent autonomously searches available skill repositories, composes complementary procedures, and adapts them to the current dataset and protocol without manual user selection \citep{Gao2026CompositionalSkillRouting}. Within the ongoing investigation, execution outcomes and review feedback guide the testing and refinement of these domain methodological skills \citep{Wang2023Voyager,Zheng2025SkillWeaver,Yang2026AutoSkill,Lin2026MUSEAutoskill}, so that improved analytical procedures can be reused reliably across later iterations of the same long-horizon study (and optionally retained for future tasks).

\subsection{Domain Instantiations Across the Research Lifecycle}
\label{sec:domain-instantiations-v2}

Biological research spans activities with fundamentally different evaluation signals, ranging from scalar benchmark metrics in algorithm design to pairwise judge rankings in hypothesis evolution, DAG-level statistical diagnostics in multi-omics analysis, and claim-to-evidence audits in manuscript drafting. CytoCoScientist therefore instantiates the unified state $(\mathcal{E}_t, Q_t^g)$ and evidence-based control loop differently for each activity (Table~\ref{tab:domain-instantiations-v2}). Multi-source evidence acquisition expands and audits the evidence store $\mathcal{E}_t$, whereas hypothesis evolution, algorithm and program evolution, bioinformatics data analysis, and scientific manuscript writing evolve candidate lineages in $Q_t^g$. Across activities, newly verified evidence in $\mathcal{E}_t$ triggers re-evaluation of existing candidates in $Q_t^g$, while unresolved questions in $Q_t^g$ launch targeted retrieval or follow-up analyses, allowing within-task RSI to operate across the full research lifecycle. Below, we detail multi-source evidence acquisition and grounding (Section~\ref{sec:method-evidence-acquisition}) and hypothesis evolution and experimental design (Section~\ref{sec:method-hypothesis-evolution}); corresponding instantiations for algorithm and program evolution, bioinformatics data analysis, and scientific manuscript writing, together with the implementation and model setup, appear in SI Section~\ref{sec:si-domain-instantiations} (including representative standalone manuscripts generated for both open-ended bioinformatics inquiry and algorithm-development studies in SI Figures~\ref{fig:si-writing-escc-manuscript}--\ref{fig:si-writing-coding-manuscript}).

\begin{table}[!htbp]
\centering\small
\caption{\textbf{Delegated subagent tasks, target states, and outputs across five research activities.} The main agent invokes activities adaptively in any order and delegates each task in column~2 to a subagent running in an isolated context. In bioinformatics data analysis, the DAG records dependencies between analysis steps, while sibling candidates represent alternative versions of the same step.}
\label{tab:domain-instantiations-v2}
\begin{tabularx}{\linewidth}{@{}>{\raggedright\arraybackslash}p{0.23\linewidth}*{2}{>{\raggedright\arraybackslash}X}@{}}
\toprule
\textbf{Research activity} & \textbf{Delegated subagent tasks} & \textbf{Target state and outputs} \\
\midrule
Multi-source evidence acquisition & Literature search, passage retrieval, web evidence collection, evidence review & Evidence store $\mathcal{E}_t$: literature and web materials, retrieved passages, source links, and evidence review records \\
\addlinespace[0.5em]
Hypothesis evolution and experimental design & Hypothesis generation, hypothesis evaluation, hypothesis evolution, cumulative review, experimental design & Hypothesis search space $Q_t^g$: candidate hypotheses, derivation lineages, evaluation and cumulative review records, and associated experimental protocols \\
\addlinespace[0.5em]
Algorithm and program evolution & Plan formulation, program generation, program evolution, execution-based evaluation, cumulative review, candidate revalidation & Program search space $Q_t^g$: modification plans and code versions, derivation lineages, execution results, evaluation and revalidation records \\
\addlinespace[0.5em]
Bioinformatics data analysis & Analysis planning, analysis execution, analysis evolution, result evaluation, cumulative review & Analysis search space $Q_t^g$: analysis scripts and candidate versions, step dependencies and derivation relations, analysis results, evaluation and cumulative review records \\
\addlinespace[0.5em]
Scientific manuscript writing & Outline planning, manuscript generation, manuscript assembly, manuscript evaluation, manuscript evolution & Manuscript search space $Q_t^g$: outlines and drafts at subsection, section, and manuscript levels, version lineages, evaluation and revision records \\
\bottomrule
\end{tabularx}
\end{table}

\subsubsection{Multi-Source Evidence Acquisition and Grounding}\label{sec:method-evidence-acquisition}

Credible claims and analytical interpretations depend on traceable evidence. CytoCoScientist gathers evidence from literature passages, database and documentation extracts, and empirical outputs (SI Section~\ref{sec:si-evidence-acquisition}). Literature search builds a deduplicated corpus and combines dense semantic retrieval with lexical matching through reciprocal rank fusion \citep{Lewis2020RetrievalaugmentedGeneration3cb08b0e9ded,Lala2023PaperQA,Skarlinski2024PaperQA2,DPR2020}. Database records and software documentation supply annotations and method-specific constraints, while analysis or experimental readouts contribute empirical observations. Each extraction retains its source and the conditions relevant to its interpretation.

Retrieved passages and external documents undergo question-specific review before updating $\mathcal E_t$. The reviewer assesses source reliability and relevance, identifying supporting evidence, conflicting findings, and unresolved gaps. Missing details prompt supplementary retrieval, while conflicting conclusions lead the main agent to examine differences in study conditions. When literature gaps persist, the main agent launches targeted analyses, proposes discriminative experiments, or narrows the corresponding manuscript claims.

\subsubsection{Hypothesis Evolution and Experimental Design}\label{sec:method-hypothesis-evolution}

\begin{figure}[H]
\centering
\includegraphics[width=\linewidth]{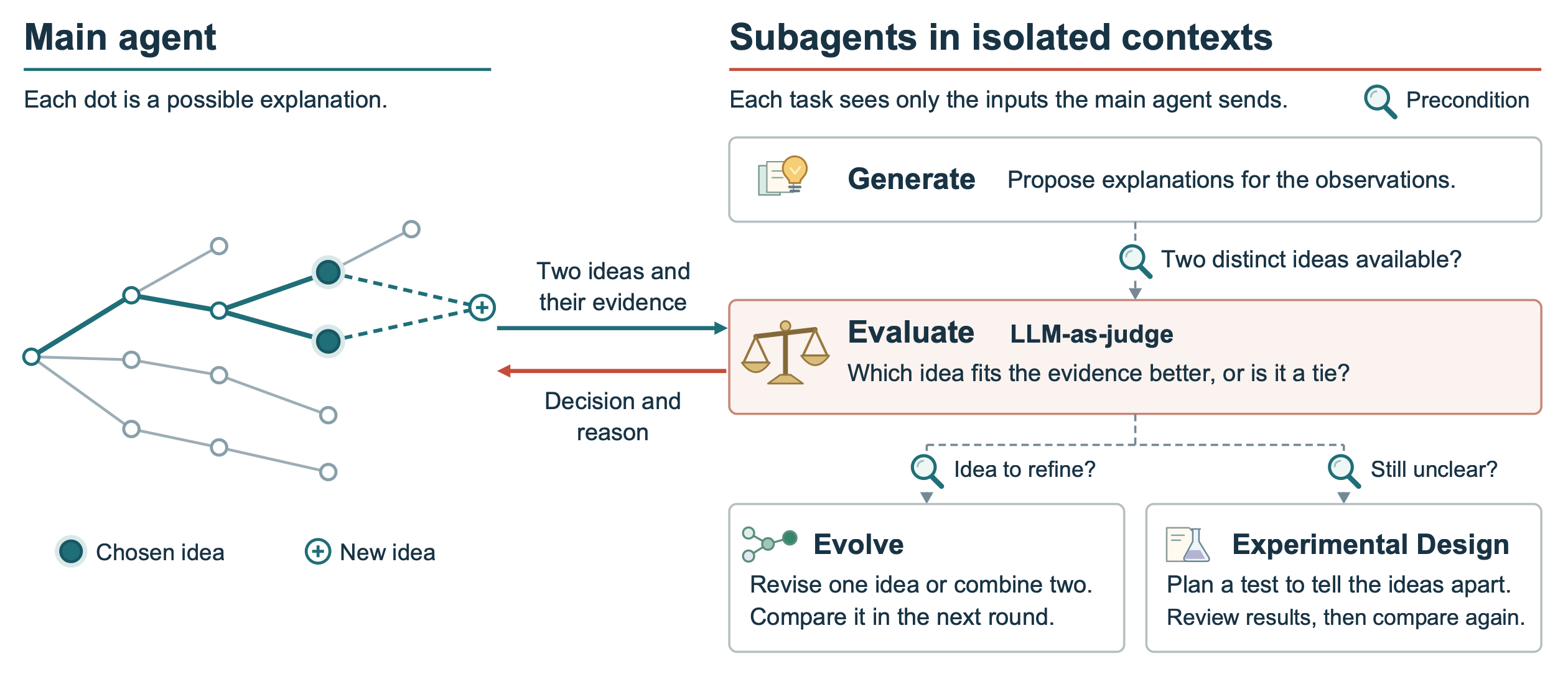}
\caption{\textbf{Iterative hypothesis evolution and discriminative experimental design over a persistent search space.}
This figure expands the hypothesis exploration activity in Figure~\ref{fig:framework-overview}. The main agent (left) maintains the lineage graph of candidate hypotheses and delegates one task per call to subagents running in isolated contexts (right), where dashed grey arrows with magnifying-glass icons indicate the preconditions for each task. The colored arrows illustrate a single Evaluate call: the main agent passes two candidate hypotheses and their selected evidence to an LLM-as-judge subagent and receives a decision and rationale, which update cumulative Elo ratings across round-robin comparisons. In the lineage graph, hollow nodes denote candidate hypotheses, filled teal nodes (Chosen idea) denote UCB-selected parents, and dashed teal edges show two-parent combination into a new candidate marked with a plus sign (New idea). Across rounds, Generate seeds initial hypotheses, Evolve refines a single parent or combines two parents using tournament rationales and cumulative review, and Experimental Design proposes discriminative tests when current evidence cannot separate competing explanations, feeding reviewed observations back into subsequent evaluations.}
\label{fig:domain-hypothesis-evolution-v2}
\end{figure}

CytoCoScientist organizes hypothesis evolution and experimental design around a persistent hypothesis search space (Figure~\ref{fig:domain-hypothesis-evolution-v2}). For a research question $g$, initial generation draws on current evidence to propose a batch of hypotheses with distinct explanations or testable predictions, which enter the lineage graph $G_t^g$ after semantic deduplication \citep{Gottweis2026AcceleratingScientific3906c8d9a2f0,Takahara2026HEP}.

Because untested hypotheses lack a direct empirical metric, an LLM-as-judge tournament compares all eligible candidates in round-robin pairs under the same research question and current evidence. Each comparison returns a win, loss, or draw with an accompanying rationale, updating cumulative Elo ratings that track relative evidential support across the candidate pool. A cumulative review synthesizes supporting findings, counterevidence, and unresolved disagreements to guide further exploration.

In subsequent rounds, UCB combines the comparative ratings with parent-selection counts to balance promising and less-explored hypotheses \citep{UCB2002,Gottweis2026AcceleratingScientific3906c8d9a2f0}. Guided by current evidence, comparison rationales, and the cumulative review, the evolution step refines a single parent or combines two compatible parents. New candidates and their derivation edges enter $G_t^g$ and are evaluated alongside existing candidates in subsequent tournament rounds.

When evidence cannot resolve competing predictions, the main agent invokes discriminative experimental design to specify tests, controls, and outcome interpretations \citep{Roohani2024BioDiscoveryAgent,Huang2025Popper}, then either launches a computational analysis or outputs a protocol for external execution. Once empirical results are obtained and pass layered review, the reviewed observations enter $\mathcal E_t$ to re-evaluate competing hypotheses in $Q_t^g$ or drive a new round of autonomous inquiry.

\section{Experiments}\label{sec:experiments}

Autonomous biological research spans two complementary regimes: quantitative bioinformatics algorithm development under sealed evaluators and open-ended bioinformatics inquiry without a fixed scalar objective. We evaluate CytoCoScientist across five long-horizon biological studies spanning single-omics to multi-omics settings along a progression of task openness and statistical difficulty (Table~\ref{tab:exp-overview}). Within \emph{bioinformatics algorithm development}, two studies progress from single-omics operator synthesis in single-cell RNA-seq denoising (Section~\ref{sec:exp:denoising}) to multi-omics clinical survival modeling in lung adenocarcinoma (Section~\ref{sec:exp:luad}), testing whether the system can move beyond fixed model templates and prioritize methodological validity.

Within \emph{open-ended bioinformatics inquiry}, three studies evaluate autonomous discovery without a fixed scalar objective: hypothesis formulation and cross-cohort exploration in sepsis mortality (Section~\ref{sec:exp:sepsis}), single-omics analysis of paired esophageal squamous cell carcinoma (ESCC) lymph nodes (Section~\ref{sec:exp:escc}), and five-omics integrative analysis in acute aortic dissection (Section~\ref{sec:exp:anzhen}). All five studies run end to end under fixed model parameters through the same evidence-coupled within-task RSI loop, varying in task openness, evaluation regime, and data-access constraints. Each subsection presents both the verified algorithmic or biological findings and the revision trajectories that shaped them (task contracts and full audits appear in SI Sections~\ref{sec:si-case-denoising}--\ref{sec:si-case-anzhen}).

\begin{table}[!htbp]
\centering\small
\caption{\textbf{Evaluation matrix of the five long-horizon biological studies.} The five studies span quantitative algorithm development and open-ended bioinformatics inquiry from single-omics to multi-omics settings, isolating distinct layers of within-task recursive self-improvement.}
\label{tab:exp-overview}
\begin{tabularx}{\linewidth}{@{}>{\raggedright\arraybackslash}p{0.17\linewidth}>{\raggedright\arraybackslash}p{0.21\linewidth}>{\raggedright\arraybackslash}p{0.27\linewidth}>{\raggedright\arraybackslash}X@{}}
\toprule
\textbf{Study} & \textbf{Research task} & \textbf{Core methodological challenge} & \textbf{Within-task RSI capability demonstrated} \\
\midrule
\multicolumn{4}{@{}l}{\textit{Bioinformatics algorithm development (sealed quantitative evaluators)}} \\
\addlinespace[0.3em]
scRNA-seq denoising (Sec.~\ref{sec:exp:denoising}) & Single-omics algorithm development & Balancing dropout recovery against oversmoothing when fixed-family shrinkage saturates & \emph{Structural operator synthesis}: diagnosing pooling limits and pivoting from empirical Bayes shrinkage to support-restricted linear graph propagation. \\
\addlinespace[0.5em]
LUAD survival (Sec.~\ref{sec:exp:luad}) & Multi-omics algorithm development & Small-sample right-censored survival modeling under assay scale mismatch and leakage risk & \emph{Validity-over-metric governance}: rejecting a higher-scoring early-concatenation baseline to build a leak-free, rank-aligned late-fusion pipeline. \\
\midrule
\multicolumn{4}{@{}l}{\textit{Open-ended bioinformatics inquiry (no fixed scalar objective)}} \\
\addlinespace[0.3em]
Sepsis mortality (Sec.~\ref{sec:exp:sepsis}) & Open-ended inquiry \& cross-cohort discovery & Broad clinical direction with no preset hypothesis; reference fragility in single blood draws & \emph{Study-design revision}: autonomously sourcing public cohorts, pivoting from cross-sectional contrasts to within-patient trajectories, and auditing replicability. \\
\addlinespace[0.5em]
ESCC metastasis (Sec.~\ref{sec:exp:escc}) & Single-omics bioinformatics analysis & Confounding among cell-state composition, intrinsic expression, and filtering-driven patient loss & \emph{Deconfounding and identifiability}: decoupling intrinsic state shifts from composition, fixing patient denominators, and defining sequencing boundaries. \\
\addlinespace[0.5em]
Aortic dissection (Sec.~\ref{sec:exp:anzhen}) & Multi-omics integrative analysis & Small-$n$ five-omics alignment where group shifts diverge from individual-level concordance & \emph{Cross-layer calibration and retraction}: testing conditional and participant-level support and retracting an unsupported chromatin-contact claim. \\
\bottomrule
\end{tabularx}
\end{table}

\subsection{Single-Cell Expression Denoising via Graph Operator Evolution}\label{sec:exp:denoising}

Single-cell RNA sequencing captures only a small fraction of the transcripts present in each cell, leaving pervasive molecular dropout and sampling variance that obscure weak expression signals~\cite{Grun2014NoiseModels}. Recovering the underlying transcriptional state requires borrowing statistical strength across related cells, yet naive neighborhood smoothing blurs fine-grained cell-state boundaries and leaks background counts into unexpressed genes~\cite{Hou2020ImputationBenchmark}. Resolving this trade-off between dropout recovery and oversmoothing requires jointly designing representation learning, neighborhood graph construction, and count-space estimation rather than tuning isolated hyperparameters. We evaluated CytoCoScientist on the OpenProblems single-cell denoising benchmark, which scores reconstruction accuracy via molecular cross-validation on held-out transcript molecules~\cite{Batson2019MolecularCrossValidation,Luecken2025OpenProblems}. Starting from three tissue count matrices with no pre-supplied denoising code, this study tests whether CytoCoScientist can evolve a single-modality mathematical operator from scratch across tens of thousands of cells.

\begin{figure}[!htbp]
  \centering
  \includegraphics[width=\linewidth]{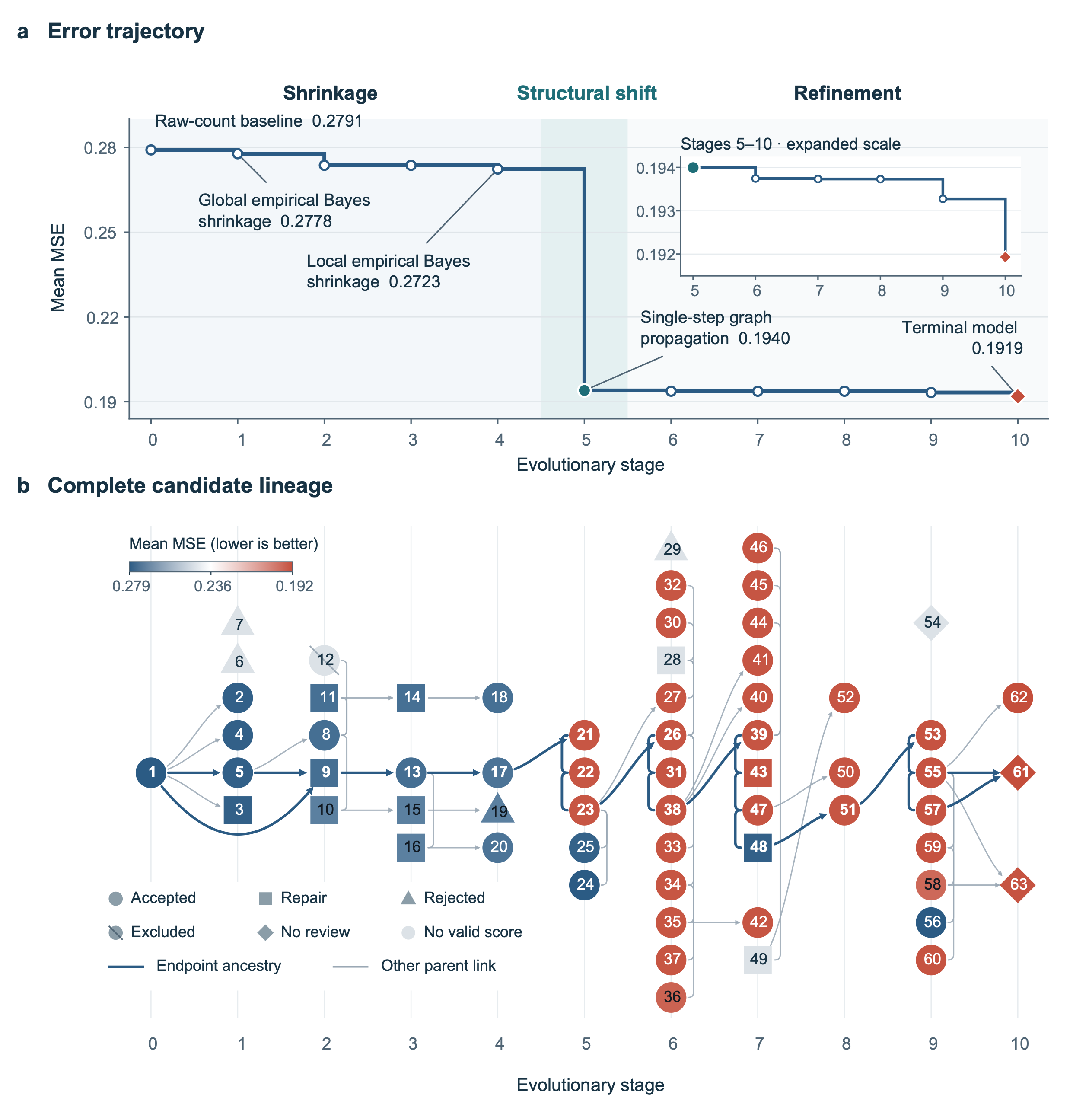}
  \caption{\textbf{Evolutionary search trajectory of CytoCoScientist on single-cell expression denoising.} \textbf{a,} Equally weighted mean MSE ($\downarrow$) across lung, pancreas, and PBMC over evolutionary stages~0--10 (inset zooms into stages~5--10). The trajectory progresses from the stage~0 raw-count baseline ($0.2791$) through global and local empirical Bayes shrinkage (stages~1--4) to a structural transition at stage~5 that introduces single-step graph propagation ($0.1940$). Subsequent mixture and prior refinements reach $0.1919$ at stage~10 (decision audit in SI Table~\ref{tab:si-denoising-decisions}). \textbf{b,} Candidate lineage graph of all 63 archived nodes indexed in chronological order (56 with valid numerical scores). Node color encodes mean MSE (gray indicates no numerical score). Node shape encodes review status: circles, accepted ($n=44$); squares, revision requested (\textit{Repair}, $n=11$), not necessarily completed; triangles, rejected ($n=4$); crossed circles, duplicate exclusion ($n=1$); and diamonds, no review entry ($n=3$). Blue edges trace the ancestry of the stage~10 champion model; gray edges denote other parent--child transitions.}
  \label{fig:exp-denoising-search}
\end{figure}

Under the task contract (SI Section~\ref{sec:si-denoising-task}), CytoCoScientist received three frozen molecular-split training matrices from OpenProblems Denoising v1.0.0 across lung, pancreas, and peripheral blood mononuclear cells (PBMC). Candidate programs read only these training matrices, applied a single cross-tissue prediction rule without dataset-specific tuning, and returned finite, non-negative denoised matrices preserving all cell and gene axes. A sealed evaluator log-normalized predicted and held-out molecular counts to library size $10^4$ and scored each candidate by the equally weighted three-tissue mean squared error ($\overline{\mathrm{MSE}}$; dataset dimensions and protocol in SI Section~\ref{sec:si-denoising-evaluation} and SI Equations~\ref{eq:si-denoising-normalization}--\ref{eq:si-denoising-mse}). Drawing on literature in count noise modeling, empirical Bayes shrinkage, and manifold graph diffusion, CytoCoScientist explored 63 candidate implementations across 10 evolutionary stages (Figure~\ref{fig:exp-denoising-search}; search settings and decision audit in SI Tables~\ref{tab:si-denoising-config} and~\ref{tab:si-denoising-decisions}). Rather than tuning parameters within a single preset template, the trajectory progressed through three distinct algorithmic regimes: global empirical Bayes shrinkage, local neighborhood priors, and direct support-restricted linear graph propagation with residual remixing.

At stage~1, CytoCoScientist formulated a global empirical Bayes (EB) shrinkage estimator that pulled positive-library cells toward a training-derived top-128 abundant-gene prior with pseudocount strength $\alpha=50$, lowering $\overline{\mathrm{MSE}}$ from $0.2791$ to $0.2778$ while replacing an early sparse-list concatenation prototype with preallocated compressed sparse row (CSR) buffers to guarantee scalable memory usage (SI Table~\ref{tab:si-denoising-decisions}). Across stages~2--4, subsequent branches built on this memory-safe foundation by pairing a 16-dimensional randomized singular value decomposition (SVD) embedding on $\log(1+10^4 R)$-transformed highly variable genes with a 16-neighbor mutual nearest-neighbor (MNN) graph to construct cell-specific local priors, further lowering $\overline{\mathrm{MSE}}$ to $0.2737$ ($\alpha=200$) and $0.2723$ ($\alpha=400$; SI Table~\ref{tab:si-denoising-decisions}). Yet increasing $\alpha$ further yielded diminishing returns. Inspecting stages~1--4 together exposed the underlying statistical bottleneck: because neighborhood expression entered solely as a pseudocount prior weighted by $\alpha/(L_i+\alpha)$, its influence decayed rapidly in cells with large library sizes $L_i$, preventing empirical Bayes shrinkage from pooling effectively across high-count cells (SI Section~\ref{sec:si-denoising-search}).

At stage~5, a decisive architectural pivot overcame this pooling bottleneck. Retaining the validated 16-dimensional SVD representation and reciprocal MNN topology from stage~4, CytoCoScientist replaced local empirical Bayes shrinkage with direct single-step linear graph propagation, reducing $\overline{\mathrm{MSE}}$ sharply from $0.2723$ to $0.1940$ (lung $0.1803$, pancreas $0.2167$, and PBMC $0.1850$; Figure~\ref{fig:exp-denoising-search}a and SI Table~\ref{tab:si-denoising-decisions}). Across stages~6--10, the search held this single-step propagation backbone fixed while refining the outer graph-to-observed mixture from $70{:}30$ to $72{:}28$ and introducing a weak training-derived global prior ($\epsilon=0.0111$) to stabilize ultra-sparse genes, reaching a terminal $\overline{\mathrm{MSE}}$ of $0.1919$ (SI Section~\ref{sec:si-denoising-terminal} and SI Table~\ref{tab:si-denoising-decisions}).

\begin{figure}[!htbp]
  \centering
  \includegraphics[width=\linewidth,height=0.68\textheight,keepaspectratio]{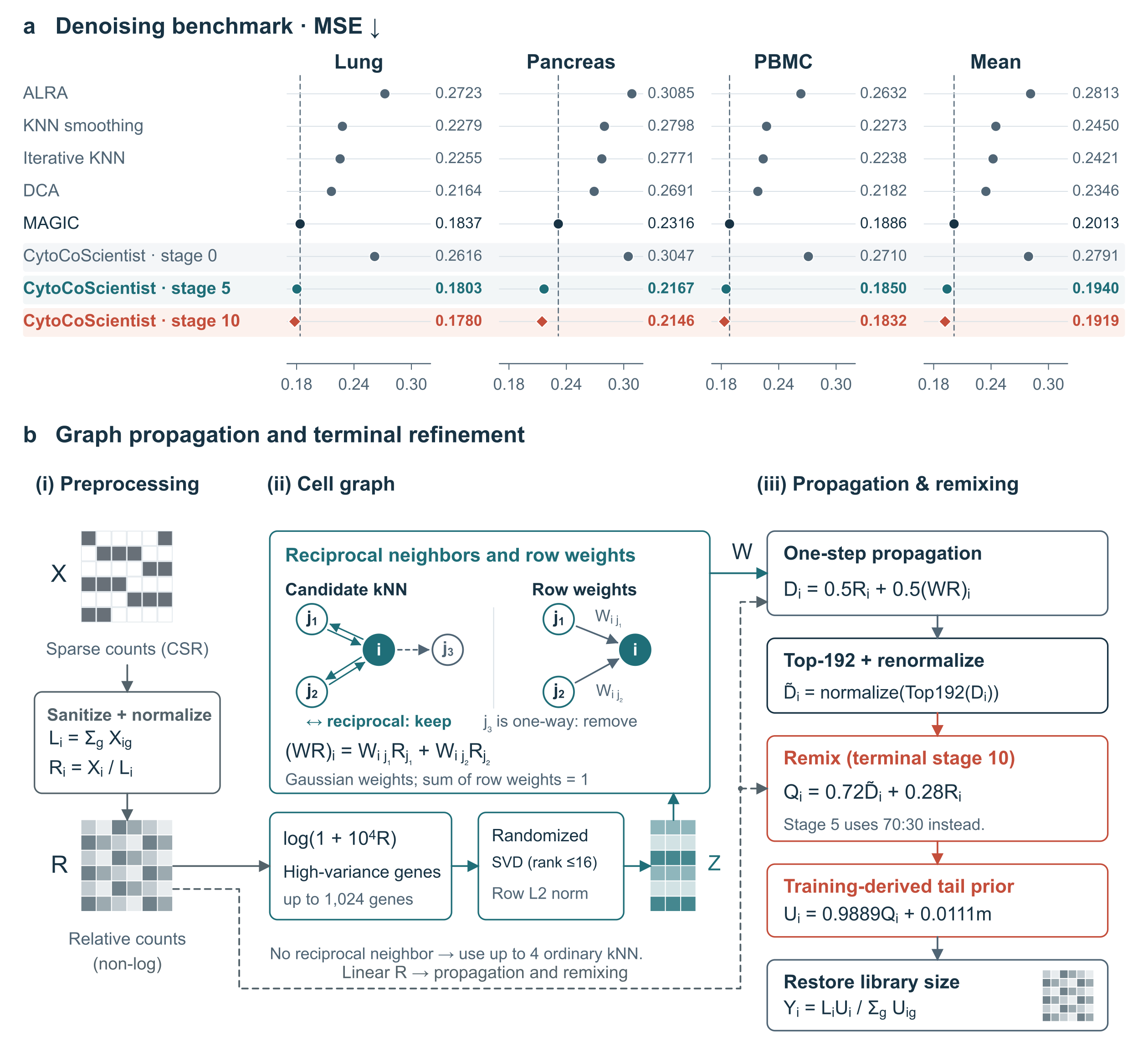}
  \caption{\textbf{Benchmark performance and architecture of the evolved single-cell graph-propagation denoiser.}
\textbf{a,} Per-tissue and equally weighted mean MSE ($\downarrow$) across lung, pancreas, and PBMC on the OpenProblems molecular cross-validation benchmark (SI Table~\ref{tab:si-denoising-results}). Published OpenProblems leaderboard baselines are compared against CytoCoScientist milestones measured by the local frozen evaluator: the stage~0 raw-count baseline ($0.2791$), stage~5 graph-propagation milestone ($0.1940$, circles), and stage~10 terminal model ($0.1919$, diamonds; provenance and confirmation scope in SI Section~\ref{sec:si-denoising-evaluation}). Vertical dashed lines mark the top external baseline (MAGIC, $0.2013$).
\textbf{b,} Architecture of the evolved estimator. Log-normalized relative counts identify high-variance genes and $L_2$-normalized randomized-SVD embeddings $Z$ to construct a reciprocal mutual nearest-neighbor (MNN) graph. Paired teal arrows retain reciprocal neighbors $j_1, j_2$ and prune one-way neighbor $j_3$ (fallback uses up to four ordinary kNN). Count propagation and remixing operate directly on unlogged relative counts $R$. Row-normalized Gaussian weights aggregate neighbor profiles (arrows labeled $W_{ij}$) into a single-step update, followed by Top-192 support restriction ($\mathcal{T}_{192}$), $72{:}28$ observed-profile residual remixing ($70{:}30$ at stage~5), blending with a training-derived global prior ($\epsilon=0.0111$), and exact library-size restoration (full equations, fallback rules, and memory guards in SI Section~\ref{sec:si-denoising-method}).}
  \label{fig:exp-denoising-results}
\end{figure}

Inspecting the confirmed stage~10 estimator independently of its search history highlights three complementary inductive biases that balance dropout recovery against multi-step diffusion oversmoothing (Figure~\ref{fig:exp-denoising-results}b and SI Section~\ref{sec:si-denoising-method}). First, the operator decouples neighborhood discovery from expression propagation: it selects up to $1{,}024$ high-variance genes in $\log(1+10^4 R)$ space to compute $L_2$-normalized 16-dimensional SVD embeddings and row-normalized Gaussian affinities $W$ over reciprocal mutual nearest neighbors (falling back to at most four ordinary neighbors when needed; SI Section~\ref{sec:si-denoising-graph} and SI Equation~\ref{eq:si-denoising-weights}), while applying a single linear propagation step $D_i = 0.50 R_i + 0.50 \sum_{j\in\mathcal{N}(i)} W_{ij} R_j$ directly on unlogged relative counts $R_i = X_i / L_i$ to avoid exponential back-transformation bias. Second, a hard support-restriction operator $\widetilde{D}_i = \mathcal{T}_{192}(D_i)$ retains and renormalizes only the 192 largest positive entries of $D_i$, concentrating neighborhood pooling on dominant transcriptional programs rather than leaking background counts across thousands of unexpressed genes. Third, residual remixing with the unpruned observed profile, $Q_i^{(10)} = 0.72\widetilde{D}_i + 0.28 R_i$ ($70{:}30$ at stage~5), preserves every observed nonzero transcript within cell~$i$, after which blending with the normalized abundant-gene training prior $m$ ($U_i^{(10)} = 0.9889 Q_i^{(10)} + 0.0111 m$) and restoring exact library sizes $L_i$ yields the final denoised matrix (SI Sections~\ref{sec:si-denoising-propagation} and~\ref{sec:si-denoising-terminal} and SI Equations~\ref{eq:si-denoising-linear-mixture}--\ref{eq:si-denoising-terminal}). The resulting two-module implementation (\texttt{main.py} and \texttt{method.py}) passed synthetic regression assertions, serialized H5AD schema and hash verification, and deterministic same-split re-execution (SI Section~\ref{sec:si-denoising-fallback} and SI Equation~\ref{eq:si-denoising-fallback}).

We compared the stage~0 ($0.2791$), stage~5 ($0.1940$), and stage~10 ($0.1919$) milestones against reference scores from the official OpenProblems leaderboard~\cite{vanDijk2018MAGIC,OpenProblemsDenoisingResults} on the identical v1.0.0 molecular-split datasets (SI Section~\ref{sec:si-denoising-evaluation} and SI Table~\ref{tab:si-denoising-results}). Both the stage~5 and stage~10 models surpassed the top published baseline MAGIC ($0.2013$), as well as DCA ($0.2346$), Iterative KNN smoothing ($0.2421$), KNN smoothing ($0.2450$), and ALRA ($0.2813$). Relative to raw counts and MAGIC, the stage~10 estimator reduced mean error by $31.2\%$ and $4.7\%$, respectively, while improving reconstruction across all three tissues: lung ($0.1780$ vs.\ $0.1837$), pancreas ($0.2146$ vs.\ $0.2316$), and PBMC ($0.1832$ vs.\ $0.1886$; Figure~\ref{fig:exp-denoising-results}a).

Taken together, this trajectory illustrates how CytoCoScientist evolves a single-modality count operator from first principles when parameter tuning saturates. After stages~1--4 exposed why pseudocount shrinkage stalls in high-library-size cells, pivoting at stage~5 to log-space MNN discovery paired with unlogged propagation, Top-192 support restriction, and observed-profile residual remixing resolved the tension between dropout recovery and oversmoothing.
\FloatBarrier

\subsection{Multi-Omics Survival Prediction in Lung Adenocarcinoma}\label{sec:exp:luad}

Lung adenocarcinoma (LUAD) exhibits pronounced transcriptional, post-transcriptional, and epigenetic variation across patients, motivating multi-omics integration for clinical survival risk stratification~\cite{TCGA2014LUAD,Poirion2021DeepProg}. Extracting prognostic signal jointly from RNA sequencing, miRNA expression, and DNA methylation is difficult because tens of thousands of molecular features outnumber the available patients, survival outcomes are heavily right-censored, and disparate measurement scales allow high-dimensional assays to dominate naive concatenation even though no single modality carries the full prognostic signal. Unlike the single-modality count operator in Section~\ref{sec:exp:denoising}, survival prediction couples imputation, feature screening, censored regression, and cross-modal risk fusion across interdependent stages. We evaluated CytoCoScientist on the TCGA-LUAD survival benchmark ($n=420$) using frozen data manifests and a sealed evaluation interface to test whether the system can evolve a leak-free multi-omics clinical prediction pipeline.

\begin{figure}[!htbp]
  \centering
  \includegraphics[width=\linewidth]{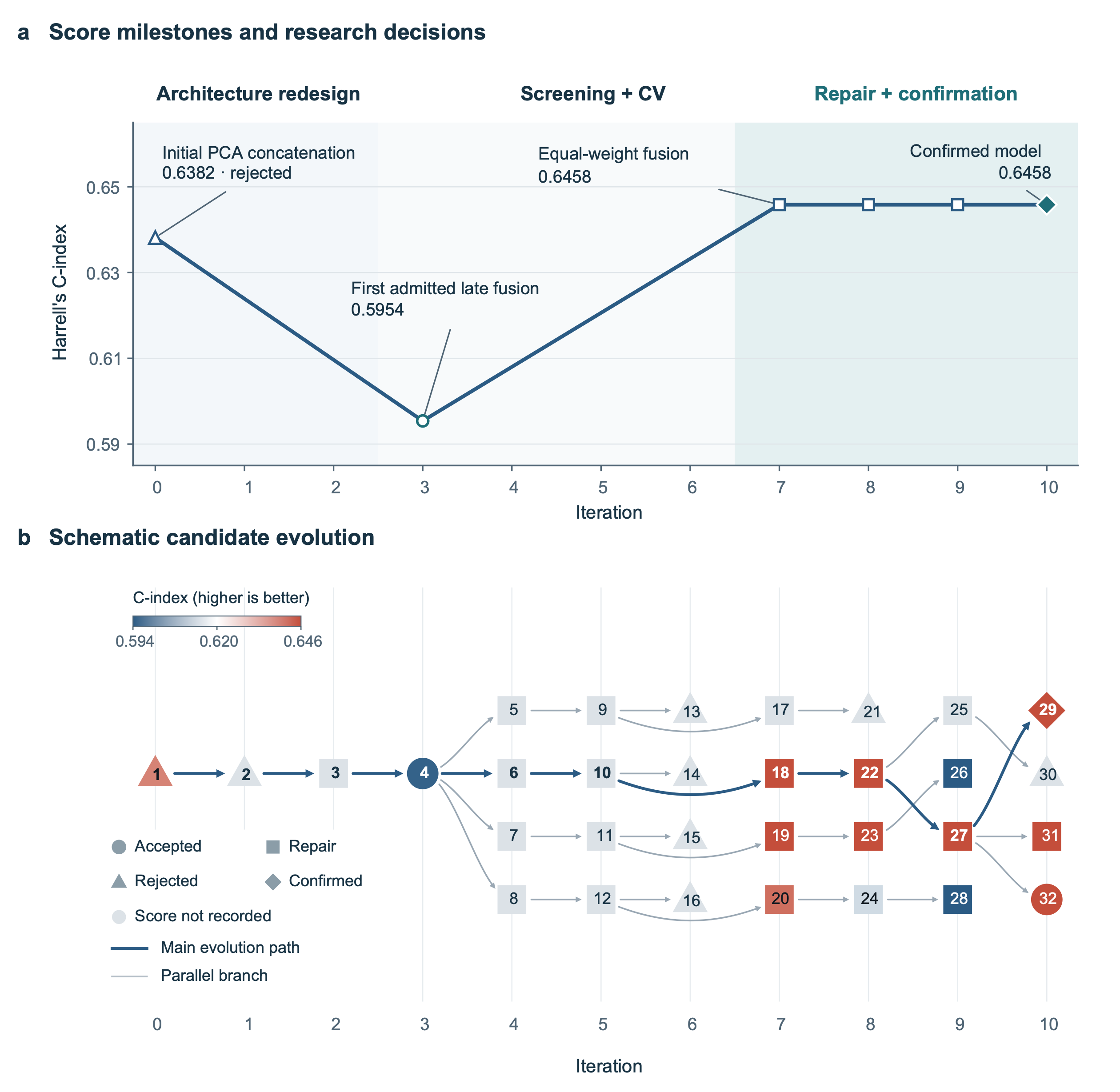}
  \caption{\textbf{Evolutionary search trajectory of CytoCoScientist on TCGA-LUAD multi-omics survival prediction.} \textbf{a,} Harrell's C-index ($\uparrow$) across key conceptual milestones over iterations~0--10 (decision audit in SI Table~\ref{tab:si-luad-decisions}). Review rejected the initial iteration~0 PCA-concatenation candidate ($0.6382$) and guided structural refactoring during iterations~1--2. The system then admitted a modular modality-specific late-fusion parent at iteration~3 ($0.5954$), reached $0.6458$ via equal-weight ECDF rank fusion at iteration~7, completed mandatory penalty-locking and numerical-invariance repairs across iterations~8--9, and passed deterministic confirmation at iteration~10 ($0.6458$). \textbf{b,} Evolution of all 32 candidate attempts indexed in iteration order across iterations~0--10. Node color encodes evaluation C-index (gray denotes no recorded score). Node shape encodes governance status: circles, Accepted ($n=3$); squares, Repair ($n=21$); triangles, Rejected ($n=8$); and diamond, confirmed terminal model. Blue arrows trace the primary conceptual progression, while gray arrows show parallel exploration and repair branches (including the iteration~7 sibling branch testing reliability-weighted fusion at $0.6411$).}
  \label{fig:exp-luad-search}
\end{figure}

Under the task contract (SI Section~\ref{sec:si-luad-task}), the 420 TCGA-LUAD patients were partitioned 80:20 into 336 training patients and 84 evaluation patients (31 observed mortality events, $1{,}368$ comparable patient pairs, and zero patient overlap). Candidate programs had to learn all missing-value imputation, feature screening, scaling, survival modeling, and cross-modal calibration parameters strictly within the 336-patient training partition before outputting continuous mortality risk scores for the 84 evaluation patients. A sealed evaluator scored predictions by Harrell's concordance index (C-index)~\cite{Harrell1982} under deterministic seed~11 (SI Section~\ref{sec:si-luad-protocol} and SI Equations~\ref{eq:si-luad-pairs}--\ref{eq:si-luad-cindex}). Starting with no prior survival code, CytoCoScientist surveyed regularized Cox proportional hazards modeling~\cite{Cox1972,Verweij1994}, tied-event partial likelihood formulations~\cite{Breslow1974}, and late-fusion survival frameworks~\cite{Poirion2021DeepProg}, then explored 32 candidate implementations across iterations~0--10 (Figure~\ref{fig:exp-luad-search}; search settings and decision audit in SI Section~\ref{sec:si-luad-search} and SI Tables~\ref{tab:si-luad-config} and~\ref{tab:si-luad-decisions}). Because small-sample survival metrics are easily inflated by scale distortion or subtle leakage, candidate evolution separated numerical score return from methodological admissibility, progressing across three stages: rejecting uncalibrated early concatenation, establishing a modular modality-specific late-fusion backbone, and hardening rank-based fusion through penalty locking and numerical-invariance checks.

At iteration~0, the initial candidate implemented a canonical early-concatenation baseline that extracted up to eight principal components per modality from $1{,}500$ RNA, $300$ miRNA, and $1{,}500$ methylation high-variance features and fed them into a single ridge-Cox model ($\lambda=0.5$). Although this baseline returned an evaluator C-index of $0.6382$, scientific review rejected it so that uncalibrated early concatenation would not anchor subsequent branches (Figure~\ref{fig:exp-luad-search}a,b and SI Tables~\ref{tab:si-luad-comparisons} and~\ref{tab:si-luad-decisions}). Guided by this critique across iterations~1--2, the system replaced early concatenation with a decoupled modality-specific architecture at iteration~3, fitting each assay with an independent ridge-penalized Cox model ($\lambda=0.25$, fixed feature cap $192$) followed by normalized-risk late fusion. Review admitted the iteration~3 candidate, which had a C-index of $0.5954$, as the parent for subsequent late-fusion variants (SI Table~\ref{tab:si-luad-decisions}).

Building on this modular parent, parallel candidate batches across iterations~4--10 tested survival-aware feature screening, training-side three-fold cross-validation over modality feature caps ($K_m\in\{96,192\}$) and ridge penalties ($\lambda\in\{0.1,0.5,2.0\}$), and competing late-fusion rules (Figure~\ref{fig:exp-luad-search}b and SI Table~\ref{tab:si-luad-decisions}). At iteration~7, candidates using parameter-free equal-weight rank fusion reached a C-index of $0.6458$, outperforming a concurrent sibling branch that blended equal weights with out-of-fold modality reliability weights estimated on the 336 training patients ($0.6411$; SI Tables~\ref{tab:si-luad-comparisons} and~\ref{tab:si-luad-decisions}), so subsequent branches retained equal-weight fusion. Even after reaching $0.6458$, the lineage remained under revision across iterations~7--9 until two methodological requirements were satisfied: locking the shared ridge penalty at $\lambda=2.0$ after confirming that training-only selectors consistently chose $2.0$ across all three assays, and adding unit self-test assertions verifying exact coefficient and prediction invariance under permutations of tied survival times (alongside output-path safety checks; SI Section~\ref{sec:si-luad-search}). Only after passing all invariance checks was the candidate admitted at iteration~10, where deterministic re-execution reproduced an exact C-index of $0.6458$ (score difference $0.0000$).

\begin{figure}[!htbp]
  \centering
  \includegraphics[width=\linewidth]{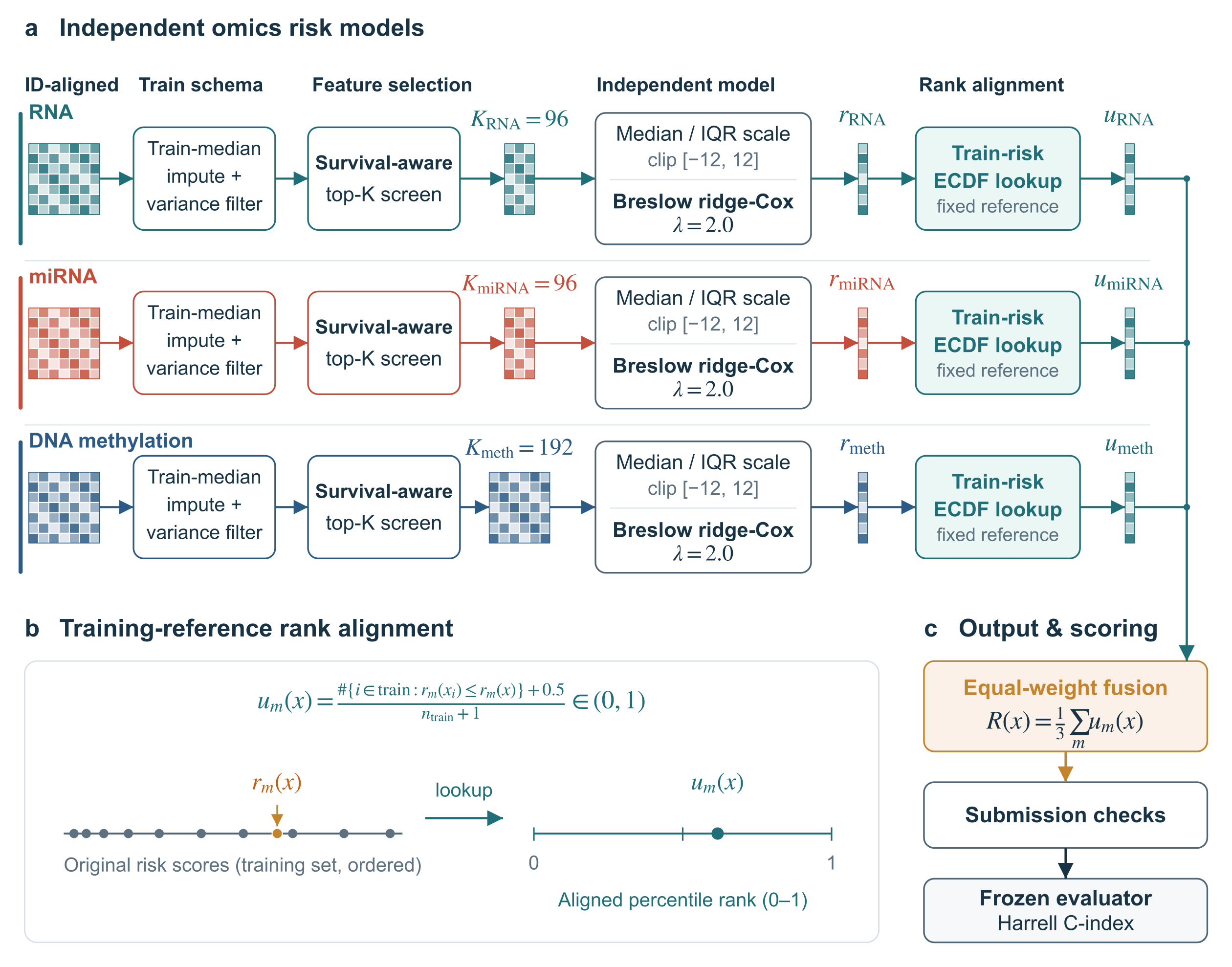}
  \caption{\textbf{Architecture of the confirmed TCGA-LUAD multi-omics survival model.}
  \textbf{a,} Modality-specific representation and ridge-Cox fitting. Patient-aligned RNA, miRNA, and DNA methylation matrices undergo independent training-median imputation, variance filtering, and survival-aware correlation screening against $z_i=e_i-0.35\,\mathrm{zscore}(\log(1+t_i))$ on the 336-patient training partition. Event-stratified three-fold cross-validation selects feature caps ($K_{\mathrm{RNA}}=96$, $K_{\mathrm{miRNA}}=96$, $K_{\mathrm{meth}}=192$), followed by training-median/IQR robust scaling, clipping to $[-12,12]$, and independent Breslow ridge-Cox fitting ($\lambda=2.0$).
  \textbf{b,} Cross-modal ECDF rank alignment. To prevent high-variance assays from dominating fusion, each modality's linear predictor $r_m(x)$ is mapped onto its continuity-corrected training-reference percentile $u_m(x)\in(0,1)$ relative to the $n_{\mathrm{train}}=336$ training scores.
  \textbf{c,} Equal-weight late fusion and evaluation. The three percentile ranks are averaged equally into the composite risk score $R(x)$ (full mathematical specifications in SI Section~\ref{sec:si-luad-model}). The output is audited for schema and identifier integrity and evaluated by the frozen Harrell's C-index evaluator on the 84-patient evaluation split (31 events, $1{,}368$ comparable pairs; self-test verification and confirmation scope in SI Section~\ref{sec:si-luad-search}).}
  \label{fig:exp-luad-results}
\end{figure}

The confirmed iteration~10 program implements this pipeline through modality-specific screening, independent ridge-Cox modeling, and cross-modal percentile alignment (Figure~\ref{fig:exp-luad-results} and SI Section~\ref{sec:si-luad-model}). Within each patient-aligned assay, the pipeline discards zero-variance features, imputes missing entries using training-set medians (SI Section~\ref{sec:si-luad-alignment}), and ranks features across the 336 training patients by absolute Pearson correlation with a composite survival-aware target, $z_i = e_i - 0.35\,\mathrm{zscore}(\log(1+t_i))$, which assigns higher positive scores to early uncensored deaths and lower scores to long-term survivors. Event-stratified three-fold cross-validation on the training partition then selects modality-specific feature caps from $\{96, 192\}$, allocating $K_{\mathrm{RNA}}=96$, $K_{\mathrm{miRNA}}=96$, and $K_{\mathrm{meth}}=192$ to grant twice the feature capacity to the broader DNA methylation array while keeping transcriptomic representations compact (SI Section~\ref{sec:si-luad-screening} and SI Equations~\ref{eq:si-luad-screening}--\ref{eq:si-luad-cap}). Each selected matrix is centered by its training median, scaled by its training interquartile range (with a standard-deviation fallback), clipped to $[-12,12]$, and fit with an independent $L_2$-penalized Cox proportional hazards model ($\lambda=2.0$)~\cite{Cox1972,Verweij1994} via L-BFGS-B under Breslow's grouped tied-event partial likelihood~\cite{Breslow1974} (SI Section~\ref{sec:si-luad-cox} and SI Equations~\ref{eq:si-luad-scaling}--\ref{eq:si-luad-cox}). Because linear predictors $r_m(x)=x_m^\top\beta_m$ from separate high-dimensional assays exhibit incompatible variances and tail behavior, the pipeline maps each modality's risk onto its continuity-corrected empirical cumulative distribution function (ECDF) relative to the 336 training scores, $u_m(x) = (\text{rank}_{\mathrm{train}}(r_m(x)) + 0.5)/(n_{\mathrm{train}}+1)\in(0,1)$, before parameter-free equal-weight averaging into $R(x) = (u_{\mathrm{RNA}}(x)+u_{\mathrm{miRNA}}(x)+u_{\mathrm{meth}}(x))/3$ (SI Section~\ref{sec:si-luad-fusion} and SI Equations~\ref{eq:si-luad-fusion}--\ref{eq:si-luad-subtype}; output serialization guards for \path{submission.csv} and nine built-in regression self-test suites).

On the identical 336/84 split and Harrell pair-comparison rule, a matched local reproduction of DeepProg~\cite{Poirion2021DeepProg} integrating RNA, miRNA, and DNA methylation achieved a C-index of $0.5950$ (SI Section~\ref{sec:si-luad-protocol} and SI Table~\ref{tab:si-luad-comparisons}). The confirmed CytoCoScientist model reached a C-index of $0.6458$ (three-group log-rank $\chi^2=5.7267$, $p=0.0571$), outperforming matched DeepProg by $+0.0508$ and the rejected iteration~0 early-concatenation baseline ($0.6382$) by $+0.0077$ (computed from unrounded scores; Figure~\ref{fig:exp-luad-search}a and SI Table~\ref{tab:si-luad-comparisons}).

In small-sample survival modeling, pure metric maximization easily rewards fragile shortcuts. Rejecting the uncalibrated early-concatenation baseline ($0.6382$) forced a temporary dip to a modular late-fusion architecture ($0.5954$) at iteration~3, from which combining modality-specific ridge-Cox models with parameter-free ECDF percentile alignment reached $0.6458$ through structural regularization rather than added parameters while passing tied-time invariance checks at iteration~10. Paired with single-cell denoising, these two benchmarks show that CytoCoScientist can evolve both count-space mathematical operators and multi-stage clinical prediction pipelines; we next examine open-ended multi-omics discovery where no fixed scalar objective is available.

\FloatBarrier

\subsection{Mortality-Associated Molecular Shifts in Sepsis}\label{sec:exp:sepsis}

This case evaluates whether CytoCoScientist can autonomously identify tractable scientific questions when given a broad research direction rather than a preset hypothesis. In exploratory bioinformatics, the starting point is rarely a single ready-made claim awaiting verification, but rather scattered observations across published studies and public patient datasets. Researchers must first isolate unexplained patterns, formulate an initial judgment, and determine what distinct observation would follow if a competing explanation held instead. That step governs which datasets to retrieve next, which comparisons to construct, and when an initial interpretation must be revised. The initial research objective was to identify measurable molecular changes associated with differences in sepsis mortality risk (SI Section~\ref{sec:si-sepsis-s1}). Prior studies have linked plasma protein abundance to sepsis mortality~\cite{Palmowski2025SepsisProteome}, and serial blood sampling has shown that circulating amino acid concentrations fluctuate over the course of illness~\cite{Su2015SepsisAminoAcids}. Even when a molecule is elevated in non-survivors, however, that shift may simply accompany greater illness severity or impaired organ clearance. Which leads merit deeper inquiry, and which observations can separate competing explanations, were left for CytoCoScientist to formulate and test autonomously during exploration.

Starting from the initial research objective and the supplied Chinese Multi-omics Advances in Sepsis (CMAISE; PRJCA006118) cohort, CytoCoScientist independently identified additional studies and public data sources and assessed whether they supported the proposed analyses. Two rounds of search generated 30 and 24 candidate hypotheses, respectively. Within each round, a large language model (LLM) conducted pairwise reviews, and CytoCoScientist selected the top four candidates by Elo rating for patient-level bioinformatics analysis. Elo ranking prioritized which hypotheses to test (Figure~\ref{fig:sepsis-dag}; SI Tables~\ref{tab:si-sepsis-candidates} and~\ref{tab:si-sepsis-analyses}).

\begin{figure}[!htbp]
\centering
\includegraphics[width=\linewidth]{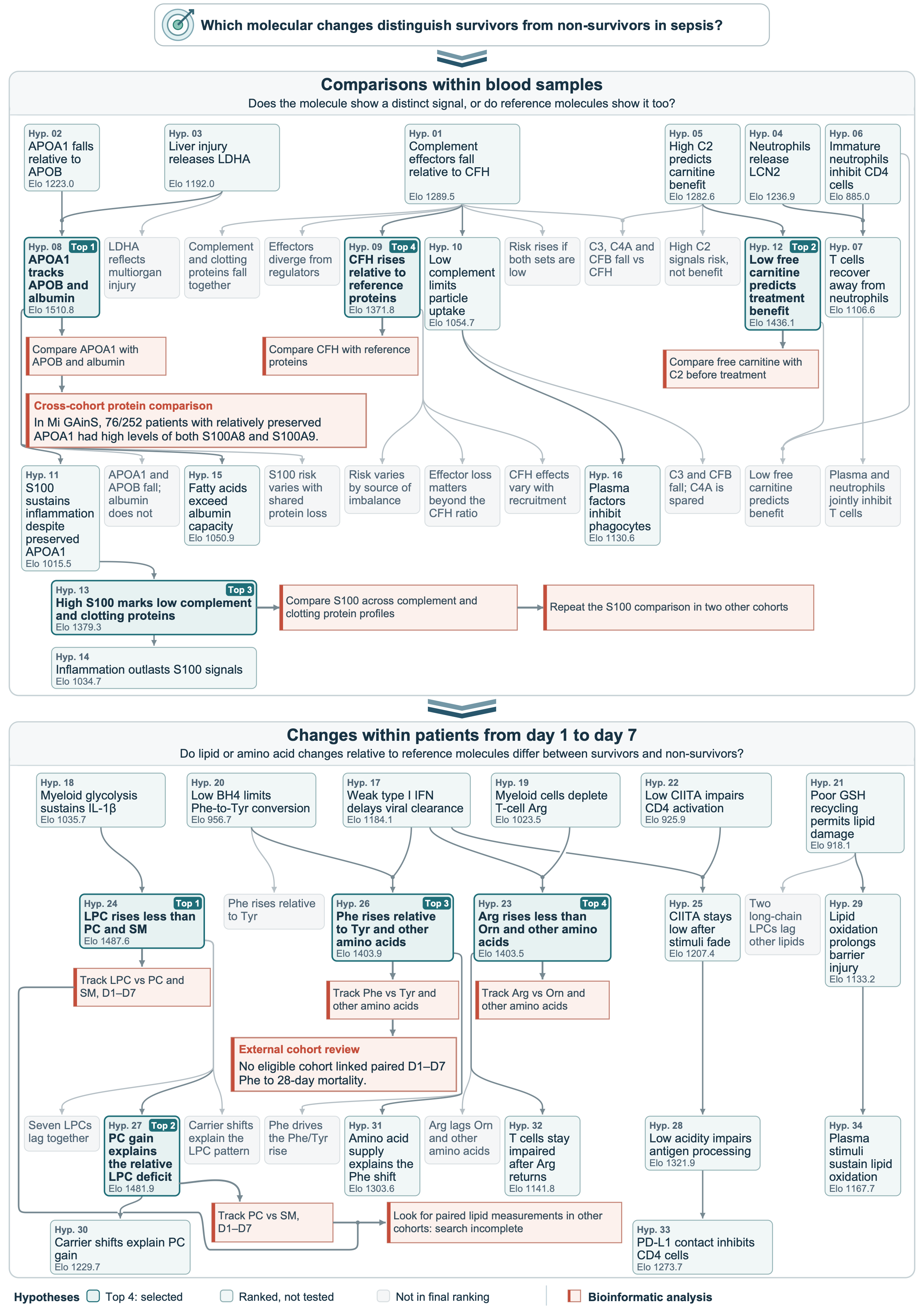}
\caption{\textbf{From relative levels in a single blood sample to relative changes within the same patient.}}
\label{fig:sepsis-dag}
\end{figure}

\begin{figure}[!htbp]
\ContinuedFloat
\captionsetup{labelformat=continuedfigure}
\caption[]{Upper and lower panels show Rounds 1 and 2, respectively. Nodes within each horizontal tier belong to the same generation, and edges trace candidate derivation. In each round, only the four highest-ranked hypotheses were selected for numerical analysis. The legend distinguishes these from the other ranked hypotheses and from candidates outside the final Elo ranking. Identifiers follow order of appearance, with Top and Elo indicating within-round rank and rating. Separate analysis boxes summarize initial analyses, cross-cohort follow-ups, and external data-eligibility audits. The central arrow indicates that Round 1 findings and limitations entered Round 2 via outer-loop meta-review as shared context, without direct node-to-node derivation.}
\end{figure}

\begin{figure}[!htbp]
\centering
\includegraphics[width=\linewidth]{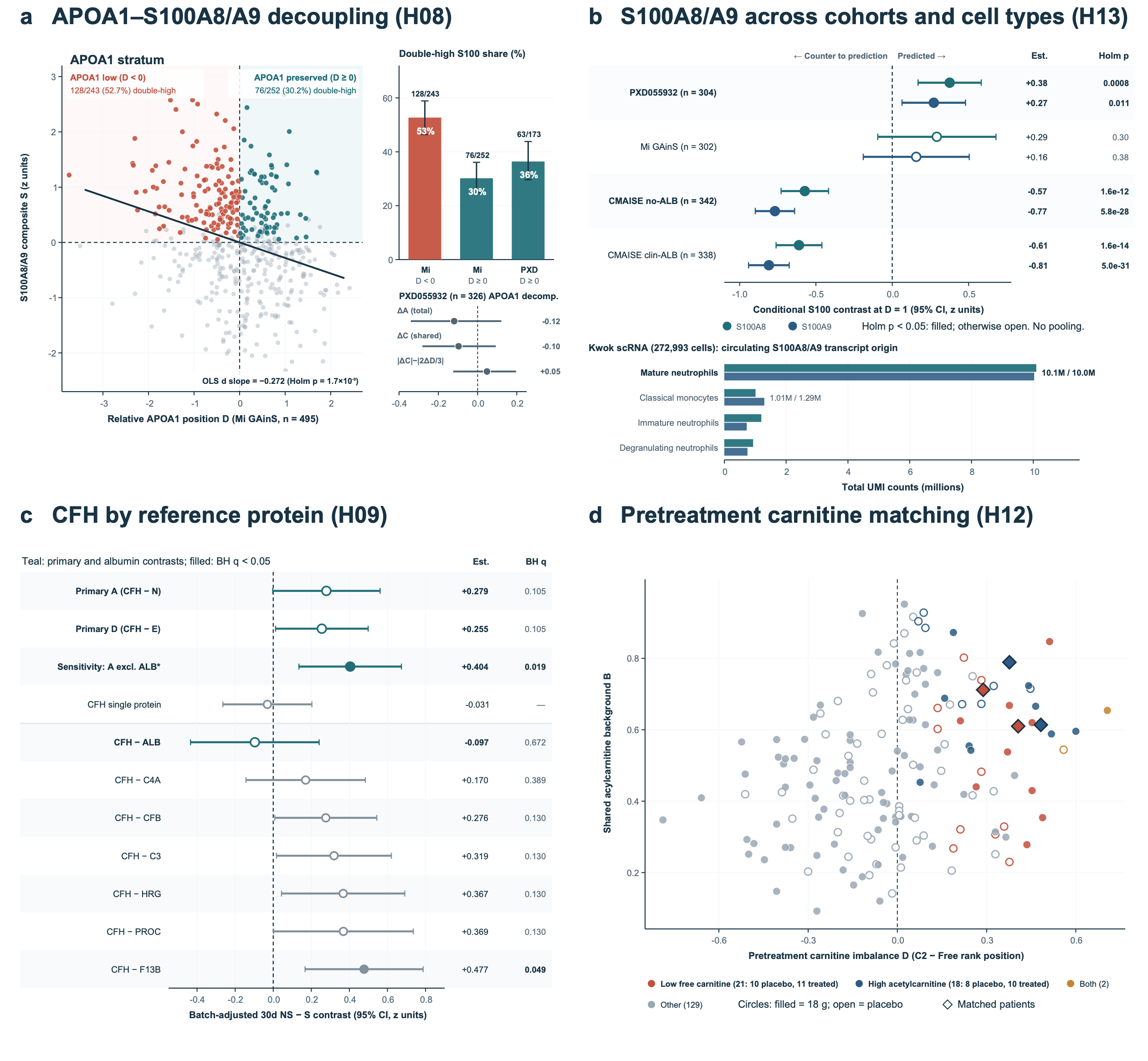}
\caption{\textbf{Round 1 tests of four single-blood-draw hypotheses across plasma proteomics and pretreatment carnitines.} \textbf{a,} Relative APOA1 position versus S100A8/A9 composite level in Mi GAinS ($n = 495$) alongside proportions of double-high S100 patients in Mi GAinS and PXD055932 ($n = 326$). \textbf{b,} Conditional S100A8 and S100A9 contrasts between two complement--hemostasis backgrounds evaluated separately across three proteomics cohorts; lower panel shows captured \textit{S100A8} and \textit{S100A9} transcript totals across peripheral blood cell types in the whole-blood single-cell atlas of Kwok et al.~\cite{Kwok2023SepsisSingleCell}. \textbf{c,} Batch-adjusted 30-day mortality contrasts for CFH relative to albumin versus other circulating reference proteins in PXD055932 ($n = 310$). \textbf{d,} Pretreatment carnitine patterns across randomized arms in RACE/CaPS~\cite{Puskarich2021RACECaPS} ($n = 170$); only two pairs met matching criteria, with zero placebo controls in the matched low-free-carnitine stratum.}
\label{fig:sepsis-round1}
\end{figure}

The four hypotheses selected in Round 1 evaluated molecular relationships within a single blood draw: three compared plasma proteins, and one examined pretreatment acylcarnitine structure. The top-ranked hypothesis (H08) proposed that low APOA1 and high S100A8/A9 might reflect a broader change in circulating proteins rather than an inflammatory response driven by selective APOA1 depletion. To distinguish these explanations, CytoCoScientist asked whether both S100 proteins could still be high in patients whose APOA1 was preserved relative to apolipoprotein B and albumin. In the public plasma proteomics cohort of Palmowski et al.\ (PXD055932)~\cite{Palmowski2025SepsisProteome}, CytoCoScientist found that 173 patients fell into this relatively preserved APOA1 stratum, yet 63 still exhibited double-high S100. Re-evaluating the same boundary in the public GAinS cohort of Mi et al.~\cite{Mi2024GAinSProteome} yielded the same pattern: double-high S100 occurred more frequently when APOA1 was relatively low ($128/243$, $53\%$), yet persisted in $76/252$ ($30\%$) patients with relatively preserved APOA1 (Figure~\ref{fig:sepsis-round1}a). Across both cohorts, elevated S100A8 and S100A9 persisted despite relatively preserved APOA1, showing that relatively low APOA1 was not necessary for this pattern, although neither cohort establishes a shared acute-phase response as its main cause, and the retrieved Mi GAinS files lacked per-patient mortality linkage.

Cross-cohort and reference-set checks exposed further fragility in single-draw protein contrasts. In H13, CytoCoScientist contrasted two proteomic backgrounds: coordinated depletion of both complement and hemostasis proteins versus depletion of complement effectors relative to complement factor H (CFH). Evaluated in the same direction across cohorts, conditional S100A8 and S100A9 differences were positive in PXD055932 yet reversed to negative in CMAISE (Figure~\ref{fig:sepsis-round1}b). In H09, the direction of the 30-day mortality contrast for CFH flipped depending on whether albumin or hemostasis proteins served as the reference (Figure~\ref{fig:sepsis-round1}c). For the carnitine hypothesis (H12) in the Rapid Administration of Carnitine in Sepsis / L-Carnitine Pharmacometabolomics in Sepsis (RACE/CaPS) trial~\cite{Puskarich2021RACECaPS}, CytoCoScientist first audited whether pretreatment low free carnitine and elevated acetylcarnitine could be separated and compared across randomized arms at matched molecular backgrounds. Only two pairs met the prespecified matching criteria, leaving zero placebo controls on the low-free-carnitine side (Figure~\ref{fig:sepsis-round1}d).

CytoCoScientist found that S100A8 and S100A9 could both be high even when APOA1 was relatively preserved. A single blood draw could reveal this pattern, but not whether a target molecule and its reference molecules changed together over time. Following outer-loop meta-review (Section~\ref{sec:method-meta-review}) of the first-round findings and their limitations, CytoCoScientist generated 24 further hypotheses under the same mortality-focused goal. All four selected by Elo ranking compared changes from day 1 to day 7 within individual patients. Each selected candidate asked whether target lipids or amino acids diverged from a patient's own reference molecules over the first week and whether those relative shifts differed between subsequent non-survivors and survivors. CytoCoScientist retrieved Supplementary Tables S1 and S2 of the Ferrario et al.\ ALBIOS biobank study~\cite{Ferrario2016ALBIOSMetabolomics}, reconstructing paired day-1 and day-7 plasma measurements for 20 patients with severe septic shock. All four analyses used paired samples from the same 20 patients. The three lipid and phenylalanine analyses used 28-day mortality as their primary outcome; the arginine analysis used 90-day mortality.

\begin{figure}[!htbp]
\centering
\includegraphics[width=\linewidth]{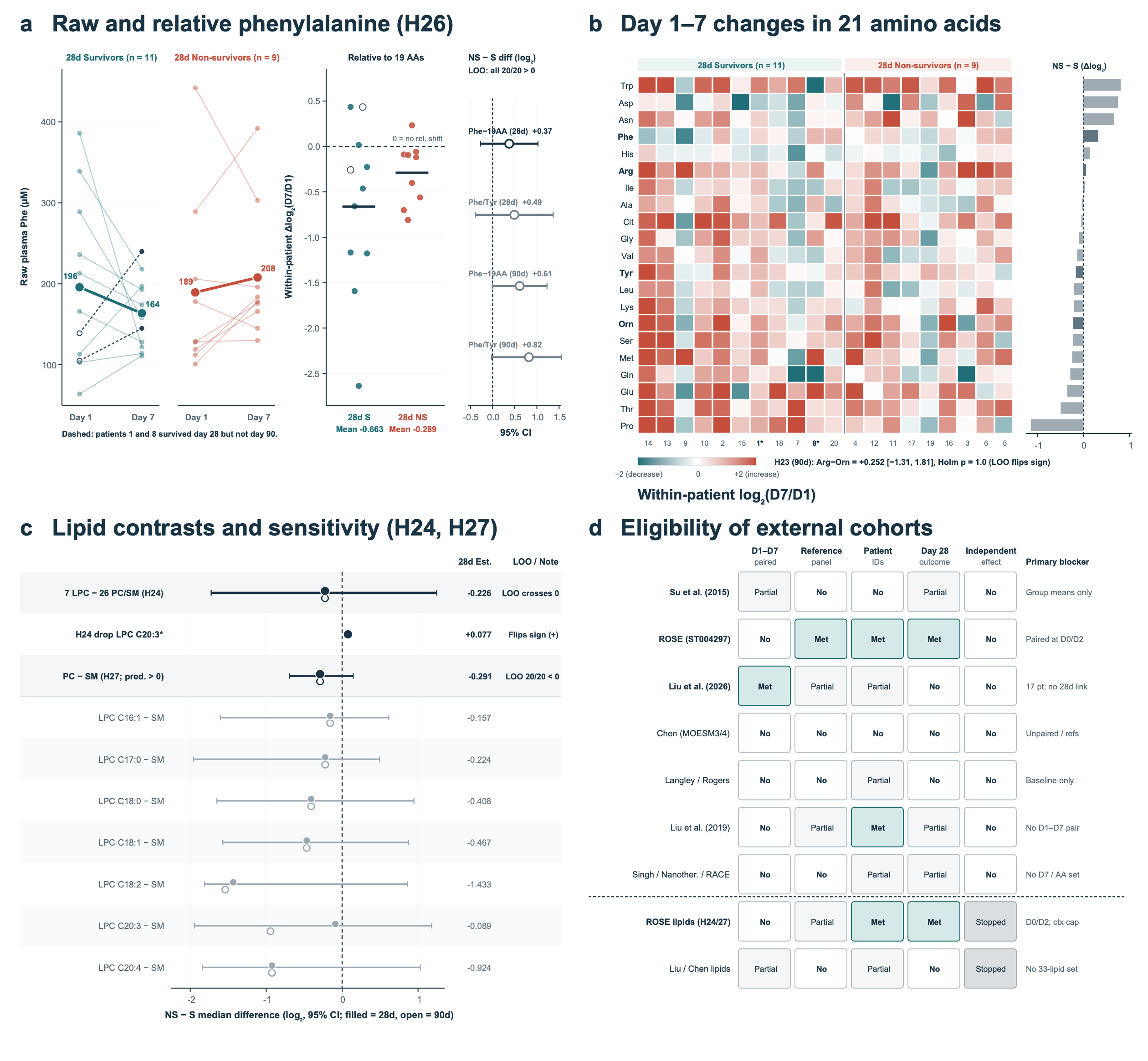}
\caption{\textbf{Within-patient first-week metabolite trajectories relative to reference panels and external cohort eligibility.} \textbf{a,} Raw paired day-1 to day-7 plasma phenylalanine concentrations, within-patient changes relative to the fixed 19-amino-acid reference (excluding phenylalanine and tyrosine), and outcome contrasts (NS minus S: non-survivors minus survivors) for 20 Ferrario/ALBIOS patients. Relative phenylalanine fell on average in both groups, less in non-survivors; the 95\% interval for the between-group difference included zero. \textbf{b,} For the same patients, a 21-amino-acid $\Delta\log_2(\text{D7}/\text{D1})$ heatmap compares phenylalanine, arginine, and their reference panels. \textbf{c,} Two lipid hypotheses (H24 and H27) test whether the observed contrasts depend on the reference lipids, individual species, or individual patients. \textbf{d,} File-level eligibility audit of external public cohorts across sampling windows, reference panels, patient keys, and 28-day outcomes, showing that none supplied all fields needed to repeat the paired comparison.}
\label{fig:sepsis-round2}
\end{figure}

The third-ranked Round 2 candidate (H26) asked whether phenylalanine rose more than a patient's reference amino acids in non-survivors. Stratified by 28-day outcome, mean raw plasma phenylalanine rose from $189$ to $208~\mu\text{M}$ in the 9 non-survivors while falling from $196$ to $164~\mu\text{M}$ in the 11 survivors. Across the 21 measured amino acids, the fixed 19-amino-acid reference excluded both phenylalanine and tyrosine. For each patient, CytoCoScientist computed the $\log_2(\text{D7}/\text{D1})$ change of phenylalanine minus the median $\log_2(\text{D7}/\text{D1})$ change across those 19 reference amino acids. Group means of this within-patient relative metric were negative in both outcome groups ($-0.289$ in non-survivors and $-0.663$ in survivors; Figure~\ref{fig:sepsis-round2}a). Thus, whereas raw concentrations rose on average in non-survivors and fell in survivors, referencing each patient against the fixed 19-amino-acid panel reframes the group contrast as a smaller average relative decline in non-survivors: the non-survivor minus survivor difference was $+0.374$ $\log_2$ units (95\% CI $-0.276$ to $1.025$), leaving any mortality association unresolved. Evaluated separately alongside this reference panel, the phenylalanine-to-tyrosine ratio showed a similarly imprecise contrast ($+0.489$, 95\% CI $-0.390$ to $1.368$).

The same within-patient reference framework did not yield consistent support for higher-ranked candidates. The top-ranked lipid candidate (H24) predicted impaired relative recovery of lysophosphatidylcholines (LPCs) in non-survivors. Although the primary group difference ($-0.226$) matched the predicted sign, dropping a single species (LPC C20:3) flipped the estimate to $+0.077$. The second-ranked candidate (H27) probed the reference directly: if apparent LPC deficits were driven by phosphatidylcholine (PC) accumulation, PC should increase more than sphingomyelin (SM) in non-survivors. Instead, the observed $\text{PC} - \text{SM}$ contrast was negative ($-0.291$) and stayed negative under all 20 leave-one-patient-out deletions, though its confidence interval still crossed zero. The arginine comparison (H23) proved similarly imprecise (Figure~\ref{fig:sepsis-round2}b,c; SI Table~\ref{tab:si-sepsis-r2}).

To assess whether the phenylalanine pattern could be replicated independently, CytoCoScientist audited nine external public sources. Supplementary Table S4 of Su et al.~\cite{Su2015SepsisAminoAcids} reported day-1 and day-7 outcome-stratified group means but lacked patient-level paired values required to compute within-patient relative changes, whereas the ROSE longitudinal acute respiratory distress syndrome (ARDS) metabolomics cohort (Metabolomics Workbench ST004297) sampled paired plasma only at days 0 and 2. None of the nine inspected public sources supplied all fields required to repeat the prespecified comparison (Figure~\ref{fig:sepsis-round2}d; SI Table~\ref{tab:si-sepsis-external}).

Together, these two rounds show how within-task recursive self-improvement (RSI) operates in open-ended clinical-omics analysis. In Round 1, testing whether double-high S100 persisted when APOA1 was relatively preserved showed across two cohorts that low APOA1 is not necessary for this pattern, while exposing the fragility of single-blood-draw contrasts. Following outer-loop meta-review (Section~\ref{sec:method-meta-review}), all four Elo-selected Round 2 analyses shifted to within-patient changes from day 1 to day 7. Mean phenylalanine rose from day 1 to day 7 in non-survivors. Yet relative to each patient's 19 reference amino acids, phenylalanine fell on average in both outcome groups, with a smaller relative fall in non-survivors. Under the same mortality-focused goal, CytoCoScientist revised not only which molecules to examine but what to compare: relative levels in one blood sample, then relative changes within the same patient over time. Because the 20-patient cohort leaves the mortality association unresolved, the next step requires independent paired day-1 and day-7 amino acid measurements linked to patient outcomes (SI Section~\ref{sec:si-sepsis-s3-5}).
\FloatBarrier

\subsection{Single-Cell B-Lineage Analysis in Esophageal Squamous Cell Carcinoma Metastasis}\label{sec:exp:escc}

A common task in bioinformatics is to use one type of molecular measurement across patients and tissues to identify disease-associated changes that recur across individuals. In single-cell transcriptomics, thousands of cells can reveal which cell states contribute to a tissue-level signal, but testing whether that signal recurs requires comparable tissue samples from different patients. Studies of esophageal squamous cell carcinoma (ESCC) have described tertiary lymphoid structures in primary tumors~\cite{Ling2022ESCCTLS}, profiled B-cell states in primary and metastatic lesions~\cite{Jia2023ESCCSingleCell}, and mapped the microenvironments of metastatic and tumor-free lymph nodes~\cite{Guo2025ESCCLymphNode}. These observations leave open how to interpret an expression difference across the B lineage: it could reflect a shift in the proportions of cell states, a change in expression within a state, or both. For example, some human germinal-center (GC) B-cell subsets normally express high levels of cell-cycle transcripts~\cite{Holmes2020GerminalCenterBCells}. We therefore tested whether CytoCoScientist could select comparable samples from a multi-tissue cohort, identify a signal without a prespecified gene, and design a patient-level comparison to distinguish these explanations.

CytoCoScientist began from the gene count matrix, existing cell annotations, and sample metadata of 110 tissue sequencing libraries across 40 patients. Thirteen patients contributed both a metastatic lymph node (MET) and a tumor-free lymph node (LYMPH), and the 12 paired patients profiled on the C4 platform formed the primary cohort, totaling $102{,}676$ B and plasma cells. The search trajectory and finalized workflow are documented in SI Figures~\ref{fig:si-escc-dag} and~\ref{fig:si-escc-workflow} and SI Table~\ref{tab:si-escc-decisions}. All tests treated patients, rather than individual cells, as independent replicates: paired tissues were contrasted within each patient before assessing directional consistency across patients (cohort and cell coverage in SI Figure~\ref{fig:si-escc-coverage}).

\begin{figure}[!htbp]
\centering
\includegraphics[width=\linewidth]{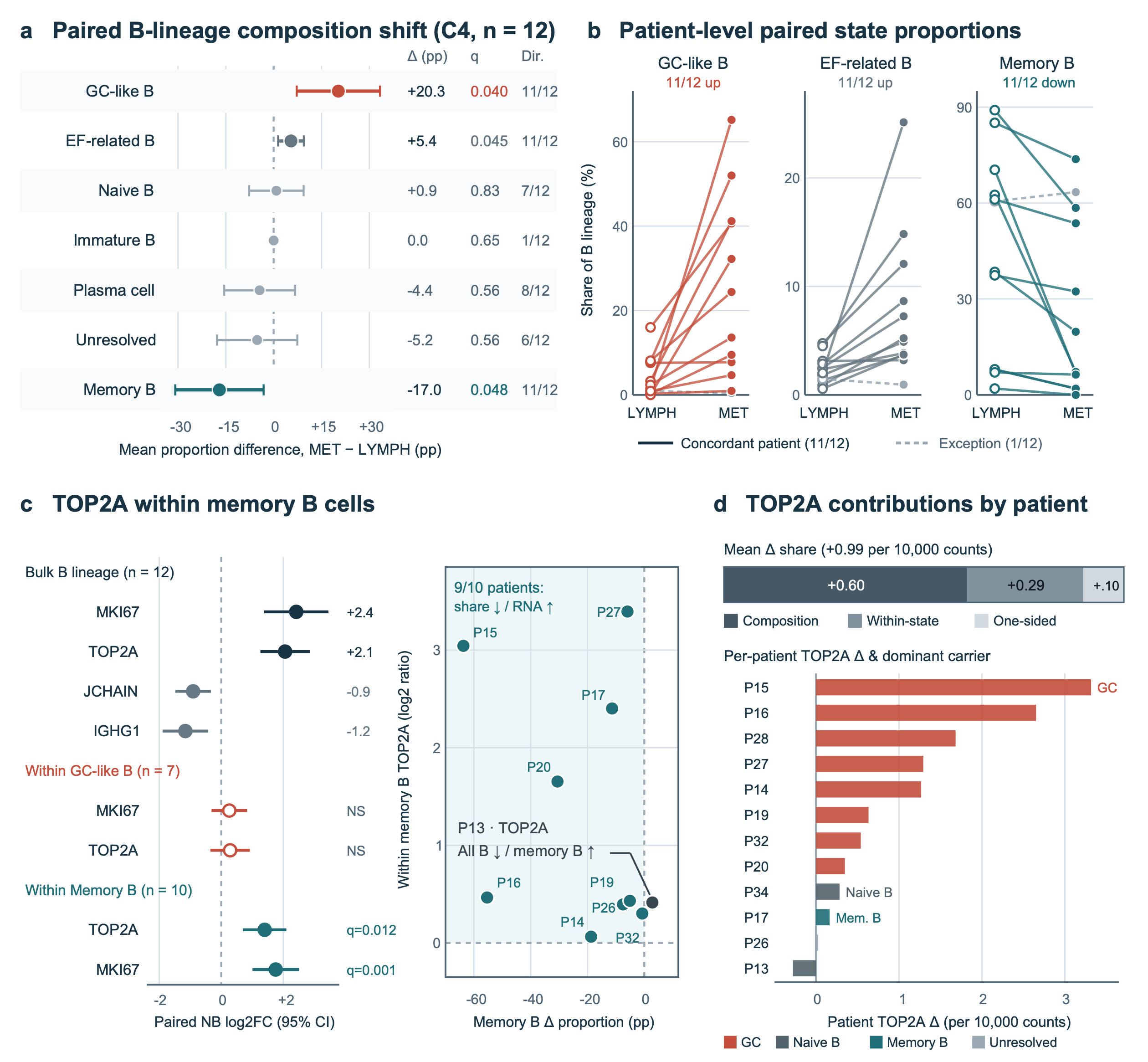}
\caption{\textbf{Memory B cells decline in proportion in metastatic lymph nodes yet express higher \textit{TOP2A}.} \textbf{a,} Mean paired proportion differences (MET minus LYMPH, percentage points) with 95\% confidence intervals across seven B-lineage states ($n = 12$ C4 patients), normalized to all captured B and plasma cells per patient and tissue. \textbf{b,} Paired patient-level proportions of GC-like, extrafollicular (EF)-related, and memory B cells, with lines linking tissues from the same patient. \textbf{c,} Paired differential expression estimates for representative genes in the bulk B lineage and within individual states (left), alongside per-patient memory B-cell proportion shifts versus within-state \textit{TOP2A} changes highlighting P13, where bulk and memory B-cell \textit{TOP2A} diverge (right). \textbf{d,} Linear count-share decomposition of the mean \textit{TOP2A} tissue difference ($+0.99$ per $10{,}000$ total counts) into cellular composition ($+0.60$), shared within-state expression ($+0.29$), and one-sided state remainder ($+0.10$).}
\label{fig:escc-decoupling}
\end{figure}

In metastatic lymph nodes, GC-like B cells increased by a mean of $20.3$ percentage points, while memory B cells declined by $17.0$ percentage points, with each shift observed in 11 of 12 patients (Figure~\ref{fig:escc-decoupling}a,b). Across the pooled B lineage, \textit{MKI67} and \textit{TOP2A} were also elevated in metastatic nodes, with paired negative-binomial $\log_2$ fold changes of $+2.42$ and $+2.06$. Because bulk averages readily conflate an expansion of cycling-rich cells with cell-intrinsic upregulation, CytoCoScientist moved beyond bulk contrasts to compare paired tissues within each cell state per patient. Within the same patient, bulk and within-state trends can diverge: in patient P13, \textit{TOP2A} declined across the bulk B lineage yet rose inside memory B cells (Figure~\ref{fig:escc-decoupling}c and SI Figure~\ref{fig:si-escc-heatmap}).

This pattern extended beyond P13. Among the 10 patients with sufficient memory B cells in both tissues, within-state \textit{TOP2A} rose in all 10, and 9 simultaneously showed a lower memory B-cell proportion (Figure~\ref{fig:escc-decoupling}c; within-state $\log_2$ fold change $+1.40$, adjusted $q = 0.012$). \textit{TYMS}, \textit{CCNB2}, \textit{UBE2C}, and \textit{CDC20} shifted in the same direction in 8 to 9 of 10 patients. By contrast, only 7 patient pairs were eligible for GC-like B cells, and the confidence intervals for within-state \textit{MKI67} and \textit{TOP2A} crossed zero, leaving the magnitude of GC-intrinsic change unresolved. Decomposing the total \textit{TOP2A} RNA count-share difference on a linear scale revealed positive contributions from both cellular composition shifts and expression changes within shared states, alongside a small remainder from states detected on only one side (Figure~\ref{fig:escc-decoupling}d and SI Figure~\ref{fig:si-escc-diagnostics}). Compositional remodeling can therefore account for part of the bulk difference, but it cannot explain why \textit{TOP2A} expression still increases inside memory B cells as their proportion declines.

\begin{figure}[!htbp]
\centering
\includegraphics[width=\linewidth]{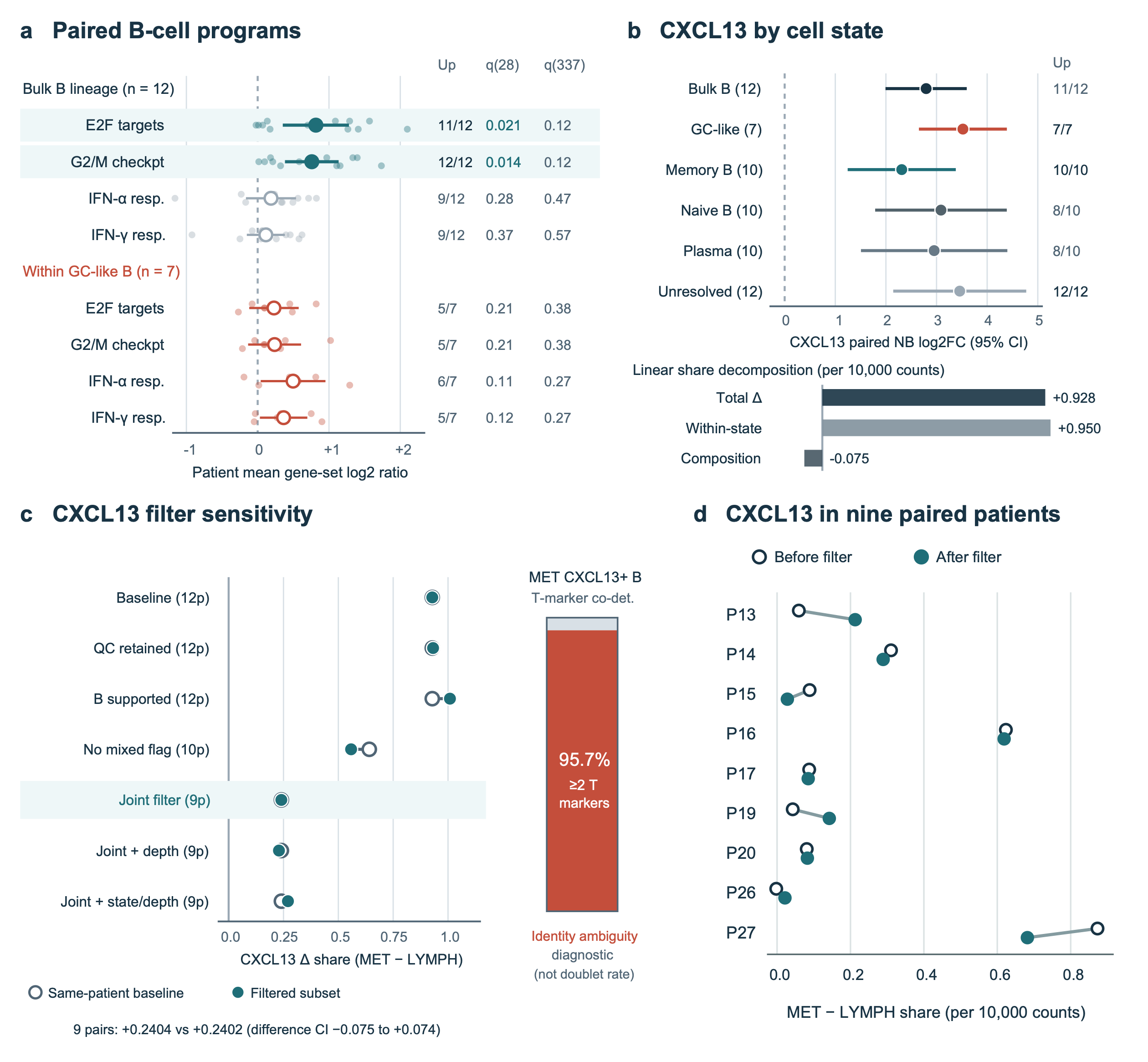}
\caption{\textbf{Patient-level pathway consistency and cellular-origin audit of \textup{CXCL13} in metastatic lymph nodes.} \textbf{a,} Patient-level exact sign-flip comparisons of Hallmark cell-cycle and interferon gene sets in the bulk B lineage and within GC-like B cells (focused and full-family testing scopes in SI Figure~\ref{fig:si-escc-pathway-audit}). \textbf{b,} Paired differential expression and linear count-share decomposition for \textup{CXCL13}. \textbf{c,} Cohort-mean \textup{CXCL13} tissue differences before and after each filtering rule, computed over the matched subset of patients retained by each filter (left), alongside T-cell marker co-detection among \textup{CXCL13}-positive cells originally annotated as B cells (right). \textbf{d,} Per-patient MET-minus-LYMPH \textup{CXCL13} count-share differences (per $10{,}000$ total counts) before (open circles) and after (filled circles) joint filtering across the 9 retained patients.}
\label{fig:escc-pathway-cxcl13}
\end{figure}

A second candidate, \textup{CXCL13}, exposed a different interpretive boundary: the cell label under which RNA is captured does not establish the cell type that transcribed it. \textup{CXCL13} increased in the bulk B lineage in 11 of 12 patients, as well as within eligible GC-like (7/7) and memory B (10/10) labels (Figure~\ref{fig:escc-pathway-cxcl13}b). To test whether this signal depended on records with ambiguous cell identity, the initial analysis applied a joint filter combining cell quality, B-cell identity markers, and mixed-lineage flags. After joint filtering, 9 of the 12 patient pairs remained eligible for analysis.

During review, CytoCoScientist withdrew an intermediate analysis that dropped sparse shared cell states and distorted state weights. Following outer-loop meta-review (Section~\ref{sec:method-meta-review}), it compared pre- and post-filtering results within the same nine retained patients (SI Figure~\ref{fig:si-escc-dag} and SI Table~\ref{tab:si-escc-decisions}). The mean MET-minus-LYMPH \textup{CXCL13} RNA count-share difference was $+0.2404$ before filtering and $+0.2402$ afterward (Figure~\ref{fig:escc-pathway-cxcl13}c and SI Figure~\ref{fig:si-escc-cxcl13-sens}). Although the overall mean remained nearly unchanged, individual responses varied: the difference increased after filtering in P13 and decreased in P27 (Figure~\ref{fig:escc-pathway-cxcl13}d).

CytoCoScientist subsequently ran two follow-up audits in the same cohort. The first tested whether single-gene cycle differences extended to functionally coherent gene sets. Ranked by full-gene differential expression, both cell-cycle and interferon-response gene sets appeared enriched in GC-like B cells. Yet high gene-level rankings do not ensure consistent direction across patients. Recomputing mean gene-set expression differences within each patient showed that bulk B-lineage E2F targets rose in 11 of 12 patients and the G2/M checkpoint gene set rose in all 12. Both passed multiple-testing correction across the 28 post-exploration focused tests, whereas GC-like cell-cycle and interferon gene sets did not; across all 337 Hallmark tests, none reached $q < 0.05$ (Figure~\ref{fig:escc-pathway-cxcl13}a and SI Figure~\ref{fig:si-escc-pathway-audit}). The evidence therefore supports consistent bulk B-lineage cell-cycle expression in this cohort rather than a confirmed GC-intrinsic program shift.

The second audit tested whether a stable 9-patient mean sufficed to attribute \textup{CXCL13} to B cells themselves. Among metastatic-node cells originally annotated as B cells with detectable \textup{CXCL13}, $95.7\%$ co-detected at least two T-cell markers (Figure~\ref{fig:escc-pathway-cxcl13}c and SI Section~\ref{sec:si-escc-s5}), pointing to pervasive cell-identity ambiguity. Although B-cell CXCL13 production has been reported in chronic obstructive pulmonary disease~\cite{Litsiou2013BCellCXCL13}, scRNA-seq alone only establishes elevated \textup{CXCL13} RNA captured under B-cell labels; resolving whether these transcripts originate inside intact B cells or adjacent cells requires in situ localization with cell-boundary resolution on paired tissue sections.

Together, these revisions show how within-task RSI reshapes interpretation in open-ended single-cell discovery. Holding cell state fixed separated intrinsic memory B-cell \textit{TOP2A} elevation from lineage-wide compositional remodeling (Figure~\ref{fig:escc-decoupling}c). Outer-loop meta-review prompted an assessment of \textup{CXCL13} filtering sensitivity within the same patient set, followed by an audit of cellular origin (Figure~\ref{fig:escc-pathway-cxcl13}c,d). Where sequencing data reached an identifiability boundary, CytoCoScientist translated the unresolved questions into targeted follow-up study designs, including paired-section RNA localization for \textup{CXCL13} and tissue-normalized cell density measurements for GC-like B cells (SI Table~\ref{tab:si-escc-follow-up}).

\subsection{Five-Omics Profiling in Acute Aortic Dissection}\label{sec:exp:anzhen}

\begin{figure}[!htbp]
\centering
\includegraphics[width=\linewidth]{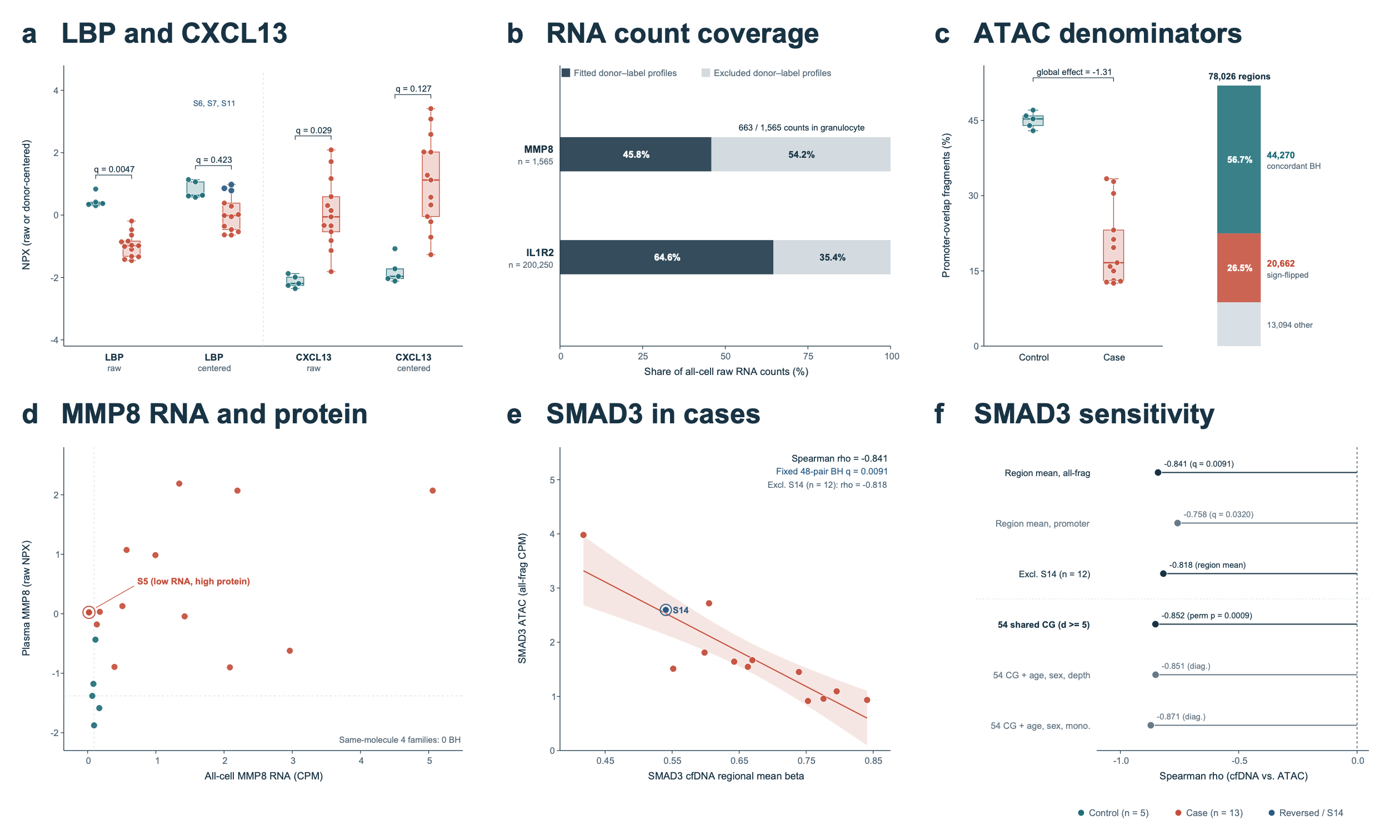}
\caption{\textbf{Group-level convergence does not imply across-patient correlation: the \textit{MMP8} counterexample and the \textit{SMAD3} regional inverse association.} \textbf{a,} Per-participant plasma LBP and CXCL13 distributions ($n = 18$) on raw NPX and individual panel-centered NPX scales (dashed lines mark the 5-control observed range); on the raw scale all 13 cases fall below or above all controls, whereas panel centering reverses LBP direction in S6, S7, and S11. \textbf{b,} Raw PBMC RNA count breakdown for \textit{MMP8} and \textit{IL1R2} across all 41 cell labels, partitioned by inclusion in subpopulation differential models; 848 of $1{,}565$ \textit{MMP8} counts ($54.2\%$) originate from unmodeled donor--label combinations. \textbf{c,} Global PBMC ATAC promoter-fragment fraction (global effect $-1.31$, left) and directional comparison across $78{,}026$ regions between whole-fragment and promoter-assigned denominators ($20{,}662$ sign flips, right). \textbf{d,} Participant-aligned all-cell \textit{MMP8} RNA CPM versus raw plasma MMP8 NPX ($n = 18$), with red circle marking the low-RNA, high-protein counterexample S5. \textbf{e,} Across-case inverse association between regional cfDNA methylation mean and whole-fragment PBMC ATAC within the fixed \textit{SMAD3} window ($n = 13$ cases; $\rho = -0.841$, 48-pair family $q = 0.0091$; excluding S14, $\rho = -0.818$). \textbf{f,} Stability of the \textit{SMAD3} cross-material association across ATAC denominators, S14 exclusion, 54 shared strand-specific CpG records ($\rho = -0.852$, permutation $p = 0.0009$), and individual covariate checks.}
\label{fig:anzhen-single-dual}
\end{figure}

\begin{figure}[!htbp]
\centering
\includegraphics[width=\linewidth]{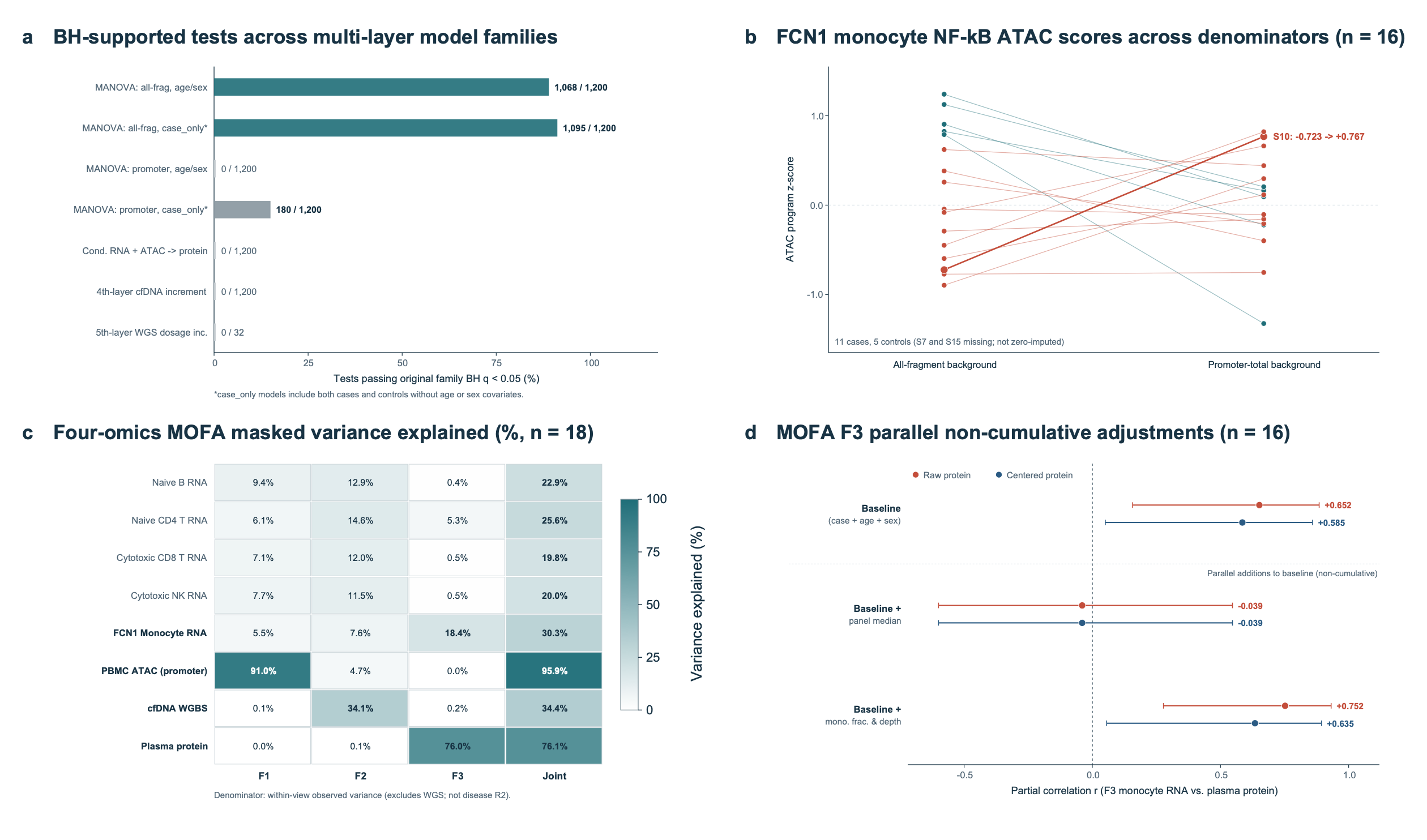}
\caption{\textbf{Group-level three-layer differences do not establish conditional associations, and four-omics F3 coupling attenuates after plasma panel-median adjustment.} \textbf{a,} Counts of tests passing Benjamini--Hochberg correction ($q < 0.05$) with exact family denominators across multi-layer models: MANOVA with case status as the sole predictor or with additional adjustment for age and sex under whole-fragment and promoter-assigned ATAC denominators ($1{,}200$ tests each), conditional RNA and ATAC association with plasma protein ($1{,}200$ tests), fourth-layer cfDNA increment ($1{,}200$ tests), and fifth-layer WGS dosage increment (32 estimable tests). \textbf{b,} Comparison of whole-fragment versus promoter-assigned ATAC scores for the FCN1 classical-like monocyte NF-$\kappa$B program across 16 observed participants (11 cases, 5 controls; S7 and S15 missing), highlighting the sign flip in S10 ($-0.723$ to $+0.767$). \textbf{c,} Within-view variance explained ($R^2$) across 8 views in 18-participant four-omics MOFA (excluding WGS), recomputed over observed non-missing entries. \textbf{d,} Partial correlation coefficients with approximate 95\% confidence intervals from Fisher's $z$-transformation between F3 monocyte RNA and plasma protein projections ($n = 16$) under three parallel, non-cumulative adjustment models.}
\label{fig:anzhen-three-four}
\end{figure}

\begin{figure}[!htbp]
\centering
\includegraphics[width=\linewidth]{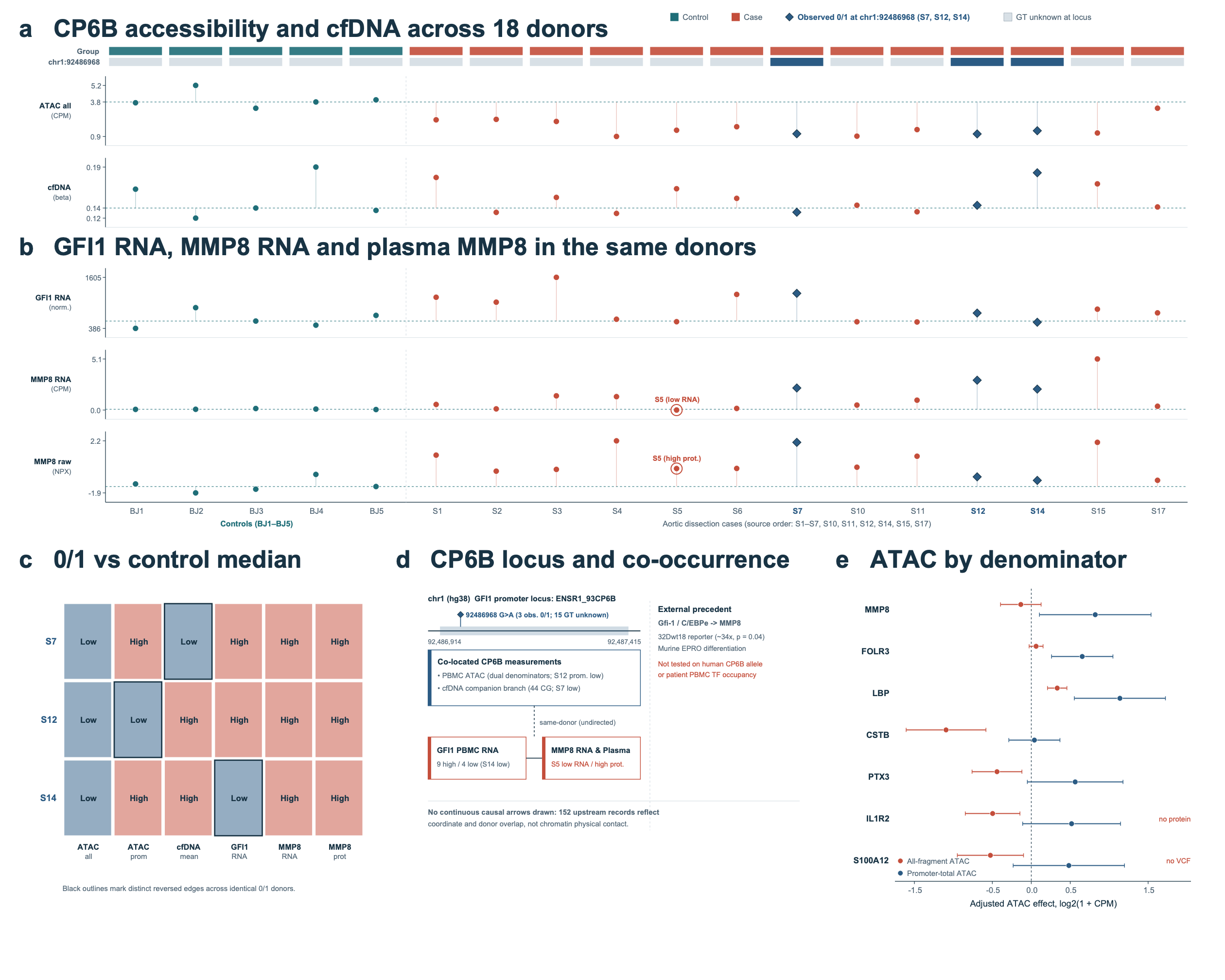}
\caption{\textbf{All three CP6B heterozygous cases show elevated \textit{MMP8} RNA and MMP8 protein but diverge in upstream readouts.} \textbf{a,b,} Participant-aligned profiles in fixed cohort order (BJ1--BJ5; S1--S7, S10--S12, S14, S15, S17) showing CP6B whole-fragment ATAC CPM, regional cfDNA methylation mean, normalized \textit{GFI1} RNA counts, all-cell \textit{MMP8} RNA CPM, and raw plasma MMP8 NPX (full dual-denominator and dual-scale values in SI Table~\ref{tab:si-anzhen-cp6b}). Dark blue diamonds mark \texttt{chr1:92486968 G>A} heterozygous carriers S7, S12, and S14 (genotypes unknown in the other 15 participants); red circles align the low-RNA, high-protein counterexample S5 across both \textit{MMP8} tracks. \textbf{c,} Direction matrix relative to control medians across six readouts for the three heterozygous carriers, with black boxes marking each carrier's discordant upstream readout. \textbf{d,} Undirected evidence diagram for CP6B--\textit{GFI1}--\textit{MMP8}, grouping sequence, ATAC, and cfDNA branches within CP6B, linking within-donor readouts by undirected edges, and separating the external Gfi-1/C/EBP$\varepsilon$ functional precedent. \textbf{e,} Adjusted ATAC effects and model confidence intervals across both denominators for the seven candidate windows (cfDNA effects and $q$ values in SI Table~\ref{tab:si-anzhen-seven}), marking missing observed layers for \textit{IL1R2} (no protein assay) and \textit{S100A12} (no VCF record).}
\label{fig:anzhen-cp6b}
\end{figure}

In small clinical cohorts, adding another omics modality provides a new view of disease without adding an independent patient, so two molecules that rise in cases on average may not be elevated in the same individuals. To test whether CytoCoScientist could separate group-level shifts from within-participant associations and revise its interpretation when the two diverged, we conducted an acute aortic dissection study based on the initial research objective (building on prior multi-omics~\cite{Zheng2023AorticMultiomics} and neutrophil extracellular trap~\cite{Zhao2024AorticDissection} studies) to identify candidate biomarkers, pathways, and follow-up priorities from peripheral blood (original objective in SI Section~\ref{sec:si-anzhen-s1}). The cohort comprises 13 patients sampled after symptom onset and before surgery and 5 unpaired healthy controls, each contributing five modalities from a single blood draw: PBMC single-cell RNA and ATAC (from the same donors, though not the same cells), $1{,}249$ plasma proteins, circulating cell-free DNA (cfDNA) methylation, and whole-genome sequencing (WGS). Auditing case--control differences against within-participant and locus-aligned readouts separated three outcomes: an across-case inverse cfDNA--ATAC association at \textit{SMAD3} (Figure~\ref{fig:anzhen-single-dual}), multi-layer group differences that did not survive denominator switching or translate into conditional associations once case status or global plasma levels were controlled (Figure~\ref{fig:anzhen-three-four}), and a participant-localized five-layer candidate (CP6B--\textit{GFI1}--\textit{MMP8}) whose physical chromatin-contact claim was retracted after outer-loop meta-review (Figure~\ref{fig:anzhen-cp6b}; cohort audit and full search trajectory in SI Section~\ref{sec:si-anzhen-s1-2}, SI Figures~\ref{fig:si-anzhen-dag}--\ref{fig:si-anzhen-qc}, and SI Tables~\ref{tab:si-anzhen-decisions} and~\ref{tab:si-anzhen-combos}).

In the analysis pooling all PBMCs, \textit{MMP8} RNA exceeded the control median in 12 of 13 cases, and \textit{IL1R2} RNA exceeded it in 11 of 13. The age- and sex-adjusted $\log_2$ fold changes were $+3.58$ for \textit{MMP8} ($q = 0.0075$) and $+5.57$ for \textit{IL1R2} ($q = 3.1 \times 10^{-5}$; Figure~\ref{fig:anzhen-single-dual}b; SI Section~\ref{sec:si-anzhen-s4} and SI Table~\ref{tab:si-anzhen-rna25}). Because elevated MMP8 has been reported in dissection tissue and serum~\cite{Zhang2022MMP8AorticDissection}, CytoCoScientist aligned PBMC RNA with matched plasma protein readouts by participant. Although plasma MMP8 exceeded the control median in all 13 cases on both protein scales, case S5 displayed elevated plasma MMP8 alongside below-median PBMC \textit{MMP8} RNA (Figure~\ref{fig:anzhen-single-dual}d). Elevated circulating MMP8 thus cannot be attributed uniformly to concurrent PBMC transcription in a single blood draw, which conflates de novo synthesis with intracellular store release, non-PBMC sources, and altered clearance.

Single-omics direction and significance also depended heavily on the chosen reference scale, making dual-reference auditing essential before cross-layer inference. On raw NPX, plasma LBP fell below the 5-control range in all 13 cases ($q = 0.0047$) and CXCL13 exceeded it in all 13 ($q = 0.0294$; Figure~\ref{fig:anzhen-single-dual}a), yet subtracting each participant's full-panel median reduced LBP below the control median to 10 of 13 cases and removed significance ($q = 0.423$). Two external dissection cohorts reported elevated rather than reduced plasma LBP~\cite{Cheng2020AADProteomics,Yesitayi2025LBPDissection}. PBMC ATAC exhibited an analogous denominator dependence: across $78{,}026$ comparable promoter regions, $20{,}662$ flipped from negative to positive case--control effects when normalized to promoter-assigned rather than whole-fragment counts, showing that both denominators capture relative share rather than absolute accessibility (Figure~\ref{fig:anzhen-single-dual}c). Leave-S14-out protein sensitivity, cfDNA heteroskedasticity-robust modeling, and WGS unknown-genotype rules are detailed in SI Section~\ref{sec:si-anzhen-s3-1}, SI Table~\ref{tab:si-anzhen-protein48}, and SI Figures~\ref{fig:si-anzhen-rna25-protein48}b and~\ref{fig:si-anzhen-prl-il1r2}a,b.

Across all 48 prespecified cross-material pairs, only the \textit{SMAD3} region showed a consistent across-case inverse association between plasma cfDNA methylation and PBMC chromatin accessibility. Within the fixed $\sim 2\text{ kb}$ \textit{SMAD3} window, higher cfDNA methylation tracked with lower PBMC accessibility across the 13 cases under whole-fragment ATAC normalization ($\rho = -0.841$, 48-pair adjusted $q = 0.0091$; Figure~\ref{fig:anzhen-single-dual}e), and the negative correlation held when switching to promoter-assigned normalization, restricting to 54 shared strand-specific CpG records covered across all cases ($\rho = -0.852$), or excluding S14 (Figure~\ref{fig:anzhen-single-dual}f; full sensitivity table in SI Table~\ref{tab:si-anzhen-smad3}). Although pathogenic \textit{SMAD3} variants cause syndromic aortic aneurysms and dissections~\cite{vandeLaar2011SMAD3}, plasma cfDNA derives from multiple tissues~\cite{Moss2018cfDNAAtlas} and downstream \textit{SMAD7} RNA and TGFB1 protein comparisons did not pass correction. \textit{SMAD3} therefore remains the clearest lead for replication, at the level of an across-case cross-material association.

Beyond \textit{SMAD3}, group-level convergence across two, three, and four modalities did not translate into individual-level or conditional cross-material coupling. In systematic RNA--protein testing, none of $16{,}822$ estimable gene-matched pairs passed multiple-testing correction; neutrophil degranulation was concordantly enriched at the group level in both layers, yet all $208$ across-case tests linking immune-process RNA scores to plasma MMP8 and LBP remained non-significant (SI Section~\ref{sec:si-anzhen-s3-2}, SI Table~\ref{tab:si-anzhen-208}, and SI Figure~\ref{fig:si-anzhen-mofa-208}d). In three-layer pathway modeling across RNA, ATAC, and plasma protein, $1{,}068$ of $1{,}200$ age- and sex-adjusted MANOVA tests showed case--control group differences under whole-fragment ATAC normalization, whereas switching to the promoter-assigned denominator reduced significant tests to $0/1{,}200$ (illustrated by S10's score sign flip in Figure~\ref{fig:anzhen-three-four}b). Once case status was controlled, none of the $1{,}200$ conditional RNA--ATAC--protein tests or fourth-layer cfDNA increment tests passed correction under any model specification (Figure~\ref{fig:anzhen-three-four}a; SI Table~\ref{tab:si-anzhen-multilayer} and SI Figure~\ref{fig:si-anzhen-nfkb-mylk}a--c). Likewise, in an unsupervised four-omics MOFA branch~\cite{Argelaguet2018MOFA} whose variance explained was recomputed over observed non-missing entries after outer-loop meta-review (Section~\ref{sec:method-meta-review}), factor F3 jointly explained $76.01\%$ of plasma protein variance and $18.40\%$ of FCN1 monocyte RNA variance (Figure~\ref{fig:anzhen-three-four}c; SI Table~\ref{tab:si-anzhen-mofa-r2}). However, the covariate-adjusted partial correlation between F3 monocyte RNA and plasma protein projections dropped from $r = 0.652$ to $-0.039$ once each participant's plasma panel median was controlled, whereas it remained positive under monocyte-proportion and sequencing-depth adjustment; these data therefore do not support monocyte-specific coupling (Figure~\ref{fig:anzhen-three-four}d; SI Table~\ref{tab:si-anzhen-f3-pcor}).

In the \textit{GFI1} promoter-annotated region CP6B (\texttt{ENSR1\_93CP6B}), three cases (S7, S12, and S14) carried the heterozygous variant \texttt{chr1:92486968 G>A}. All three had PBMC \textit{MMP8} RNA and plasma MMP8 protein above the control medians. The upstream measurements differed by carrier: cfDNA methylation was below the control median in S7, promoter-normalized ATAC in S12, and \textit{GFI1} RNA in S14 (Figure~\ref{fig:anzhen-cp6b}a--c and SI Table~\ref{tab:si-anzhen-cp6b}).

Outer-loop meta-review prompted a re-examination of the 152 records used to link these measurements. The records documented genomic-interval overlaps or linked measurements from the same donor. CytoCoScientist withdrew the physical chromatin-contact interpretation and redrew the evidence diagram to show those relationships and the previously published cellular evidence for Gfi-1/C/EBP$\varepsilon$ regulation of \textit{MMP8}~\cite{KhannaGupta2007Gfi1} as separate sources (Figure~\ref{fig:anzhen-cp6b}d). CP6B genotypes were unknown for the other 15 participants, and none of the 32 genetic-dosage comparisons across six candidate loci reached significance (Figure~\ref{fig:anzhen-three-four}a and SI Table~\ref{tab:si-anzhen-multilayer}). The pattern in the three carriers identifies CP6B as a candidate for follow-up.

Figure~\ref{fig:anzhen-cp6b}e and SI Section~\ref{sec:si-anzhen-s5} also report candidates with discordant measurements across layers (\textit{FOLR3}, \textit{CSTB}, and \textit{PTX3}) or unmeasured layers (\textit{IL1R2} and \textit{S100A12}), and the single-case structural hypothesis for S14 \textit{FBN1} \texttt{p.R1539*} (SI Figures~\ref{fig:si-anzhen-seven} and~\ref{fig:si-anzhen-fbn1} and SI Tables~\ref{tab:si-anzhen-seven} and~\ref{tab:si-anzhen-fbn1}).

Outer-loop meta-review prompted recomputation of MOFA variance strictly over observed entries and retraction of the CP6B physical-contact claim (SI Table~\ref{tab:si-anzhen-decisions} also documents a correction of age and sex mismatches in the earlier cfDNA model). The dual-reference analyses identified which results depended on the choice of reference. We proposed two priorities for follow-up: replicate the \textit{SMAD3} regional cfDNA--ATAC association in an independent dissection cohort, and resolve carrier versus non-carrier genotypes at CP6B across all participants before testing allelic effects (full follow-up plan in SI Table~\ref{tab:si-anzhen-followup}).
\FloatBarrier

\section{Limitations}\label{sec:limitations}

The five case studies evaluate CytoCoScientist as a complete system. In future work, we will benchmark CytoCoScientist against other systems under matched computational budgets and use ablation studies to measure how individual components affect performance.

\section{Conclusion}
\label{sec:conclusion}

We presented CytoCoScientist, an autonomous biological research system that achieves evidence-coupled recursive self-improvement (RSI) under fixed model parameters. In long-horizon biological research, retrieved literature and empirical checks continually reshape how candidate solutions are evaluated, while revised candidates redirect what evidence must be gathered next. Rather than treating autonomous research as a fixed workflow or an isolated code-generation loop, CytoCoScientist anchors decision-making in a persistent research state that couples explicit candidate search spaces, evidence-based control with outer-loop meta-review, and isolated subagent contexts paired with a revisable timeline, provenance-linked memory, and reusable domain skills. Separating result generation from evidential admissibility allows the system to sustain exploration over dozens of interdependent decisions without amplifying data leakage, compositional confounding, or superseded interpretations.

Across five long-horizon biological studies, CytoCoScientist demonstrates both quantitative algorithm development and open-ended bioinformatics inquiry. In algorithm development, CytoCoScientist outperforms established baseline algorithms on OpenProblems single-cell denoising ($4.7\%$ lower mean squared error than MAGIC) and TCGA-LUAD survival modeling ($+0.0508$ C-index over matched DeepProg) while rejecting a higher-scoring but uncalibrated early-concatenation model. In open-ended inquiry, it pivots from single-draw proteomics to within-patient longitudinal trajectories in sepsis mortality, decouples intrinsic memory B-cell \textit{TOP2A} elevation from compositional remodeling while auditing \textup{CXCL13} cellular origin in paired esophageal squamous cell carcinoma (ESCC) lymph nodes, and isolates a cross-material \textit{SMAD3} association while retracting an unsupported chromatin-contact claim across five blood-derived modalities in acute aortic dissection, translating unresolved computational leads into concrete plans for experimental validation.

As AI systems take on increasingly long-horizon scientific problems, progress will depend less on generating more hypotheses or code in isolation than on governing how evidence reshapes inquiry over time. CytoCoScientist demonstrates that recursive self-improvement can operate directly on scientific reasoning and methodology, turning empirical feedback, rigorous auditing, and even negative findings into cumulative progress across a study. Extending this evidence-grounded paradigm into closed-loop dry- and wet-lab research cycles and multi-study programs will allow RSI to accelerate biological discovery and advance autonomous inquiry across the empirical sciences.

\section*{Declarations}
\phantomsection
\addcontentsline{toc}{section}{Declarations}
\subsection*{Data availability}
Public data analyzed in this study are available from the sources cited in the Supplementary Information: the OpenProblems Denoising v1.0.0 benchmark files, TCGA-LUAD, the plasma proteomics cohorts PXD055932 and PXD039875, RACE/CaPS, the Ferrario/ALBIOS supplementary tables, whole-blood expression datasets GSE95233, GSE54514, and GSE272769, and the Kwok et al.\ single-cell atlas (Zenodo 7924238). The in-house ESCC and acute aortic dissection cohorts are subject to data-provider authorization and are not publicly redistributable; derived results needed to evaluate the reported analyses are provided in the Supplementary Information.
\subsection*{Code availability}
The case-level code for the five studies, including the analysis pipelines, candidate implementations, and evaluation procedures, is available at \url{https://github.com/Cyto-System/Cyto.CoScientist}. For the two algorithm-development studies, this repository includes the pinned environment of the single-cell denoising implementation (\texttt{anndata}~0.10.9, \texttt{numpy}~1.26.4, \texttt{scipy}~1.13.1, \texttt{h5py}~3.12.1, and \texttt{pandas}), the \texttt{numpy}, \texttt{pandas}, and \texttt{scipy} dependencies of the TCGA-LUAD survival pipeline, and the DeepProg reproduction harness pinned to upstream commit \texttt{e2c9e3c71344b1b470235371c3f0cea8c9440b16}. 
\subsection*{Use of AI}
The initial manuscript draft and related research materials were generated using CytoCoScientist and subsequently reviewed and substantially revised by the authors. The authors verified all analyses, references, figures, interpretations, and final text, and take full responsibility for the content of the manuscript.

\FloatBarrier
\phantomsection
\addcontentsline{toc}{section}{References}
\putbib[references]
\end{bibunit}
\appendix
\begin{bibunit}[plainnat]
\makeatletter
\def\@extra@b@citeb{.si}
\def\@extra@binfo{.si}
\@startbibunitorrelax
\makeatother
\renewcommand{\thelstlisting}{S\arabic{lstlisting}}
\section{Benchmarks for Scientific and Life Science Agents}\label{sec:si-related-work}\label{sec:si-related-evaluation}

SI Table~\ref{tab:related-benchmarks} summarizes representative scientific-agent benchmarks, including general research tasks and bioinformatics tasks. Comparisons require distinguishing the guidance supplied to an agent from the outputs being evaluated; task duration or the number of stages alone does not establish how open-ended a task is. For example, BixBench3 supplies research objectives and methodological guidance, PaperBench provides a target paper for replication, and FML-bench separates search strategies from execution infrastructure \citep{Koch2026BixBench3,Starace2025PaperBench,Zou2026FMLBench}. Scores variously emphasize answer correctness, computational results, evidential support, verification ability, or research outputs and should therefore be interpreted within the corresponding task conditions.

\begingroup
\small
\setlength{\tabcolsep}{4pt}
\renewcommand{\arraystretch}{1.12}
\begin{longtable}{@{}>{\raggedright\arraybackslash}p{0.23\textwidth}>{\raggedright\arraybackslash}p{0.46\textwidth}>{\raggedright\arraybackslash}p{\dimexpr0.31\textwidth-4\tabcolsep\relax}@{}}
\caption{\textbf{Representative benchmarks for scientific and bioinformatics agents.} Grouped by primary evaluation role, with task conditions, evaluation procedures, and evaluation focus.}\label{tab:related-benchmarks}\\
\toprule
\textbf{Benchmark} & \textbf{Task and evaluation} & \textbf{Evaluation focus} \\
\midrule
\endfirsthead
\multicolumn{3}{@{}l}{\small\tablename~\thetable{} (continued)}\\
\toprule
\textbf{Benchmark} & \textbf{Task and evaluation} & \textbf{Evaluation focus} \\
\midrule
\endhead
\midrule
\multicolumn{3}{r@{}}{\small Continued on next page}\\
\endfoot
\bottomrule
\endlastfoot

\multicolumn{3}{@{}l}{\textbf{Biological research capabilities and analytical judgment}}\\*
LAB-Bench (2024) \citep{Laurent2024LABBench}
& Biology research questions involving literature, data, and experimental knowledge; evaluates answer correctness.
& Biological research knowledge and evidence interpretation. \\
\addlinespace[0.5em]
LABBench2 (2026) \citep{Laurent2026LABBench2}
& Extends biological research evaluation to more realistic contexts and more demanding tasks.
& Biological research capabilities in complex settings. \\
\addlinespace[0.5em]
GeneBench-Pro (2026) \citep{OpenAI2026GeneBenchPro}
& Computational biology tasks with controlled data and known answers assess analytical choices, hypothesis revision, and interpretation.
& Analytical decisions under controlled conditions. \\
\addlinespace[0.5em]

\multicolumn{3}{@{}l}{\textbf{Bioinformatics workflows and biological conclusions}}\\*
BixBench (2025) \citep{Mitchener2025BixBench}
& Biological datasets and questions require multi-step computation and interpretation; evaluates answers.
& Biological data analysis and answer correctness. \\
\addlinespace[0.5em]
BixBench3 (2026) \citep{Koch2026BixBench3}
& Study-scale tasks supply raw data, an objective, and methodological guidance; output artifacts are graded against published study artifacts.
& Sustained execution and study-scale outputs. \\
\addlinespace[0.5em]
BioAgent Bench (2026) \citep{Fa2026BioAgentBench}
& RNA-seq, variant calling, and metagenomics workflows assess pipeline progress, output validity, and robustness to input perturbations.
& Bioinformatics workflow execution and robustness. \\
\addlinespace[0.5em]
scBench-Long (2026) \citep{Diks2026ScBenchLong}
& Long-horizon single-cell tasks derive biological conclusions from raw or near-raw data without prescribed methods; uses controlled answers and trajectory rubrics.
& Long-horizon single-cell analysis and supported conclusions. \\
\addlinespace[0.5em]
SpatialBench-Long (2026) \citep{Diks2026SpatialBenchLong}
& Spatial biology tasks derive claims from complex measurements and experimental context; uses deterministic answer grading and process rubrics.
& Spatial biology analysis and evidence interpretation. \\
\addlinespace[0.5em]

\multicolumn{3}{@{}l}{\textbf{Hypothesis generation and research verification}}\\*
BioVerge (2025) \citep{Yang2025BioVerge}
& Biomedical hypothesis generation from literature and structured information; studies iterative self-evaluation and proposal quality.
& Hypothesis generation, self-evaluation, and proposal quality. \\
\addlinespace[0.5em]
HypoBench (2025) \citep{Liu2025HypoBench}
& Scientific hypothesis generation across synthetic and real-world tasks; evaluates plausibility, novelty, and explanatory power.
& Hypothesis quality and comparison of candidate explanations. \\
\addlinespace[0.5em]
DiscoveryBench (2024) \citep{Majumder2024DiscoveryBench}
& Data-driven discovery tasks formulate and verify scientific hypotheses from datasets and domain context.
& Data-driven hypothesis formulation and verification. \\
\addlinespace[0.5em]
BioDSA-1K (2025) \citep{Wang2025BioDSA1K}
& Hypothesis validation with study-derived evidence, including insufficient-evidence cases; assesses decisions, evidence--conclusion alignment, reasoning, and code executability.
& Alignment between evidence and conclusions. \\
\addlinespace[0.5em]
SPOT (2025) \citep{Son2025SPOT}
& Scientific research verification tasks assess automated systems' ability to identify research errors.
& Research error detection and verification. \\
\addlinespace[0.5em]

\multicolumn{3}{@{}l}{\textbf{Scientific computation, method improvement, and reproducibility}}\\*
MLE-bench (2024) \citep{Chan2024MLEBench}
& Machine-learning engineering tasks drawn from Kaggle competitions; evaluates predictive performance.
& Machine-learning engineering and predictive performance. \\
\addlinespace[0.5em]
ScienceAgentBench (2024) \citep{Chen2024ScienceAgentBench}
& Data-driven scientific programming tasks use execution checks and task-specific output evaluation.
& Scientific program execution and task completion. \\
\addlinespace[0.5em]
EXP-Bench (2025) \citep{Kon2025EXPBench}
& Research questions and starter repositories support evaluation of experimental design, execution, and analysis.
& Research experiment design and implementation. \\
\addlinespace[0.5em]
CORE-Bench (2024) \citep{Siegel2024COREBench}
& Reproduces computational results from supplied research code and data and answers questions about the results.
& Computational reproducibility and result verification. \\
\addlinespace[0.5em]
MLGym-Bench (2025) \citep{Nathani2025MLGym}
& Open-ended machine-learning tasks involve ideas, implementation, experiments, analysis, and iterative improvement over baselines.
& Exploration and improvement of computational solutions. \\
\addlinespace[0.5em]
FML-bench (2026) \citep{Zou2026FMLBench}
& Research tasks separate search strategies from shared execution infrastructure; evaluates final performance and process-level search behavior.
& Search dynamics and exploration strategies. \\
\addlinespace[0.5em]

\multicolumn{3}{@{}l}{\textbf{Research stages and complete research projects}}\\*
PaperBench (2025) \citep{Starace2025PaperBench}
& Supplies a target machine-learning paper for replication from scratch, including code development and experiment execution; uses hierarchical rubrics developed with paper authors.
& End-to-end research replication. \\
\addlinespace[0.5em]
MLR-Bench (2025) \citep{Chen2025MLRBench}
& Open-ended machine-learning research evaluated across ideas, proposals, experiments, manuscripts, and final outputs.
& Multi-stage research and overall output quality. \\
\addlinespace[0.5em]
ResearchClawBench (2026) \citep{Xu2026ResearchClawBench}
& Provides literature and raw data while withholding the target paper; evaluates end-to-end scientific outputs with expert-defined criteria.
& Complete research outputs from supplied materials. \\
\addlinespace[0.5em]
AstaBench (2025) \citep{Bragg2025AstaBench}
& A suite of scientific research activities and end-to-end tasks with standardized tools, environments, and resource-cost accounting.
& Research capabilities and resource use. \\
\addlinespace[0.5em]

\end{longtable}
\endgroup

We evaluate CytoCoScientist across five biological studies spanning algorithm development and open-ended bioinformatics inquiry. Single-cell expression denoising and lung adenocarcinoma multi-omics survival prediction assess iterative computational method improvement under sealed quantitative evaluators. Sepsis mortality cross-cohort exploration, single-omics paired lymph-node profiling in esophageal squamous cell carcinoma, and five-omics integrative analysis in acute aortic dissection examine how the system organizes analyses, compares competing explanations, and revises judgments around open biological questions. Together, these studies document candidate exploration, review feedback, and research-state updates to show how evidence-driven revision shapes subsequent research decisions.

\FloatBarrier
\section{Domain Instantiations and Computational Setup}
\label{sec:si-domain-instantiations}

\subsection{Multi-Source Evidence Acquisition}
\label{sec:si-evidence-acquisition}

\subsubsection{Literature Search, Passage Retrieval, and Review}\label{sec:si-literature-review}

\begin{figure}[!htbp]
\centering
\includegraphics[width=\linewidth]{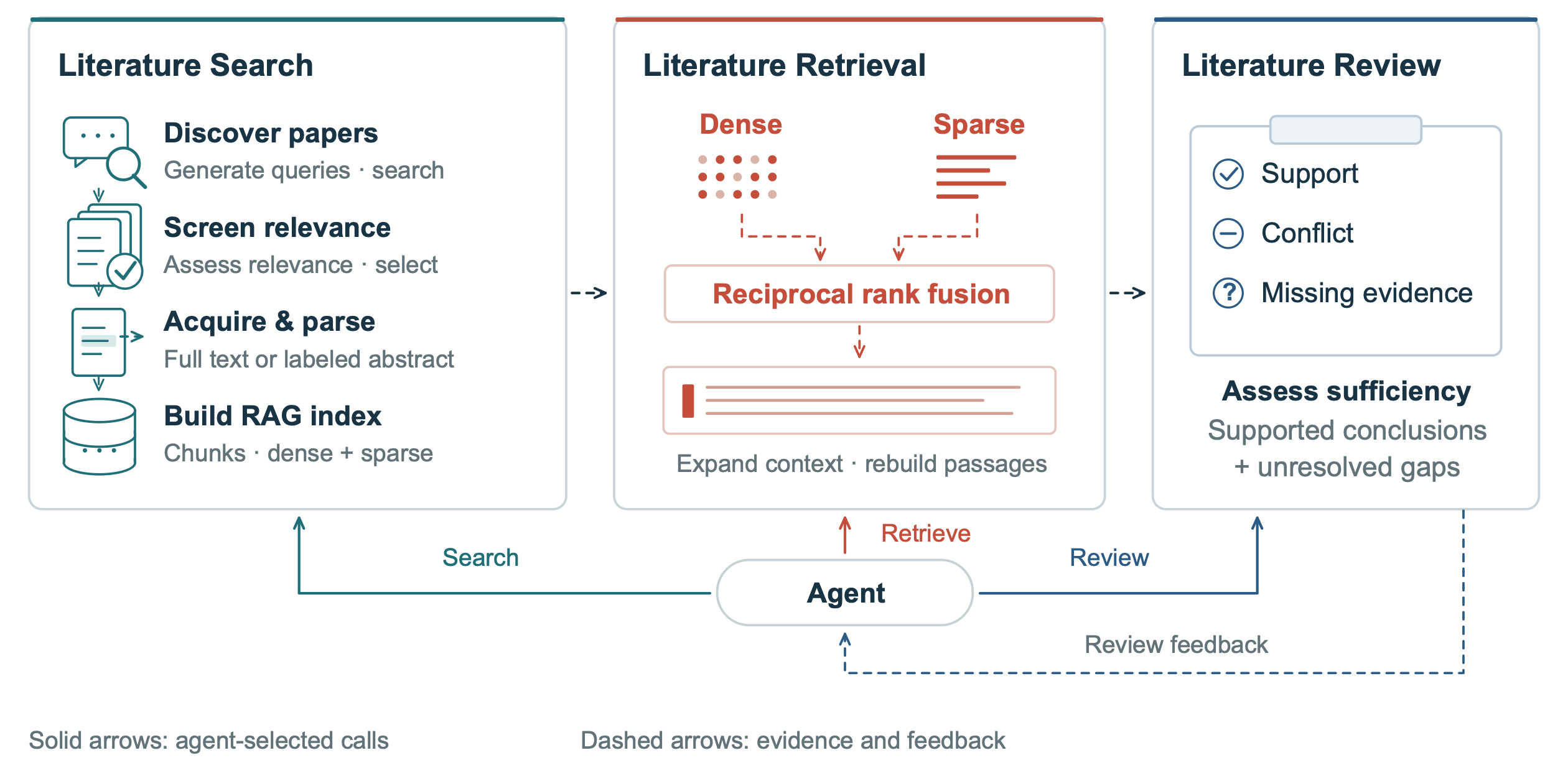}
\caption{\textbf{Literature evidence acquisition and review workflow.} The main agent adaptively invokes Paper Search, hybrid Passage Retrieval, and Evidence Review. Search constructs a deduplicated, reusable paper corpus; retrieval extracts provenance-linked passages; and review classifies evidential support before updating $\mathcal{E}_t$ or flagging coverage gaps to trigger query refinement and downstream activities.}
\label{fig:domain-literature-flow-v2}
\end{figure}

Complementing the overview in Section~\ref{sec:method-evidence-acquisition}, the scholarly path builds a reusable, provenance-indexed corpus (SI Figure~\ref{fig:domain-literature-flow-v2}), whereas the web path extracts primary documentation on demand and audits it in place. The system searches multiple scholarly sources using complementary queries and can retrieve specific papers by identifier. After cross-source identity resolution and deduplication, an LLM screens candidate titles, abstracts, and metadata within the acquisition budget, removing retracted articles, treating venue and citation counts as auxiliary context, and separating direct tests of the target relationship from broader methodological background. Accessible full texts are segmented along section and sentence boundaries and indexed with both dense embeddings and sparse lexical representations, while abstract-only records are tagged explicitly so downstream synthesis respects their restricted scope.

For active sub-questions, retrieval merges dense semantic search with sparse lexical matching through reciprocal rank fusion, deduplicating top segments and expanding them to neighboring sentences to yield self-contained passages with exact citation and section provenance. For each research question, the agent reviews the retrieved passages together to assess whether the available evidence supports an answer and what gaps remain. Passages are classified as direct support, partial support, conflicting findings, or missing coverage, discounting repeated passages and derivative reviews as independent corroboration.

\subsubsection{Web Acquisition and Review}\label{sec:si-web-review}

Web acquisition targets authoritative technical documentation, package manuals, database schemas, and public benchmark specifications. The system deduplicates candidate URLs, extracts page text with retrieval timestamps and truncation flags, and operates directly on extracted artifacts without a persistent vector index. Web Evidence Review checks source authority, version currency, and independence: official documentation supports software interfaces, parameters, and data schemas. Reviewed pages update $\mathcal{E}_t$ or trigger follow-up queries, and admissible extractions remain addressable by URL and retrieval timestamp for later inspection.

\subsection{Algorithm and Program Evolution}\label{sec:si-program-evolution}

\subsubsection{Plan-Constrained Modification and Cumulative Retrospectives}\label{sec:si-program-planning}

Program evolution operates over a persistent lineage graph of complete executable implementations, governed by explicit editable code boundaries, held-out evaluation splits, and isolated evaluators that compute scalar utility and optional behavioral descriptors. While the active search policy (Section~\ref{sec:evolution-selection-v2}) selects which admissible parent version to expand, code modification is constrained by a three-role planning step (Strategist--Skeptic--Planner): the Strategist proposes architectural, representation, or objective modifications from bottleneck diagnostics and literature evidence; the Skeptic audits proposals against prior retrospectives to filter out redundant trials, data leakage risks, or confounded multi-factor edits; and the Planner specifies the exact modules to edit, controls to hold fixed, and validation checks required to isolate each change.

Across batches, settled candidate outcomes (including accepted improvements, valid negative ablations, and diagnosed failures) are distilled into a cumulative program retrospective. This retrospective compares intended plan specifications against actual code diffs and empirical metrics, recording which modifications succeeded under specific data regimes and which implementation patterns triggered validation or review failures.

\subsubsection{Failure Isolation, Parent Admissibility, and Revalidation}\label{sec:si-program-admissibility}

Code execution and admissibility gating separate three verification regimes:
\begin{enumerate}
\item \textbf{Candidate runtime failure vs.\ evaluator infrastructure failure:} Deterministic pre-screening checks that candidate diffs respect editable scope and input/output contracts, execute without exceptions, and produce valid predictions free of NaNs or collapsed outputs before isolated benchmark scoring. When execution fails, the runtime separates \emph{candidate code defects} (routed to targeted repair with source code and tracebacks) from \emph{evaluator or environment faults} (retried or flagged without penalizing algorithmic utility).
\item \textbf{Parent eligibility gating via admissibility review:} Raw benchmark scores do not enter the search pool automatically. Post-execution review audits the code diff, logs, and outputs for data leakage, split peeking, or metric gaming. Only admissible candidates (including methodologically sound negative ablations) enter the parent-eligible pool and update utility rankings; rejected candidates record failure diagnostics and increment parent failure penalties without corrupting rankings, while repairable implementation bugs remain in a repair state until resolved.
\item \textbf{Candidate revalidation:} The selected implementation is rerun under the same benchmark conditions, and its score is compared with the original result.
\end{enumerate}

\subsection{Bioinformatics Data Analysis}\label{sec:si-bioinformatics}

\subsubsection{Analysis Planning, Execution, and Evolution on the Analysis DAG}\label{sec:si-bioinformatics-dependencies}

CytoCoScientist represents the structure of a bioinformatics investigation as a directed acyclic graph (DAG). Each node represents an analysis defined by a biological question, specified inputs, and an evaluable output, while each directed edge indicates that a downstream analysis consumes an upstream result (for example, a cell-type-specific expression test depending on a preceding cell annotation). The distinction between extending a route and revising a step is central: a downstream node builds on an earlier result to address a further question, whereas sibling nodes represent competing implementations of the same analytical step.

\textbf{Analysis planning} identifies which scientific uncertainty to address next based on the research goal, available results, result evaluations, and prior cumulative reviews. The main agent acts on a proposed direction in two ways. \textbf{Analysis execution} introduces and runs an analysis not yet represented in the DAG, either opening a new route or extending an existing one downstream. \textbf{Analysis evolution} revisits an existing step and proposes an alternative; when both versions start from the same upstream data, the original and revised analyses are recorded as siblings. An earlier result can therefore motivate a revision without becoming a computational input to it.

For example, an initial analysis might compare gene expression between clinical groups at the single-cell level. If evaluation flags donor replication as a concern, the agent returns to the same annotated counts to run a donor-level pseudobulk comparison. This is an analysis evolution step: both candidates test the same biological contrast but differ in their unit of inference. Because the pseudobulk test does not consume the cell-level differential-expression table, linking it as a downstream child would misrepresent the dependency. By contrast, a pathway enrichment analysis computed from the pseudobulk gene statistics is a downstream extension via analysis execution. In both cases, the system records each executed analysis alongside its outputs so that every interpretation remains traceable to the computation that produced it.

\subsubsection{Result Evaluation and Cumulative Review}\label{sec:si-bioinformatics-branching}

\textbf{Result evaluation} assesses whether an individual result supports its stated interpretation, examining the unit of inference, sample coverage, controls, uncertainty, and diagnostics relevant to the comparison. Across sibling alternatives, the scientific question is not simply which specification yields the stronger signal, but what their comparison reveals about the finding. In the example above, signal attenuation after donor-level aggregation narrows the interpretation of the cell-level contrast; conversely, agreement indicates robustness to the choice of inference unit within the same cohort. A well-supported null result can thus guide the investigation as directly as a positive finding.

\textbf{Cumulative review} steps back from individual nodes to the overall research goal, synthesizing evaluated results across routes to separate conclusions backed by complementary evidence from those resting on a single cohort, upstream result, or modeling choice. It also surfaces discrepancies and evidence gaps that matter for the primary question. These findings feed directly into the next analysis planning decision: the main agent can extend a supported route via analysis execution, revisit a consequential modeling choice via analysis evolution, or retain a narrower conclusion when available data cannot separate competing explanations. The DAG thus records both how the investigation progressed and where an analytical change altered what could be concluded.

\subsection{Scientific Manuscript Writing}\label{sec:si-writing}

The writing workflow maintains links between manuscript drafts and the evidence used to support them \citep{Lewis2020RetrievalaugmentedGeneration3cb08b0e9ded}. Section-level and manuscript-level reviews assess whether the claims remain consistent with that evidence. When support is insufficient, the main agent can seek additional evidence or narrow the claim to what the existing results justify. Specifically, quality control relies on three mechanisms:
\begin{enumerate}
\item \textbf{Two-tier hierarchical review:} Section Review audits individual assembled sections for internal argumentative coherence, fidelity to cited figures and tables, citation accuracy, and methodological clarity; Manuscript Review audits the full paper for global narrative alignment, notation consistency, and strict agreement across the abstract, results, and conclusion.
\item \textbf{Version-bound review:} Review verdicts apply to the specific section and manuscript versions inspected; editing a subsection invalidates prior reviews of the affected assembly, requiring re-review before completion gating (Section~\ref{sec:method-completion-v2}).
\item \textbf{Evidence-guided revision:} When review identifies an overgeneralized claim, missing statistical control, or unsupported citation, the critique routes the main agent either back to bioinformatics analysis or literature acquisition, or to targeted text revision that narrows the claim.
\end{enumerate}

\begin{figure}[!htbp]
\centering
\includegraphics[width=\linewidth,height=0.83\textheight,keepaspectratio]{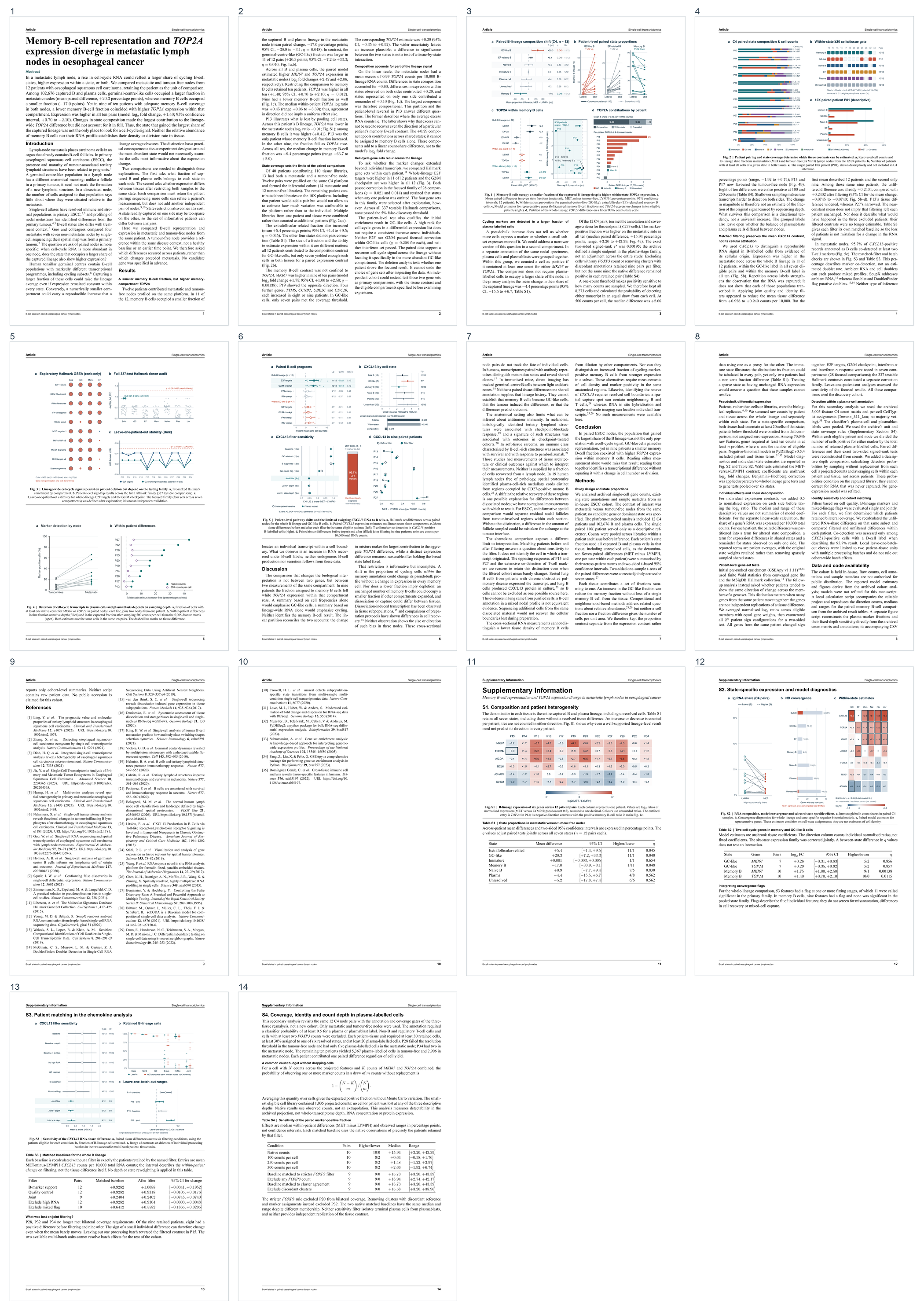}
\caption{\textbf{Representative standalone manuscript generated by CytoCoScientist from the ESCC paired lymph-node study.} All 14 pages of the assembled article (\emph{Memory B-cell representation and TOP2A expression diverge in metastatic lymph nodes in oesophageal cancer}) are shown in reading order from left to right. Pages 1--10 comprise the abstract, introduction, results with four multi-panel figures, discussion, methods, and references; pages 11--14 contain the manuscript's companion Supplementary Information (sensitivity tables and diagnostic figures).}
\label{fig:si-writing-escc-manuscript}
\end{figure}

SI Figure~\ref{fig:si-writing-escc-manuscript} shows a representative standalone manuscript assembled by CytoCoScientist's writing workflow from the ESCC paired lymph-node study (SI Section~\ref{sec:si-case-escc}). Rather than reporting a sequential execution log, the workflow organizes the evidence into a structured research article centered on the divergence between reduced memory B-cell representation and elevated within-compartment \textit{TOP2A} expression (observed in 9 of 10 evaluable patients), integrating multi-panel figures, methods, literature citations, and companion supplementary analyses.

\begin{figure}[!htbp]
\centering
\includegraphics[trim=0mm 75mm 0mm 0mm,clip,width=0.90\linewidth]{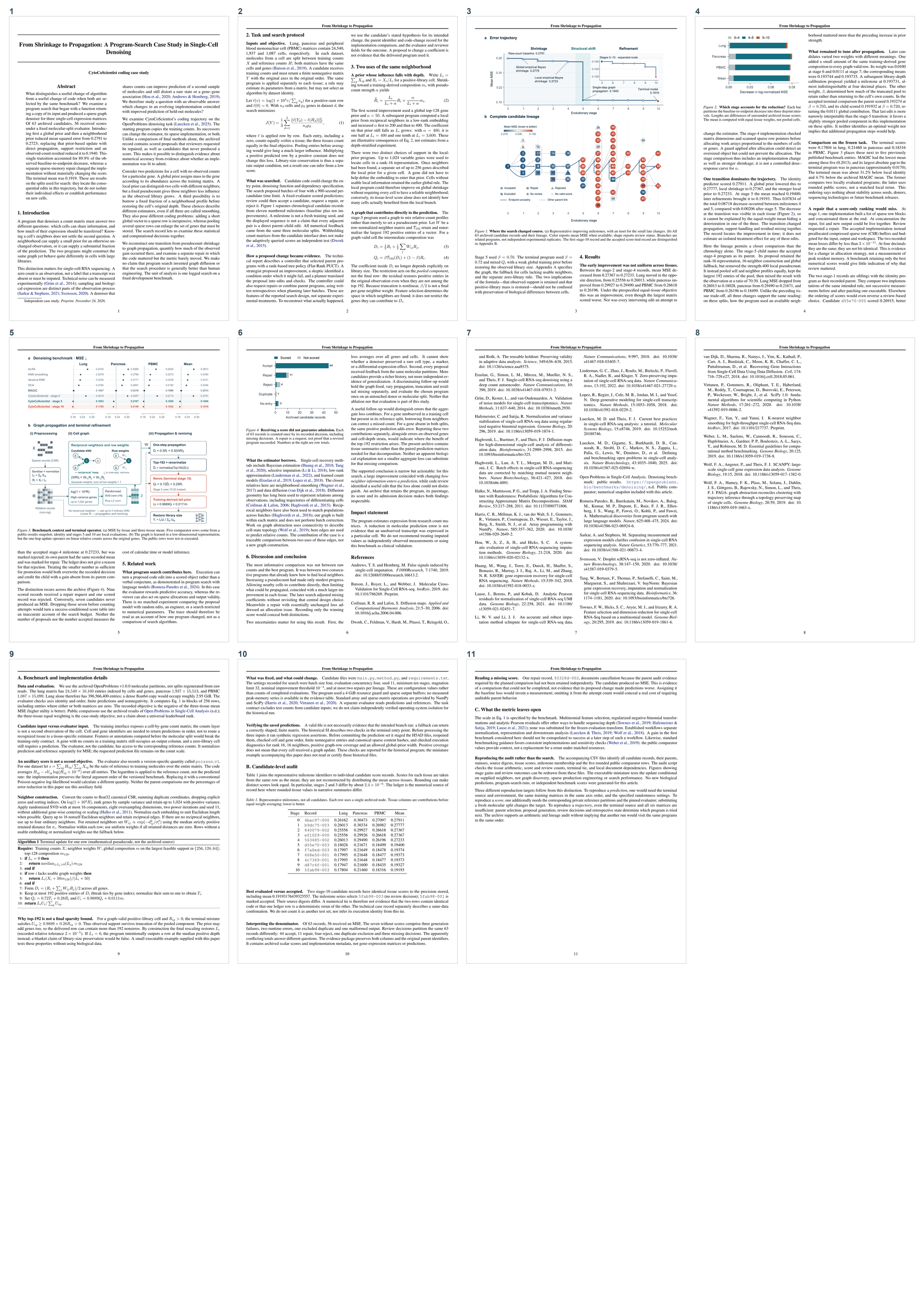}
\caption{\textbf{Representative standalone methods manuscript generated by CytoCoScientist from the OpenProblems single-cell denoising study.} All 11 pages of the complete manuscript, \emph{From Shrinkage to Propagation: A Program-Search Case Study in Single-Cell Denoising}, are shown in reading order from left to right. Pages 1--8 comprise the abstract, introduction, search protocol, operator formulation, results with four figures, related work, discussion, and references; pages 9--11 contain the companion appendices (implementation details with Algorithm~1, candidate-level audit table, and metric scope).}
\label{fig:si-writing-coding-manuscript}
\end{figure}

SI Figure~\ref{fig:si-writing-coding-manuscript} further illustrates a standalone methods manuscript generated by the same writing workflow for the OpenProblems single-cell denoising study (SI Section~\ref{sec:si-case-denoising}). The manuscript describes the transition from neighborhood shrinkage to direct graph propagation alongside benchmark results and candidate-lineage analyses. Its appendix provides the final operator algorithm, candidate-level audit, and evaluation protocol.

\FloatBarrier
\subsection{Implementation and Model Setup}\label{sec:si-implementation}

All five studies ran under fixed model parameters within a single model configuration. The central main agent, the outer-loop meta-reviewer, and every LLM-based review and gating role, including R3 domain review, completion gating, and the Elo comparison judge, were instantiated with GPT-6.1 Sol; generation and analysis subagents used GPT-6.1 Terra and GPT-6.1 Luna (Table~\ref{tab:si-model-stack}). All model calls used the providers' standard API interfaces with default decoding settings, and no model weights changed within a study.
Literature and passage retrieval combined dense embeddings from Qwen3-Embedding-8B with sparse lexical representations (fastembed, Qdrant/bm25) through reciprocal rank fusion, matching the hybrid retrieval described in Section~\ref{sec:si-literature-review}.

\begin{table}[!htbp]
\centering\small
\renewcommand{\arraystretch}{1.12}
\caption{\textbf{Model roles and retrieval backends of the fixed computational stack.} All LLM-based review and gating roles run on GPT-6.1 Sol, whereas generation and analysis subagents use GPT-6.1 Terra and GPT-6.1 Luna. No model weights changed within any study.}\label{tab:si-model-stack}
\begin{tabularx}{\linewidth}{@{}>{\raggedright\arraybackslash}p{0.30\linewidth}>{\raggedright\arraybackslash}p{0.24\linewidth}>{\raggedright\arraybackslash}X@{}}
\toprule
\textbf{Role} & \textbf{Model} & \textbf{Usage} \\
\midrule
Main agent; outer-loop meta-review; all LLM review and gating roles (R3 domain review, completion gating, Elo comparison) & GPT-6.1 Sol & Activity selection, candidate revision, trajectory meta-review, and admissibility or ranking judgments \\
\addlinespace[0.5em]
Generation and analysis subagents & GPT-6.1 Terra; GPT-6.1 Luna & Isolated task contexts returning distilled summaries to the main agent \\
\addlinespace[0.5em]
Dense retrieval embeddings & Qwen3-Embedding-8B & Literature and passage embeddings for hybrid retrieval \\
\addlinespace[0.5em]
Sparse retrieval & fastembed (Qdrant/bm25) & Lexical matching fused with dense retrieval via reciprocal rank fusion \\
\bottomrule
\end{tabularx}
\end{table}

\FloatBarrier
\section{Single-Cell Expression Denoising via Graph Operator Evolution}\label{sec:si-case-denoising}

\subsection{Task Prompt and Data Rules}\label{sec:si-denoising-task}\label{sec:si-case-denoising-input}

\begin{caseprompt}{Goal.md | Single-Cell RNA-seq Denoising}
\textbf{Goal.} Synthesize a general-purpose single-cell RNA-seq denoising pipeline across the frozen molecular-split datasets from lung, pancreas, and peripheral blood mononuclear cells (PBMC) to minimize equally weighted mean squared error ($\overline{\mathrm{MSE}}$) across the three tissues. Starting without pre-supplied denoising code, the system may evolve the complete representation, neighborhood graph, count-propagation operator, regularization, and hyperparameters.

\textbf{Inputs and data rules.} The pipeline receives three training H5AD count matrices (cells by genes, non-negative raw molecular counts with exact cell and gene identifiers). Candidate methods must use only training counts and axis identities; held-out molecular test files remain sealed inside the frozen evaluator, and dataset identifiers may not be used to select tissue-specific rules or hyperparameters.

\textbf{Outputs and execution.} Output one denoised H5AD matrix per tissue with finite, non-negative estimates on the original training-count scale, preserving exact dimensions and cell/gene ordering under deterministic seed~11 with bounded sparse memory usage.
\end{caseprompt}

\subsection{Benchmark Dataset, Metrics, and Baselines}\label{sec:si-denoising-evaluation}
The OpenProblems Denoising v1.0.0 benchmark provides three training-count matrices of contrasting scale and cellular heterogeneity: lung ($24{,}540\text{ cells}\times 16{,}160\text{ genes}$), pancreas ($1{,}937\text{ cells}\times 15{,}513\text{ genes}$), and PBMC ($1{,}087\text{ cells}\times 15{,}099\text{ genes}$). We used the official frozen molecular-split files without regenerating splits locally. Under molecular cross-validation, observed UMI molecules within each cell are partitioned binomially between training and held-out matrices sharing identical cell and gene axes~\cite{Batson2019MolecularCrossValidation}. Held-out counts remained sealed inside the evaluator while their numerical MSEs supplied selection feedback during search; reported metrics therefore reflect this fixed development split.

For any nonnegative cell-expression row vector $v$, the frozen evaluator first applies library-size log-normalization:
\begin{equation}
\mathcal{L}(v)_g=\log\!\left(1+\frac{10^{4}v_g}{s(v)}\right),\qquad
s(v)=\begin{cases}\sum_g v_g,&\sum_g v_g>0,\\1,&\text{otherwise}.\end{cases}
\label{eq:si-denoising-normalization}
\end{equation}
For dataset $d\in\{\text{lung},\text{pancreas},\text{PBMC}\}$ with held-out count matrix $H^{(d)}$, predicted matrix $Y^{(d)}$, $n_d$~cells, and $p_d$~genes, the dataset-specific and equally weighted mean MSE are defined as
\begin{equation}
\mathrm{MSE}_d=\frac{1}{n_dp_d}\sum_{i,g}
\left[\mathcal{L}(H^{(d)}_i)_g-\mathcal{L}(Y^{(d)}_i)_g\right]^2,
\qquad
\overline{\mathrm{MSE}}=\frac{1}{3}\sum_d\mathrm{MSE}_d.
\label{eq:si-denoising-mse}
\end{equation}
Every matrix entry (including zeros) contributes to $\mathrm{MSE}_d$, which is evaluated in 256-row blocks to bound evaluator memory overhead. Prior to numerical scoring, the evaluator verifies exact cell/gene axis alignment and finite nonnegativity, and returns $-\overline{\mathrm{MSE}}$ as the candidate search utility.

\begin{table}[!htbp]
\centering
\footnotesize
\setlength{\tabcolsep}{3pt}
\caption{\textbf{Benchmark comparison of CytoCoScientist evolutionary milestones against published OpenProblems denoising baselines across lung, pancreas, and PBMC datasets.} Figure~\ref{fig:exp-denoising-results}a plots these baselines alongside the stage~0 raw-count baseline, stage~5 milestone, and stage~10 terminal model (the endpoint in Figure~\ref{fig:exp-denoising-search}a). Per-tissue MSEs ($\downarrow$) are rounded to four decimal places, with equally weighted mean MSE computed prior to rounding.}
\label{tab:si-denoising-results}
\begin{tabularx}{\linewidth}{@{}>{\raggedright\arraybackslash}p{0.23\linewidth}>{\raggedright\arraybackslash}Xrrrr@{}}
\toprule
\textbf{Provenance} & \textbf{Method} & \textbf{Lung $\downarrow$} & \textbf{Pancreas $\downarrow$} & \textbf{PBMC $\downarrow$} & \textbf{Mean MSE $\downarrow$}\\
\midrule
Published OpenProblems & ALRA & 0.2723 & 0.3085 & 0.2632 & 0.2813\\
Published OpenProblems & KNN smoothing & 0.2279 & 0.2798 & 0.2273 & 0.2450\\
Published OpenProblems & Iterative KNN smoothing & 0.2255 & 0.2771 & 0.2238 & 0.2421\\
Published OpenProblems & DCA & 0.2164 & 0.2691 & 0.2182 & 0.2346\\
Published OpenProblems & MAGIC & 0.1837 & 0.2316 & 0.1886 & 0.2013\\
\midrule
Local frozen evaluator & No-denoising identity baseline & 0.2616 & 0.3047 & 0.2710 & 0.2791\\
Local frozen evaluator & CytoCoScientist, stage 5 (graph propagation) & 0.1803 & 0.2167 & 0.1850 & 0.1940\\
Local frozen evaluator & \textbf{CytoCoScientist, stage 10 (terminal model)} & 0.1780 & 0.2146 & 0.1832 & 0.1919\\
\bottomrule
\end{tabularx}
\end{table}

Table~\ref{tab:si-denoising-results} lists the local frozen-evaluator scores (seed~11) alongside reference scores for ALRA~\cite{Linderman2022ALRA}, KNN smoothing and iterative KNN smoothing~\cite{Wagner2017KNN}, DCA~\cite{Eraslan2019DCA}, and MAGIC~\cite{vanDijk2018MAGIC} from the official OpenProblems leaderboard~\cite{OpenProblemsDenoisingResults}. When the leaderboard lists multiple hyperparameter variants for a baseline, we report the single uniform configuration with the lowest three-tissue mean MSE (\texttt{approximate, reversed normalization} for MAGIC; \texttt{sqrt norm, reversed normalization} for ALRA).

\subsection{Evolutionary Search Trajectory Across 63 Candidates}\label{sec:si-denoising-search}
Starting from the raw-count baseline ($\overline{\mathrm{MSE}}=0.2791$) with no pre-supplied code, CytoCoScientist surveyed single-cell count noise modeling, empirical Bayes shrinkage, and manifold graph diffusion before evolving 63 candidate implementations across stages~0--10 (Figure~\ref{fig:exp-denoising-search}; search settings in Table~\ref{tab:si-denoising-config}). Of the 63 archived nodes, 56 produced valid numerical scores and 7 failed prior to scoring (3 code-generation failures, 2 runtime errors, 1 duplicate exclusion, and 1 malformed output); review logged 44 accepted candidates, 11 targeted repairs, 4 rejections, 1 duplicate exclusion, and 3 records terminated before review. Table~\ref{tab:si-denoising-decisions} audits the representative algorithmic and review-driven transitions.

\begingroup
\footnotesize
\setlength{\tabcolsep}{2.5pt}
\renewcommand{\arraystretch}{1.12}
\begin{longtable}{@{}>{\raggedright\arraybackslash}p{0.11\linewidth}>{\raggedright\arraybackslash}p{0.20\linewidth}>{\raggedright\arraybackslash}p{0.25\linewidth}>{\raggedright\arraybackslash}p{0.19\linewidth}>{\raggedright\arraybackslash}p{0.09\linewidth}>{\raggedright\arraybackslash}p{\dimexpr0.16\linewidth-10\tabcolsep\relax}@{}}
\caption{\textbf{Audit of representative algorithmic and review-driven transitions along the OpenProblems denoising trajectory.} Each entry maps a recorded hypothesis proposal or review directive to its executed code modification, per-tissue and mean MSE ($\downarrow$), and governance status.}
\label{tab:si-denoising-decisions}\\
\toprule
\textbf{Stage \& family} & \textbf{Recorded proposal or review signal} & \textbf{Executed code modification} & \textbf{Lung / Panc / PBMC $\downarrow$} & \textbf{$\overline{\mathrm{MSE}}$ $\downarrow$} & \textbf{Govern\-ance status}\\
\midrule
\endfirsthead
\multicolumn{6}{@{}l}{\small\tablename~\thetable{} (continued)}\\
\toprule
\textbf{Stage \& family} & \textbf{Recorded proposal or review signal} & \textbf{Executed code modification} & \textbf{Lung / Panc / PBMC $\downarrow$} & \textbf{$\overline{\mathrm{MSE}}$ $\downarrow$} & \textbf{Govern\-ance status}\\
\midrule
\endhead

Stage 0\newline Raw-count baseline & Initial zero-code baseline with no denoising applied. & Evaluate canonicalized raw training counts directly. & 0.2616 / 0.3047 / 0.2710 & 0.2791 & Accepted\\
\addlinespace[2pt]
Stage 1\newline Global EB shrinkage & Shrink positive-library cells toward top-128 training-gene prior ($\alpha=50$). & Implement $Y_i = \frac{L_i}{L_i+50}(X_i + 50 m^{(128)})$ with block-list CSR assembly. & 0.2603 / 0.3033 / 0.2697 & 0.2778 & Repair\\
\addlinespace[2pt]
Stage 1\newline Memory-guard repair & Review flagged simultaneous live memory of input, block list, and \texttt{vstack} output. & Preallocate CSR output buffers and include input, output, and workspace in 4~GiB guard. & 0.2603 / 0.3033 / 0.2697 & 0.2778 & Accepted\\
\addlinespace[2pt]
Stage 2\newline Local MNN shrinkage & Construct 16-D SVD on $\log(1+10^4 R)$ HVGs and $k=16$ reciprocal MNN prior ($\alpha=200$). & Compute cell-specific top-256 neighborhood prior and shrink with $\alpha=200$ ($\alpha=50$ fallback). & 0.2564 / 0.2990 / 0.2656 & 0.2737 & Accepted\\
\addlinespace[2pt]
Stage 4\newline Guarded local shrinkage & Increase local shrinkage to $\alpha=400$ with shape-only pre-allocation guards. & Enforce scalar \texttt{indptr} memory checks before axis-length allocations at $\alpha=400$. & 0.2551 / 0.2977 / 0.2642 & 0.2723 & Accepted\\
\addlinespace[2pt]
Stage 5\newline One-step graph propagation & Replace local shrinkage with direct linear relative-count propagation and Top-192 truncation. & Compute $D_i = 0.5 R_i + 0.5 \sum_j W_{ij}R_j$, apply $\widetilde{D}_i=\mathcal{T}_{192}(D_i)$, and remix $0.70\widetilde{D}_i + 0.30 R_i$. & 0.1803 / 0.2167 / 0.1850 & 0.1940 & Accepted\\
\addlinespace[2pt]
Stages 6--9\newline Global prior \& mixture & Inject training-derived global prior ($\epsilon\in[0.01,0.0111]$) and test $0.705$ outer mix. & Blend dynamic top-256/128/64 training prior $m$ into graph rows and raise outer mix to $0.705$. & 0.1795 / 0.2159 / 0.1844 & 0.1933 & Accepted\\
\addlinespace[2pt]
\textbf{Stage 10\newline Terminal refinement} & Increase outer graph-to-observed mixture to $0.720$ while retaining $\epsilon=0.0111$ prior. & Set $Q_i^{(10)}=0.72\widetilde{D}_i + 0.28 R_i$ and $U_i^{(10)}=0.9889 Q_i^{(10)} + 0.0111 m$ with output verification. & 0.1780 / 0.2146 / 0.1832 & 0.1919 & Confirmed$^{\dagger}$\\
\bottomrule
\end{longtable}
\vspace{2pt}
\par\noindent\footnotesize $^{\dagger}$At stage~10, the admitted candidate and its deterministic program confirmation run achieved identical dataset-level and mean MSEs ($0.1919$), supplying the verified terminal source bundle.
\endgroup

\begin{table}[!htbp]
\centering\small
\setlength{\tabcolsep}{3pt}
\renewcommand{\arraystretch}{1.12}
\caption{\textbf{Search runtime settings for the OpenProblems denoising study.}}
\label{tab:si-denoising-config}
\begin{tabularx}{\linewidth}{@{}>{\raggedright\arraybackslash}p{0.42\linewidth}>{\raggedright\arraybackslash}X@{}}
\toprule
\textbf{Setting} & \textbf{Value}\\
\midrule
Candidate batch size / evaluation concurrency & 4 / 4\\
Minimum stages / stagnation limit & 10 / 32 iterations\\
Minimum improvement threshold / repair limit & $0.0001$ / 2 repairs per lineage\\
Per-candidate timeout / random seed & 900~s / seed~11\\
Editable files & \texttt{main.py}, \texttt{method.py}, \texttt{requirements.txt}\\
\bottomrule
\end{tabularx}
\end{table}

\subsection{Final Evolved Denoising Algorithm}\label{sec:si-denoising-method}
The denoising implementation uses \texttt{anndata}~0.10.9, \texttt{numpy}~1.26.4, \texttt{scipy}~1.13.1, \texttt{h5py}~3.12.1, and \texttt{pandas}.

To make the final evolved estimator self-contained without requiring traversal of historical wrapper files, we specify the complete terminal computation executed across \texttt{method.py} and \texttt{main.py}. Both modules enforce strict shape, axis, and nonnegativity invariants without conditioning prediction rules on dataset identifiers.

\subsubsection{Decoupled Representation Learning and Reciprocal MNN Graph}\label{sec:si-denoising-graph}
Each training matrix $X\in\mathbb{R}_{\ge 0}^{n\times p}$ is first sanitized into canonical float32 CSR format by summing duplicate coordinates, stripping explicit zeros, sorting column indices, and verifying finite nonnegativity. For each cell $i$ with positive library size $L_i=\sum_{g=1}^p X_{ig}>0$, let $R_{ig}=X_{ig}/L_i$ denote its relative transcript composition. Neighbor discovery operates in log-normalized space $A_{ig}=\log(1+10^4 R_{ig})$, ranking genes by sample variance across cells and selecting up to $1{,}024$ positive-variance genes. The selected submatrix is projected via randomized singular value decomposition (SVD; up to $16$ components, oversampling $8$, two power iterations, seed~$11$, without additional gene-wise centering or unit-variance scaling)~\cite{Halko2011RandomizedSVD}, followed by row-wise $L_2$ normalization onto the unit hypersphere.

Within this embedding, the algorithm queries up to $16$ non-self nearest neighbors per cell and retains reciprocal mutual nearest neighbors (MNNs) to limit smoothing across distinct cell states; cells with zero reciprocal neighbors fall back to at most $4$ ordinary nearest neighbors. Given retained Euclidean distances $d_{ij}$ and cell-specific adaptive bandwidth $\sigma_i=\mathrm{median}_{j\in\mathcal{N}(i):d_{ij}>0}(d_{ij})$, row-normalized Gaussian affinities are assigned as
\begin{equation}
W_{ij}=\frac{\exp(-d_{ij}^2/\sigma_i^2)}{\sum_{\ell\in\mathcal{N}(i)}\exp(-d_{i\ell}^2/\sigma_i^2)},
\label{eq:si-denoising-weights}
\end{equation}
using uniform weights when all retained distances are zero. Whereas feature selection and $W$ operate in $\log(1+10^4 R)$ space on at most $1{,}024$ genes, expression propagation operates in linear relative-count space $R$ across all $p$ genes.

\subsubsection{Count Propagation, Top-192 Support, and Residual Remixing}\label{sec:si-denoising-propagation}
For every cell $i$ with a valid graph neighborhood, the stage~5 backbone (\texttt{method.py}) computes an equal self-and-neighbor relative-count mixture followed by hard support restriction:
\begin{align}
D_i &= 0.5000 R_i + 0.5000 \sum_{j\in\mathcal{N}(i)} W_{ij} R_j,\label{eq:si-denoising-linear-mixture}\\
\widetilde{D}_i &= \mathcal{T}_{192}(D_i).
\label{eq:si-denoising-propagation}
\end{align}
Here $\mathcal{T}_{192}$ retains only the 192 largest positive entries of $D_i$ (breaking ties deterministically by gene index) and renormalizes them to sum to one, concentrating neighborhood pooling onto dominant transcriptional programs instead of leaking low-level counts across thousands of unexpressed genes. At stage~5, remixing $\widetilde{D}_i$ with the untruncated observed profile $R_i$ at a $70{:}30$ ratio ($Q_i = 0.7000\widetilde{D}_i + 0.3000 R_i$) and rescaling by $L_i$ achieved $\overline{\mathrm{MSE}}=0.1940$.

\subsubsection{Terminal Outer-Mixture Refinement and Global Prior}\label{sec:si-denoising-terminal}
At stage~10 (\texttt{main.py}), the outer graph-to-observed mixture shifts to $\beta=0.7200$ and incorporates a weak global prior ($\epsilon=0.0111$) estimated from the training matrix of each dataset by normalizing total counts over the most abundant genes (selecting the largest memory-feasible support $K\in\{256,128,64\}$). For each graph-valid cell $i$, the terminal prediction is
\begin{align}
Q_i^{(10)} &= 0.7200\widetilde{D}_i + 0.2800 R_i,\label{eq:si-denoising-outer-mixture}\\
U_i^{(10)} &= 0.9889 Q_i^{(10)} + 0.0111 m,\qquad
Y_i^{(10)} = L_i\frac{U_i^{(10)}}{\sum_g U_{ig}^{(10)}}.
\label{eq:si-denoising-terminal}
\end{align}
Because $R_i$ restores all observed nonzero transcripts and $m$ contributes up to $K$ abundant genes, each predicted row $Y_i^{(10)}$ preserves all observed nonzero entries while conserving exact library size $L_i$ for positive-library cells (relative mass tolerance $2\times 10^{-5}$).

\subsubsection{Non-Graph Fallback Rules and Memory Bounds}\label{sec:si-denoising-fallback}
Cells without a valid embedding norm or normalized neighborhood weights bypass Eq.~\eqref{eq:si-denoising-terminal} and use a bounded global empirical Bayes rule based on the normalized top-128 training-count prior $m^{(128)}$:
\begin{equation}
Y_i^{\mathrm{fallback}} = \frac{L_i}{L_i+50}X_i + \frac{50 L_i}{L_i+50}m^{(128)} \quad (L_i>0),\qquad
Y_i^{\mathrm{zero}} = L_{\mathrm{med}} m^{(128)} \quad (L_i=0),
\label{eq:si-denoising-fallback}
\end{equation}
where $L_{\mathrm{med}}$ is the median library size among positive-library training cells. Across datasets spanning $1{,}087$ to $24{,}540$ cells, preallocated CSR output buffers, block-wise neighbor queries, and pre-computed memory bounds keep peak working memory within 4~GiB.

\FloatBarrier
\section{Multi-Omics Survival Prediction in Lung Adenocarcinoma}\label{sec:si-case-luad}

\subsection{Task Prompt and Data Rules}\label{sec:si-luad-task}\label{sec:si-case-luad-input}

\begin{caseprompt}{Goal.md | Multi-Omics Survival Prediction in Lung Adenocarcinoma}
\textbf{Goal.} Synthesize and evolve a multi-omics survival risk prediction pipeline across 420 lung adenocarcinoma (LUAD) patients to maximize Harrell's concordance index (C-index) on the 84-patient evaluation split, integrating RNA sequencing, miRNA expression, and DNA methylation beta values. Starting without pre-supplied survival code, the system may evolve feature screening, normalization, cross-omics integration, survival modeling, and regularization.

\textbf{Inputs and data rules.} The training partition supplies omics matrices, patient manifests, and overall survival endpoints (\texttt{OS.time}, \texttt{OS}) for 336 patients; the evaluation partition supplies omics matrices for 84 patients without survival labels. All imputation, feature screening, scaling, dimensionality reduction, model fitting, and cross-modal calibration parameters must be learned strictly within the 336-patient training partition while evaluation outcomes remain sealed inside the frozen evaluator.

\textbf{Outputs and evaluation.} Output \path{submission.csv} containing finite continuous mortality \texttt{risk\_score} and categorical \texttt{risk\_subtype} assignments for all 84 evaluation patients in exact manifest order under deterministic seed~11. Under Harrell's C-index, pairs with tied observed times are excluded, distinct-time pairs are comparable when the shorter-time patient experienced an observed death (\texttt{OS=1}), and tied predictions receive 0.5 credit.
\end{caseprompt}

\subsection{Dataset, Evaluation Metric, and Baselines}\label{sec:si-luad-protocol}
The TCGA-LUAD multi-omics cohort comprised 420 patients~\cite{TCGA2014LUAD} partitioned 80:20 into 336 training patients and 84 evaluation patients (31 observed mortality events, $1{,}368$ comparable patient pairs, zero patient overlap). The split was generated prior to the study under seed~2127 with joint stratification by event indicator (\texttt{OS}) and observed survival-time quartile (\texttt{OS.time}). During evolution, the 336-patient training set supported model fitting and internal cross-validation while the frozen evaluator scored predictions on the 84-patient partition under seed~11 to guide candidate selection.

For predicted continuous mortality risk scores $R=(R_1,\dots,R_{n_{\mathrm{eval}}})$ across patients with observed times $t_i\ge 0$ and event indicators $e_i\in\{0,1\}$, both the frozen evaluator and external baseline scorer applied Harrell's pairwise concordance rule~\cite{Harrell1982}. The set of comparable pairs $\mathcal{P}$ consists of all unordered pairs $(i,j)$ with strictly unequal observed times in which the shorter-time patient experienced an uncensored death:
\begin{equation}
\mathcal{P} = \bigl\{(i,j) : (t_i < t_j \text{ and } e_i = 1) \;\text{or}\; (t_j < t_i \text{ and } e_j = 1)\bigr\}.
\label{eq:si-luad-pairs}
\end{equation}
Ordering each pair $(i,j)\in\mathcal{P}$ such that $t_{\mathrm{early}} < t_{\mathrm{late}}$ and $e_{\mathrm{early}}=1$, Harrell's C-index is
\begin{equation}
\mathrm{C\text{-}index}(R) = \frac{1}{|\mathcal{P}|}\sum_{(i,j)\in\mathcal{P}}
\Bigl(\mathbb{I}\bigl[R_{\mathrm{early}} > R_{\mathrm{late}}\bigr] + 0.5\,\mathbb{I}\bigl[R_{\mathrm{early}} = R_{\mathrm{late}}\bigr]\Bigr),
\label{eq:si-luad-cindex}
\end{equation}
with $|\mathcal{P}|=1{,}368$ across the 84 evaluation patients. Alongside C-index, which served as the sole search utility, the evaluator logged auxiliary three-group log-rank statistics across the submitted \texttt{risk\_subtype} strata (\texttt{low}, \texttt{intermediate}, \texttt{high}). Deterministic re-execution of the iteration~10 model reproduced $\mathrm{C\text{-}index}=0.6458$ ($\Delta=0.0000$; log-rank $\chi^2=5.7267$, $p=0.0571$).

To establish a matched domain benchmark on the identical 336/84 split and concordance rule, we reproduced DeepProg~\cite{Poirion2021DeepProg} from its official open-source release (commit \texttt{e2c9e3\allowbreak{}c71344\allowbreak{}b1b470\allowbreak{}235371\allowbreak{}c3f0ce\allowbreak{}a8c944\allowbreak{}0b16}) under seed~10045 (10 ensemble iterations, 5 internal cross-validation folds, 3 Gaussian-mixture subtypes, 50 selected features per modality, 100-dimensional projection, 10 training epochs, with training-side rank normalization and correlation reduction). Fitting was restricted to the 336 training patients, and the posterior probability of the highest-risk subtype $P(\text{class } 0 \mid x)$ was oriented toward higher mortality hazard via a univariate Cox slope fitted solely on the training set ($\hat{\beta}_{\mathrm{Cox}}=1.0966$). Evaluated on the same 84 patients, matched DeepProg achieved a C-index of $0.5950$ (log-rank $\chi^2=2.0273$, $p=0.3629$).

\begin{table}[!htbp]
\centering\small
\setlength{\tabcolsep}{3pt}
\renewcommand{\arraystretch}{1.12}
\caption{\textbf{Benchmark comparison of the external DeepProg baseline, the initial iteration~0 candidate, and key CytoCoScientist evolutionary milestones} on the frozen TCGA-LUAD evaluation partition (84 patients, 31 events, 1,368 comparable pairs). Evaluator scores and admissibility-review governance statuses are recorded separately.}\label{tab:si-luad-comparisons}
\begin{tabularx}{\linewidth}{@{}>{\raggedright\arraybackslash}p{0.18\linewidth}>{\raggedright\arraybackslash}X>{\raggedright\arraybackslash}p{0.09\linewidth}>{\raggedright\arraybackslash}p{0.16\linewidth}>{\raggedright\arraybackslash}p{0.15\linewidth}@{}}
\toprule
\textbf{Provenance} & \textbf{Method / candidate variant} & \textbf{C-index $\uparrow$} & \textbf{Log-rank $\chi^2$ ($p$)} & \textbf{Govern\-ance status} \\
\midrule
External baseline (local reproduction) & DeepProg~\cite{Poirion2021DeepProg} (10-iteration ensemble, 3 subtypes, Cox-calibrated risk) & 0.5950 & 2.0273 ($0.3629$) & External reference \\
CytoCoScientist, iteration 0 & Early PCA concatenation + ridge-Cox ($\le 8$ PCs/modality, $\lambda=0.5$) & 0.6382 & --- & Rejected \\
CytoCoScientist, iteration 3 & Modality-specific ridge-Cox + normalized-risk late fusion ($K_m=192, \lambda=0.25$) & 0.5954 & --- & Accepted \\
CytoCoScientist, iteration 7 (branch A) & Modality-specific ridge-Cox + out-of-fold reliability-weighted fusion & 0.6411 & --- & Repair \\
CytoCoScientist, iteration 7 (branch B) & Modality-specific ridge-Cox + training-ECDF equal-weight fusion (pre-repair) & 0.6458 & --- & Repair \\
CytoCoScientist, iteration 10 (terminal) & \textbf{Confirmed modality-specific Breslow ridge-Cox + ECDF equal-weight fusion ($\lambda=2.0$)} & 0.6458 & 5.7267 ($0.0571$) & Confirmed \\
\bottomrule
\end{tabularx}
\end{table}

\subsection{Evolutionary Search Trajectory Across 32 Candidates}\label{sec:si-luad-search}
Starting without pre-supplied modeling code, CytoCoScientist surveyed $L_2$-penalized Cox regression~\cite{Cox1972,Verweij1994}, Breslow's tied-event partial likelihood~\cite{Breslow1974}, and late-fusion multi-omics survival modeling~\cite{Poirion2021DeepProg}, then generated 32 candidate attempts across iterations~0--10 (1 candidate per iteration during iterations~0--3 and 4 parallel candidates per iteration during iterations~4--10; search settings in Table~\ref{tab:si-luad-config}). Across all 32 attempts, review accepted 3 candidates, routed 21 to targeted revision, and rejected 8 outright (Figure~\ref{fig:exp-luad-search}). Table~\ref{tab:si-luad-decisions} audits the primary algorithmic transitions and review-driven repairs.

\begingroup
\small
\setlength{\tabcolsep}{3pt}
\renewcommand{\arraystretch}{1.12}
\begin{longtable}{@{}>{\raggedright\arraybackslash}p{0.14\linewidth}>{\raggedright\arraybackslash}p{0.25\linewidth}>{\raggedright\arraybackslash}p{0.30\linewidth}>{\raggedright\arraybackslash}p{0.09\linewidth}>{\raggedright\arraybackslash}p{\dimexpr0.22\linewidth-8\tabcolsep\relax}@{}}
\caption{\textbf{Audit of representative algorithmic transitions and review-driven revisions along the TCGA-LUAD evolutionary trajectory.} Each row maps an archived hypothesis proposal or review signal to the executed code modification, evaluation C-index ($\uparrow$), and governance status.}\label{tab:si-luad-decisions}\\
\toprule
\textbf{Iteration \& regime} & \textbf{Hypothesis proposal or review signal} & \textbf{Executed code modification} & \textbf{C-index $\uparrow$} & \textbf{Govern\-ance status} \\
\midrule
\endfirsthead
\multicolumn{5}{@{}l}{\small\tablename~\thetable{} (continued)}\\
\toprule
\textbf{Iteration \& regime} & \textbf{Hypothesis proposal or review signal} & \textbf{Executed code modification} & \textbf{C-index $\uparrow$} & \textbf{Govern\-ance status} \\
\midrule
\endhead

Iteration 0: Initial PCA--Cox candidate & Synthesize initial early-concatenation baseline and conduct R3 structural audit of cross-omics PCA concatenation. & Median-impute each assay, retain $1{,}500$ RNA / $300$ miRNA / $1{,}500$ meth high-variance features, extract $\le 8$ PCs per assay, and fit a concatenated Cox model ($\lambda=0.5$). & 0.6382 & Rejected \\
Iterations 1--2: Early restructuring & Explore modality-decoupled representations and resolve initial interface/scaling defects. & Replace cross-omics PCA concatenation with separate per-modality survival estimators and refine input schema handling. & --- & Rejected / Repair \\
Iteration 3: First admitted late-fusion parent & Decouple modality fitting via independent ridge-Cox models and normalized-risk late fusion. & Fit three separate ridge-Cox models ($K_m=192, \lambda=0.25$) on training-imputed assays and average normalized modality risk scores. & 0.5954 & Accepted \\
Iterations 4--6: Survival screening \& CV capacity & Introduce survival-aware correlation screening and training-side 3-fold CV over $K_m\in\{96,192\}$ and $\lambda\in\{0.1,0.5,2.0\}$. & Rank features by $|r(x_g, z)|$ with $z_i=e_i-0.35\,\mathrm{zscore}(\log(1+t_i))$; apply median/IQR scaling ($[-12,12]$) and training-ECDF rank mapping. & --- & Repair / Rejected \\
Iteration 7 (branch A): Reliability-weighted fusion & Weight modality ECDF risks by out-of-fold validation C-index reliability blended with equal weights. & Estimate non-negative out-of-fold modality reliability weights on the 336 training patients and form weighted average of ECDF risks. & 0.6411 & Repair \\
Iteration 7 (branch B): Equal-weight ECDF fusion & Combine modality-specific ECDF percentile risks with exact equal weights ($1/3$ each) to prevent weight overfitting. & Retain 3-fold CV feature-cap selection ($96, 96, 192$) and average training-ECDF percentile risks equally across RNA, miRNA, and methylation. & 0.6458 & Repair \\
Iterations 8--9: Penalty locking \& I/O hardening & R3 repair directives requiring fixed $\lambda=2.0$, output path guards, and tied-time invariance tests. & Lock $\lambda=2.0$ after verifying unanimous training selection; add input/output file-alias guards and tied-time permutation self-tests. & --- & Repair / Rejected \\
\textbf{Iteration 10: Champion admission \& confirmation} & Final R0--R3 verification of repaired equal-weight pipeline and deterministic same-data confirmation. & Commit self-auditing \texttt{main.py} ($\lambda=2.0$, CV caps $96/96/192$, ECDF equal-weight fusion) and re-verify under seed~11. & 0.6458 & Accepted / Confirmed \\
\bottomrule
\end{longtable}
\endgroup

\begin{table}[!htbp]
\centering\small
\setlength{\tabcolsep}{3pt}
\renewcommand{\arraystretch}{1.12}
\caption{\textbf{Search runtime settings for the TCGA-LUAD multi-omics survival study.}}\label{tab:si-luad-config}
\begin{tabularx}{\linewidth}{@{}>{\raggedright\arraybackslash}p{0.44\linewidth}>{\raggedright\arraybackslash}X@{}}
\toprule
\textbf{Setting} & \textbf{Value} \\
\midrule
Candidate batch size (iterations 0--3 / 4--10) & 1 / 4 \\
Minimum iterations / stagnation threshold & 10 / 32 iterations \\
Minimum improvement threshold & $0.0001$ \\
Outer split / evaluator / DeepProg seeds & $2127$ / $11$ / $10045$ \\
Editable files & \texttt{main.py}, \texttt{train\_and\_predict.py}, \texttt{requirements.txt} \\
\bottomrule
\end{tabularx}
\end{table}

\subsection{Final Evolved Multi-Omics Survival Model}\label{sec:si-luad-model}
The survival pipeline uses \texttt{numpy}, \texttt{pandas}, and \texttt{scipy}.

\subsubsection{Manifest Alignment and Feature Mapping}\label{sec:si-luad-alignment}
For each modality $m\in\{\mathrm{RNA},\mathrm{miRNA},\mathrm{meth}\}$, input tables are indexed by canonicalized patient identifiers and aligned to the ordered training manifest ($n_{\mathrm{train}}=336$) and evaluation manifest ($n_{\mathrm{eval}}=84$). Training survival endpoints $(t_i, e_i)$ are validated to enforce finite $t_i\ge 0$, binary $e_i\in\{0,1\}$, and at least 10 aligned training patients. On the 336-patient training matrix of modality $m$, features with zero finite observations are removed, missing or non-finite entries are imputed using the training-set feature median $m_{mg}^{\mathrm{imp}}$, and features whose post-imputation sample variance satisfies $\operatorname{Var}_{\mathrm{train}}(x_{mg})\le 10^{-12}$ are discarded. The retained feature list $\mathcal{F}_m$ and training medians $\{m_{mg}^{\mathrm{imp}}\}_{g\in\mathcal{F}_m}$ define the frozen modality schema. When constructing the evaluation matrix for modality $m$, the pipeline verifies that at least $50\%$ of $\mathcal{F}_m$ is present in the evaluation table (\texttt{MIN\_FEATURE\_OVERLAP} $=0.50$); any missing schema columns or non-finite cells in the evaluation matrix are filled strictly with the training medians $m_{mg}^{\mathrm{imp}}$. Because schema filtering and median imputation are computed once across the full 336-patient training partition before internal cross-validation, evaluation data remain strictly isolated, while internal three-fold cross-validation refits the subsequent supervised screening, robust scaling, and Cox optimization inside each fold.

\subsubsection{Survival-Aware Screening and Feature-Cap Selection}\label{sec:si-luad-screening}
To screen high-dimensional molecular features on a fitting subset of $n_{\mathrm{fit}}$ training patients ($n_{\mathrm{fit}}\approx 224$ inside a cross-validation fold; $n_{\mathrm{fit}}=336$ for final refitting) without leaking evaluation outcomes, the pipeline constructs a composite survival-aware target $z_i$ combining event status and log-transformed follow-up time:
\begin{equation}
z_i = e_i - 0.35\,\frac{\log(1+t_i)-\overline{\log(1+t)}}{\operatorname{sd}\{\log(1+t)\}+\epsilon},
\qquad \epsilon=10^{-12},
\label{eq:si-luad-screening}
\end{equation}
which assigns larger positive values to early uncensored deaths and smaller values to long-term censored or surviving patients. Centering each feature column $\tilde{x}_{ig}=x_{ig}-\bar{x}_g$ and the target $\tilde{z}_i=z_i-\bar{z}$ on the fitting set, each feature $g\in\mathcal{F}_m$ is scored by its regularized absolute Pearson correlation, and the top $K_m^{\mathrm{eff}}$ features are retained via stable mergesort:
\begin{equation}
s_{mg} = \frac{\bigl|\sum_{i=1}^{n_{\mathrm{fit}}}\tilde{x}_{ig}\tilde{z}_i\bigr|}{\sqrt{\sum_{i=1}^{n_{\mathrm{fit}}}\tilde{x}_{ig}^2 \sum_{i=1}^{n_{\mathrm{fit}}}\tilde{z}_i^2}+\epsilon},
\qquad
K_m^{\mathrm{eff}} = \min\bigl\{K_m,\,|\mathcal{F}_m|,\,\max(12,\,2n_{\mathrm{fit}})\bigr\}.
\label{eq:si-luad-cap}
\end{equation}
Candidate feature caps $K_m\in\{96, 192\}$ are compared using deterministic 3-fold cross-validation on the 336-patient training set: indices of censored ($e_i=0$) and uncensored ($e_i=1$) training patients in manifest order are interleaved round-robin ($\mathrm{position} \bmod 3$) into three folds. Within each fold, screening, robust scaling, and ridge-Cox fitting ($\lambda=2.0$) are refitted on the fold training split and scored by Harrell's C-index (Eq.~\eqref{eq:si-luad-cindex}) on the fold validation split, selecting the cap maximizing mean validation C-index (breaking ties toward $96$). On TCGA-LUAD, this selected $K_{\mathrm{RNA}}=96$, $K_{\mathrm{miRNA}}=96$, and $K_{\mathrm{meth}}=192$.

\subsubsection{Outlier-Clipped Scaling and Grouped Ridge-Cox Optimization}\label{sec:si-luad-cox}
Each selected feature $g$ is centered on the fitting set by its median $c_{mg}=\operatorname{median}(x_{\cdot g})$, scaled by its interquartile range $\mathrm{IQR}_{mg} = P_{75}(x_{\cdot g}) - P_{25}(x_{\cdot g})$ (falling back to training standard deviation $\sigma_{mg}$ or $1.0$), and winsorized to $[-12, 12]$:
\begin{equation}
\widehat{x}_{ig} = \operatorname{clip}\!\left(\frac{x_{ig}-c_{mg}}{s_{mg}},\, -12.0,\, 12.0\right),
\qquad
s_{mg} = \begin{cases}
\mathrm{IQR}_{mg}, & \mathrm{IQR}_{mg} > 10^{-8},\\
\sigma_{mg}, & \mathrm{IQR}_{mg} \le 10^{-8} \text{ and } \sigma_{mg} > 10^{-8},\\
1.0, & \text{otherwise}.
\end{cases}
\label{eq:si-luad-scaling}
\end{equation}
For each modality $m$, an independent $L_2$-penalized Cox proportional hazards model~\cite{Cox1972,Verweij1994} under grouped Breslow tie handling~\cite{Breslow1974} minimizes the negative average partial log-likelihood over observed training events $N_e = \max\bigl(1, \sum_{i=1}^{n_{\mathrm{fit}}} e_i\bigr)$ with ridge penalty $\lambda = 2.0$. Letting $\tau_1 > \dots > \tau_G$ denote distinct observed times, $D_\tau = \{i : t_i = \tau, e_i = 1\}$ with multiplicity $d_\tau = |D_\tau|$, $\mathcal{R}_\tau = \{j : t_j \ge \tau\}$, and stabilized linear predictors $\eta_{im} = \operatorname{clip}(\widehat{x}_{im}^\top\beta_m, -40.0, 40.0)$, the objective is
\begin{equation}
\mathcal{J}_m(\beta_m) = \frac{1}{N_e}\sum_{\tau : d_\tau > 0}
\left[d_\tau\log\!\left(\sum_{j\in\mathcal{R}_\tau}\exp(\eta_{jm})\right) - \sum_{i\in D_\tau}\eta_{im}\right]
+ \frac{\lambda}{2}\lVert\beta_m\rVert_2^2,
\qquad \lambda = 2.0.
\label{eq:si-luad-cox}
\end{equation}
Equation~\eqref{eq:si-luad-cox} is minimized from $\beta_m^{(0)}=\mathbf{0}$ using analytical gradients via L-BFGS-B with box bounds $\beta_{mg}\in[-8.0, 8.0]$, after which continuous modality risk scores are evaluated as $r_m(x) = \widehat{x}_m^\top\hat{\beta}_m$.

\subsubsection{Percentile Alignment, Late Fusion, and Risk Subtypes}\label{sec:si-luad-fusion}
To align incompatible modality variances and tails before fusion, each patient's linear predictor $r_m(x)$ is mapped onto the continuity-corrected empirical cumulative distribution function (ECDF) of the 336 training scores $\{r_m(x_i)\}_{i=1}^{n_{\mathrm{train}}}$ and averaged equally across the three assays:
\begin{equation}
u_m(x)=\frac{\#\bigl\{i\in\text{train}:r_m(x_i)\leq r_m(x)\bigr\}+0.5}{n_{\mathrm{train}}+1}\in(0,1),
\qquad
R(x)=\frac{u_{\mathrm{RNA}}(x)+u_{\mathrm{miRNA}}(x)+u_{\mathrm{meth}}(x)}{3}.
\label{eq:si-luad-fusion}
\end{equation}
For the required categorical \texttt{risk\_subtype} column in \path{submission.csv}, evaluation patients are partitioned into tertiles by their average-tie rank fraction within the 84-patient cohort:
\begin{equation}
a_i = \frac{\operatorname{rank}_{\mathrm{eval}}(R_i)}{n_{\mathrm{eval}}+1},
\qquad
\operatorname{subtype}_i = \begin{cases}
\texttt{low}, & a_i \le 1/3,\\
\texttt{intermediate}, & 1/3 < a_i \le 2/3,\\
\texttt{high}, & a_i > 2/3.
\end{cases}
\label{eq:si-luad-subtype}
\end{equation}

\FloatBarrier
\newlength{\sepsisTableWidth}
\section{Mortality-Associated Molecular Shifts in Sepsis}\label{sec:si-case-sepsis}

This supplement supports Section~\ref{sec:exp:sepsis} and Figures~\ref{fig:sepsis-dag}--\ref{fig:sepsis-round2}. It details the research goal, data provenance, and candidate selection rules (SI Section~\ref{sec:si-sepsis-s1}), Round 1 single-blood-draw hypotheses and analyses (SI Section~\ref{sec:si-sepsis-s2}), Round 2 within-patient day-1 to day-7 analyses, external cohort audits, and the next-step validation protocol (SI Section~\ref{sec:si-sepsis-s3}), followed by complete supplementary tables covering all 54 candidate hypotheses, eight initial analyses, two cross-cohort follow-up analyses, two external data-eligibility audits, and sensitivity checks (SI Tables~\ref{tab:si-sepsis-candidates}--\ref{tab:si-sepsis-sensitivity}).

\subsection{Research Goal, Data Sources, and Selection Rules}\label{sec:si-sepsis-s1}

\subsubsection{Research Goal and Testability Requirements}\label{sec:si-sepsis-s1-1}

\begin{caseprompt}{Goal.md | Mortality-Associated Molecular Patterns in Sepsis}
Focusing on mortality risk in sepsis patients, combine literature evidence, reference datasets, and accessible transcriptomic, proteomic, metabolomic, and other omics measurements to propose candidate hypotheses testable in patient data. The core questions are: which measurable molecular patterns are associated with differences in patient mortality risk, and which observations align better with the proposed biological explanation rather than merely reflecting illness severity, blood-cell composition, organ clearance capacity, or treatment background? Do not restrict infection source, site, or pathogen; do not pre-specify a dominant mechanism; and do not treat predictive gain from a new biomarker or panel as the sole criterion of value.

Every hypothesis proposed for numerical testing must state at least one discriminative prediction whose outcome can shift judgment between the primary explanation and competing explanations. CytoCoScientist should locate appropriate measurements for each question by proactively retrieving independent sepsis studies and public datasets, verifying target-molecule coverage, patient and death counts, sampling and outcome time points, necessary covariates, access conditions, and cohort overlap before deciding which predictions are testable and which can only be answered partially.
\end{caseprompt}

\subsubsection{Data Provenance, Analysis Samples, and Outcome Windows}\label{sec:si-sepsis-s1-2}

Starting with the initially supplied CMAISE dataset, CytoCoScientist independently identified additional studies and data sources and assessed whether they supported the proposed analyses. The tables below distinguish data available for reanalysis from sources assessed only for eligibility. SI Table~\ref{tab:si-sepsis-sources} summarizes the cohorts used for patient-level quantitative analyses, their effective sample sizes, outcome windows, and analytical scope.

\begingroup
\footnotesize
\setlength{\tabcolsep}{3.5pt}
\renewcommand{\arraystretch}{1.15}
\setlength{\LTleft}{0pt}
\setlength{\LTright}{\fill}
\setlength{\LTcapwidth}{\linewidth}
\captionsetup{singlelinecheck=false,justification=raggedright}
\setlength{\sepsisTableWidth}{\dimexpr\linewidth-6\tabcolsep\relax}
\begin{longtable}{@{}>{\raggedright\arraybackslash}p{0.21\sepsisTableWidth}>{\raggedright\arraybackslash}p{0.17\sepsisTableWidth}>{\raggedright\arraybackslash}p{0.28\sepsisTableWidth}>{\raggedright\arraybackslash}p{0.34\sepsisTableWidth}@{}}
\caption{\textbf{Data provenance, effective patient denominators, and analytical boundaries in the sepsis case.}}\label{tab:si-sepsis-sources}\\
\toprule
\textbf{Data source} & \textbf{Provenance} & \textbf{Retrieved material and sample size} & \textbf{Question addressed and boundaries} \\
\midrule
\endfirsthead
\multicolumn{4}{@{}l}{\footnotesize\textbf{Table~\thetable\ (continued)}}\\
\toprule
\textbf{Data source} & \textbf{Provenance} & \textbf{Retrieved material and sample size} & \textbf{Question addressed and boundaries} \\
\midrule
\endhead
\midrule
\multicolumn{4}{r@{}}{\footnotesize\emph{Continued on next page}}\\
\endfoot
\bottomrule
\endlastfoot
\textbf{CMAISE} (PRJCA006118) & \textbf{Initially supplied dataset (restricted access)} & Plasma proteomics, clinical records, and whole-blood RNA. Primary protein branch: $n = 342$ (44 in-hospital deaths); clinical-albumin auxiliary branch: $n = 338$ (43 deaths). & H08, H13: cross-cohort protein comparisons and auxiliary RNA interpretation. The DIA proteomics matrix lacks usable albumin; clinical albumin is a separate assay. Endpoint is in-hospital mortality (distinct from fixed-day mortality). \\
Palmowski (PXD055932; \cite{Palmowski2025SepsisProteome}) & Autonomously retrieved by CytoCoScientist & First-draw patient-level plasma proteomics: $n = 334$ (130 30-day deaths). Complete target subsets: H08 ($n = 326$), H13 ($n = 304$), H09 ($n = 310$). & Compares target and reference proteins within the same blood draw and tests associations with 30-day mortality; all three subsets derive from the same cohort. \\
RACE/CaPS (DOI: 10.1111/cts.13088) & Autonomously retrieved by CytoCoScientist & Pretreatment serum acylcarnitines for 236 patients; randomized dose comparison covers placebo ($n = 70$) vs.\ 18~g L-carnitine ($n = 100$). & H12: separates low free carnitine from elevated acetylcarnitine. Comparing treatment arms requires verified molecular-background overlap and certified 90-day follow-up. \\
Mi GAinS (PXD039875; \cite{Mi2024GAinSProteome}) & Autonomously retrieved by CytoCoScientist & Plasma proteomics; $n = 495$ with complete 5 targets (H08) and $n = 302$ with complete 10 targets (H13). & Checks APOA1--S100 co-occurrence and conditional S100 contrasts across proteomic backgrounds; retrieved files lacked per-patient mortality linkage. \\
GEO whole-blood RNA (GSE95233, GSE54514, GSE272769) & Autonomously retrieved by CytoCoScientist & Whole-blood expression data from the three studies were analyzed separately and kept distinct from the initially supplied CMAISE RNA data. & H13: checks transcriptional \textit{S100A8} and \textit{S100A9} contrasts (whole-blood RNA abundance only). \\
Kwok whole-blood scRNA-seq (Zenodo 7924238) & Autonomously retrieved by CytoCoScientist & Single-cell RNA counts and cell-type annotations across $272{,}993$ peripheral blood cells. & H13: profiles captured \textit{S100A8} and \textit{S100A9} transcripts by peripheral blood cell type. \\
Ferrario/ALBIOS (\cite{Ferrario2016ALBIOSMetabolomics}) & Autonomously retrieved Supplementary Tables S1 and S2 of the Ferrario study & 20 severe septic-shock patients selected from 997 ALBIOS biobank participants; paired day-1 and day-7 plasma concentrations linked by patient ID to 28-day (9 deaths) and 90-day (11 deaths) outcomes. & Shared across H24, H27, H26, and H23 to test within-patient lipid and amino acid shifts. Blood draws are anchored to shock diagnosis and mortality follow-up to enrollment; restricted to patients surviving to day 7. \\
\end{longtable}
\endgroup

\subsubsection{Candidate Ranking and Analysis Selection}\label{sec:si-sepsis-s1-3}

Within each round, CytoCoScientist used an LLM to perform pairwise comparisons across candidate hypotheses along four criteria: alignment with the research goal, scientific quality, mechanistic specificity, and novelty alongside testability. For each pair, the reviewer returned a win, loss, or tie judgment with justification, which CytoCoScientist used to update Elo ratings and select the top four candidates per round. Elo ratings reflect within-round relative preference for testing priority.

Only after selection did CytoCoScientist test predictions against patient data, first verifying whether the required contrast was estimable and then reporting effect direction, magnitude, and uncertainty, while checking robustness across alternative references or independent cohorts whenever suitable data existed. Non-estimable contrasts and opposite-direction estimates are reported separately.

Hypothesis identifiers (H01--H34) follow the order of first appearance among ranked candidates and remain fixed across Elo ranks; unranked candidates fell outside the final Elo ranking. SI Sections~\ref{sec:si-sepsis-s2} and~\ref{sec:si-sepsis-s3} summarize each selected hypothesis, its discriminative prediction, and its data and endpoint from logs recorded before numerical testing. The summaries retain the target panels, discriminative criteria, and endpoints. The analyses and results follow each summary.

\subsection{Round 1: Within-Blood-Sample Molecular Comparisons}\label{sec:si-sepsis-s2}

Starting from 6 initial hypotheses, CytoCoScientist generated 30 candidates across 4 generations, of which 16 entered final Elo ranking. Because pre-analysis eligibility checks identified unresolved normalization and follow-up documentation gaps across candidates, CytoCoScientist proceeded with exploratory numerical analyses on the top four ranked candidates. Three protein hypotheses started from first-draw plasma samples in PXD055932, and one carnitine hypothesis used pretreatment samples from RACE/CaPS. Each proteomics cohort was analyzed on its published scale ($\log_2$ transformation with plate adjustment for positive intensities in Mi GAinS; published per-protein standardized values without re-logging in CMAISE), and mortality endpoints were never pooled across cohorts.

\subsubsection{H08: Shared Protein Remodeling versus Selective APOA1 Depletion}\label{sec:si-sepsis-s2-1}

\begin{caseprompt}{Hypothesis H08 | Pre-analysis summary (Round 1, rank 1, Elo 1510.8)}
\textbf{Hypothesis.} In early adult sepsis, mortality-associated co-occurrence of low APOA1 with elevated S100A8 and S100A9 arises primarily from shared acute-phase plasma protein remodeling (co-varying with APOB and ALB) rather than selective APOA1 deficiency driving inflammation. Motivated by discordant literature reports (elevated APOA1 in non-survivors in Langley et al., DOI: 10.1126/scitranslmed.3005893; severity-linked lipoprotein declines and S100 increases in Mi et al., DOI: 10.1126/scitranslmed.adh0185), this replaces a selective-deficiency claim with a competing shared-response explanation.

\textbf{Discriminative prediction.} Mortality-associated APOA1 shifts should track primarily with shared APOB and ALB variation rather than an APOA1-specific negative deviation, and coordinated S100A8/A9 elevation should persist among patients whose APOA1 is relatively preserved. Concentration of S100 elevation and mortality risk exclusively within APOA1-specific negative deviation would instead favor an APOA1-selective process.

\textbf{Data and endpoint.} First-draw plasma proteomics from PXD055932 (326 patients with joint coverage of APOA1, APOB, ALB, S100A8, and S100A9) linked to 30-day mortality. Standardized contrasts reflect within-cohort relative position rather than absolute concentration ratios or HDL neutralization function.
\end{caseprompt}

\paragraph{Initial analysis}
Across 326 patients in PXD055932, CytoCoScientist used APOB and albumin as references to decompose APOA1 variation into a component shared with the references and an APOA1-specific relative deviation. Among 173 patients with non-negative relative APOA1 deviation, 63 still exhibited both S100A8 and S100A9 above their respective standardized zeros (defined by sample-standardized cutoffs rather than clinical thresholds). The difference in absolute contribution to the group contrast between shared and specific components was $+0.048$ (95\% CI $-0.124$ to $0.196$), which could not establish that the shared component dominated (SI Figure~\ref{fig:si-sepsis-r1-initial}a).

\paragraph{Cross-cohort follow-up}
In Mi GAinS ($n = 495$), double-high S100 occurred in $128/243$ ($52.7\%$) patients with relatively low APOA1 and in $76/252$ ($30.2\%$) patients with relatively preserved APOA1; the negative conditional association between relative APOA1 deviation and S100 persisted after plate adjustment ($-0.248$, Holm $p = 1.12 \times 10^{-6}$). Double-high S100 is thus more frequent alongside relatively low APOA1 but is not restricted to it. Retrieved Mi GAinS files lacked per-patient mortality linkage, while in CMAISE standardized APOA1 was higher in in-hospital non-survivors ($+0.584$). Neither observation provides independent confirmation of low APOA1 as a general mortality phenotype (SI Figure~\ref{fig:si-sepsis-r1-followup}a).

\subsubsection{H12: Low Free Carnitine versus Acetylcarnitine Accumulation}\label{sec:si-sepsis-s2-2}

\begin{caseprompt}{Hypothesis H12 | Pre-analysis summary (Round 1, rank 2, Elo 1436.1)}
\textbf{Hypothesis.} Among septic-shock patients sampled before L-carnitine therapy, the candidate treatment response linked to elevated acetylcarnitine relative to free carnitine stems primarily from relative deficiency in free carnitine rather than selective acetylcarnitine accumulation (contextualizing exploratory interactions in Puskarich et al., DOI: 10.1111/cts.13088, against renal/hepatic disposal effects in Langley et al., DOI: 10.1126/scitranslmed.3005893, and Chung et al., DOI: 10.1097/ccm.0000000000003517).

\textbf{Discriminative prediction.} At comparable acetylcarnitine--free-carnitine imbalance and matched background across the remaining acylcarnitines, favorable 90-day mortality differences for 18~g L-carnitine versus placebo should concentrate in patients with low free carnitine and preserved acetylcarnitine rather than those with high acetylcarnitine and preserved free carnitine, evaluated jointly on absolute and relative risk scales.

\textbf{Data and endpoint.} Pretreatment serum acylcarnitines from RACE/CaPS ($n = 236$; 18~g L-carnitine vs.\ placebo) linked to 90-day mortality, conditioned on verifying patient-level molecular-background overlap and resolving the discrepancy between 125 reconstructed deaths and 124 published deaths.
\end{caseprompt}

\paragraph{Initial analysis}
RACE/CaPS contained 236 pretreatment profiles, with 70 placebo and 100 18~g patients in the prespecified dose contrast. CytoCoScientist defined candidate patterns from pretreatment free carnitine, acetylcarnitine, and the remaining acylcarnitines, then checked randomized-arm coverage among patients with matched molecular backgrounds. Requiring imbalance and shared-background differences $\le 0.10$ and per-analyte background RMSD $\le 0.20$ yielded only 2 matched pairs, with both free-carnitine-low patients in the 18~g arm (0 placebo controls). Reconstructed follow-up deaths also differed from the published count ($125/236$ vs.\ $124/236$). CytoCoScientist therefore halted at comparability and identifiability checks, leaving the treatment effect non-estimable (SI Figure~\ref{fig:si-sepsis-r1-initial}b).

\subsubsection{H13: S100 Elevation across Combined Complement--Hemostasis Backgrounds}\label{sec:si-sepsis-s2-3}

\begin{caseprompt}{Hypothesis H13 | Pre-analysis summary (Round 1, rank 3, Elo 1379.3)}
\textbf{Hypothesis.} In early adult sepsis, coordinated elevation of S100A8 and S100A9 marks a cross-system complement--hemostasis host response (jointly low C3, C4A, CFB, HRG, PROC, and F13B) rather than every instance of low complement effectors relative to CFH (integrating severity-linked S100/complement patterns in Mi et al., DOI: 10.1126/scitranslmed.adh0185, and Palmowski et al., DOI: 10.1016/j.ebiom.2024.105508, against conflicting complement reports in Langley et al., DOI: 10.1126/scitranslmed.3005893, and de Nooijer et al., DOI: 10.1186/s13054-023-04344-6).

\textbf{Discriminative prediction.} At comparable deviation of C3, C4A, and CFB relative to CFH, coordinated S100A8 and S100A9 elevation should be stronger when complement effectors and hemostasis proteins (HRG, PROC, F13B) are jointly low, and weaker when CFH is uniquely elevated with all six proteins preserved; secondary checks test whether 30-day mortality risk also concentrates in the cross-system background.

\textbf{Data and endpoint.} First-draw plasma proteomics from PXD055932 across the 10-protein panel (S100A8, S100A9, C3, C4A, CFB, CFH, HRG, PROC, F13B, ALB) linked to 30-day mortality, with cross-cohort replication in Mi GAinS and CMAISE.
\end{caseprompt}

\paragraph{Initial analysis}
CytoCoScientist contrasted the two model-specified proteomic backgrounds and verified whether actual patients populated their neighborhoods. Across 304 complete-target patients in PXD055932, conditional S100A8 and S100A9 differences between the two backgrounds were $+0.376$ and $+0.272$ standardized units (Holm-adjusted $p = 0.00077$ and $0.0114$), yet the cross-system neighborhood was sparse ($5/304$) and the background mortality odds-ratio ratio was $1.035$ (95\% CI $0.632$ to $1.695$), indicating no mortality-risk separation (SI Figure~\ref{fig:si-sepsis-r1-initial}c).

\begin{figure}[p]
\centering
\includegraphics[width=\linewidth]{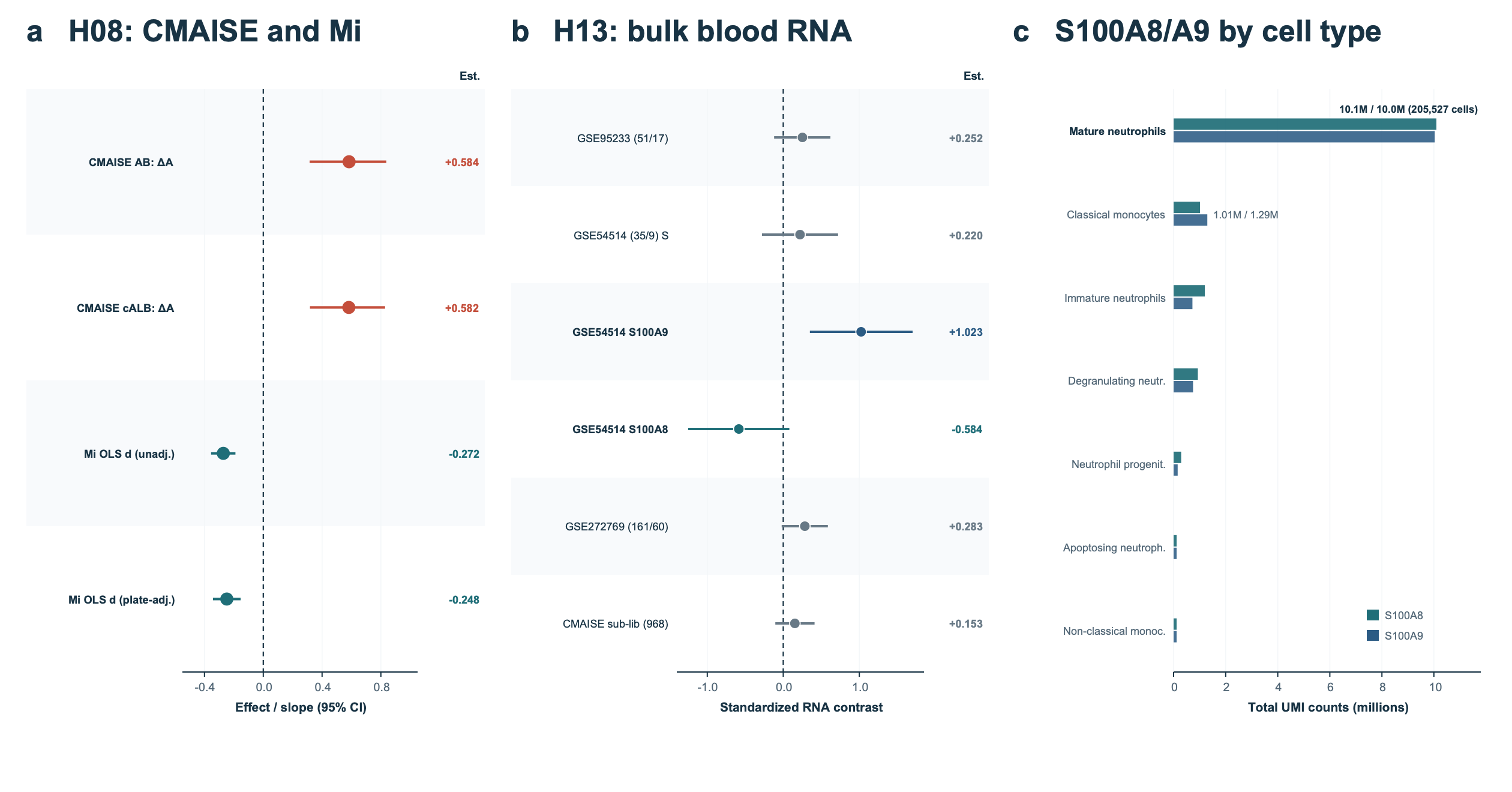}
\caption{\textbf{Round 1 cross-cohort protein comparisons and auxiliary whole-blood and single-cell RNA checks.} \textbf{a,} H08 contrasts in CMAISE ($n = 342 / 338$) and unadjusted versus plate-adjusted OLS slopes in Mi GAinS ($n = 495$). \textbf{b,} Auxiliary whole-blood bulk RNA S100 contrasts across four cohorts. \textbf{c,} Captured \textit{S100A8} and \textit{S100A9} UMI totals across top peripheral blood cell types in the Kwok single-cell atlas ($272{,}993$ cells). Cross-cohort plasma protein contrasts for H13 appear in Figure~\ref{fig:sepsis-round1}b.}
\label{fig:si-sepsis-r1-followup}
\end{figure}

\paragraph{Cross-cohort follow-up and auxiliary RNA checks}
Evaluated in the same direction, Mi GAinS ($n = 302$) yielded positive point estimates whose 95\% intervals both crossed zero ($+0.292$ and $+0.157$), whereas CMAISE reversed to negative ($-0.573$ and $-0.769$ in the primary no-albumin branch; $-0.611$ and $-0.808$ with clinical albumin). CytoCoScientist also checked four whole-blood bulk RNA cohorts and the Kwok single-cell atlas. For the H13 auxiliary RNA analysis, the full CMAISE CPM matrix was excluded because it contained non-finite values, and exploratory results from the fixed CMAISE RNA subset are reported separately. In the Kwok single-cell dataset ($272{,}993$ cells), counts and metadata were not completely aligned cell-for-cell; mature neutrophils ($205{,}527$ cells) contributed the largest share of captured \textit{S100A8}/\textit{S100A9} transcripts, though cellular RNA abundance does not establish the source of circulating plasma protein (Figure~\ref{fig:sepsis-round1}b; SI Figure~\ref{fig:si-sepsis-r1-followup}b,c).

\subsubsection{H09: Reference-Set Decomposition of Complement Factor H}\label{sec:si-sepsis-s2-4}

\begin{caseprompt}{Hypothesis H09 | Pre-analysis summary (Round 1, rank 4, Elo 1371.8)}
\textbf{Hypothesis.} In early adult sepsis, the 30-day mortality-associated pattern of low complement effectors (C3, C4A, CFB) relative to complement factor H (CFH) may be driven by positive deviation of regulatory CFH relative to other circulating proteins rather than coordinated effector depletion, representing a regulator-side restriction mechanism distinct from effector-supply deficiency.

\textbf{Discriminative prediction.} In normalized abundance space, 30-day mortality differences should concentrate in consistent positive deviation of CFH relative to both the three effectors and multiple non-complement comparators (ALB, HRG, PROC, F13B), without coordinated negative deviation of effectors relative to non-complement comparators. Consistent effector depletion across references, or results flipping sign by comparator protein, would weaken this regulator-side explanation.

\textbf{Data and endpoint.} First-draw plasma proteomics from PXD055932 ($n = 310$ patients with complete coverage of C3, C4A, CFB, CFH, ALB, HRG, PROC, and F13B) linked to 30-day mortality; plasma proteomics resolves relative abundance structure rather than intact CFH inhibitory activity.
\end{caseprompt}

\begin{figure}[!htbp]
\centering
\includegraphics[width=\linewidth]{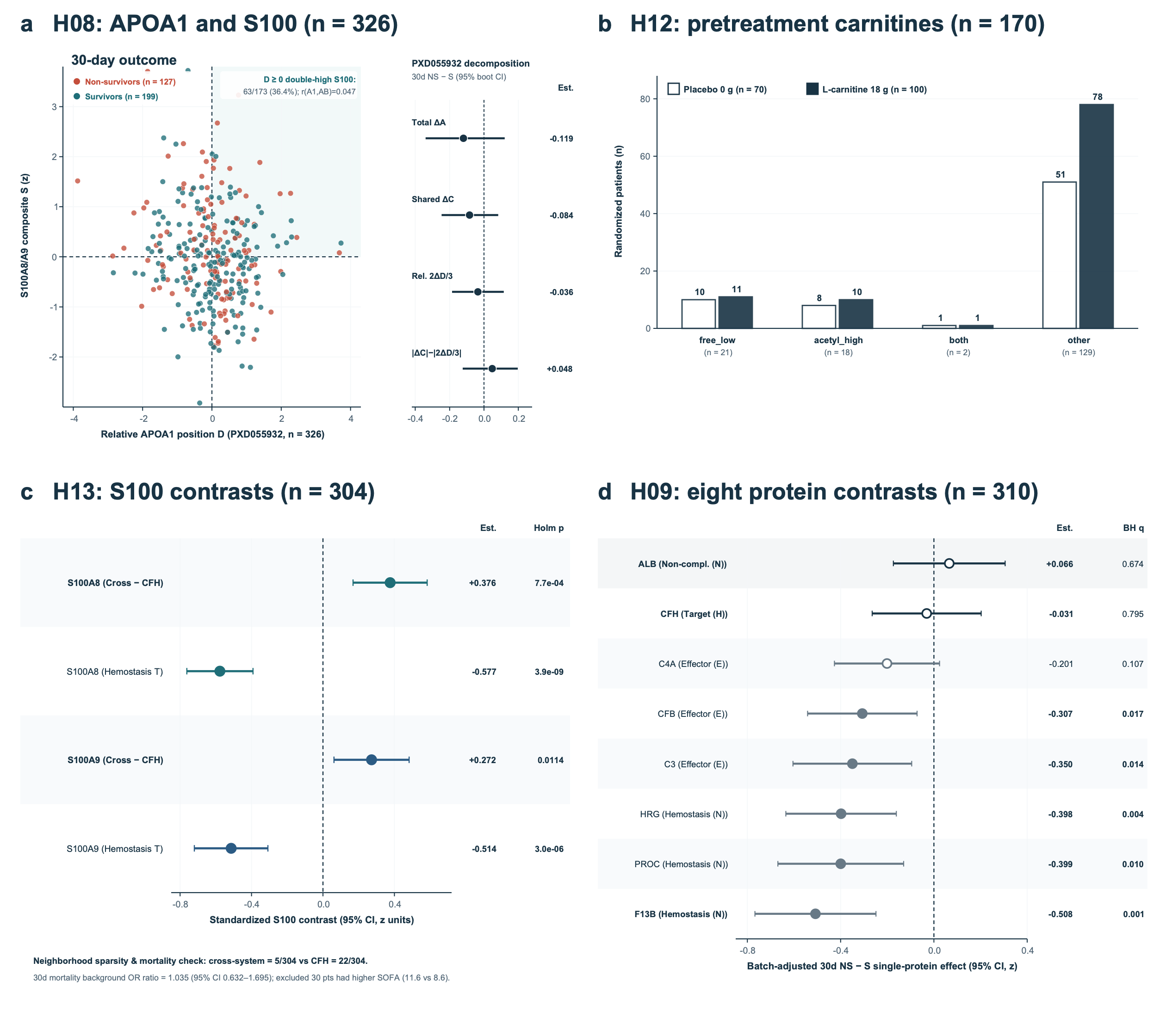}
\caption{\textbf{Initial patient-level analyses of the four Round 1 hypotheses.} \textbf{a,} Relative APOA1 deviation versus S100A8/A9 composite level ($n = 326$, colored by 30-day mortality) and shared-versus-relative decomposition in PXD055932. \textbf{b,} Pretreatment acylcarnitine pattern counts by randomized arm in RACE/CaPS ($n = 170$) and local-overlap identifiability gate. \textbf{c,} Primary conditional S100A8 and S100A9 contrasts in PXD055932 ($n = 304$). \textbf{d,} Single-protein 30-day mortality contrasts across all eight proteins in the H09 panel in PXD055932 ($n = 310$).}
\label{fig:si-sepsis-r1-initial}
\end{figure}

\paragraph{Initial analysis}
Across 310 patients in PXD055932, prespecified CFH contrasts against the non-complement reference (A) and complement-effector reference (D) were $+0.279$ and $+0.255$ (both BH $q = 0.105$). Single-protein CFH showed a near-zero mortality difference ($-0.031$), and pairwise contrasts flipped sign by reference protein: $-0.097$ against ALB versus $+0.367$ to $+0.477$ against HRG, PROC, and F13B (SI Figure~\ref{fig:si-sepsis-r1-initial}d). Excluding ALB post hoc ($+0.404$, $q = 0.019$) serves as a sensitivity check alongside the prespecified contrast.

\clearpage

\clearpage

\subsection{Round 2: First-Week Within-Patient Metabolite Changes}\label{sec:si-sepsis-s3}

Following outer-loop meta-review (Section~\ref{sec:method-meta-review}) of Round 1 hypotheses, findings, and limitations, CytoCoScientist generated 24 new candidates across 3 generations from 6 fresh starting hypotheses (18 ranked), without direct node-to-node derivation across rounds. Under the same mortality-focused research goal, all four candidates selected by Elo ranking in this round compared day-1 to day-7 metabolite changes within individual patients.

All four analyses used the 20 severe septic-shock patients from the Ferrario/ALBIOS subcohort~\cite{Ferrario2016ALBIOSMetabolomics} with paired day-1 and day-7 plasma measurements. For each patient, CytoCoScientist computed $\log_2(\text{D7}/\text{D1})$ per metabolite, subtracted that patient's reference summary, and estimated non-survivor minus survivor differences. H24, H27, and H26 used 28-day outcomes (9 deaths, 11 survivors), whereas H23 used 90-day outcomes (11 deaths, 9 survivors). All four analyses evaluate the same 20 paired patients across 28-day and 90-day follow-up windows and apply only to patients who survived to day 7.

\subsubsection{H24: First-Week Relative Lysophosphatidylcholine Trajectories}\label{sec:si-sepsis-s3-1}

\begin{caseprompt}{Hypothesis H24 | Pre-analysis summary (Round 2, rank 1, Elo 1487.6)}
\textbf{Hypothesis.} Among severe septic-shock patients surviving to day 7, first-week circulating lipid shifts involve selective relative under-recovery of lysophosphatidylcholines (LPCs) rather than uniform broad lipid fold changes driven by fluid balance or shared concentration scaling. This contracts the unmeasurable glycolysis--IL1B parent hypothesis into a testable within-patient lipid-selectivity question.

\textbf{Discriminative prediction.} Across seven prespecified LPC species (LPC a C16:1, C17:0, C18:0, C18:1, C18:2, C20:3, C20:4), the within-patient $\log_2(\text{D7}/\text{D1})$ change minus the median $\log_2(\text{D7}/\text{D1})$ across 26 fixed phosphatidylcholine and sphingomyelin (PC/SM) references should be lower in 28-day non-survivors than in survivors, with per-species consistency verified alongside the seven-member summary.

\textbf{Data and endpoint.} Paired day-1 and day-7 plasma p180 concentrations ($\mu\text{M}$) for 20 severe septic-shock patients (lactate $> 4$~mmol/L, SOFA $> 8$, 7--14-day post-shock ICU stay) in Supplementary Tables S1 and S2 of Ferrario et al.\ (2016; DOI: 10.1038/srep20391), with 28-day mortality primary (9 deaths, 11 survivors) and 90-day mortality secondary.
\end{caseprompt}

\paragraph{Initial analysis}
Referenced to the 26 PC/SM median per patient, the 28-day non-survivor minus survivor median difference was $-0.226$ $\log_2$ units (95\% CI $-1.724$ to $1.247$, all BH $q = 1.0$). Although the point estimate aligned with the predicted direction, leave-one-patient-out deletion crossed zero ($[-0.510, +0.007]$) and removing LPC C20:3 reversed the sign to $+0.077$. Current data cannot establish selective LPC under-recovery. The joint external lipid audit shared with H27 is reported after H27 (SI Figure~\ref{fig:si-sepsis-r2-initial}a).

\subsubsection{H27: Reference-Side Phosphatidylcholine versus Sphingomyelin Decomposition}\label{sec:si-sepsis-s3-2}

\begin{caseprompt}{Hypothesis H27 | Pre-analysis summary (Round 2, rank 2, Elo 1481.9)}
\textbf{Hypothesis.} Because the mixed PC/SM reference in H24 is not biologically inert, smaller day-1 to day-7 increases in the seven LPCs relative to PC/SM in 28-day non-survivors may reflect selective net accumulation on the phosphatidylcholine (PC) side rather than growth restriction on the LPC side.

\textbf{Discriminative prediction.} Splitting the fixed 33-lipid panel into three arms (7 LPCs, 20 PCs, 6 SMs), within-patient median $\text{PC} - \text{SM}$ $\log_2(\text{D7}/\text{D1})$ should be more positive in 28-day non-survivors than in survivors, while $\text{LPC} - \text{SM}$ shows no commensurate coordinated negative shift. Coordinated negative $\text{LPC} - \text{SM}$ without $\text{PC} - \text{SM}$ divergence would instead support LPC-side restriction.

\textbf{Data and endpoint.} Same 20 paired severe septic-shock patients and fixed 33-lipid p180 panel from Supplementary Tables S1 and S2 of Ferrario et al.\ (2016; DOI: 10.1038/srep20391), with 28-day mortality primary and 90-day mortality secondary.
\end{caseprompt}

\paragraph{Initial analysis}
CytoCoScientist computed each patient's median change across PCs and across SMs separately. The 28-day non-survivor minus survivor median difference for $\text{PC} - \text{SM}$ was $-0.291$ (95\% CI $-0.692$ to $0.146$), remaining negative across all 20 leave-one-patient-out deletions ($[-0.320, -0.203]$). The observed direction opposed the predicted PC-accumulation mechanism, though the primary confidence interval still crossed zero and cannot rule out the prediction (SI Figure~\ref{fig:si-sepsis-r2-initial}b).

\paragraph{Joint external lipid audit (H24 and H27)}
H24 and H27 shared one external lipid eligibility audit. ROSE provided paired plasma only at days 0 and 2, and inspected Liu records lacked complete linkage across the fixed 33-lipid panel and per-patient 28-day outcomes. The joint audit terminated without identifying an eligible independent lipid dataset; all patient-level lipid estimates therefore derive from Ferrario ($n = 20$; SI Figure~\ref{fig:si-sepsis-external-audit}a).

\subsubsection{H26: Relative First-Week Phenylalanine Trajectories}\label{sec:si-sepsis-s3-3}

\begin{caseprompt}{Hypothesis H26 | Pre-analysis summary (Round 2, rank 3, Elo 1403.9)}
\textbf{Hypothesis.} Among severe septic-shock patients surviving to day 7, first-week shifts in phenylalanine (Phe) relative to tyrosine (Tyr) and other amino acids reflect selective net balance changes in production, conversion, or clearance rather than uniform broad amino acid scaling. This contracts the unmeasurable BH4/eNOS parent hypothesis into a testable within-patient amino-acid selectivity contrast.

\textbf{Discriminative prediction.} In 28-day non-survivors, both within-patient $\log_2(\text{D7}/\text{D1})$ of Phe minus the median $\log_2(\text{D7}/\text{D1})$ across the fixed 19 reference amino acids (excluding Phe and Tyr) and $\Delta\log_2(\text{Phe}/\text{Tyr})$ should be more positive than in survivors. Uniform scaling across amino acids, or a ratio shift driven solely by falling Tyr, would weaken a Phe-selective interpretation.

\textbf{Data and endpoint.} Paired day-1 and day-7 plasma concentrations ($\mu\text{M}$) of Phe, Tyr, and 19 reference amino acids for the 20 severe septic-shock patients in Supplementary Tables S1 and S2 of Ferrario et al.\ (2016; DOI: 10.1038/srep20391), with 28-day mortality primary and 90-day mortality secondary.
\end{caseprompt}

\paragraph{Initial analysis}
Stratified by 28-day outcome, mean raw plasma Phe rose from $189$ to $208~\mu\text{M}$ in the 9 non-survivors and declined from $196$ to $164~\mu\text{M}$ in the 11 survivors. Across the 21 measured amino acids, referencing each patient's $\log_2(\text{D7}/\text{D1})$ Phe change to the median $\log_2(\text{D7}/\text{D1})$ across the fixed 19 reference amino acids (excluding both Phe and Tyr) yielded negative group means in both outcome groups ($-0.289$ in non-survivors versus $-0.663$ in survivors). The $+0.374$ $\log_2$-unit between-group difference (non-survivor minus survivor; 95\% CI $-0.276$ to $1.025$) therefore reflects a smaller average relative decline in non-survivors, with a wide interval leaving the mortality association unresolved. Evaluating the separate Phe/Tyr ratio contrast yielded $+0.489$ (95\% CI $-0.390$ to $1.368$).

\begin{figure}[p]
\centering
\includegraphics[width=\linewidth]{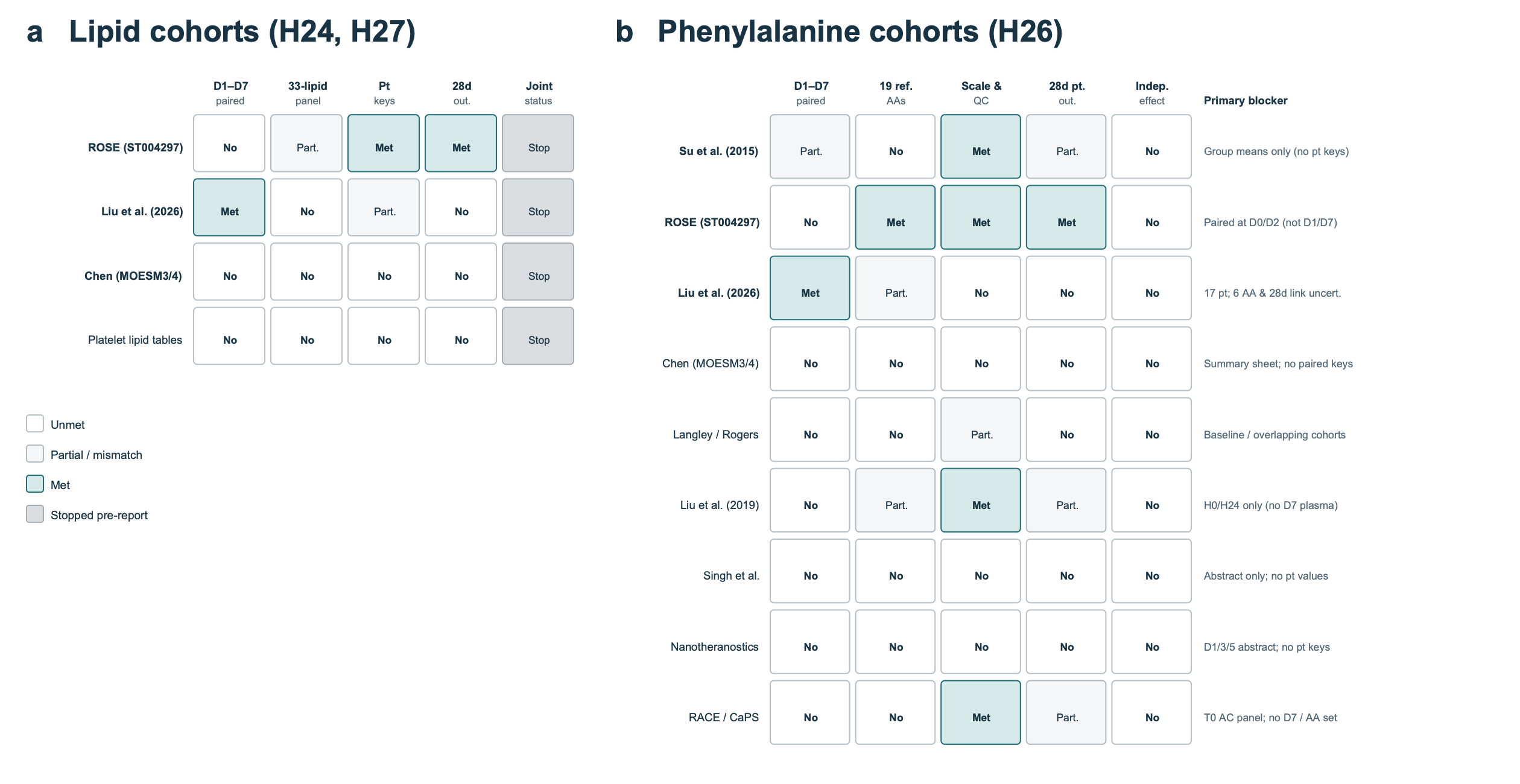}
\caption{\textbf{File-level external cohort eligibility audit for repeating first-week within-patient contrasts.} \textbf{a,} Joint lipid eligibility audit shared by H24 and H27 (stopped prior to completion). \textbf{b,} Nine external public sources audited for H26 phenylalanine across D1--D7 pairing, fixed 19-amino-acid coverage, scale/QC, and per-patient 28-day outcomes. Panels summarize data availability and eligibility for independent replication.}
\label{fig:si-sepsis-external-audit}
\end{figure}

\paragraph{External cohort eligibility audit}
Repeating this contrast independently requires per-patient paired day-1 and day-7 concentrations, reference amino acids, and linked 28-day outcomes. Across nine audited external sources, Supplementary Table S4 of Su et al.~\cite{Su2015SepsisAminoAcids} reported only outcome-stratified group means, ROSE sampled days 0 and 2, and none of the remaining inspected sources provided complete patient-level linkage; no independent effect estimate was therefore obtained (SI Table~\ref{tab:si-sepsis-external} and SI Figure~\ref{fig:si-sepsis-external-audit}b).

\begingroup
\footnotesize
\setlength{\tabcolsep}{3.5pt}
\renewcommand{\arraystretch}{1.15}
\setlength{\LTleft}{0pt}
\setlength{\LTright}{\fill}
\setlength{\LTcapwidth}{\linewidth}
\captionsetup{singlelinecheck=false,justification=raggedright}
\setlength{\sepsisTableWidth}{\dimexpr\linewidth-4\tabcolsep\relax}
\begin{longtable}{@{}>{\raggedright\arraybackslash}p{0.25\sepsisTableWidth}>{\raggedright\arraybackslash}p{0.34\sepsisTableWidth}>{\raggedright\arraybackslash}p{0.41\sepsisTableWidth}@{}}
\caption{\textbf{External public sources audited for independent day-1 to day-7 phenylalanine validation (H26).} Availability reflects inspected public files only and summarizes cohort eligibility for replication.}\label{tab:si-sepsis-external}\\
\toprule
\textbf{External source} & \textbf{Retrieved material} & \textbf{Reason prespecified Phe estimate was not obtained} \\
\midrule
\endfirsthead
\multicolumn{3}{@{}l}{\footnotesize\textbf{Table~\thetable\ (continued)}}\\
\toprule
\textbf{External source} & \textbf{Retrieved material} & \textbf{Reason prespecified Phe estimate was not obtained} \\
\midrule
\endhead
\midrule
\multicolumn{3}{r@{}}{\footnotesize\emph{Continued on next page}}\\
\endfoot
\bottomrule
\endlastfoot
Su et al.\ (2015; \cite{Su2015SepsisAminoAcids}) & Multi-timepoint group means at D1 and D7 (Supplementary Table S4 of the Su study) & No patient-level values, sample keys, or D1--D7 patient overlap. \\
ROSE (ST004297) & Linkable early paired plasma samples & Sampled at D0/D2 rather than D1/D7. \\
Liu et al.\ (2026) & 17-patient four-timepoint structure with candidate Phe/Tyr measurements & Incomplete reference scale/coverage and uncertified per-patient 28-day outcome linkage. \\
Chen (MOESM3/4) & Summary feature tables comparing sepsis vs.\ healthy controls & No paired patient sample keys or 28-day outcomes. \\
Langley / Rogers & Early cohorts and summary records (partial cohort overlap) & First-week paired per-patient measurements and sample-to-outcome linkage unverified. \\
Liu et al.\ (2019) & 51 patients with H0/H24 pairs & Both draws occur within the first 24 hours; no D7 pair. \\
Singh et al. & Abstract reporting sample counts at two time points & No verifiable patient-level paired values, analyte concentrations, or 28-day labels obtained. \\
Nanotheranostics study & Abstract noting pre/post-treatment days 1, 3, and 5 & Time points, patient keys, and outcomes do not support target contrast. \\
RACE/CaPS & Pretreatment acylcarnitines and trial metadata & Outcome summaries cannot establish day-7 plasma sample linkage. \\
\end{longtable}
\endgroup

\subsubsection{H23: Relative First-Week Arginine Trajectories}\label{sec:si-sepsis-s3-4}

\begin{caseprompt}{Hypothesis H23 | Pre-analysis summary (Round 2, rank 4, Elo 1403.5)}
\textbf{Hypothesis.} Among severe septic-shock patients surviving to day 7, first-week arginine (Arg) trajectories involve selective relative under-recovery compared with ornithine (Orn) and broadly measured amino acids rather than uniform fold scaling. This contracts the unmeasurable ARG1--T-cell parent hypothesis into a testable substrate--product balance contrast.

\textbf{Discriminative prediction.} In 90-day non-survivors, both $\Delta\log_2(\text{Arg}/\text{Orn})$ and within-patient $\log_2(\text{D7}/\text{D1})$ of Arg minus the median across the fixed 19 reference amino acids (excluding Arg and Orn) should be lower than in survivors. Uniform amino acid scaling or an opposite direction would weaken the selective under-recovery explanation.

\textbf{Data and endpoint.} Paired day-1 and day-7 plasma concentrations ($\mu\text{M}$) of Arg, Orn, and 19 reference amino acids for the 20 severe septic-shock patients in Supplementary Tables S1 and S2 of Ferrario et al.\ (2016; DOI: 10.1038/srep20391), with 90-day mortality primary (11 deaths, 9 survivors) and 28-day mortality secondary.
\end{caseprompt}

\begin{figure}[p]
\centering
\includegraphics[width=\linewidth]{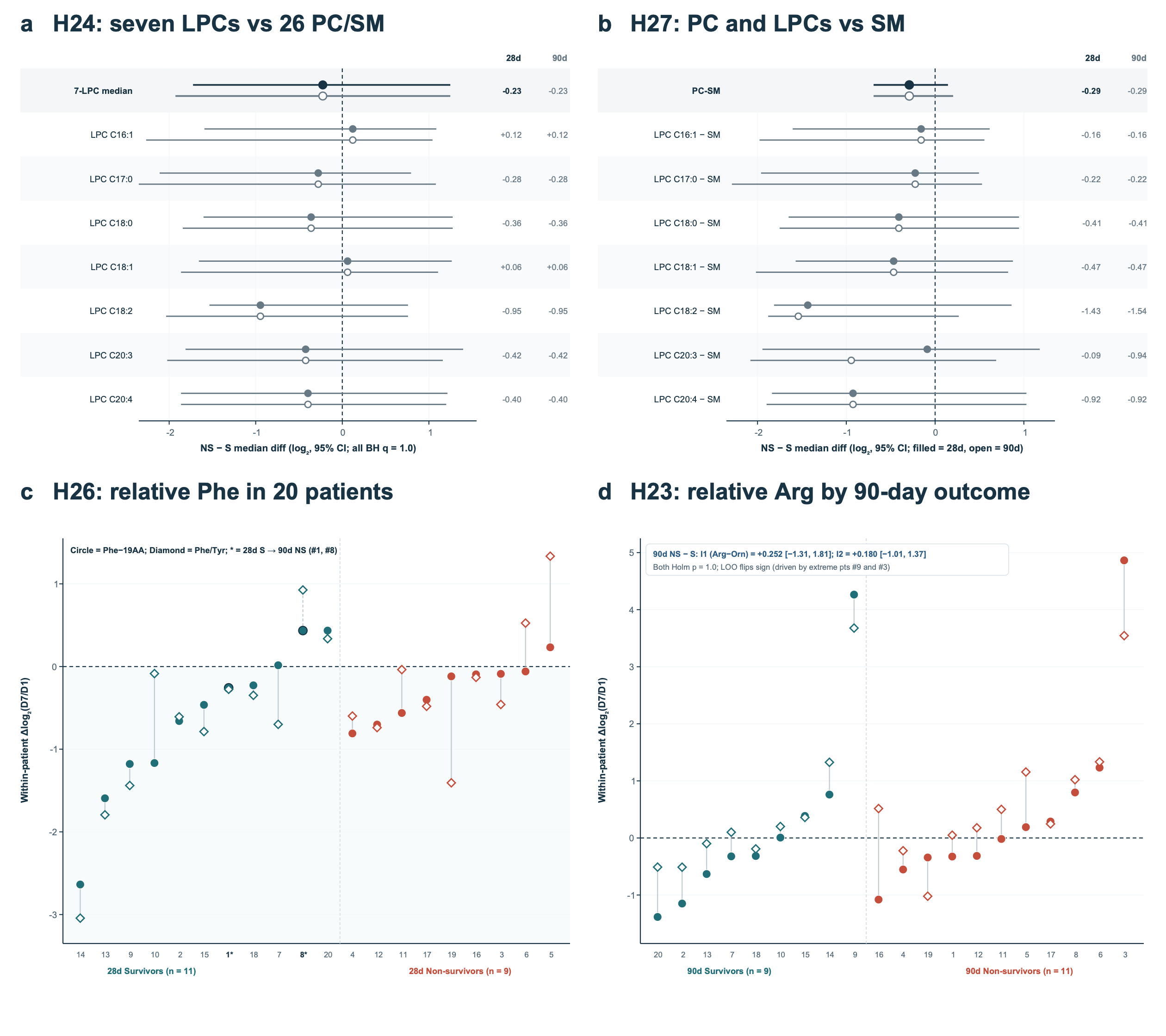}
\caption{\textbf{Within-patient day-1 to day-7 metabolite changes across all four Round 2 hypotheses in Ferrario/ALBIOS ($n = 20$).} \textbf{a,} H24: seven LPC species and their median relative to 26 PC/SM references (28-day and 90-day outcomes). \textbf{b,} H27: PC relative to SM and individual LPC species relative to SM (28-day and 90-day). \textbf{c,} H26: patient-level $\Delta\log_2(\text{D7}/\text{D1})$ for Phe minus the fixed 19-amino-acid reference (excluding Phe and Tyr) and for Phe/Tyr (asterisks mark patients \#1 and \#8 who flip from 28-day survivor to 90-day non-survivor). \textbf{d,} H23: patient-level 90-day Arg minus Orn (I1) and Arg minus the fixed 19-amino-acid reference (excluding Arg and Orn; I2).}
\label{fig:si-sepsis-r2-initial}
\end{figure}

\paragraph{Initial analysis}
Across 11 non-survivors and 9 survivors at 90 days, the non-survivor minus survivor mean difference was $+0.252$ (95\% CI $-1.308$ to $1.811$) relative to Orn and $+0.180$ ($-1.013$ to $1.372$) relative to the fixed 19 reference amino acids (excluding Arg and Orn). Both point estimates opposed the predicted recovery deficit, intervals were wide, and leave-one-patient-out deletion flipped sign, preventing determination of whether relative arginine under-recovery is present or absent (SI Figure~\ref{fig:si-sepsis-r2-initial}d).

\clearpage

\clearpage

\clearpage

\subsubsection{Proposed Follow-Up Validation for Phenylalanine (H26)}\label{sec:si-sepsis-s3-5}

\paragraph{Question to resolve}
In an independent cohort with paired day-1 and day-7 plasma, does phenylalanine (Phe) shift more positively relative to the fixed 19 reference amino acids (excluding Phe and Tyr) in 28-day non-survivors than in survivors? Here ``more positively'' denotes the between-group contrast: even when both groups show negative averages on the within-patient relative metric, a smaller relative decline in non-survivors matches the predicted direction. Individual Phe and tyrosine (Tyr) trajectories must also be inspected separately to distinguish rising Phe from falling Tyr as the driver of any ratio shift.

\paragraph{Current baseline}
In Ferrario ($n = 20$), non-survivor minus survivor mean differences were $+0.374$ $\log_2$ units (95\% CI $-0.276$ to $1.025$) relative to the fixed 19-amino-acid reference and $+0.489$ ($-0.390$ to $1.368$) for Phe/Tyr. Both point estimates aligned with the predicted direction, but both 95\% CIs crossed zero. Supplementary Table S4 of Su et al.~\cite{Su2015SepsisAminoAcids} reports only outcome-stratified group mean concentrations and cannot reconstruct within-patient relative changes; none of the other audited external sources supported independent recalculation (SI Table~\ref{tab:si-sepsis-external}).

\paragraph{Next step and decision branches}
Prioritize de-identified records from patients outside the Ferrario cohort or independent biobanks with clinical follow-up and banked paired plasma. The next step depends on the available material:
\begin{enumerate}
  \item \emph{Qualified paired concentration records already exist:} reanalyze directly without new assays.
  \item \emph{Only banked paired plasma is available:} quantify Phe, Tyr, and the fixed 19 reference amino acids on a single assay platform with batch and limit-of-detection (LOD) logging, with laboratory personnel blinded to clinical outcomes.
  \item \emph{Only Phe and Tyr records qualify:} test the second contrast only ($\Delta\log_2(\text{Phe}/\text{Tyr})$) and explicitly bound scope; do not claim replication of the fixed 19-amino-acid primary contrast.
\end{enumerate}
If a new cohort must be built, pre-specify the minimum interpretable effect or target confidence-interval precision before inspecting outcome splits, size the required effective paired sample accordingly, and include a buffer for early deaths and missing day-7 samples. Across all branches, verify patient IDs, sample IDs, baseline anchoring events, actual draw timestamps, 28-day follow-up origins and completeness, and the explicit reason for any missing day-7 blood draw (distinguishing early death, early discharge, and missed sampling to define the applicable population). Audit concentration units, sample matrices, low-value/LOD flags, and missingness codes, and link time-aligned clinical covariates (severity scores, hepatic and renal function, nutritional support, and renal replacement therapy) where available.

Before viewing outcome splits, freeze the sampling window, missingness handling, reference list, and two primary contrasts. For each patient, compute $\log_2(\text{D7}/\text{D1})$ per amino acid, then compute: (1) $\Delta\log_2(\text{Phe})$ minus the median $\Delta\log_2$ across the fixed 19 reference amino acids (excluding Phe and Tyr), and (2) $\Delta\log_2(\text{Phe}) - \Delta\log_2(\text{Tyr})$, i.e., $\Delta\log_2(\text{Phe}/\text{Tyr})$. Report 28-day non-survivor minus survivor mean differences and 95\% CIs for both primary contrasts with Holm correction across the two tests. Report paired and unpaired patient counts separately by outcome group, and present individual Phe and Tyr trajectories alongside the primary contrasts.

\paragraph{How different outcomes change judgment}
Concordant positive contrasts with sufficient precision on both primary metrics in independent patients would justify prioritizing mechanistic follow-up of this relative shift, without establishing mechanism directly. If only the Phe/Tyr ratio changes and is driven primarily by declining Tyr, the ``selective Phe accumulation'' interpretation should be revised. A well-precisioned negative contrast would deprioritize H26, whereas wide confidence intervals would preserve uncertainty. Until qualified independent patient-level data are obtained, H26 remains an unvalidated exploratory candidate.

\subsection{Complete SI Tables}\label{sec:si-sepsis-tables}

SI Tables~\ref{tab:si-sepsis-candidates} and~\ref{tab:si-sepsis-analyses} index all 54 candidate hypotheses together with the eight initial analyses, two cross-cohort follow-up analyses, and two external data-eligibility audits; SI Tables~\ref{tab:si-sepsis-r1}--\ref{tab:si-sepsis-sensitivity} consolidate numerical estimates and sensitivity analyses across both rounds. SI Table~\ref{tab:si-sepsis-external} appears above with the H26 external cohort audit.

\begin{landscape}
\begingroup
\footnotesize
\setlength{\tabcolsep}{3.5pt}
\renewcommand{\arraystretch}{1.12}
\setlength{\LTleft}{0pt}
\setlength{\LTright}{\fill}
\setlength{\LTcapwidth}{\linewidth}
\captionsetup{singlelinecheck=false,justification=raggedright}
\setlength{\sepsisTableWidth}{\dimexpr\linewidth-4\tabcolsep\relax}
\begin{longtable}{@{}>{\raggedright\arraybackslash}p{0.10\sepsisTableWidth}>{\raggedright\arraybackslash}p{0.70\sepsisTableWidth}>{\raggedright\arraybackslash}p{0.20\sepsisTableWidth}@{}}
\caption{\textbf{Summary of all 54 pre-analysis candidate hypotheses generated across Rounds 1 and 2 and their proposed data sources.} Listed in order of appearance; entries are condensed from CytoCoScientist's original pre-analysis hypothesis statements while preserving all candidate mechanisms, target panels, and scopes. Identifiers match Figure~\ref{fig:sepsis-dag}; unranked candidates correspond to gray nodes. Right column lists only the data source identified when each hypothesis was proposed.}\label{tab:si-sepsis-candidates}\\
\toprule
\textbf{Node} & \textbf{Pre-analysis hypothesis summary} & \textbf{Proposed data source} \\
\midrule
\endfirsthead
\multicolumn{3}{@{}l}{\footnotesize\textbf{Table~\thetable\ (continued)}}\\
\toprule
\textbf{Node} & \textbf{Pre-analysis hypothesis summary} & \textbf{Proposed data source} \\
\midrule
\endhead
\midrule
\multicolumn{3}{r@{}}{\footnotesize\emph{Continued on next page}}\\
\endfoot
\bottomrule
\endlastfoot
\multicolumn{3}{@{}l}{\textit{Round 1 (30 candidates: 16 ranked, 14 unranked)}} \\
\addlinespace[0.25em]
\textbf{H01} & Selective depletion of complement effectors (C3, C4A, CFB) relative to CFH and non-complement proteins impairs opsonophagocytic reserve and pathogen clearance, raising 30-day mortality risk. & PXD055932 \\
\textbf{H02} & Selective reduction of APOA1 relative to APOB weakens circulating bacterial-lipid neutralization and anti-inflammatory buffering, aggravating tissue injury and 30-day mortality risk. & PXD055932 \\
\textbf{H03} & Elevated plasma LDHA marks hepatocellular leakage (co-varying with ALDOB and TALDO1) and lost hepatic metabolic capacity, raising 30-day mortality risk, rather than reflecting glycolytic flux. & PXD055932 \\
\textbf{H04} & Elevated plasma LCN2 co-occurring with S100A8 and S100A9 marks persistent neutrophil-associated inflammatory release and bystander tissue injury (higher 30-day mortality) rather than solely impaired renal clearance. & PXD055932 \\
\textbf{H05} & Selective pretreatment acetylcarnitine accumulation relative to free carnitine and other acylcarnitines marks a carnitine-remediable mitochondrial acyl-buffering defect that lowers 90-day mortality under L-carnitine therapy. & RACE/CaPS pretreatment acylcarnitines \\
\textbf{H06} & Emergency granulopoiesis expands IL1R2$^+$ immature neutrophils that suppress CD4$^+$ T cells, delaying adaptive immune recovery and raising mortality risk beyond mature neutrophil counts alone. & Kwok whole-blood scRNA-seq metadata (Zenodo 7924238) \\
Unranked & Mortality-associated low complement abundance reflects coordinated selective declines of C3, C4A, and CFB relative to CFH and non-complement proteins due to differential turnover, rather than uniform plasma protein reduction. & PXD055932 \\
Unranked & Mortality-associated low complement abundance belongs to a shared complement--hemostasis response (jointly low C3, C4A, CFB, CFH, HRG, PROC, and F13B with preserved ALB), attenuating when hemostasis proteins remain preserved. & PXD055932 \\
\textbf{H07} & CD4$^+$ T-cell hyporesponsiveness in granulocyte-rich sepsis is maintained by ongoing neutrophil and extracellular-milieu exposure (reversible upon removal) rather than irreversible early cell-intrinsic T-cell injury. & PXD055932 \\
Unranked & High plasma LDHA accompanying 30-day mortality reflects hypoperfusion-associated multi-tissue injury (tracking muscle MB alongside hepatic ALDOB and TALDO1) rather than isolated hepatocyte leakage. & PXD055932 \\
\textbf{H08} & \textbf{(Selected R1 Top 1)} Co-occurrence of low APOA1 with elevated S100A8/A9 arises primarily from shared acute-phase plasma protein remodeling (co-varying with APOB and ALB), with S100 elevation persisting even when APOA1 is relatively preserved. & PXD055932 \\
Unranked & Day-1 mortality-associated complement shifts represent coordinated relative divergence of C3, C4A, and CFB below CFH and non-complement comparators rather than low absolute C3 alone. & PXD055932 \\
\textbf{H09} & \textbf{(Selected R1 Top 4)} Apparent low complement effectors relative to CFH is driven by positive deviation of regulatory CFH relative to effectors and non-complement proteins, potentially restricting pathogen opsonization. & PXD055932 \\
Unranked & Inflammation-associated remodeling includes a shared apolipoprotein axis (jointly low APOA1 and APOB with preserved ALB) co-occurring with elevated S100A8/A9, distinct from uniform three-protein decline or APOA1-only loss. & PXD055932 \\
\textbf{H10} & Complement-related opsonophagocytic impairment requires ongoing circulating complement restriction to persist (reversible upon late relief) rather than irreversible early phagocyte-intrinsic injury. & PXD055932 \\
\textbf{H11} & Extracellular S100A8/A9 amplification actively maintains inflammation and mortality risk independently of selective APOA1 deficiency, persisting when APOA1 is relatively preserved. & PXD055932 \\
Unranked & Pretreatment acetylcarnitine elevation marks a high-baseline-risk metabolic state rather than a carnitine-specific defect; larger absolute 90-day benefit on 18~g carnitine may reflect a constant relative effect on higher baseline risk. & RACE/CaPS pretreatment acylcarnitines \\
Unranked & Low complement effectors (C3, C4A, CFB relative to CFH) mark higher 30-day mortality risk only when hemostasis proteins (HRG, PROC, F13B relative to ALB) are concurrently low. & PXD055932 \\
Unranked & Relative complement imbalance carries higher 30-day mortality risk when driven by coordinated effector and hemostasis declines (C3, C4A, CFB, HRG, PROC, F13B) than when driven by isolated positive CFH deviation. & PXD055932 \\
Unranked & Mortality differences associated with coordinated S100A8/A9 elevation depend on jointly low APOA1, APOB, and ALB (broad remodeling) rather than APOA1-unique deviation, weakening when all three proteins are preserved. & PXD055932 \\
\textbf{H12} & \textbf{(Selected R1 Top 2)} Candidate L-carnitine treatment response stems primarily from relative free-carnitine deficiency rather than selective acetylcarnitine accumulation at matched imbalance and acylcarnitine background. & RACE/CaPS pretreatment acylcarnitines \\
Unranked & Complement imbalance involves discordant intra-effector shifts (coordinated low C3 and CFB with preserved C4A relative to non-complement comparators), reflecting unequal turnover across complement components. & PXD055932 \\
\textbf{H13} & \textbf{(Selected R1 Top 3)} Coordinated S100A8 and S100A9 elevation primarily marks a cross-system complement--hemostasis response (jointly low C3, C4A, CFB, HRG, PROC, F13B) rather than isolated CFH elevation with those six proteins preserved. & PXD055932 \\
Unranked & Coordinated negative deviation of all three complement effectors (C3, C4A, CFB) below non-complement comparators is required for higher 30-day mortality risk, whereas isolated positive CFH deviation with preserved effectors is insufficient. & PXD055932 \\
Unranked & Exogenous L-carnitine survival benefit strengthens as baseline free carnitine decreases (remediable substrate limitation), whereas elevated acetylcarnitine primarily marks metabolic load or impaired organ disposal. & RACE/CaPS pretreatment acylcarnitines \\
Unranked & Adverse host-defense effects of elevated CFH relative to complement effectors occur only when the infecting pathogen actively recruits CFH to its surface to evade opsonization. & PXD055932 \\
\textbf{H14} & Early extracellular S100 signaling initiates a self-sustaining myeloid pro-inflammatory secretory state that persists after S100 removal, making enrollment S100 elevation an accompanying readout rather than an ongoing maintenance requirement. & PXD055932 \\
\textbf{H15} & Shared APOA1--APOB--ALB variation reflects disease course, whereas endothelial injury is mediated independently by insufficient albumin fatty-acid binding capacity elevating unbound long-chain fatty acids (even when ALB abundance is preserved). & PXD055932, CMAISE \\
\textbf{H16} & Low complement abundance and opsonophagocytic impairment arise in parallel from a shared disease process; phagocyte failure is maintained by an unidentified circulating inhibitory milieu rather than insufficient opsonization. & PXD055932 \\
Unranked & Sustained CD4$^+$ T-cell hyporesponsiveness requires simultaneous support from both neutrophils and the circulating soluble milieu, with neither alone sufficient to maintain suppression. & PXD055932 \\
\midrule
\multicolumn{3}{@{}l}{\textit{Round 2 (24 candidates: 18 ranked, 6 unranked)}} \\
\addlinespace[0.25em]
\textbf{H17} & Impaired leukocyte type-I interferon responses weaken antiviral defense, sustain viral burden, and raise mortality risk specifically in patients with active viral infection or reactivation. & GSE272769 \\
\textbf{H18} & Sustained myeloid HIF1A glycolytic programming coupled to IL1B output delays endothelial barrier recovery and increases mortality risk under persistent microbial stimulation, beyond circulating lactate alone. & CMAISE \\
\textbf{H19} & Myeloid ARG1-mediated arginine consumption depletes extracellular arginine, suppressing T-cell proliferation and infection control and raising mortality risk independently of renal clearance or nutrition. & CMAISE \\
\textbf{H20} & Oxidative-stress-induced tetrahydrobiopterin (BH4) deficiency impairs both phenylalanine hydroxylation (raising Phe/Tyr) and eNOS coupling, promoting microvascular dysfunction and mortality in cofactor-deficient patients. & CMAISE \\
\textbf{H21} & Insufficient glutathione regeneration impairs GPX4-dependent lipid-peroxide clearance under oxidative stress, sensitizing endothelium to ferroptosis-related leakage and raising mortality risk. & CMAISE \\
\textbf{H22} & Persistent suppression of intracellular monocyte CIITA antigen-presentation programming impairs MHC class~II-dependent CD4$^+$ T-cell activation and infection clearance, beyond total monocyte counts or bulk HLA RNA. & GSE272769 \\
\textbf{H23} & \textbf{(Selected R2 Top 4)} In day-7-surviving severe septic-shock patients, 90-day non-survivors show selective relative under-recovery of arginine from D1 to D7 compared with ornithine and 19 reference amino acids rather than uniform fold scaling. & Ferrario/ALBIOS plasma metabolomics (Ferrario Supplementary Tables S1, S2) \\
\textbf{H24} & \textbf{(Selected R2 Top 1)} In day-7-surviving severe septic-shock patients, 28-day non-survivors show selective relative under-recovery of seven LPC species from D1 to D7 compared with 26 PC/SM references rather than uniform lipid scaling. & Ferrario/ALBIOS plasma metabolomics (Ferrario Supplementary Tables S1, S2) \\
\textbf{H25} & Once monocyte antigen-presentation hyporesponsiveness is established, non-recovery of intracellular CIITA programming maintains impaired CD4$^+$ activation even after exogenous inhibitory cues are removed. & GSE272769, Ferrario \\
Unranked & Differential net handling of aromatic amino acids (rather than uniform amino acid concentration shifts) produces selective D1-to-D7 phenylalanine accumulation relative to tyrosine in subsequent non-survivors. & Ferrario/ALBIOS plasma metabolomics (Ferrario Supplementary Tables S1, S2) \\
\textbf{H26} & \textbf{(Selected R2 Top 3)} In day-7-surviving severe septic-shock patients, 28-day non-survivors exhibit larger D1-to-D7 upward shifts in phenylalanine relative to both tyrosine and 19 reference amino acids than survivors, departing from uniform scaling. & Ferrario/ALBIOS plasma metabolomics (Ferrario Supplementary Tables S1, S2) \\
Unranked & First-week circulating lipid shifts involve selective under-recovery of long-chain lysophosphatidylcholines (LPC a C17:0 and C18:0) relative to PCs and SMs in 28-day or 90-day non-survivors. & Ferrario/ALBIOS plasma metabolomics (Ferrario Supplementary Tables S1, S2) \\
\textbf{H27} & \textbf{(Selected R2 Top 2)} Apparent relative LPC deficits against the mixed PC/SM reference in 28-day non-survivors reflect selective net PC accumulation relative to SM rather than coordinated LPC growth restriction relative to SM. & Ferrario/ALBIOS plasma metabolomics (Ferrario Supplementary Tables S1, S2) \\
Unranked & First-week amino acid divergence in 90-day non-survivors reflects Arg-specific under-growth relative to the fixed broad reference while Orn tracks the reference, rather than selective Orn growth. & Ferrario/ALBIOS plasma metabolomics (Ferrario Supplementary Tables S1, S2) \\
Unranked & Aromatic amino acid divergence in 28-day non-survivors requires a selective upward Phe trajectory relative to both Tyr and the majority of reference amino acids, rather than falling Tyr or reference shifts alone. & Ferrario/ALBIOS plasma metabolomics (Ferrario Supplementary Tables S1, S2) \\
\textbf{H28} & In hyporesponsive monocytes retaining surface HLA class~II, persistent dysfunction after exogenous washout is maintained by impaired endolysosomal acidification and antigen processing rather than CIITA transcriptional suppression. & GSE272769 \\
Unranked & Smaller D1-to-D7 LPC gains relative to PC/SM in 28-day non-survivors reflect coordinated target-side under-growth across all seven LPCs rather than faster reference-side PC or SM expansion. & Ferrario/ALBIOS plasma metabolomics (Ferrario Supplementary Tables S1, S2) \\
\textbf{H29} & Ongoing iron-dependent membrane phospholipid peroxidation maintains established endothelial barrier failure even after oxidative stimuli wane or intracellular glutathione recovers. & CMAISE \\
\textbf{H30} & Any selective D1-to-D7 plasma PC increase in 28-day non-survivors arises from shifts in circulating lipoprotein carrier abundance or class composition rather than intra-carrier PC enrichment. & Ferrario/ALBIOS plasma metabolomics (Ferrario Supplementary Tables S1, S2) \\
\textbf{H31} & Any upward D1-to-D7 Phe shift relative to Tyr and reference amino acids in 28-day non-survivors depends primarily on compositional differences in delivered exogenous amino acid nutrition rather than supply-independent handling. & Ferrario/ALBIOS plasma metabolomics (Ferrario Supplementary Tables S1, S2) \\
\textbf{H32} & Early local arginine limitation initiates an intracellular T-cell hyporesponsive state that persists after arginine restoration and exogenous washout, no longer requiring ongoing arginine deficiency. & Ferrario/ALBIOS plasma metabolomics (Ferrario Supplementary Tables S1, S2) \\
\textbf{H33} & Persistent monocyte antigen-presentation failure with intact HLA class~II and endolysosomal processing is maintained by contact-dependent inhibition via monocyte surface PD-L1 rather than impaired antigen processing. & GSE272769 \\
Unranked & Any mortality-associated D1-to-D7 departure across the seven LPCs reflects shared shifts in circulating lipoprotein and albumin carrier distribution rather than coordinated depletion of intra-carrier LPC loading. & Ferrario/ALBIOS plasma metabolomics (Ferrario Supplementary Tables S1, S2) \\
\textbf{H34} & Iron-dependent endothelial phospholipid peroxidation is a reversible mediator requiring continuous circulating stimulation (which also suffices to re-induce barrier failure in undamaged endothelium) rather than a cell-autonomous state. & CMAISE \\
\end{longtable}
\endgroup
\end{landscape}

\begingroup
\footnotesize
\setlength{\tabcolsep}{3.5pt}
\renewcommand{\arraystretch}{1.15}
\setlength{\LTleft}{0pt}
\setlength{\LTright}{\fill}
\setlength{\LTcapwidth}{\linewidth}
\captionsetup{singlelinecheck=false,justification=raggedright}
\setlength{\sepsisTableWidth}{\dimexpr\linewidth-6\tabcolsep\relax}
\begin{longtable}{@{}>{\raggedright\arraybackslash}p{0.20\sepsisTableWidth}>{\raggedright\arraybackslash}p{0.23\sepsisTableWidth}>{\raggedright\arraybackslash}p{0.41\sepsisTableWidth}>{\raggedright\arraybackslash}p{0.16\sepsisTableWidth}@{}}
\caption{\textbf{Summary of the eight initial analyses, two cross-cohort follow-up analyses, and two external data-eligibility audits.} Each round includes four initial analyses; Round 1 adds two cross-cohort protein/RNA follow-ups (H08, H13), and Round 2 adds two external eligibility audits (H24+H27 shared; H26 separate) which identified no eligible independent cohorts for effect estimation.}\label{tab:si-sepsis-analyses}\\
\toprule
\textbf{Node and step} & \textbf{Dataset} & \textbf{Comparison and stopping point} & \textbf{Figure} \\
\midrule
\endfirsthead
\multicolumn{4}{@{}l}{\footnotesize\textbf{Table~\thetable\ (continued)}}\\
\toprule
\textbf{Node and step} & \textbf{Dataset} & \textbf{Comparison and stopping point} & \textbf{Figure} \\
\midrule
\endhead
\midrule
\multicolumn{4}{r@{}}{\footnotesize\emph{Continued on next page}}\\
\endfoot
\bottomrule
\endlastfoot
H08 $\cdot$ Initial & PXD055932 ($n = 326$) & Double-high S100 persists among relatively preserved APOA1 ($63/173$); whether the shared component dominates remains unresolved. & Fig.~\ref{fig:sepsis-round1}a; SI Fig.~\ref{fig:si-sepsis-r1-initial}a \\
H08 $\cdot$ Follow-up & Mi GAinS; CMAISE & Mi compares double-high rates and continuous slope; CMAISE reports both albumin branches separately. & Fig.~\ref{fig:sepsis-round1}a; SI Fig.~\ref{fig:si-sepsis-r1-followup}a \\
H12 $\cdot$ Initial & RACE/CaPS ($n = 170$) & Evaluates pretreatment distribution and local overlap; halts before estimating 90-day treatment effect. & Fig.~\ref{fig:sepsis-round1}d; SI Fig.~\ref{fig:si-sepsis-r1-initial}b \\
H13 $\cdot$ Initial & PXD055932 ($n = 304$) & Estimates conditional S100 contrast between two model-defined proteomic backgrounds. & Fig.~\ref{fig:sepsis-round1}b; SI Fig.~\ref{fig:si-sepsis-r1-initial}c \\
H13 $\cdot$ Follow-up & Mi GAinS, CMAISE; bulk \& scRNA & Cross-cohort plasma protein direction differs (reverses in CMAISE); RNA provides a separate measurement layer. & Fig.~\ref{fig:sepsis-round1}b; SI Fig.~\ref{fig:si-sepsis-r1-followup}b,c \\
H09 $\cdot$ Initial & PXD055932 ($n = 310$) & CFH contrast depends on reference protein; single-protein and prespecified reference results reported separately. & Fig.~\ref{fig:sepsis-round1}c; SI Fig.~\ref{fig:si-sepsis-r1-initial}d \\
H24 $\cdot$ Initial & Ferrario/ALBIOS ($n = 20$, D1/D7 paired; 28d 9/11) & 7 LPC relative to 26 PC/SM sensitive to individual patients and LPC C20:3 removal. & Fig.~\ref{fig:sepsis-round2}c; SI Fig.~\ref{fig:si-sepsis-r2-initial}a \\
H27 $\cdot$ Initial & Ferrario/ALBIOS ($n = 20$, D1/D7 paired; 28d 9/11) & $\text{PC} - \text{SM}$ point estimate opposes predicted PC gain; confidence interval crosses zero. & Fig.~\ref{fig:sepsis-round2}c; SI Fig.~\ref{fig:si-sepsis-r2-initial}b \\
H26 $\cdot$ Initial & Ferrario/ALBIOS ($n = 20$, D1/D7 paired; 28d 9/11) & Separates raw Phe concentrations, within-patient relative changes (vs.\ fixed 19 AAs excluding Phe/Tyr), and outcome group differences. & Fig.~\ref{fig:sepsis-round2}a,b; SI Fig.~\ref{fig:si-sepsis-r2-initial}c \\
H23 $\cdot$ Initial & Ferrario/ALBIOS ($n = 20$, D1/D7 paired; 90d 11/9) & Arg relative to Orn and fixed 19 AAs (excluding Arg/Orn) imprecise and sensitive to single-patient removal. & Fig.~\ref{fig:sepsis-round2}b; SI Fig.~\ref{fig:si-sepsis-r2-initial}d \\
H24+H27 $\cdot$ Audit (shared) & External lipid sources & Audits 33-lipid panel, pairing window, and per-patient mortality; stopped prior to completion without independent effect estimates. & Fig.~\ref{fig:sepsis-round2}d; SI Fig.~\ref{fig:si-sepsis-external-audit}a \\
H26 $\cdot$ Audit & 9 external Phe sources & Audits 9 public sources; none support prespecified independent effect estimation. & Fig.~\ref{fig:sepsis-round2}d; SI Fig.~\ref{fig:si-sepsis-external-audit}b \\
\end{longtable}
\endgroup

\begingroup
\footnotesize
\setlength{\tabcolsep}{3.5pt}
\renewcommand{\arraystretch}{1.15}
\setlength{\LTleft}{0pt}
\setlength{\LTright}{\fill}
\setlength{\LTcapwidth}{\linewidth}
\captionsetup{singlelinecheck=false,justification=raggedright}
\setlength{\sepsisTableWidth}{\dimexpr\linewidth-6\tabcolsep\relax}
\begin{longtable}{@{}>{\raggedright\arraybackslash}p{0.17\sepsisTableWidth}>{\raggedright\arraybackslash}p{0.19\sepsisTableWidth}>{\raggedright\arraybackslash}p{0.34\sepsisTableWidth}>{\raggedright\arraybackslash}p{0.30\sepsisTableWidth}@{}}
\caption{\textbf{Round 1 numerical summary: within-blood-sample molecular contrasts.} Parentheses denote 95\% confidence intervals; protein-level contrasts are reported separately from mortality-risk estimates.}\label{tab:si-sepsis-r1}\\
\toprule
\textbf{Branch and cohort} & \textbf{Effective sample} & \textbf{Primary estimates (95\% CI)} & \textbf{Interpretation} \\
\midrule
\endfirsthead
\multicolumn{4}{@{}l}{\footnotesize\textbf{Table~\thetable\ (continued)}}\\
\toprule
\textbf{Branch and cohort} & \textbf{Effective sample} & \textbf{Primary estimates (95\% CI)} & \textbf{Interpretation} \\
\midrule
\endhead
\midrule
\multicolumn{4}{r@{}}{\footnotesize\emph{Continued on next page}}\\
\endfoot
\bottomrule
\endlastfoot
H08 $\cdot$ PXD055932 & 326 (127 30d deaths) & $|\Delta C| - |2\Delta D/3| = +0.048$ ($-0.124, 0.196$); $D \ge 0$ double-high S100: $63/173$. & Shared dominance unresolved; double-high S100 is not restricted to relatively low APOA1. \\
H08 $\cdot$ Mi, CMAISE & Mi 495; CMAISE primary 342 (44 in-hospital deaths) / clinical-ALB auxiliary 338 (43) & Mi double-high: $128/243$ ($D < 0$) vs.\ $76/252$ ($D \ge 0$). CMAISE standardized APOA1 higher in non-survivors ($\Delta A = +0.584$). & Mi tests co-occurrence without mortality linkage; CMAISE does not support low APOA1 as a universal mortality marker. \\
H12 $\cdot$ RACE/CaPS & Placebo 70; 18~g 100 & Only 2 matched pairs under primary criteria ($0/2$ placebo on \texttt{free\_low}); reconstructed follow-up $125/236$ vs.\ published $124/236$. & Local overlap and follow-up certification insufficient; treatment effect not estimated. \\
H13 $\cdot$ PXD055932 & 304 (116 30d deaths) & Cross-system vs.\ CFH: S100A8 $+0.376$, S100A9 $+0.272$ $z$; mortality OR ratio $1.035$ ($0.632, 1.695$). & Positive protein contrast without 30-day mortality-risk separation. \\
H13 $\cdot$ Mi, CMAISE & Mi 302; CMAISE 342 / auxiliary 338 & Same direction: Mi $+0.292$ ($-0.096, 0.680$), $+0.157$ ($-0.191, 0.505$). CMAISE primary: $-0.573$ ($-0.729, -0.417$), $-0.769$ ($-0.905, -0.633$). & Mi intervals cross zero; CMAISE estimates reverse direction relative to PXD055932. \\
H13 $\cdot$ Auxiliary RNA & 4 bulk cohorts; Kwok $272{,}993$ cells & Mature neutrophils ($205{,}527$ cells) contribute largest captured \textit{S100A8}/\textit{S100A9} UMI totals; eligibility of the fixed CMAISE RNA sub-library remains to be verified. & Cellular RNA profile only; plasma protein origin unresolved. \\
H09 $\cdot$ PXD055932 & 310 (118 30d deaths) & CFH vs.\ non-complement $+0.279$, vs.\ effectors $+0.255$ (both BH $q = 0.105$); vs.\ ALB $-0.097$, vs.\ HRG/PROC/F13B $+0.367$ to $+0.477$. & Direction depends on reference choice; drop-ALB result is a post-hoc sensitivity check. \\
\end{longtable}
\endgroup

\begingroup
\footnotesize
\setlength{\tabcolsep}{3.5pt}
\renewcommand{\arraystretch}{1.15}
\setlength{\LTleft}{0pt}
\setlength{\LTright}{\fill}
\setlength{\LTcapwidth}{\linewidth}
\captionsetup{singlelinecheck=false,justification=raggedright}
\setlength{\sepsisTableWidth}{\dimexpr\linewidth-6\tabcolsep\relax}
\begin{longtable}{@{}>{\raggedright\arraybackslash}p{0.14\sepsisTableWidth}>{\raggedright\arraybackslash}p{0.15\sepsisTableWidth}>{\raggedright\arraybackslash}p{0.37\sepsisTableWidth}>{\raggedright\arraybackslash}p{0.34\sepsisTableWidth}@{}}
\caption{\textbf{Round 2 numerical summary: within-patient day-1 to day-7 relative metabolite shifts in Ferrario/ALBIOS ($n = 20$).} All four analyses evaluate the same 20 paired patients across 28-day and 90-day follow-up windows. Entries report non-survivor minus survivor differences in within-patient relative $\log_2(\text{D7}/\text{D1})$ shifts (95\% CI in parentheses).}\label{tab:si-sepsis-r2}\\
\toprule
\textbf{Branch} & \textbf{Window (NS/S)} & \textbf{Contrast and between-group estimate} & \textbf{Assessment of original prediction} \\
\midrule
\endfirsthead
\multicolumn{4}{@{}l}{\footnotesize\textbf{Table~\thetable\ (continued)}}\\
\toprule
\textbf{Branch} & \textbf{Window (NS/S)} & \textbf{Contrast and between-group estimate} & \textbf{Assessment of original prediction} \\
\midrule
\endhead
\midrule
\multicolumn{4}{r@{}}{\footnotesize\emph{Continued on next page}}\\
\endfoot
\bottomrule
\endlastfoot
H24 $\cdot$ LPC & 28d (9/11) & 7 LPC relative to 26 PC/SM median diff.: $-0.226$ ($-1.724, 1.247$). & Point estimate matches predicted smaller relative gain in NS, but sign flips under LOO or removing LPC C20:3 ($+0.077$). \\
H27 $\cdot$ PC & 28d (9/11) & $\text{PC} - \text{SM}$ median diff.: $-0.291$ ($-0.692, 0.146$). & Opposes predicted greater PC gain in NS (LOO 20/20 negative), though CI crosses zero and cannot exclude it. \\
H26 $\cdot$ Phe & 28d (9/11) & Phe vs.\ fixed 19 AAs (excl.\ Phe, Tyr) mean diff.: $+0.374$ ($-0.276, 1.025$); separate Phe/Tyr diff.: $+0.489$ ($-0.390, 1.368$). & Group means of the relative metric are negative in both outcomes (NS $-0.289$, S $-0.663$); positive diff.\ reflects smaller relative decline in NS, with both CIs crossing zero. \\
H23 $\cdot$ Arg & 90d (11/9) & Arg vs.\ Orn mean diff.: $+0.252$ ($-1.308, 1.811$); vs.\ fixed 19 AAs (excl.\ Arg, Orn): $+0.180$ ($-1.013, 1.372$). & Point estimates oppose predicted recovery deficit; wide CIs and LOO sign flips preclude determination. \\
\end{longtable}
\endgroup

\begingroup
\footnotesize
\setlength{\tabcolsep}{3.5pt}
\renewcommand{\arraystretch}{1.15}
\setlength{\LTleft}{0pt}
\setlength{\LTright}{\fill}
\setlength{\LTcapwidth}{\linewidth}
\captionsetup{singlelinecheck=false,justification=raggedright}
\setlength{\sepsisTableWidth}{\dimexpr\linewidth-8\tabcolsep\relax}
\begin{longtable}{@{}>{\raggedright\arraybackslash}p{0.21\sepsisTableWidth}>{\raggedright\arraybackslash}p{0.14\sepsisTableWidth}>{\raggedright\arraybackslash}p{0.23\sepsisTableWidth}>{\raggedright\arraybackslash}p{0.13\sepsisTableWidth}>{\raggedright\arraybackslash}p{0.29\sepsisTableWidth}@{}}
\caption{\textbf{Reference-set, covariate, and leave-one-patient-out (LOO) sensitivity summary.} Leave-one-patient-out (LOO) checks evaluate single-patient influence within the same sample (each Ferrario LOO estimate uses $n = 19$ patients). H12 yielded no estimable treatment effect for sensitivity testing.}\label{tab:si-sepsis-sensitivity}\\
\toprule
\textbf{Branch and primary metric} & \textbf{Primary est.} & \textbf{LOO / sensitivity range} & \textbf{Sign stability} & \textbf{Key sensitivity takeaway} \\
\midrule
\endfirsthead
\multicolumn{5}{@{}l}{\footnotesize\textbf{Table~\thetable\ (continued)}}\\
\toprule
\textbf{Branch and primary metric} & \textbf{Primary est.} & \textbf{LOO / sensitivity range} & \textbf{Sign stability} & \textbf{Key sensitivity takeaway} \\
\midrule
\endhead
\midrule
\multicolumn{5}{r@{}}{\footnotesize\emph{Continued on next page}}\\
\endfoot
\bottomrule
\endlastfoot
\textbf{H08} (Mi APOA1-to-S100 slope) & $-0.272$ & Plate-adjusted $-0.248$ (Holm $p = 1.12 \times 10^{-6}$) & Stable ($-$) & Plate adjustment preserves both negative conditional slope and $D \ge 0$ double-high patients. \\
\textbf{H09} (PXD CFH vs.\ non-compl.\ A) & $+0.279$ ($q = 0.105$) & Drop ALB: $+0.404$ ($q = 0.019$) & Reference-dependent & ALB ($-0.097$) opposes HRG/PROC/F13B ($+0.367$ to $+0.477$). \\
\textbf{H13} (PXD S100A8 / A9) & $+0.376 / +0.272$ $z$ & \textbf{S100A9} attenuates after SOFA adjustment or dropping HRG; reverses to negative in CMAISE & Cross-cohort flip & Single-cohort positive contrast does not generalize across cohorts. \\
\textbf{H24} (Ferrario 7 LPC $-$ PC/SM) & $-0.226$ & LOO $[-0.510, +0.007]$; drop LPC C20:3 $+0.077$ & Unstable (flips) & Sensitive to individual patients and a single LPC species. \\
\textbf{H27} (Ferrario PC $-$ SM) & $-0.291$ & LOO $[-0.320, -0.203]$ (20/20 $< 0$) & Stable ($-$) & Consistently opposite to predicted PC gain; primary CI crosses zero. \\
\textbf{H26} (Ferrario Phe $-$ 19 AAs, excl.\ Phe/Tyr) & $+0.374$ (90d $+0.610$) & LOO 20/20 $> 0$; 90d bootstrap CI $> 0$ while Welch CI crosses 0 & Stable ($+$) & Direction stable under single-patient deletion within sample; primary CI still crosses zero. \\
\textbf{H23} (Ferrario Arg $-$ Orn) & $+0.252$ & LOO crosses zero / flips sign & Unstable (flips) & Cannot establish whether relative arginine recovery is impaired. \\
\end{longtable}
\endgroup

\FloatBarrier
\newlength{\esccTableWidth}
\section{Single-Cell B-Lineage Analysis in Esophageal Cancer Metastasis}\label{sec:si-case-escc}

This supplement supports Section~\ref{sec:exp:escc} and Figures~\ref{fig:escc-decoupling}--\ref{fig:escc-pathway-cxcl13}. It presents the task prompt and analysis rules (SI Section~\ref{sec:si-escc-s1}), patient cohort and analysis samples (SI Section~\ref{sec:si-escc-s1-2}), search trajectory and core research decisions (SI Section~\ref{sec:si-escc-s2}), bioinformatics analysis methods (SI Section~\ref{sec:si-escc-s3}), cohort coverage and model diagnostics (SI Section~\ref{sec:si-escc-s4}), and pathway auditing alongside \textup{CXCL13} sensitivity analysis (SI Section~\ref{sec:si-escc-s5}). It concludes with proposed follow-up validation studies (SI Table~\ref{tab:si-escc-follow-up}).

\subsection{Task Prompt and Analysis Rules}\label{sec:si-escc-s1}\label{sec:si-case-escc-input}

The task specifies within-patient paired lymph-node comparisons across B and plasma cells without pre-specifying target genes, a dominant biological response, or a statistical pipeline, permitting direct reporting of patient-specific or unsupported patterns.

\begin{caseprompt}{Goal.md | Paired Lymph-Node B-Lineage Analysis in ESCC}
This project aims to use single-cell transcriptomic data from 110 tissue samples across 40 patients with esophageal squamous cell carcinoma (ESCC) to investigate B-cell and plasma-cell responses associated with lymph-node metastasis. For patients with paired lymph-node samples in the cohort, compare metastatic lymph nodes (MET) against tumor-free lymph nodes (LYMPH) within the same patient to characterize differences across cellular composition, cell states, and gene expression, identifying which features recur across multiple patients and which vary by individual.

This study is framed as an exploratory discovery analysis. Based on observed differences in the data, prioritize cell subpopulations, candidate genes, and biological processes of interest, and formulate biological questions testable in independent samples or follow-up experiments. Do not pre-specify which B-cell response dominates; if different patients exhibit distinct response patterns, or if existing data are insufficient to support a shared pattern, report this faithfully.
\end{caseprompt}

\subsection{Patient Cohort and Analysis Samples}\label{sec:si-escc-s1-2}\label{sec:si-case-escc-cohort}

This case uses the initially supplied ESCC single-cell transcriptomic dataset, comprising 110 tissue sequencing libraries from 40 patients. Analysis begins from the gene count matrix, cell-state annotations, and patient--tissue pairing metadata. Across the full cohort, 13 patients have both metastatic lymph-node (MET) and tumor-free lymph-node (LYMPH) samples, enabling within-patient tissue comparisons. The in-house count matrix, cell annotations, and sample metadata are not publicly available.

The primary analysis includes the 12 paired patients profiled on the C4 platform, comprising 26 lymph-node libraries (14 MET and 12 LYMPH) and $102{,}676$ B and plasma cells. Statistical comparisons treat the patient as the independent unit: multiple libraries from the same tissue of the same patient do not count as independent replicates. Each cell-state analysis includes only patients whose cell counts meet the eligibility threshold in both tissues, with actual pair counts reported alongside each result. Three lymph-node libraries from 1 additional patient profiled on the 10X platform serve solely as a descriptive reference and are excluded from the primary analysis.

\subsection{Search Trajectory and Core Research Decisions}\label{sec:si-escc-s2}

\begin{figure}[!htbp]
\centering
\includegraphics[width=\linewidth]{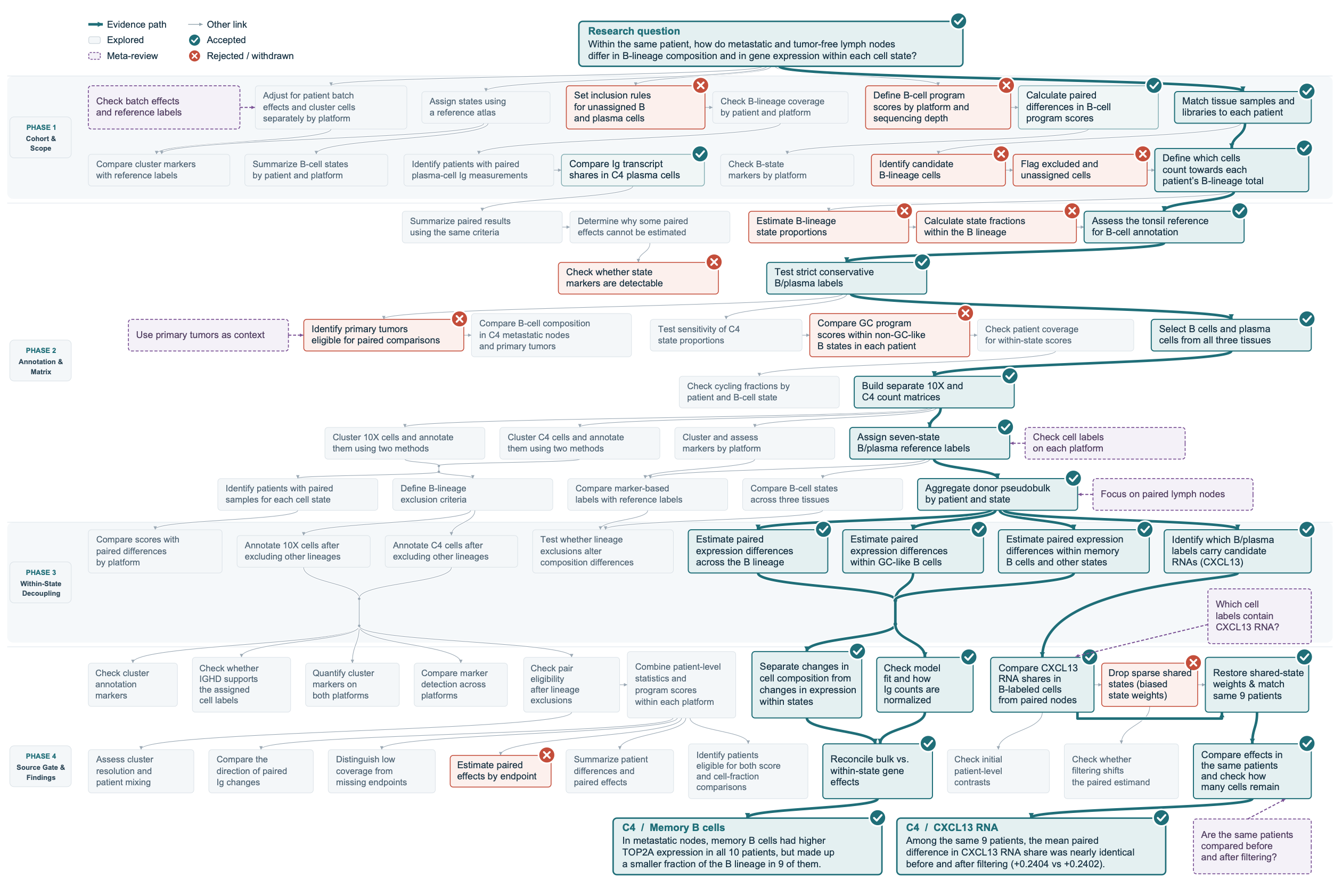}
\caption{\textbf{CytoCoScientist search trajectory in the ESCC paired lymph-node case.} Branches represent candidate analytical paths explored around the research goal, thick dark teal lines trace the final evidence-supporting path, terracotta crosses ($\times$) mark withdrawn or rejected variants, and dashed purple boxes mark outer-loop meta-review checkpoints that guided subsequent analyses.}
\label{fig:si-escc-dag}
\end{figure}

During analysis, CytoCoScientist separated two distinct types of change: shifts in the proportion of each state within the B lineage, and gene expression differences within a fixed cell state. Along this main line of inquiry, CytoCoScientist evaluated paired expression across the bulk B lineage and six cell states, checked directional consistency in every patient, withdrew an intermediate version that altered state weights by dropping sparse shared states, and, prompted by outer-loop meta-review, recomputed pre- and post-filtering \textup{CXCL13} differences over the same set of retained patients (SI Figure~\ref{fig:si-escc-dag} and SI Table~\ref{tab:si-escc-decisions}).

Subsequent within-cohort follow-up analyses (pre-ranked gene-set enrichment, patient-level pathway testing, and T-cell marker co-detection in \textup{CXCL13}-positive B-labeled cells) evaluated across-patient pathway consistency and B-cell identity resolution within the C4 cohort.

\begingroup
\footnotesize
\setlength{\tabcolsep}{3pt}
\renewcommand{\arraystretch}{1.15}
\setlength{\LTleft}{0pt}
\setlength{\LTright}{\fill}
\setlength{\LTcapwidth}{\linewidth}
\captionsetup{singlelinecheck=false,justification=raggedright}
\setlength{\esccTableWidth}{\dimexpr\linewidth-10\tabcolsep\relax}
\begin{longtable}{@{}>{\raggedright\arraybackslash}p{0.16\esccTableWidth}>{\raggedright\arraybackslash}p{0.16\esccTableWidth}>{\raggedright\arraybackslash}p{0.17\esccTableWidth}>{\raggedright\arraybackslash}p{0.20\esccTableWidth}>{\raggedright\arraybackslash}p{0.17\esccTableWidth}>{\raggedright\arraybackslash}p{0.14\esccTableWidth}@{}}
\caption{\textbf{Core research decisions and revisions in the ESCC paired lymph-node case.}}\label{tab:si-escc-decisions}\\
\toprule
\textbf{Research decision} & \textbf{Prior formulation} & \textbf{Proposed change or control} & \textbf{Observed evidence / feedback} & \textbf{Action and status} & \textbf{Role in final conclusion} \\
\midrule
\endfirsthead
\multicolumn{6}{@{}l}{\footnotesize\textbf{Table~\thetable\ (continued)}}\\
\toprule
\textbf{Research decision} & \textbf{Prior formulation} & \textbf{Proposed change or control} & \textbf{Observed evidence / feedback} & \textbf{Action and status} & \textbf{Role in final conclusion} \\
\midrule
\endhead
\midrule
\multicolumn{6}{r@{}}{\footnotesize\emph{Continued on next page}}\\
\endfoot
\bottomrule
\endlastfoot
\textbf{Drill down} from bulk to within-state and per-patient contrasts & Bulk B-lineage \textit{MKI67}/\textit{TOP2A} elevation and GC-like expansion only & Hold seven states fixed, enforce $\ge 20$ cells per tissue, and compare within patients & P13 bulk \textit{TOP2A} falls while memory B \textit{TOP2A} rises; 10/10 memory B up, 9/10 contracting in share & Adopted; added three-term linear count-share split ($+0.60$ composition / $+0.29$ shared within-state / $+0.10$ one-sided remainder) & Core finding 1 (Figure~\ref{fig:escc-decoupling}) \\
\textbf{Withdraw intermediate variant} dropping sparse shared states & Additional removal of sparse shared states during \textup{CXCL13} sensitivity decomposition & Audit effect of dropping sparse states on state weights and total share difference & Dropping sparse states distorted state weights and biased the share decomposition & Withdrawn ($\times$); restored original state-weight definition & Discarded variant (SI Figure~\ref{fig:si-escc-dag}) \\
\textbf{Fix patient denominator} before and after \textup{CXCL13} filtering & Direct comparison of 12-patient unfiltered mean ($+0.928$) vs.\ 9-patient joint-filtered mean ($+0.240$) & Prompted by outer-loop meta-review, compare pre- and post-filtering over the exact 9 retained patients & Within the same 9 patients, mean is $+0.2404$ before vs.\ $+0.2402$ after (diff.\ 95\% CI crosses zero), with bidirectional patient shifts & Adopted; established that mean drop reflects patient turnover rather than filtering & Core finding 2 (Figure~\ref{fig:escc-pathway-cxcl13}c--d) \\
\textbf{Switch pathway audit} from pre-ranked GSEA to donor sign-flip (same-cohort follow-up) & Pre-ranked GSEA suggesting GC-like cell-cycle and interferon enrichment & Compute per-patient mean gene-set $\log_2$ ratios with synchronized sign-flip and leave-one-out tests & Only bulk B-lineage E2F and G2/M pass 28-test focused correction and leave-one-out; GC not significant; $0/337$ full Hallmark $q < 0.05$ & Adopted; withheld positive GC pathway claim & Pathway boundary (Figure~\ref{fig:escc-pathway-cxcl13}a) \\
\textbf{Add T-cell marker co-detection} in B-labeled \textup{CXCL13}$^+$ cells (same-cohort follow-up) & Stable 9-patient mean could be misread as proof of endogenous B-cell expression & Quantify fraction of MET original B-labeled \textup{CXCL13}$^+$ cells co-detecting $\ge 2$ T-cell markers & $95.7\%$ co-detect at least two T-cell markers & Adopted; bounded claim to elevated capture under B labels pending in situ localization & Source boundary (Figure~\ref{fig:escc-pathway-cxcl13}c) \\
\end{longtable}
\endgroup

\subsection{Bioinformatics Analysis Methods}\label{sec:si-escc-s3}

\begin{figure}[!htbp]
\centering
\includegraphics[width=\linewidth]{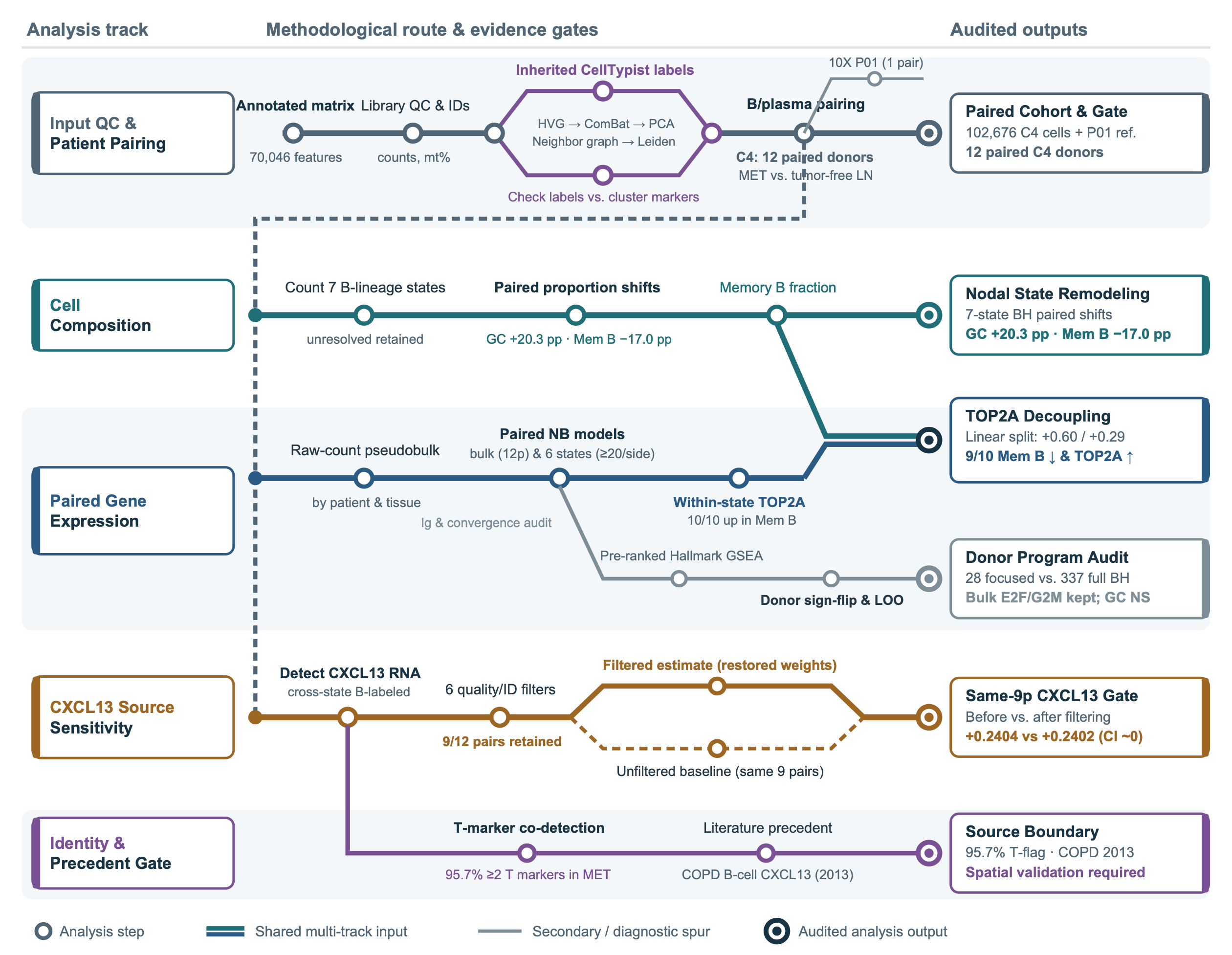}
\caption{\textbf{Finalized bioinformatics analysis workflow for ESCC paired lymph nodes.} Twelve paired patients on the C4 platform enter within-patient statistical comparisons, whereas the single 10X paired patient serves as a descriptive reference. The cell-composition and paired gene-expression tracks converge at the memory B-cell comparison, pairing cellular state proportions with within-state \textit{TOP2A} expression per patient. In GC-like B cells, neither E2F nor G2/M enrichment passed correction across the 28 focused tests. Among the nine patients retained after joint filtering, the 95\% confidence interval for the mean change in the \textup{CXCL13} tissue contrast after filtering spans zero, although individual estimates moved in both directions. T-cell marker co-detection leaves the cellular source of the RNA unresolved.}
\label{fig:si-escc-workflow}
\end{figure}

Analyses begin from the gene count matrix, existing cell-state annotations, and patient--tissue pairing metadata (SI Section~\ref{sec:si-escc-s1-2}). SI Figure~\ref{fig:si-escc-workflow} maps the four analytical steps detailed below.

\subsubsection{Paired Design and Cellular Composition}\label{sec:si-escc-s3-1}

The primary analysis includes only patients on the C4 platform with both a metastatic lymph node (MET) and a tumor-free lymph node (LYMPH), pooling counts across multiple libraries from the same tissue of the same patient. For each patient and tissue, proportions of the seven cell states use all captured B and plasma cells (including unresolved cells) as the denominator (Figure~\ref{fig:escc-decoupling}a,b). Within-patient MET-minus-LYMPH proportion differences are computed first, followed by across-patient mean differences and $t$-based 95\% confidence intervals. Two-sided one-sample $t$-tests evaluate whether mean within-patient differences deviate from zero, with multiple-testing correction applied jointly across all seven states.

\subsubsection{Within-State Differential Expression and Bulk-Difference Decomposition}\label{sec:si-escc-s3-2}

Differential expression treats patients as biological replicates~\cite{Squair2021SingleCellDE} and compares tissues across both the bulk B lineage and individual cell states~\cite{Crowell2020muscat} (Figure~\ref{fig:escc-decoupling}c). Within-state testing requires at least 20 cells of that state in both tissues of the same patient. Raw counts from eligible cells are summed within each patient and tissue to form pseudobulk profiles. Across all $70{,}046$ raw features, each contrast tests genes with counts $\ge 10$ in at least $n$ pseudobulk profiles, where $n$ is the number of paired patients in that contrast. Negative-binomial models containing patient and tissue terms (\texttt{\textasciitilde{} patient + tissue}) are fit with PyDESeq2 (v0.5.4)~\cite{Muzellec2023PyDESeq2}, using Wald tests to estimate the MET-versus-LYMPH tissue effect and reporting unshrunk $\log_2$ fold changes. Benjamini--Hochberg correction~\cite{Benjamini1995FDR} is applied separately to seven-state composition, bulk B-lineage expression, and pooled six-state expression families.

Figure~\ref{fig:escc-decoupling}d separately decomposes the \textit{TOP2A} tissue difference on the linear scale of counts per $10{,}000$ total counts. Within each patient, the total difference is partitioned into a state-composition term, a shared within-state expression term, and a one-sided state remainder that sum directly to the linear total difference, and then averaged across patients.

\subsubsection{Pathway Enrichment and Across-Patient Consistency}\label{sec:si-escc-s3-3}

In the same-cohort follow-up pathway analysis (Figure~\ref{fig:escc-pathway-cxcl13}a), genes passing convergence checks with finite differential-expression Wald statistics are ranked for pre-ranked gene-set enrichment analysis (GSEA)~\cite{Subramanian2005GSEA} against the MSigDB Hallmark collection~\cite{Liberzon2015Hallmark} using GSEApy (v1.1.11)~\cite{Fang2023GSEApy}.

To evaluate whether enrichment reflects concordant directional shifts across patients, normalized expression $\log_2$ ratios (MET versus LYMPH, pseudocount $0.5$) are computed per member gene within each patient and averaged with equal weights across set members. Two-sided exact sign-flip tests operate at the patient level, flipping the signs of all member differences simultaneously within each patient to preserve inter-gene correlations. Multiple-testing correction is applied separately across the 28 post-exploration focused comparisons and all 337 testable Hallmark comparisons, alongside leave-one-patient-out checks to evaluate sensitivity to individual patients.

\subsubsection{Cell Filtering and Source Auditing for \textup{CXCL13}}\label{sec:si-escc-s3-4}

Filtering variants for \textup{CXCL13} (Figure~\ref{fig:escc-pathway-cxcl13}b--d) are built from cell quality, B-cell identity markers, and mixed-lineage flags. Under each filter, the analysis first identifies the subset of patients retaining sufficient cells in both tissues and then recomputes the unfiltered baseline strictly over that same patient subset. Pre- and post-filtering \textup{CXCL13} RNA share differences (MET minus LYMPH, per $10{,}000$ total counts) are compared within each patient, reporting the mean change across patients, 95\% confidence interval, and individual patient directions. Marker co-detection and local batch checks appear in SI Section~\ref{sec:si-escc-s5}.

\subsection{Cohort Coverage, Model Diagnostics, and Per-Patient Expression}\label{sec:si-escc-s4}

\begin{figure}[!htbp]
\centering
\includegraphics[width=\linewidth]{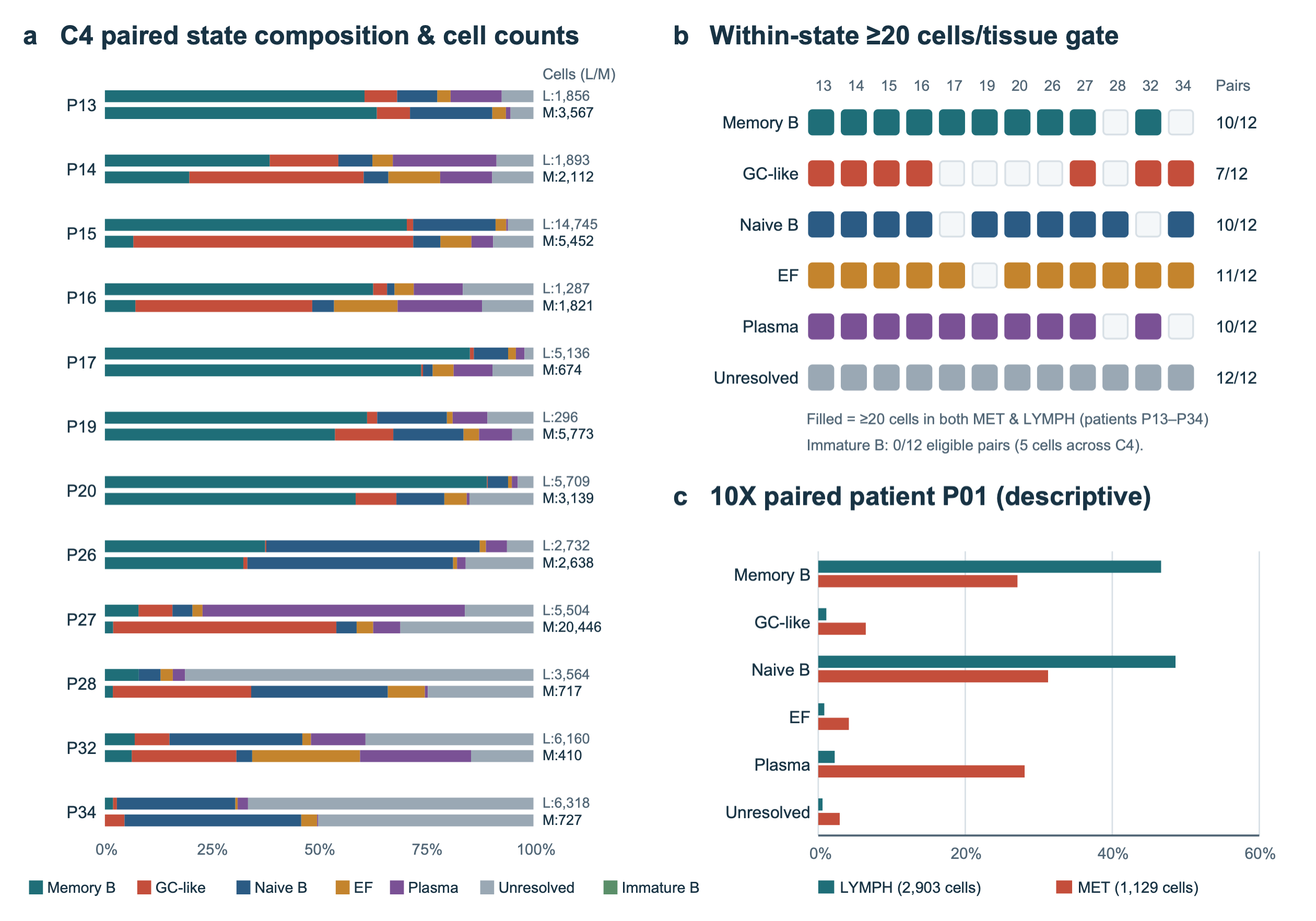}
\caption{\textbf{Patient, cell-state, and platform coverage.} \textbf{a,} Cell counts and seven-state composition across paired C4 lymph-node samples ($n = 12$ patients). \textbf{b,} Patient pairs meeting the bilateral $\ge 20$-cell eligibility threshold per cell state. \textbf{c,} Descriptive summary of the single 10X paired patient (P01), shown as a platform reference outside the primary C4 statistical cohort.}
\label{fig:si-escc-coverage}
\end{figure}

Under the denominator and bilateral $\ge 20$-cell rules in SI Sections~\ref{sec:si-escc-s3-1} and~\ref{sec:si-escc-s3-2}, patients with insufficient coverage are marked non-estimable rather than zero-imputed, yielding 12, 7, and 10 eligible patient pairs for the bulk B lineage, GC-like B cells, and memory B cells, respectively (SI Figure~\ref{fig:si-escc-coverage}). Because eligible pair counts differ across states, cross-state comparisons of statistical significance reflect differences in sample power as well as effect size.

\begin{figure}[!htbp]
\centering
\includegraphics[width=\linewidth]{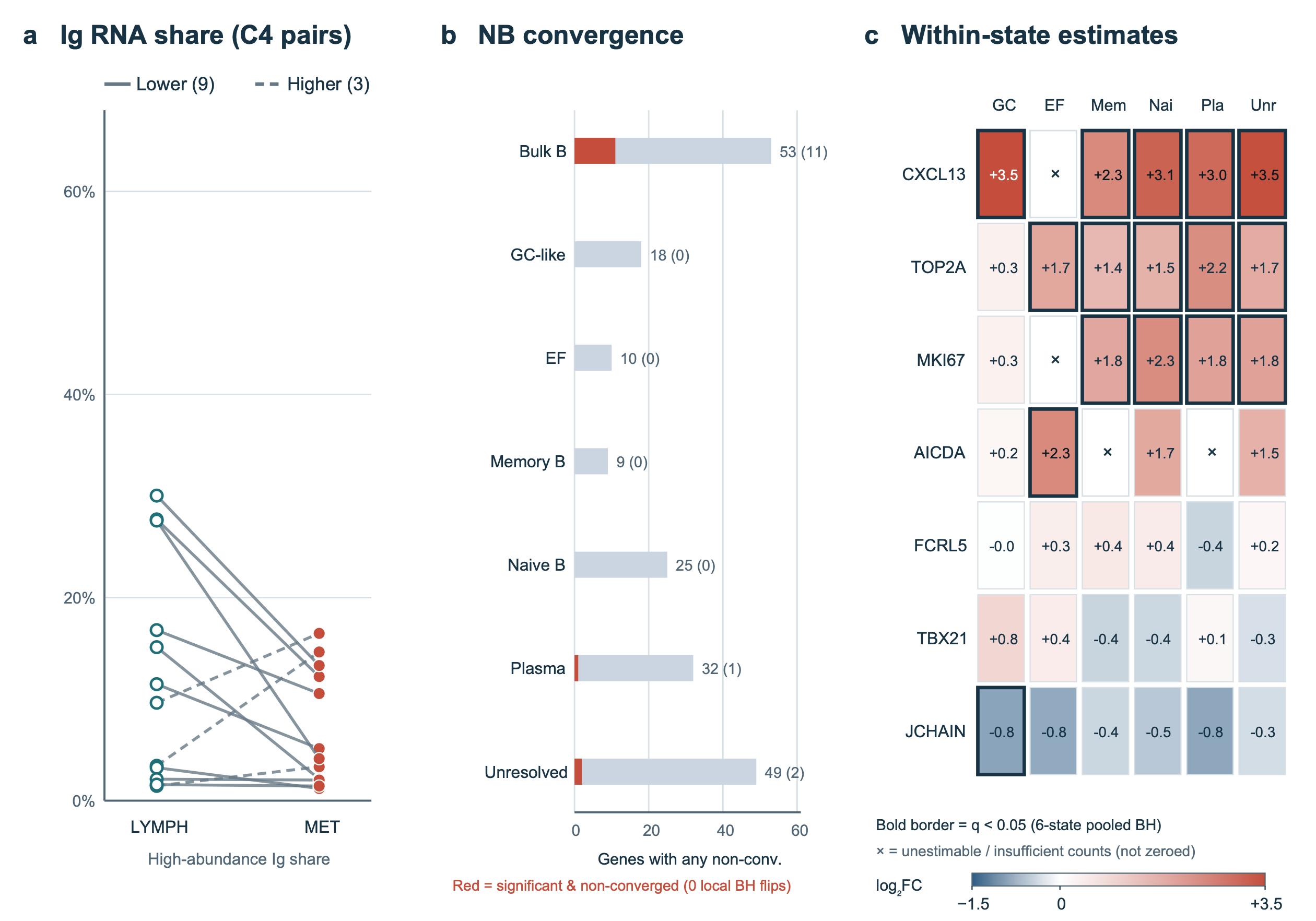}
\caption{\textbf{Model diagnostics and within-state candidate gene estimates.} \textbf{a,} Immunoglobulin RNA count shares across paired C4 lymph-node samples. \textbf{b,} Convergence diagnostics for paired negative-binomial models across bulk and within-state contrasts. \textbf{c,} Paired differential expression estimates for representative genes across individual B-lineage states.}
\label{fig:si-escc-diagnostics}
\end{figure}

\begin{figure}[!htbp]
\centering
\includegraphics[width=\linewidth]{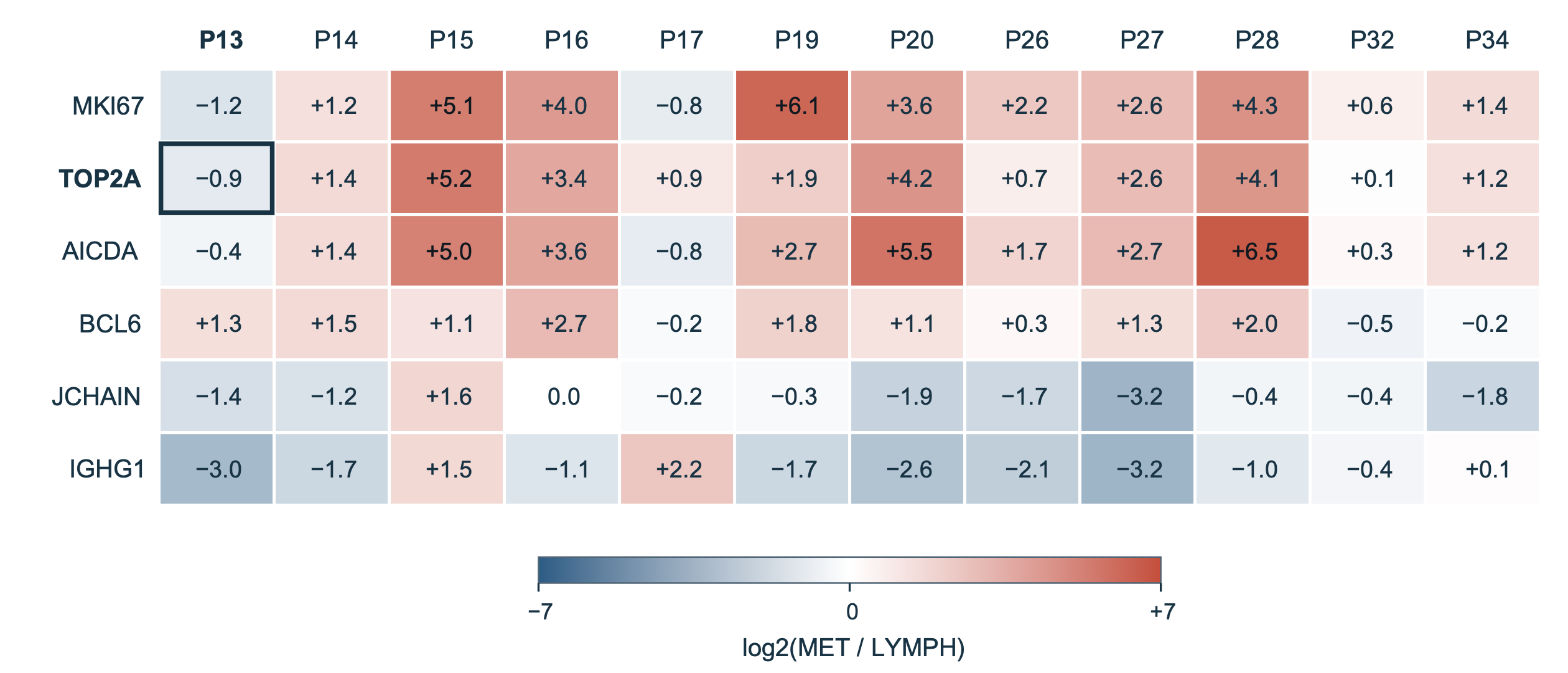}
\caption{\textbf{Paired bulk B-lineage expression differences for six representative genes across all 12 C4 patients.} Each column corresponds to one patient's paired lymph nodes, and each cell shows the per-patient normalized expression $\log_2$ ratio (MET over LYMPH, pseudocount $0.5$, rounded to one decimal place). Positive log ratios indicate higher expression in MET and negative log ratios indicate higher expression in LYMPH. The color scale is centered at zero and uses unrounded log ratios; a printed value of 0.0 need not indicate exactly equal expression. The box highlights \textit{TOP2A} in P13.}
\label{fig:si-escc-heatmap}
\end{figure}

The within-state $q$ values in Figure~\ref{fig:escc-decoupling}c are adjusted across the pooled six-state testing family, and all within-state contrasts condition on the provided cell-state annotations, leaving finer subpopulation shifts within the memory B label unmodeled (SI Figure~\ref{fig:si-escc-diagnostics}). SI Figure~\ref{fig:si-escc-heatmap} retains all 12 patients across six representative genes in the bulk B lineage: bulk \textit{MKI67} and \textit{TOP2A} decline in P13, and bulk \textit{MKI67} declines in P17, contrasting with the rise in memory B-cell \textit{TOP2A} in P13 shown in Figure~\ref{fig:escc-decoupling}c.

\subsection{Pathway Audit and \textup{CXCL13} Sensitivity Analysis}\label{sec:si-escc-s5}

\begin{figure}[!htbp]
\centering
\includegraphics[width=\linewidth]{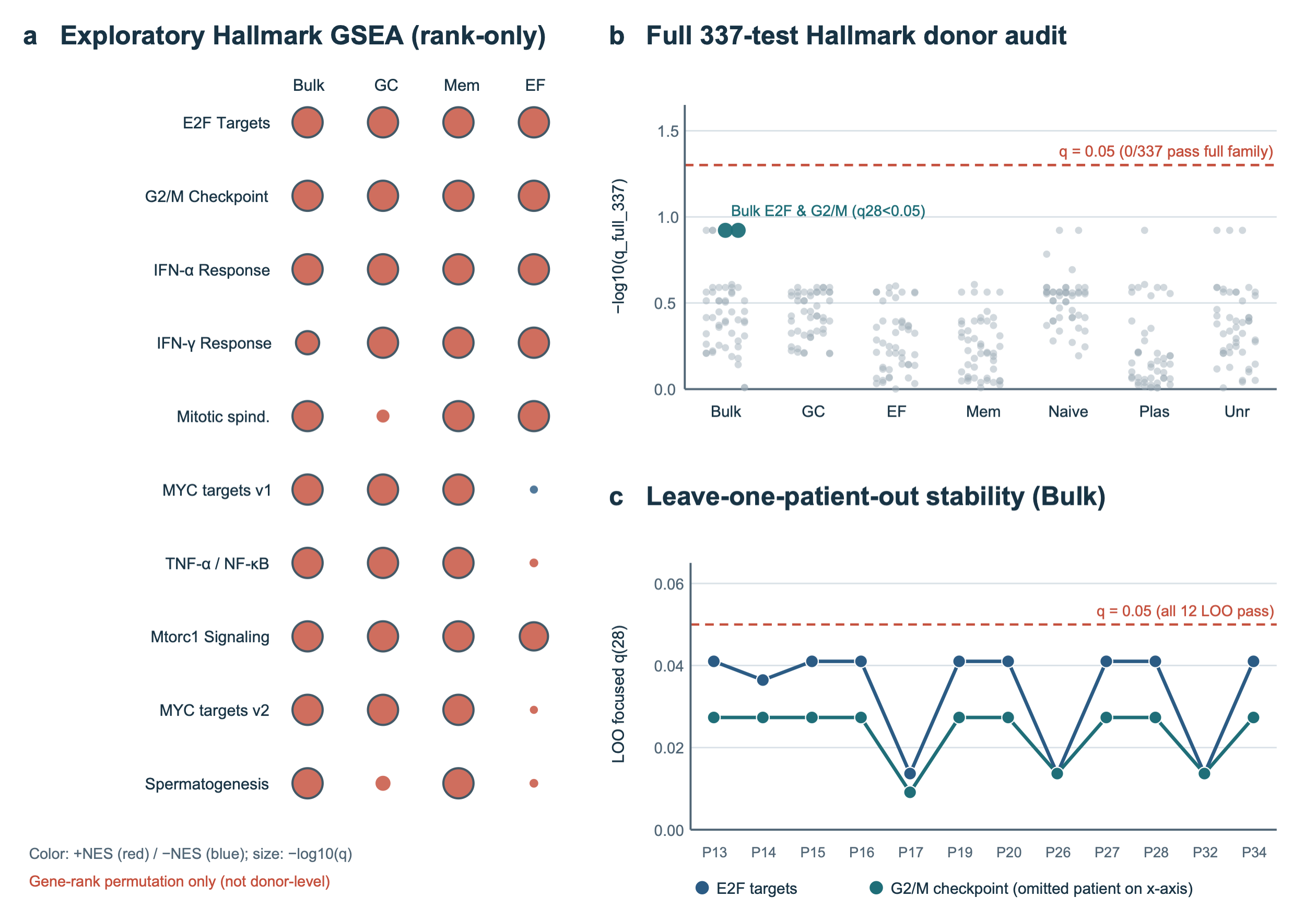}
\caption{\textbf{Pathway enrichment audit from pre-ranked GSEA to patient-level sign-flip testing.} \textbf{a,} Exploratory MSigDB Hallmark pre-ranked GSEA across bulk B-lineage and within-state contrasts. \textbf{b,} Patient-level exact sign-flip test results across the full 337-test Benjamini--Hochberg correction family. \textbf{c,} Leave-one-patient-out sensitivity checks for bulk B-lineage E2F targets and G2/M checkpoint gene sets.}
\label{fig:si-escc-pathway-audit}
\end{figure}

None of the 337 testable Hallmark comparisons reached $q < 0.05$ after full-family correction. Within the 28 post-exploration focused comparisons, bulk B-lineage E2F targets ($q = 0.021$) and G2/M checkpoint ($q = 0.014$) remained significant under leave-one-patient-out deletion, whereas both gene sets yielded $q = 0.208$ within GC-like B cells (SI Figure~\ref{fig:si-escc-pathway-audit}).

\begin{figure}[!htbp]
\centering
\includegraphics[width=\linewidth]{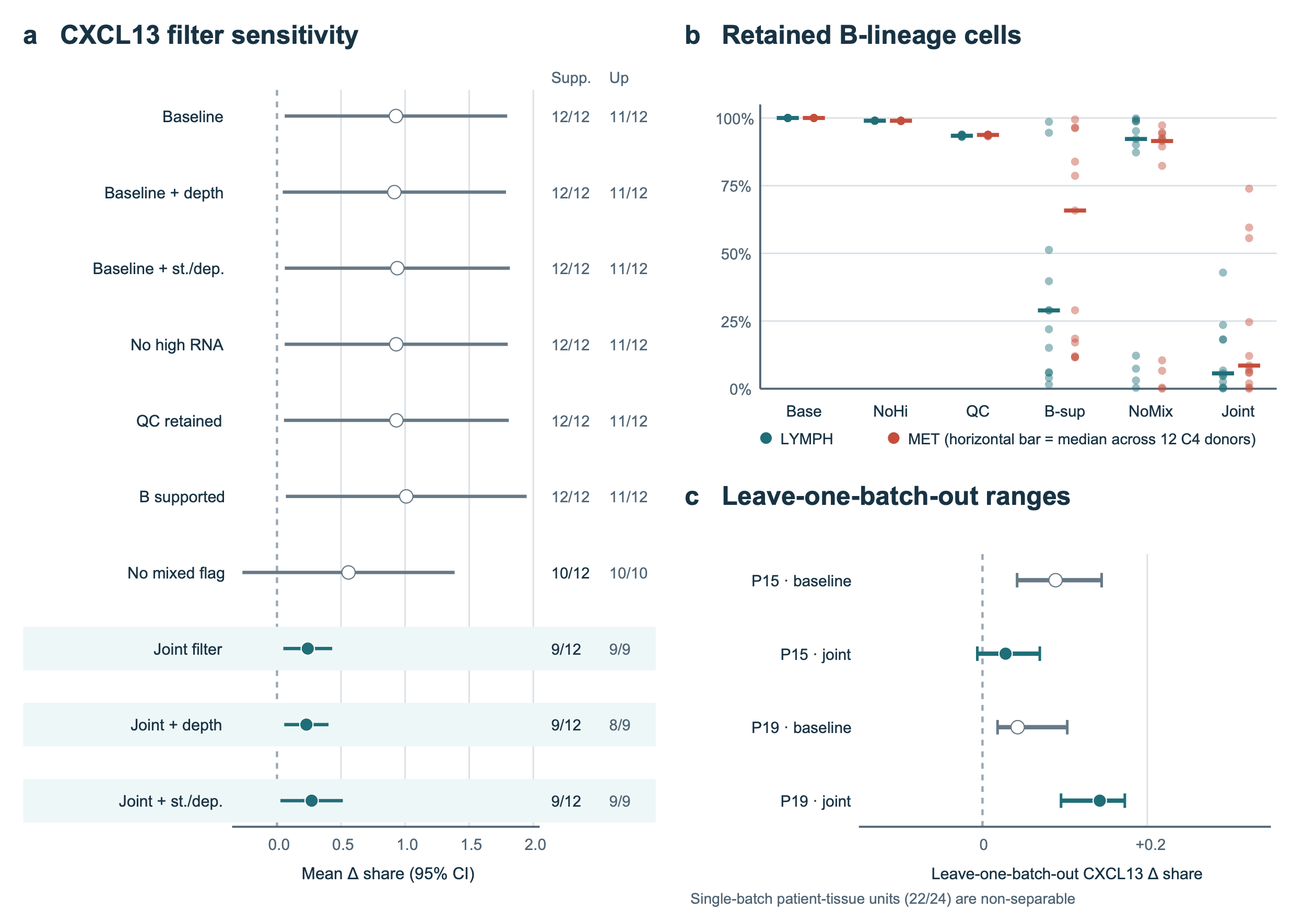}
\caption{\textbf{Sensitivity of \textup{CXCL13} estimates to cell filtering and processing batches.} \textbf{a,} Matched-patient paired count-share differences across six filtering rules. \textbf{b,} Fraction of B-lineage cells retained under each filtering condition. \textbf{c,} Range of estimates under leave-one-batch-out deletion in multi-batch patient--tissue units.}
\label{fig:si-escc-cxcl13-sens}
\end{figure}

Joint filtering reduced the number of eligible patient pairs from 12 to 9. The mean MET-minus-LYMPH \textup{CXCL13} RNA count-share difference was $+0.928$ before filtering across all 12 patients and $+0.240$ after filtering across the nine retained patients; these means therefore represent different patient sets. For each filtering variant, the unfiltered baseline was calculated using the same patients retained by that filter. Among the nine patients retained under joint filtering, the mean difference was $+0.2404$ before filtering and $+0.2402$ afterward, with a near-zero mean change (95\% CI, $-0.0745$ to $+0.0740$ per $10{,}000$ total counts), although individual changes were bidirectional. The apparent decrease from $+0.928$ to $+0.240$ thus largely reflected changes in patient composition rather than a filtering effect within the same patients. Identity auditing showed that $95.7\%$ of metastatic-node records originally annotated as B cells with detectable \textup{CXCL13} co-detected at least two T-cell markers (Figure~\ref{fig:escc-pathway-cxcl13}c), indicating ambiguous cellular origin. Final estimates retain the original state weights and discard the intermediate variant that dropped sparse shared states. In the two patient--tissue units profiled across multiple processing batches, leave-one-batch-out checks quantified local batch sensitivity (SI Figure~\ref{fig:si-escc-cxcl13-sens}).

\subsection{Follow-Up Validation Plan}\label{sec:si-escc-follow-up}

Building on the cohort analyses in SI Sections~\ref{sec:si-escc-s2} and~\ref{sec:si-escc-s5}, CytoCoScientist proposed the follow-up experiments and audits in SI Table~\ref{tab:si-escc-follow-up}. RNA localization in intact tissue and cell counts per unit of tissue would help identify the cellular source of \textup{CXCL13} and determine whether B-cell subpopulations increase in absolute abundance.

\begingroup
\footnotesize
\setlength{\tabcolsep}{3pt}
\renewcommand{\arraystretch}{1.15}
\setlength{\LTleft}{0pt}
\setlength{\LTcapwidth}{\linewidth}
\setlength{\esccTableWidth}{\dimexpr\linewidth-6\tabcolsep\relax}
\begin{longtable}{@{}>{\raggedright\arraybackslash}p{0.18\esccTableWidth}>{\raggedright\arraybackslash}p{0.25\esccTableWidth}>{\raggedright\arraybackslash}p{0.32\esccTableWidth}>{\raggedright\arraybackslash}p{0.25\esccTableWidth}@{}}
\caption{\textbf{Proposed experimental and auditing follow-up tests to distinguish competing explanations in paired ESCC lymph nodes.}}\label{tab:si-escc-follow-up}\\
\toprule
\textbf{Open question} & \textbf{What the data show} & \textbf{What to do next} & \textbf{How to interpret the result} \\
\midrule
\endfirsthead
\multicolumn{4}{@{}l}{\footnotesize\textbf{Table~\thetable\ (continued)}}\\
\toprule
\textbf{Open question} & \textbf{What the data show} & \textbf{What to do next} & \textbf{How to interpret the result} \\
\midrule
\endhead
\midrule
\multicolumn{4}{r@{}}{\footnotesize\emph{Continued on next page}}\\
\endfoot
\bottomrule
\endlastfoot
Which cells contain \textup{CXCL13} RNA? & In metastatic nodes, 95.7\% of \textup{CXCL13}-positive cells labeled as B cells also had detectable RNA from at least two T-cell markers. & In paired node sections, locate \textup{CXCL13} RNA relative to B-cell protein markers and cell boundaries. Test for the transcript in individual B cells verified by imaging during sorting. The same design can assess cell-cycle and keratin signals. & Signal inside intact B cells supports a B-cell origin, whereas signal confined to adjacent cells points to extrinsic capture. \\
Does a higher cell-cycle signal reflect a shift in cell proportions or a change within a state? & Memory B cells had higher \textit{TOP2A} RNA in all 10 evaluable pairs, although their share fell in 9 of those pairs. & Split each node into aliquots. Match sampling, digestion, storage and library preparation across tissues. Measure each state's share and, separately, RNA or protein within that state. & A difference within the same state under matched handling supports an expression change beyond composition, though both effects can coexist. \\
Does a larger share mean more cells? & GC-like B cells made up a larger share of captured B-lineage cells in metastatic nodes, without tissue-normalized cell counts. & In paired nodes from each patient, count cells of each state per unit of tissue mass or volume and record cell recovery. & A larger share with unchanged density indicates compositional shift without local expansion, whereas higher density supports greater local abundance. \\
How much does the \textup{CXCL13} contrast depend on particular patients or batches? & P28, P32 and P34 no longer met coverage requirements after filtering. In P15, omitting one batch reversed the filtered contrast; P26 showed a decrease in the original comparison. & Review the paired counts, eligible cell numbers and batch records for these patients. Retain opposite-direction results and specify patient-level review criteria before examining new samples. & This audit identifies contrasts sensitive to sparse sampling or batch handling within the cohort. \\
\end{longtable}
\endgroup

\FloatBarrier
\newlength{\anzhenTableWidth}
\section{Five-Omics Profiling in Acute Aortic Dissection}\label{sec:si-case-anzhen}

This supplement supports Section~\ref{sec:exp:anzhen} and Figures~\ref{fig:anzhen-single-dual}--\ref{fig:anzhen-cp6b}. It is organized along two tracks. First, it traces how and why CytoCoScientist revised analytical paths and interpretations, detailing the study objective and cohort specifications (SI Section~\ref{sec:si-anzhen-s1}), the search trajectory (SI Figure~\ref{fig:si-anzhen-dag}), and the evidence behind key decisions (SI Table~\ref{tab:si-anzhen-decisions}). Second, it documents the omics provenance and statistical verification of each main-text finding, mapping the full workflow (SI Figure~\ref{fig:si-anzhen-workflow}) and modeling protocols (SI Section~\ref{sec:si-anzhen-s3}). Supporting data map directly to the main figures: \textit{SMAD3} regional associations in Figure~\ref{fig:anzhen-single-dual}e,f appear in SI Table~\ref{tab:si-anzhen-smad3}; the 208 across-case process--protein tests appear in SI Table~\ref{tab:si-anzhen-208} (prespecified 8 all-cell tests in SI Figure~\ref{fig:si-anzhen-mofa-208}d); multi-layer conditional models and four-omics MOFA results for Figure~\ref{fig:anzhen-three-four} appear in SI Tables~\ref{tab:si-anzhen-multilayer}--\ref{tab:si-anzhen-f3-pcor}; and seven-candidate regional estimates and per-carrier CP6B profiles for Figure~\ref{fig:anzhen-cp6b} appear in SI Tables~\ref{tab:si-anzhen-seven}--\ref{tab:si-anzhen-cp6b}. Cohort audits, single-omics full results, and model diagnostics appear in SI Section~\ref{sec:si-anzhen-s4} (SI Tables~\ref{tab:si-anzhen-rna25}--\ref{tab:si-anzhen-f3-pcor} and SI Figures~\ref{fig:si-anzhen-qc}--\ref{fig:si-anzhen-mofa-208}), the independent \textit{FBN1} structural lead in S14 appears in SI Figure~\ref{fig:si-anzhen-fbn1} and SI Table~\ref{tab:si-anzhen-fbn1}, and the unexecuted follow-up validation plan is consolidated in SI Table~\ref{tab:si-anzhen-followup}.

\subsection{Task Prompt and Cohort Specifications}\label{sec:si-anzhen-s1}\label{sec:si-case-anzhen-input}

\subsubsection{Initial Research Objective}\label{sec:si-anzhen-s1-1}

\begin{caseprompt}{Goal.md | Five-Omics Profiling in Aortic Dissection}
Using paired plasma and peripheral blood mononuclear cell (PBMC) samples from 18 participants (13 cases and 5 healthy controls), integrate five omics layers (plasma OWWW proteomics, cfDNA methylome, whole-genome sequencing [WGS], single-cell transcriptomics [scRNA], and single-cell chromatin accessibility [scATAC]) to perform multi-omics differential analysis and cross-omics integration between patients with acute aortic dissection and healthy controls. Identify and prioritize disease-specific peripheral-blood cell subpopulations, genetic variants, DNA methylation sites, accessible chromatin regions, key genes, and plasma protein biomarkers; construct an association chain linking genetic variation, epigenetic regulation, cell-type-specific transcriptional regulation, protein expression changes, and the acute aortic dissection phenotype; determine the cell types, regulatory networks, and potential pathogenic mechanisms of key disease-related molecules; screen potential early-screening molecular biomarkers; and propose testable experimental validation plans for high-priority biomarkers and candidate pathogenic pathways.

The study proceeds in stages from single-omics analysis to cross-omics integration and then to mechanistic hypotheses and experimental validation. The five single-omics modules (OWWW, WGS, WGBS, scRNA, and scATAC) should each complete independent, thorough, and verifiable analyses before entering cross-omics integration; in principle, focus on completing one omics module at a time rather than switching frequently before a module reaches complete results. If a module cannot proceed due to missing data, quality issues, or resource constraints, record the blocking reason clearly and continue advancing the remaining modules so that a single module does not stall the overall study.
\end{caseprompt}

\subsubsection{Cohort Design and Measurement Specifications}\label{sec:si-anzhen-s1-2}\label{sec:si-case-anzhen-cohort}

This study uses the initially supplied five-omics acute aortic dissection cohort, comprising 18 independent participants: 13 cases sampled after symptom onset and before surgery (ages 19--66 years; 3 female) and 5 unpaired healthy controls (ages 29--39 years; 2 female). All five modalities per participant were collected from the same blood draw for within-participant cross-material alignment, whereas cases and controls are unpaired. Participant identifiers (BJ1--BJ5; S1--S7, S10--S12, S14, S15, S17) are not treated as technical batches. Height and BMI are missing for S4 and S12, and blood type is missing for S12; no missing values are imputed.

The five omics modalities differ in measurement targets and coverage, while all group-level inferences treat the participant as the independent unit: PBMC RNA comprises $195{,}574$ Stereo-seq segmented cell-bins (aggregated by donor rather than treating bins as replicates, covering the all-cell baseline, 25 eligible differential labels, and 41 composition-conserving labels); PBMC ATAC comprises $78{,}691$ fixed promoter-region counts ($78{,}026$ comparable regions, retaining both whole-fragment and promoter-assigned denominators, and profiled on separate cells from PBMC RNA); plasma proteomics retains $1{,}249$ assays (evaluated in parallel on raw NPX and individual panel-median-centered NPX scales, neither representing absolute concentration nor enzymatic activity); circulating cfDNA WGBS aggregates strand-specific CpG records across $78{,}691$ fixed regions ($77{,}773$ estimable regions, separating $1{,}322$ observed zeros from $19{,}139$ missing entries, and distinct from PBMC or aortic-tissue methylation); and WGS consists of 18 single-sample hg38 variant VCF files (\texttt{includeNonVariantSites=false}, \texttt{FILTER="."}), in which absent records denote unknown genotypes rather than homozygous reference calls.

Cohort baseline characteristics, PBMC RNA coverage, and the five-omics input audit are summarized in SI Figure~\ref{fig:si-anzhen-qc}. The raw sequencing data, VCFs, and complete individual-level matrices are not publicly available.

\subsection{Search Trajectory and Core Research Decisions}\label{sec:si-anzhen-s2}

Analysis began by contrasting peripheral-blood profiles between the 13 acute aortic dissection cases and 5 healthy controls and testing which group-level signals aligned across modalities within the same participants. CytoCoScientist first matched the five assay datasets by participant: the RNA branch used explicit gene-ID-corrected all-cell and per-label counts (25 of 41 labels had sufficient case and control donors for group modeling); plasma proteomics retained both raw and individually centered NPX scales; ATAC aggregated promoter counts from sequencing fragments under both normalization denominators; cfDNA tracked regional coverage, methylation values, and missingness separately; and WGS distinguished explicit genotype calls from absent VCF records. All downstream cross-omics analyses built on these audited tables. Within the cfDNA branch, outer-loop meta-review checked whether age and sex aligned with participant identities, triggering retraction and recomputation of the earlier cfDNA group contrasts.

To test whether adding modalities contributed information beyond existing layers, CytoCoScientist scored Hallmark pathways across the 25 RNA labels and sequentially tested three-layer case--control joint differences, conditional associations of RNA and ATAC with plasma protein, and fourth-layer cfDNA increments. It then audited coverage and denominators across all $4{,}800$ model rows and tested fifth-layer genetic dosage exclusively within donor subsets carrying explicit WGS calls. In parallel, an unsupervised branch fit four-omics MOFA across 8 views; when outer-loop meta-review flagged that post-training variance summaries penalized missing slots with zero placeholders, CytoCoScientist retained the fitted factors and loadings, recomputed variance explained strictly over observed non-missing entries, and compared case versus control values among top positive- and negative-loading features of F3.

In two-layer RNA--protein analysis, CytoCoScientist tested gene-matched RNA--protein correlations, evaluated signed shared-process enrichment across full rankings, audited member identities, and extracted per-donor direction tables to test whether process-level RNA rankings tracked plasma MMP8 or LBP across cases. Of 936 expected entries, 902 had observed values and 34 were missing; across all 208 estimable tests restricted to covered cases, none passed Benjamini--Hochberg (BH) correction, and this negative branch was retained in the search trajectory.

Two regional branches pursued distinct genomic leads. The \textit{SMAD3} branch reused \textit{FBN1}-related donor/label tables and matched regional cfDNA tables to evaluate 48 prespecified cross-material pairs; after detecting the inverse cfDNA--ATAC correlation in \textit{SMAD3}, CytoCoScientist tested sensitivity to 54 shared strand-specific CpG records and to S14 exclusion. The CP6B branch started from explicit WGS calls near \textit{CEBPE}/\textit{GFI1}, extracted ATAC and cfDNA readouts over the exact regulatory interval, aligned transcription-factor RNA, \textit{MMP8} RNA, and plasma MMP8 by donor, and retained the discordant upstream readout in each of the three heterozygous carriers (low cfDNA in S7, low promoter-normalized ATAC in S12, and low \textit{GFI1} RNA in S14).

\begin{figure}[!htbp]
\centering
\includegraphics[width=\linewidth]{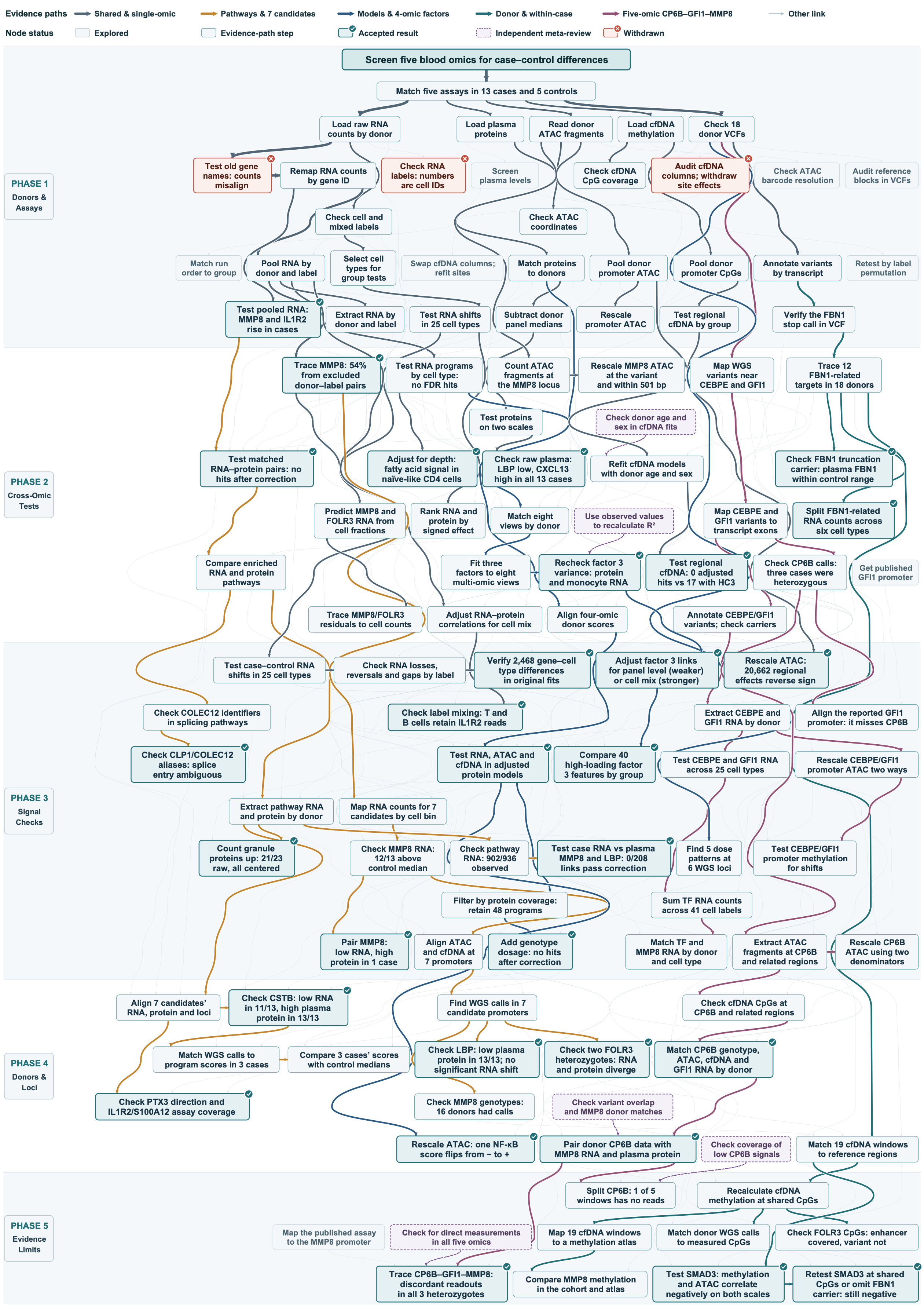}
\caption{\textbf{CytoCoScientist search trajectory across the five-omics acute aortic dissection dataset.}}
\label{fig:si-anzhen-dag}
\end{figure}

\begin{figure}[!htbp]
\ContinuedFloat
\captionsetup{labelformat=continuedfigure}
\caption[]{For readability, the diagram shows the main analysis paths, revisions, and counterexamples. Five vertical stages progress from donor alignment and cross-omics testing to signal verification, per-donor auditing, and evidence-boundary determination, retaining key revisions and counterexamples (such as S5 showing low PBMC \textit{MMP8} RNA with elevated plasma MMP8). Thick colored arrows trace five tracks: blue-gray for single-omics contrasts, ochre for shared processes and seven candidates, blue for conditional models and four-omics MOFA, teal for individual counterexamples and across-case covariation, and berry for CP6B--\textit{GFI1}--\textit{MMP8}; thin lines mark data dependencies. Green checks mark retained outputs (including null findings), red crosses mark retracted specifications, and dashed purple boxes mark outer-loop meta-review checkpoints (model scopes in SI Tables~\ref{tab:si-anzhen-combos}--\ref{tab:si-anzhen-multilayer}; finalized workflow in SI Figure~\ref{fig:si-anzhen-workflow}).}
\end{figure}

\begingroup
\footnotesize
\setlength{\tabcolsep}{3pt}
\renewcommand{\arraystretch}{1.15}
\setlength{\LTleft}{0pt}
\setlength{\LTright}{\fill}
\setlength{\LTcapwidth}{\linewidth}
\captionsetup{singlelinecheck=false,justification=raggedright}
\setlength{\anzhenTableWidth}{\dimexpr\linewidth-6\tabcolsep\relax}
\begin{longtable}{@{}>{\raggedright\arraybackslash}p{0.20\anzhenTableWidth}>{\raggedright\arraybackslash}p{0.23\anzhenTableWidth}>{\raggedright\arraybackslash}p{0.31\anzhenTableWidth}>{\raggedright\arraybackslash}p{0.26\anzhenTableWidth}@{}}
\caption{\textbf{Key evidence and revisions that changed CytoCoScientist's analytical decisions.} Audits from input verification and outer-loop meta-review are listed separately; retained candidates carry explicit evidence boundaries.}\label{tab:si-anzhen-decisions}\\
\toprule
\textbf{Audit question} & \textbf{System check} & \textbf{Decisive evidence} & \textbf{Final decision and location} \\
\midrule
\endfirsthead
\multicolumn{4}{@{}l}{\footnotesize\textbf{Table~\thetable\ (continued)}}\\
\toprule
\textbf{Audit question} & \textbf{System check} & \textbf{Decisive evidence} & \textbf{Final decision and location} \\
\midrule
\endhead
\midrule
\multicolumn{4}{r@{}}{\footnotesize\emph{Continued on next page}}\\
\endfoot
\bottomrule
\endlastfoot
Legacy RNA count reliability & Compared gene-name-misaligned legacy tables against explicit gene-ID-corrected matrices. & Legacy tables misassigned counts across genes and could not support differential testing. & Adopted explicit gene-ID-corrected matrices (SI Section~\ref{sec:si-anzhen-s1-2}, SI Figure~\ref{fig:si-anzhen-qc}). \\
Correction of age and sex mismatches in the cfDNA model & Outer-loop meta-review triggered donor identity verification, comparing original versus aligned age and sex designs. & The original design contained 13 age and 6 sex mismatches, flipping $4{,}170$ regional effects; aligned design reproduced current $77{,}773$-region results (max diff.\ $1.22 \times 10^{-14}$). & Retracted cfDNA inferences based on mismatched age and sex covariates while keeping audited modalities intact (SI Section~\ref{sec:si-anzhen-s1-2}, SI Figure~\ref{fig:si-anzhen-qc}). \\
Reference choice for plasma proteins and ATAC & Audited raw versus panel-centered protein NPX and whole-fragment versus promoter-assigned ATAC denominators in parallel. & $20{,}662/78{,}026$ ATAC regions flipped sign; raw LBP was below all controls in 13/13 cases but became non-significant after centering ($q = 0.423$). & Reported both references in parallel without selecting by significance (Figure~\ref{fig:anzhen-single-dual}a,c). \\
Process--protein tests under missing cell labels & Audited coverage across 936 expected entries, pairing only cases with both observed readouts ($\ge 8$ cases per test). & 902 entries observed and 34 missing due to low cell counts; all 208 tests estimable with $0/208$ passing BH correction. & Retained group-level process shift without claiming across-case ordering concordance (SI Table~\ref{tab:si-anzhen-208}, SI Figure~\ref{fig:si-anzhen-mofa-208}d). \\
Sensitivity of \textit{SMAD3} negative correlation to CpG coverage and S14 & After identifying the signal in the 48-pair family, restricted to shared strand-specific CpG records, excluded S14, and tested covariates. & Primary $\rho = -0.841$; 54 shared strand-specific CpG records $\rho = -0.852$; excluding S14 $\rho = -0.818$. & Retained within-cohort cross-material regional association as top replication lead (Figure~\ref{fig:anzhen-single-dual}e,f, SI Table~\ref{tab:si-anzhen-smad3}). \\
Effect of missing entries on MOFA variance explained & Outer-loop meta-review flagged zero-placeholder distortion; recomputed $R^2$ over observed entries while keeping fitted factors and loadings. & F1 explained $90.96\%$ of ATAC variance, F2 $34.11\%$ of cfDNA variance, and F3 jointly $76.01\%$ of protein and $18.40\%$ of monocyte RNA variance. & Corrected $R^2$ without refitting and ran parallel F3 sensitivity models over the same 16 donors (Figure~\ref{fig:anzhen-three-four}c,d, SI Tables~\ref{tab:si-anzhen-mofa-r2}--\ref{tab:si-anzhen-f3-pcor}). \\
Evidential status of 152 upstream CP6B records & Outer-loop meta-review prompted record-type verification; audited genomic intervals, donor identity, and per-carrier readouts. & Records reflected coordinate overlap or shared donor identity rather than physical chromatin contact; S7 (cfDNA), S12 (promoter ATAC), and S14 (\textit{GFI1} RNA) each showed a low upstream readout. & Retracted physical-contact and phasing claims; retained donor-verifiable five-layer candidate (Figure~\ref{fig:anzhen-cp6b}a--d, SI Table~\ref{tab:si-anzhen-cp6b}). \\
Structural and pathway role of S14 \textit{FBN1} \texttt{p.R1539*} & Mapped six published recombinant fragments onto \texttt{p.R1539*} and audited S14 peripheral multi-omics readouts. & N-terminal interaction domain lies upstream and C-terminal partner downstream of \texttt{p.R1539*}; no truncated protein measured and downstream peripheral RNA was inconsistent. & Retained as an S14-specific conditional structural hypothesis separate from the 13-case \textit{SMAD3} association (SI Figure~\ref{fig:si-anzhen-fbn1}, SI Table~\ref{tab:si-anzhen-fbn1}). \\
\end{longtable}
\endgroup

\subsection{Bioinformatics Analysis Methods}\label{sec:si-anzhen-s3}

SI Figure~\ref{fig:si-anzhen-workflow} maps each of the five omics modalities to its downstream analyses and key findings; statistical models, sample denominators, and multiple-testing families are detailed in SI Sections~\ref{sec:si-anzhen-s3-1}--\ref{sec:si-anzhen-s3-4}.

\begin{figure}[!htbp]
\centering
\includegraphics[width=\linewidth,height=\dimexpr\textheight-3\baselineskip\relax,keepaspectratio]{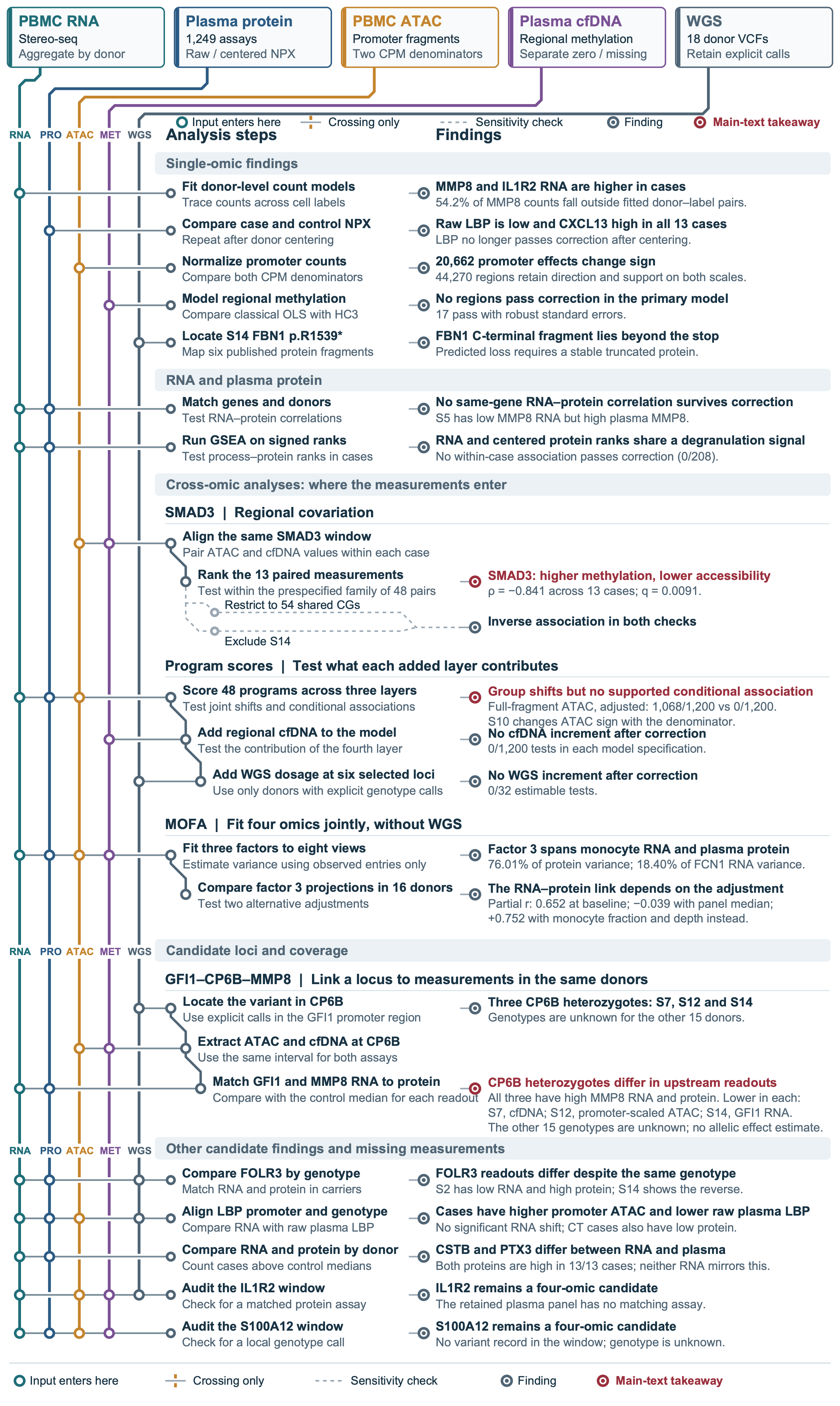}
\caption{\textbf{Omics provenance and finalized analytical workflow for the acute aortic dissection study.}}
\label{fig:si-anzhen-workflow}
\end{figure}

\begin{figure}[!htbp]
\ContinuedFloat
\captionsetup{labelformat=continuedfigure}
\caption[]{Five colored tracks trace PBMC RNA, plasma proteins, PBMC ATAC, plasma cfDNA methylation, and WGS. Open circles mark where a modality enters an analysis; a crossing without a circle is not an input. Solid lines trace primary analyses and dashed lines mark sensitivity checks. A ring with a central dot marks a finding; crimson highlights the three main-text takeaways (\textit{SMAD3} across-case inverse association, separation of three-layer group shifts from conditional associations, and divergent upstream readouts among CP6B heterozygous carriers). Cases ($n = 13$) and controls ($n = 5$) are unpaired and aligned across assays by participant or genomic interval; MOFA integrates four modalities (excluding WGS), CP6B--\textit{GFI1}--\textit{MMP8} edges denote data alignment across layers, and \textit{IL1R2} (no protein assay) and \textit{S100A12} (no window variant call) remain at four observed layers.}
\end{figure}

\subsubsection{Single-Omics Modeling, Dual Scales, and Dual Denominators}\label{sec:si-anzhen-s3-1}

For PBMC RNA, donor-level age- and sex-adjusted negative-binomial differential models (parametric dispersion trend, independent Benjamini--Hochberg correction~\cite{Benjamini1995FDR} per label) were fit across the all-cell baseline and 25 eligible cell labels, alongside raw-count conservation and modeled versus unmodeled decomposition across all 41 inherited labels. For plasma proteomics, age- and sex-adjusted linear models were fit across all $1{,}249$ retained assays on both raw NPX and individual panel-median-centered NPX (raw value minus participant median, without additional log transformation), applying BH correction within each $1{,}249$-assay family; clinical-heterogeneity sensitivity models further spanned 48 rows across the full cohort (13/5), leave-S14-out (12/5), age-overlap 29--39 years (4/5), and age-overlap excluding S14 (3/5). For PBMC ATAC, linear models (\texttt{case + age + female}) were fit to $\log_2(1+\mathrm{CPM})$ across $78{,}026$ comparable promoter regions under both whole-fragment and promoter-assigned denominators, with separate $78{,}026$-region BH correction. For circulating cfDNA, classical OLS (\texttt{case + age + female}) and HC3 heteroskedasticity-robust sensitivity models were fit to regional CpG means across $77{,}773$ estimable regions, preserving true zeros without imputing missing values; none of the $77{,}773$ regions passed BH correction under primary OLS, whereas 17 passed under HC3 heteroskedasticity-robust standard errors. For WGS, bounded local annotation and expanded locus inspection used 18 single-sample hg38 VCFs with absent entries retained as unknown genotypes.

\subsubsection{Gene-Matched RNA--Protein and Shared-Process Analyses}\label{sec:si-anzhen-s3-2}

Gene-matched two-layer analysis retained all $1{,}249$ plasma protein assays ($1{,}225$ mapped uniquely to stable gene symbols; 24 unresolved symbols left unmerged), yielding $31{,}850$ rows across 26 RNA scopes ($16{,}822$ estimable gene-matched RNA--protein comparisons in each of four testing families) across raw/centered protein scales and unadjusted/residualized Pearson correlations. Shared-process analysis used fully signed pre-ranked gene-set enrichment analysis (GSEA~\cite{Subramanian2005GSEA}; $1{,}868$ gene sets, size 10--500, exponent 1, seed 20261120, \texttt{eps=0}), corrected separately across 26 RNA scopes ($31{,}395$ eligible tests) and both protein scales (282 tests per scale); neutrophil degranulation was concordantly enriched across RNA and panel-centered protein rankings, with positive case--control effects in 21 of 23 core proteins on the raw scale and 23 of 23 on the centered scale, whereas three overlapping splicing terms containing the ambiguously mapped \textit{COLEC12} were excluded from mechanistic interpretation.

\subsubsection{Multi-Layer Conditional Models, Genetic Dosage, and MOFA}\label{sec:si-anzhen-s3-3}

Donor-aligned three- and four-layer models filtered 50 MSigDB Hallmark~\cite{Liberzon2015Hallmark} pathway programs by modality coverage (ATAC $\ge 10$ regions, protein $\ge 5$ distinct symbols, WGBS $\ge 10$ regions, and $\ge 8$ cases and $\ge 4$ controls per model), excluding Notch Signaling and Myc Targets V2 for insufficient protein coverage to yield $48\text{ programs} \times 25\text{ RNA labels} \times 2\text{ ATAC denominators} \times 2\text{ models}$ (\texttt{age\_sex}, adjusting for age and sex, and \texttt{case\_only}, using case status as the sole predictor across both cases and controls) $= 4{,}800$ rows (observed $n = 15\text{--}18$ per row, no RNA zero imputation). Each question (three-layer MANOVA, conditional RNA+ATAC association with protein, and fourth-layer WGBS increment) was BH-corrected within its $1{,}200$-test family. Bounded genetic dosage models were restricted to explicit genotype subsets across 6 candidate loci (5 distinct dosage patterns; $\ge 15$ called donors, $\ge 4$ per dosage group; outcome $n \ge 12$, $\ge 3$ per group, residual $\text{df} \ge 6$), testing $85/100$ layer associations and $32/40$ fifth-layer increments. Four-omics MOFA~\cite{Argelaguet2018MOFA} extracted 3 latent factors across 18 participants and 8 views (5 RNA labels, promoter-normalized ATAC, cfDNA, and plasma protein, excluding WGS), with post-training $R^2$ recomputed over observed non-missing entries; F3 monocyte RNA--protein projections ($n = 16$) were evaluated under three parallel, non-cumulative adjustment models (baseline; baseline + panel median; baseline + monocyte proportion and sequencing depth), reporting partial correlations and approximate 95\% confidence intervals from Fisher's $z$-transformation.

\subsubsection{Candidate Windows, Fragment Boundaries, and Across-Case Checks}\label{sec:si-anzhen-s3-4}

Dual-denominator ATAC effects, regional cfDNA effects, strand-specific CpG coverage, and WGS observations for the seven candidates and CP6B were extracted over fixed half-open hg38 genomic windows. For S14 \textit{FBN1}, six published experimental protein fragments were intersected against \texttt{p.R1539*} using 1-based closed amino-acid intervals. For \textit{SMAD3}, across-case Spearman rank correlations were BH-corrected within the prespecified 48-pair family and supplemented by a shared-site diagnostic across 54 strand-specific CpG records covered at depth $\ge 5$ in all 13 cases ($9{,}999$ permutations, seed 20261121). The 208 across-case process--protein tests crossed two processes (innate immunity and neutrophil degranulation), 26 RNA scopes, MMP8 and LBP, and both protein scales, computing Spearman $\rho$, permutation $p$ values, and 208-test BH correction over non-missing cases ($\ge 8$ cases per test).

\subsection{Cohort Audit and Multi-Layer Model Results}\label{sec:si-anzhen-s4}

\textit{MMP8} did not meet the gene-count threshold for differential-expression testing in any of the 25 modeled PBMC cell labels. Of its $1{,}565$ raw counts, $848$ ($54.2\%$) came from donor--cell-label combinations not included in these models (Figure~\ref{fig:anzhen-single-dual}b and SI Table~\ref{tab:si-anzhen-rna25}). Under the current coverage requirements, the \textit{MMP8} difference observed in the pooled PBMC analysis could therefore not be assigned to a specific modeled cell subpopulation.

\begin{figure}[!htbp]
\centering
\includegraphics[width=\linewidth]{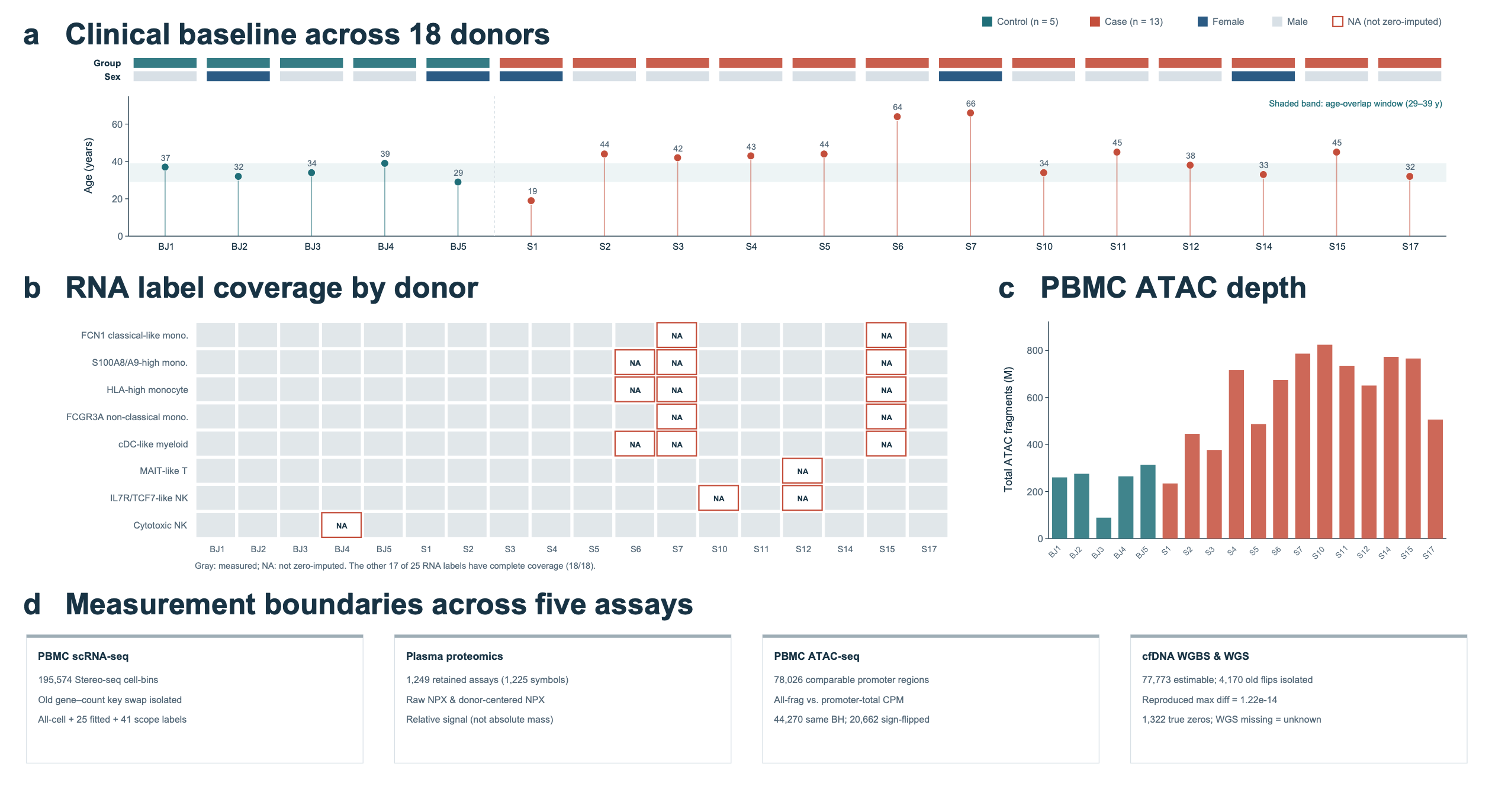}
\caption{\textbf{Cohort clinical baseline, PBMC RNA label coverage matrix, and five-omics input audit ($n = 18$).} \textbf{a,} Case/control status, sex, and age distribution in fixed participant order (BJ1--BJ5; S1--S7, S10--S12, S14, S15, S17), with light cyan shading marking the 29--39-year age-overlap window. \textbf{b,} Coverage matrix for the 8 of 25 eligible PBMC RNA labels with missing donors; white cells outlined in terracotta and labeled \texttt{NA} mark donor--label combinations excluded from modeling because of insufficient cell counts; these entries were not imputed as zero (the remaining 17 labels have complete 18/18 coverage). \textbf{c,} Total PBMC ATAC fragment depth (millions) across all 18 participants. \textbf{d,} Summary of participant alignment and measurement scope across the five omics layers. Correcting mismatches in participant age and sex in the cfDNA model changed the effect direction in $4{,}170$ regions. Numerical verification of the corrected $77{,}773$-region model yielded a maximum difference of $1.22 \times 10^{-14}$.}
\label{fig:si-anzhen-qc}
\end{figure}

\begin{figure}[!htbp]
\centering
\includegraphics[width=\linewidth]{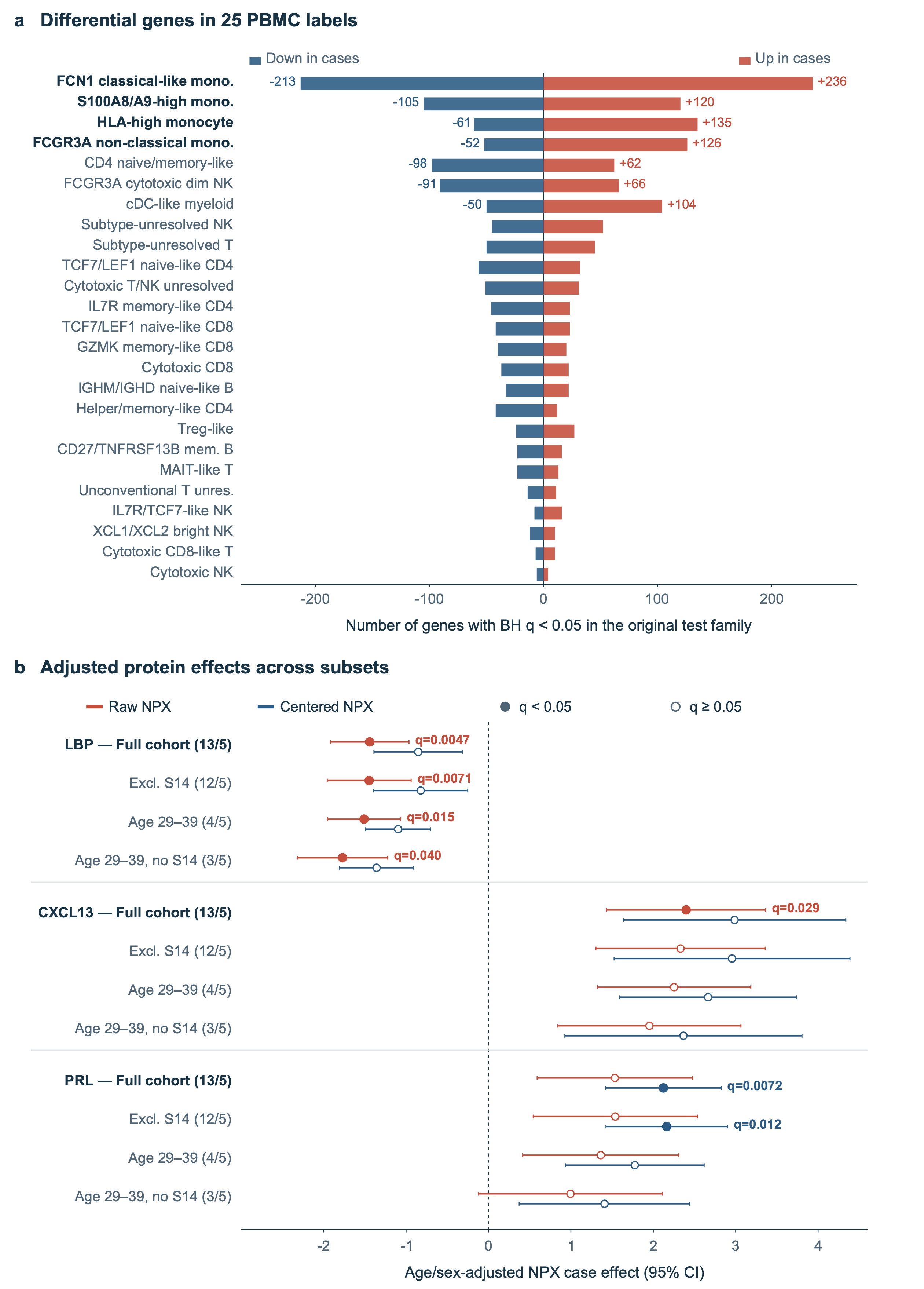}
\caption{\textbf{Differential gene counts across 25 eligible PBMC RNA labels and plasma protein clinical-heterogeneity sensitivity.} \textbf{a,} Numbers of BH-significant ($q < 0.05$) upregulated (right) and downregulated (left) genes across all 25 eligible PBMC RNA labels; the top four labels are monocyte subpopulations. \textbf{b,} Age- and sex-adjusted raw and panel-centered NPX effects with 95\% confidence intervals for plasma LBP, CXCL13, and PRL across full cohort (13/5), leave-S14-out (12/5), 29--39-year age-overlap (4/5), and age-overlap excluding S14 (3/5); filled points indicate $q < 0.05$ within the $1{,}249$-assay family.}
\label{fig:si-anzhen-rna25-protein48}
\end{figure}

\begin{figure}[!htbp]
\centering
\includegraphics[width=\linewidth]{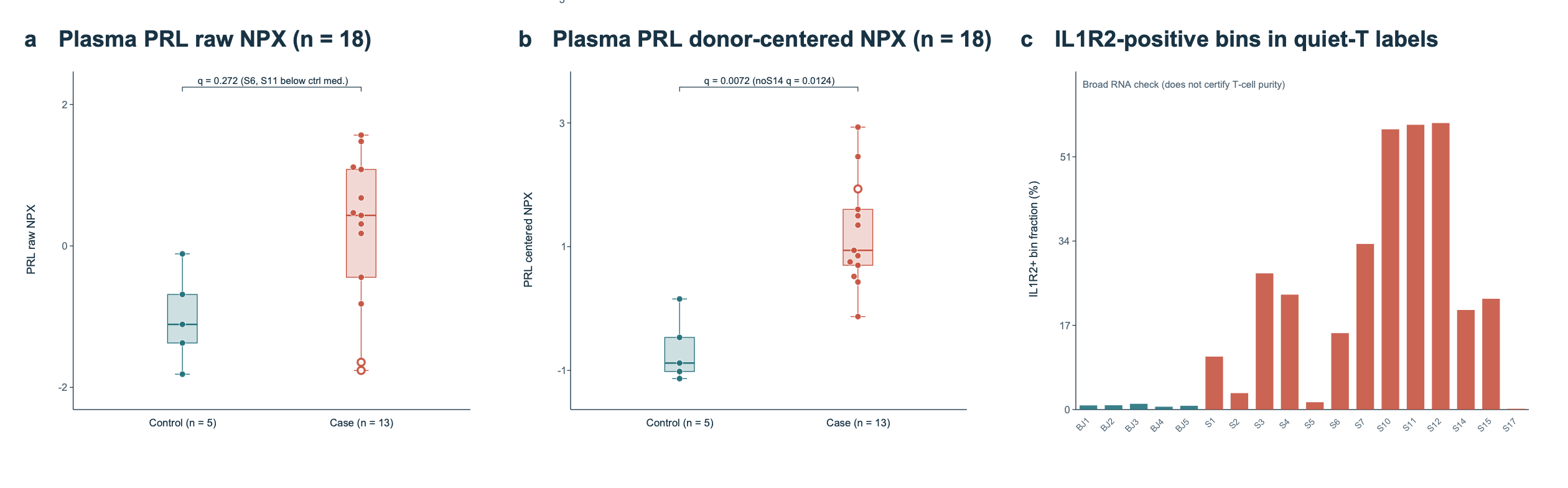}
\caption{\textbf{Dual-scale per-participant plasma PRL distributions and broad \textit{IL1R2} RNA audit.} \textbf{a,b,} Per-participant plasma PRL profiles ($n = 18$) on raw and panel-centered NPX scales, marking low raw-scale donors (S6, S11) and the distribution of high-value cases alongside S14. \textbf{c,} Fraction of \textit{IL1R2}-positive bins in the quiescent T-cell (quiet-T) label, auditing cross-label RNA capture.}
\label{fig:si-anzhen-prl-il1r2}
\end{figure}

\begin{figure}[!htbp]
\centering
\includegraphics[width=\linewidth]{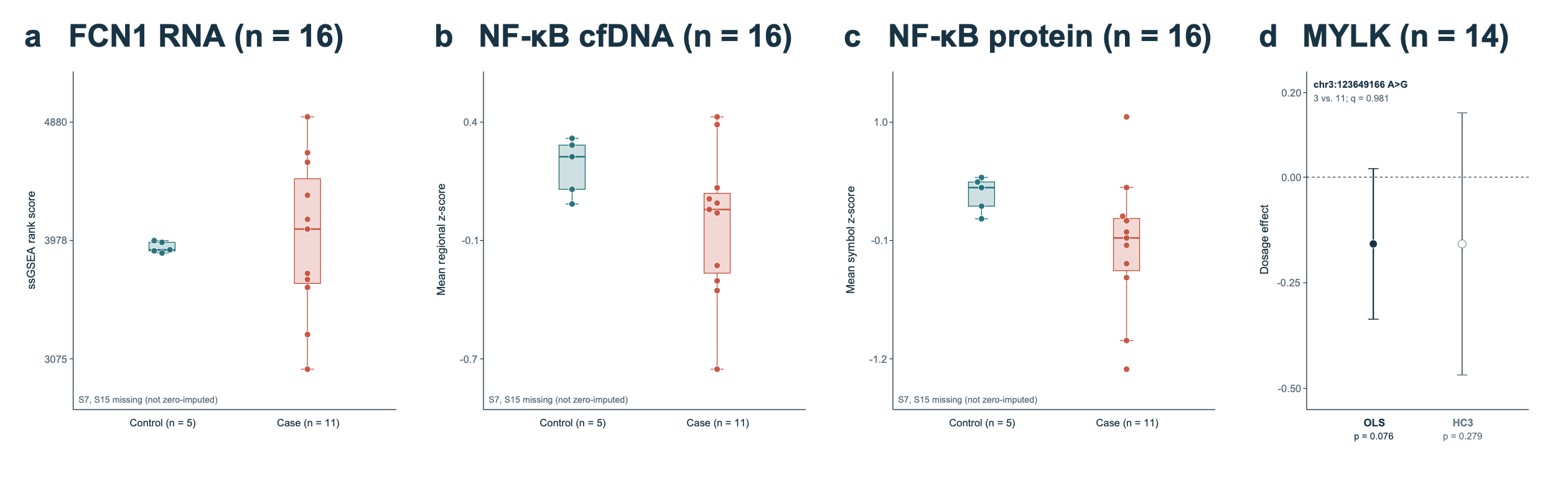}
\caption{\textbf{Remaining three-layer scores for the NF-$\kappa$B/FCN1 axis ($n = 16$) and bounded \textit{MYLK} genetic dosage example.} \textbf{a--c,} FCN1 RNA ssGSEA, cfDNA, and plasma protein program scores (S7 and S15 missing). \textbf{d,} Bounded genetic dosage regression for \textit{MYLK}/Coagulation whole-fragment ATAC ($n = 14$, genotype groups 3/11; both OLS and HC3 confidence intervals cross zero, BH $q = 0.981$).}
\label{fig:si-anzhen-nfkb-mylk}
\end{figure}

\begin{figure}[!htbp]
\centering
\includegraphics[width=\linewidth]{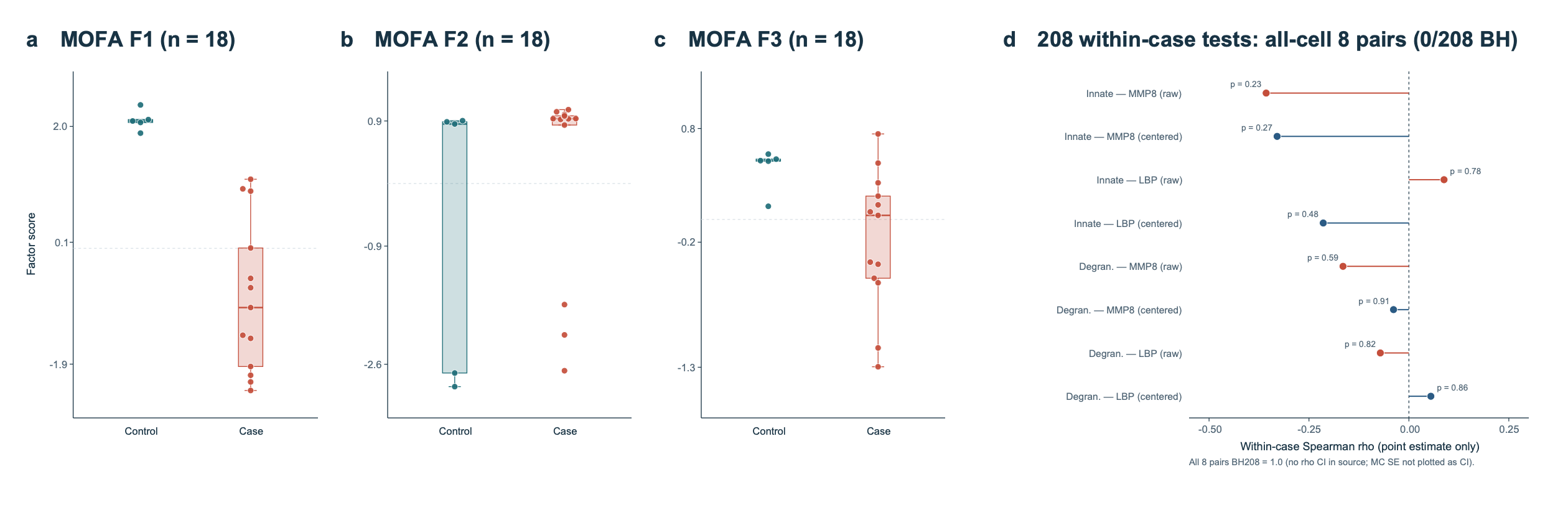}
\caption{\textbf{Per-participant MOFA factor scores ($n = 18$) and prespecified all-cell subset of the 208 across-case process--protein tests.} \textbf{a--c,} Continuous factor scores for MOFA F1, F2, and F3. \textbf{d,} Point estimates ($\rho$), permutation $p$ values, and 208-test BH-adjusted $q$ values for the 8 prespecified all-cell comparisons in SI Table~\ref{tab:si-anzhen-208} ($0/208$ significant overall).}
\label{fig:si-anzhen-mofa-208}
\end{figure}

This section provides the 18-participant clinical baseline and input audit (SI Figure~\ref{fig:si-anzhen-qc}), the 25 eligible RNA labels (SI Table~\ref{tab:si-anzhen-rna25}), the 48-row plasma protein sensitivity table (SI Table~\ref{tab:si-anzhen-protein48} and SI Figure~\ref{fig:si-anzhen-rna25-protein48}b), coverage across all 31 non-empty modality combinations (SI Table~\ref{tab:si-anzhen-combos}), the $4{,}800$-row multi-layer and bounded WGS dosage summary (SI Table~\ref{tab:si-anzhen-multilayer}), and MOFA variance explained alongside F3 parallel partial correlations (SI Tables~\ref{tab:si-anzhen-mofa-r2}--\ref{tab:si-anzhen-f3-pcor}). Individual readouts appear in SI Figure~\ref{fig:si-anzhen-prl-il1r2} (PRL and \textit{IL1R2}), SI Figure~\ref{fig:si-anzhen-nfkb-mylk} (NF-$\kappa$B/FCN1 program and \textit{MYLK} dosage example), and SI Figure~\ref{fig:si-anzhen-mofa-208} (MOFA factor scores and the 8 prespecified all-cell across-case correlations). In a secondary sensitivity analysis, the fatty-acid metabolism RNA score in TCF7/LEF1 naive-like CD4 T cells did not pass $1{,}250$-test correction in the primary age/sex model (effect $111.25$, $q = 0.9506$) but reached $q = 0.0456$ after adding sequencing depth (effect $153.40$, max leverage $0.704$). This result motivates a prespecified sequencing-depth sensitivity check in future cohorts.

\begingroup
\footnotesize
\setlength{\tabcolsep}{4pt}
\renewcommand{\arraystretch}{1.12}
\setlength{\LTleft}{0pt}
\setlength{\LTright}{\fill}
\setlength{\LTcapwidth}{\linewidth}
\captionsetup{singlelinecheck=false,justification=raggedright}
\setlength{\anzhenTableWidth}{\dimexpr\linewidth-8\tabcolsep\relax}
\begin{longtable}{@{}>{\raggedright\arraybackslash}p{0.34\anzhenTableWidth}>{\raggedleft\arraybackslash}p{0.14\anzhenTableWidth}>{\raggedleft\arraybackslash}p{0.14\anzhenTableWidth}>{\raggedleft\arraybackslash}p{0.20\anzhenTableWidth}>{\raggedright\arraybackslash}p{0.18\anzhenTableWidth}@{}}
\caption{\textbf{Donor coverage, modeled genes, and BH-significant genes ($q < 0.05$) across the 25 eligible PBMC RNA cell labels.}}\label{tab:si-anzhen-rna25}\\
\toprule
\textbf{RNA cell label} & \textbf{Cases / controls} & \textbf{Modeled genes} & \textbf{BH hits (up / down)} & \textbf{Missing donors} \\
\midrule
\endfirsthead
\multicolumn{5}{@{}l}{\footnotesize\textbf{Table~\thetable\ (continued)}}\\
\toprule
\textbf{RNA cell label} & \textbf{Cases / controls} & \textbf{Modeled genes} & \textbf{BH hits (up / down)} & \textbf{Missing donors} \\
\midrule
\endhead
\midrule
\multicolumn{5}{r@{}}{\footnotesize\emph{Continued on next page}}\\
\endfoot
\bottomrule
\endlastfoot
FCN1 classical-like monocyte & 11 / 5 & 15,680 & 449 (236 / 213) & S7, S15 \\
S100A8/S100A9-high monocyte & 10 / 5 & 12,083 & 225 (120 / 105) & S6, S7, S15 \\
HLA-high monocyte & 10 / 5 & 9,069 & 196 (135 / 61) & S6, S7, S15 \\
FCGR3A non-classical-like monocyte & 11 / 5 & 10,450 & 178 (126 / 52) & S7, S15 \\
CD4 naive/memory-like & 13 / 5 & 13,269 & 160 (62 / 98) & None \\
FCGR3A cytotoxic dim-like NK & 13 / 5 & 12,400 & 157 (66 / 91) & None \\
cDC-like, unresolved myeloid & 10 / 5 & 10,640 & 154 (104 / 50) & S6, S7, S15 \\
Unresolved NK & 13 / 5 & 11,573 & 97 (52 / 45) & None \\
Unresolved T & 13 / 5 & 10,944 & 95 (45 / 50) & None \\
TCF7/LEF1 naive-like CD4 & 13 / 5 & 13,482 & 89 (32 / 57) & None \\
Cytotoxic T/NK-like, unresolved & 13 / 5 & 11,927 & 82 (31 / 51) & None \\
IL7R memory-like CD4 & 13 / 5 & 13,912 & 69 (23 / 46) & None \\
TCF7/LEF1 naive-like CD8 & 13 / 5 & 12,050 & 65 (23 / 42) & None \\
GZMK memory-like CD8 & 13 / 5 & 11,318 & 60 (20 / 40) & None \\
Cytotoxic CD8 & 13 / 5 & 12,114 & 59 (22 / 37) & None \\
IGHM/IGHD naive-like B & 13 / 5 & 14,127 & 55 (22 / 33) & None \\
Helper/memory-like CD4 & 13 / 5 & 10,479 & 54 (12 / 42) & None \\
Treg-like & 13 / 5 & 9,523 & 51 (27 / 24) & None \\
CD27/TNFRSF13B memory-like B & 13 / 5 & 11,776 & 39 (16 / 23) & None \\
MAIT-like T & 12 / 5 & 11,970 & 36 (13 / 23) & S12 \\
Unconventional-like T, unresolved & 13 / 5 & 10,294 & 25 (11 / 14) & None \\
IL7R/TCF7-like NK & 11 / 5 & 8,346 & 24 (16 / 8) & S10, S12 \\
XCL1/XCL2 bright-like NK & 13 / 5 & 8,251 & 22 (10 / 12) & None \\
Cytotoxic CD8-like T & 13 / 5 & 9,512 & 17 (10 / 7) & None \\
Cytotoxic NK & 13 / 4 & 9,889 & 10 (4 / 6) & BJ4 \\
\end{longtable}
\endgroup

\begin{table}[!htbp]
\centering\footnotesize
\setlength{\tabcolsep}{3.5pt}
\caption{\textbf{Clinical-heterogeneity sensitivity models across 48 plasma protein specifications (effect; 1,249-assay BH $q$ value).} Primary age- and sex-adjusted effects with 95\% confidence intervals: LBP raw full cohort $-1.4417$ ($-1.9189$ to $-0.9645$, $q = 0.0047$), leave-S14-out $-1.4481$ ($-1.9571$ to $-0.9391$, $q = 0.0071$); CXCL13 raw full cohort $+2.3984$ ($1.4319$ to $3.3649$, $q = 0.0294$), leave-S14-out $+2.3310$ ($1.3037$ to $3.3583$, $q = 0.0569$); PRL centered full cohort $+2.1230$ ($1.4217$ to $2.8243$, $q = 0.0072$), leave-S14-out $+2.1639$ ($1.4257$ to $2.9022$, $q = 0.0124$).}
\label{tab:si-anzhen-protein48}
\begin{tabularx}{\linewidth}{@{}>{\raggedright\arraybackslash}Xlrrrr@{}}
\toprule
\textbf{Subcohort} & \textbf{Protein} & \shortstack[r]{\textbf{Raw}\\Unadjusted} & \shortstack[r]{\textbf{Raw}\\Age/sex adj.} & \shortstack[r]{\textbf{Centered}\\Unadjusted} & \shortstack[r]{\textbf{Centered}\\Age/sex adj.} \\
\midrule
Full cohort (13 / 5) & CXCL13 & $+2.2090$; 0.0398 & $+2.3984$; 0.0294 & $+2.8743$; 0.0742 & $+2.9869$; 0.1273 \\
Full cohort (13 / 5) & PRL & $+1.2216$; 0.1253 & $+1.5345$; 0.2723 & $+1.8869$; 0.0742 & $+2.1230$; 0.0072 \\
Full cohort (13 / 5) & LBP & $-1.4213$; 0.0010 & $-1.4417$; 0.0047 & $-0.7560$; 0.1834 & $-0.8532$; 0.4233 \\
Leave-S14-out (12 / 5) & CXCL13 & $+2.1178$; 0.0393 & $+2.3310$; 0.0569 & $+2.7967$; 0.0705 & $+2.9557$; 0.1489 \\
Leave-S14-out (12 / 5) & PRL & $+1.1484$; 0.1212 & $+1.5392$; 0.2375 & $+1.8273$; 0.0665 & $+2.1639$; 0.0124 \\
Leave-S14-out (12 / 5) & LBP & $-1.4324$; 0.0016 & $-1.4481$; 0.0071 & $-0.7535$; 0.1454 & $-0.8234$; 0.2331 \\
Ages 29--39 (4 / 5) & CXCL13 & $+2.0345$; 0.0844 & $+2.2521$; 0.0641 & $+2.4345$; 0.2273 & $+2.6651$; 0.0870 \\
Ages 29--39 (4 / 5) & PRL & $+1.2700$; 0.0917 & $+1.3624$; 0.0782 & $+1.6699$; 0.2558 & $+1.7754$; 0.1464 \\
Ages 29--39 (4 / 5) & LBP & $-1.4864$; 0.0056 & $-1.5096$; 0.0152 & $-1.0865$; 0.0736 & $-1.0967$; 0.0641 \\
Ages 29--39, no S14 (3 / 5) & CXCL13 & $+1.6115$; 0.1601 & $+1.9529$; 0.2077 & $+1.9773$; 0.3529 & $+2.3660$; 0.3344 \\
Ages 29--39, no S14 (3 / 5) & PRL & $+0.9933$; 0.2135 & $+0.9950$; 0.2395 & $+1.3592$; 0.3529 & $+1.4081$; 0.4315 \\
Ages 29--39, no S14 (3 / 5) & LBP & $-1.5526$; 0.0210 & $-1.7713$; 0.0400 & $-1.1868$; 0.1252 & $-1.3583$; 0.1286 \\
\bottomrule
\end{tabularx}\par\smallskip\raggedright Counts in parentheses are cases / controls.
\end{table}

\begin{table}[!htbp]
\centering\footnotesize
\setlength{\tabcolsep}{4pt}
\caption{\textbf{Analytical coverage and interpretive scope across all 31 non-empty modality combinations} (R = RNA, P = plasma protein, A = ATAC, M = cfDNA methylation, G = WGS).}
\label{tab:si-anzhen-combos}
\begin{tabularx}{\linewidth}{@{}>{\raggedright\arraybackslash}p{0.20\linewidth}r>{\raggedright\arraybackslash}p{0.37\linewidth}>{\raggedright\arraybackslash}X@{}}
\toprule
\textbf{Combination} & \textbf{Layers} & \textbf{Executed analyses and coverage} & \textbf{Interpretive scope} \\
\midrule
R / P / A / M / G (5 single-omics) & 1 & Independently modeled (SI Section~\ref{sec:si-anzhen-s3-1}; examples in Figure~\ref{fig:anzhen-single-dual}a--c; input audit in SI Figure~\ref{fig:si-anzhen-qc}d) & Retains modality-specific scales, denominators, missingness, and callability \\
RP & 2 & Four gene-matched RNA--protein correlation families, signed GSEA, MOFA F3, 208 across-case tests & Group shifts may coexist with 0 BH hits in individual/conditional correlation families \\
RA / PA / RM / PM & 2 & Included via three-/four-layer program models, MOFA, and candidate/fixed-locus alignment & Not genome-wide pairwise scans or single-cell co-measurement proofs \\
AM & 2 & Direct fixed-region comparison: \textit{SMAD3} across-case inverse association and 54 shared CpG records & Cross-material association; does not establish single-cell methylation repression \\
RG / PG / AG / MG & 2 & Explicit-call bounded genetic dosage, candidate windows, and locus alignment & Not population eQTLs or chromatin contacts; absent VCF rows denote unknown genotypes \\
RPA & 3 & 48 programs $\times$ 25 labels $\times$ 2 ATAC denominators $\times$ 2 models $= 4{,}800$ rows (MANOVA and conditional tests) & Group-level multivariate shifts differ from conditional increments (0 conditional BH hits) \\
RPAM & 4 & Fourth-layer cfDNA increment models, 18-participant 8-view MOFA, seven candidates & 0 WGBS increment BH hits; MOFA excludes WGS \\
RPAMG & 5 & CP6B--\textit{GFI1}--\textit{MMP8} five-layer candidate and 32 estimable fifth-layer dosage models & Retains discordant upstream edges and companion branches alongside 0 dosage BH hits \\
Remaining 9 three-layer and 4 four-layer subsets & 3--4 & Represented as observed subsets within higher-order models or candidate tables & Displayed per candidate/donor without standalone subset-specific significance claims \\
\bottomrule
\end{tabularx}
\end{table}

\begin{table}[!htbp]
\centering\footnotesize
\setlength{\tabcolsep}{4pt}
\caption{\textbf{Summary of the 4,800-row three- and four-layer joint models (1,200-test families) and bounded WGS dosage models.} Comparing across columns within a row separates group shifts from conditional coupling: under whole-fragment ATAC with age and sex adjustment, $1{,}068/1{,}200$ tests detect joint case--control differences, whereas $0/1{,}200$ detect conditional RNA+ATAC associations with plasma protein. Comparing down the MANOVA column shows denominator sensitivity: switching to promoter-assigned ATAC under the same adjustment reduces group detections from $1{,}068$ to $0$.}
\label{tab:si-anzhen-multilayer}
\begin{tabularx}{\linewidth}{@{}>{\raggedright\arraybackslash}Xrrrr@{}}
\toprule
\textbf{ATAC denominator and model} & \shortstack[r]{\textbf{Family}\\\textbf{size}} & \textbf{MANOVA} & \shortstack[r]{\textbf{Conditional}\\\textbf{RNA + ATAC}} & \shortstack[r]{\textbf{cfDNA}\\\textbf{increment}} \\
\midrule
Whole-fragment, age/sex adjusted & 1,200 & 1,068 (0.0194) & 0 (0.9990) & 0 (0.6792) \\
Whole-fragment, case status only & 1,200 & 1,095 (0.0032) & 0 (0.9745) & 0 (0.6296) \\
Promoter-assigned, age/sex adjusted & 1,200 & 0 (0.0562) & 0 (0.9986) & 0 (0.6512) \\
Promoter-assigned, case status only & 1,200 & 180 (0.0133) & 0 (0.9995) & 0 (0.6284) \\
\bottomrule
\end{tabularx}\par\smallskip\raggedright Entries for each test give the number passing BH correction (minimum $q$). In the bounded WGS dosage tests, 0/85 layer associations (minimum $q = 0.9810$) and 0/32 fifth-layer increments (minimum $q = 0.9405$) passed correction.
\end{table}

\begin{table}[!htbp]
\centering\footnotesize
\caption{\textbf{Four-omics MOFA variance explained ($R^2$) across 8 views recomputed over observed non-missing entries ($n = 18$, excluding WGS).} All 18 participants entered joint four-omics fitting; values report variance explained within each view's observed entries. Sixteen participants had estimable projections in both FCN1 monocyte RNA and plasma protein for the parallel partial-correlation checks in SI Table~\ref{tab:si-anzhen-f3-pcor}.}
\label{tab:si-anzhen-mofa-r2}
\begin{tabular}{@{}lrrrr@{}}
\toprule
\textbf{MOFA view} & \textbf{F1 $R^2$} & \textbf{F2 $R^2$} & \textbf{F3 $R^2$} & \textbf{Combined 3-factor $R^2$} \\
\midrule
IGHM/IGHD naive-like B RNA & 9.44\% & 12.90\% & 0.43\% & 22.94\% \\
TCF7/LEF1 naive-like CD4 RNA & 6.11\% & 14.59\% & 5.34\% & 25.63\% \\
Cytotoxic CD8 RNA & 7.06\% & 12.01\% & 0.53\% & 19.82\% \\
FCGR3A cytotoxic dim-like NK RNA & 7.74\% & 11.55\% & 0.49\% & 20.00\% \\
FCN1 classical-like monocyte RNA & 5.45\% & 7.58\% & 18.40\% & 30.35\% \\
PBMC ATAC (promoter-assigned) & 90.96\% & 4.74\% & 0.002\% & 95.88\% \\
cfDNA WGBS & 0.08\% & 34.11\% & 0.17\% & 34.35\% \\
Plasma protein & 0.05\% & 0.07\% & 76.01\% & 76.12\% \\
\bottomrule
\end{tabular}
\end{table}

\begin{table}[!htbp]
\centering\footnotesize
\caption{\textbf{Parallel, non-cumulative partial correlations between F3 FCN1 classical-like monocyte RNA and plasma protein projections ($n = 16$ identical participants).} The second and third models independently add distinct covariates to the baseline rather than stacking cumulatively; intervals are approximate 95\% confidence intervals from Fisher's $z$-transformation ($\pm 1.96$ standard errors back-transformed).}
\label{tab:si-anzhen-f3-pcor}
\begin{tabularx}{\linewidth}{@{}>{\raggedright\arraybackslash}Xrr@{}}
\toprule
\textbf{Adjustment model} & \shortstack[r]{\textbf{Raw protein}\\$r$ (approx.\ 95\% CI)} & \shortstack[r]{\textbf{Centered protein}\\$r$ (approx.\ 95\% CI)} \\
\midrule
Baseline (\texttt{case + age + sex}) & $+0.6517$ ($0.1571$ to $0.8849$) & $+0.5854$ ($0.0507$ to $0.8592$) \\
Baseline + individual protein panel median & $-0.0395$ ($-0.5998$ to $0.5468$) & $-0.0395$ ($-0.5998$ to $0.5468$) \\
Baseline + monocyte share and depth (non-cumulative) & $+0.7522$ ($0.2775$ to $0.9317$) & $+0.6345$ ($0.0559$ to $0.8941$) \\
\bottomrule
\end{tabularx}
\end{table}

\subsection{Candidate Evidence Cards and Independent Branches}\label{sec:si-anzhen-s5}

\begin{figure}[!htbp]
\centering
\includegraphics[width=\linewidth]{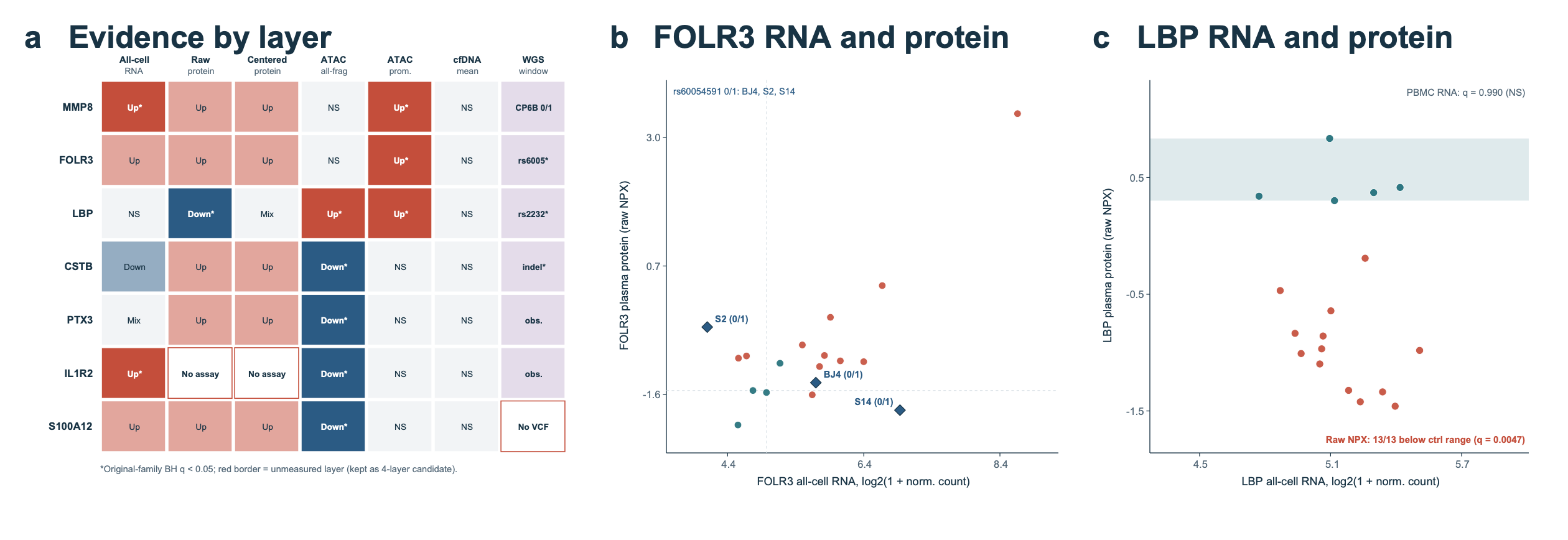}
\caption{\textbf{Five-layer directional audit of seven candidates and cross-layer decoupling in \textit{FOLR3} and \textit{LBP}.} \textbf{a,} Direction, within-family significance ($^*q < 0.05$), and missing layers (white cells outlined in terracotta: ``No assay'' for the unavailable IL1R2 protein measurement on both NPX scales, and ``No VCF'' for the unavailable \textit{S100A12} window VCF; neither label denotes a negative finding) across all-cell RNA, raw and centered plasma protein, dual-denominator ATAC, regional cfDNA, and fixed-window WGS. \textbf{b,} Participant-aligned all-cell \textit{FOLR3} RNA versus raw plasma FOLR3 protein ($n = 18$), with dark blue diamonds marking \texttt{rs60054591} heterozygous donors BJ4, S2 (low RNA, high protein), and S14 (high RNA, low protein). \textbf{c,} All-cell \textit{LBP} RNA versus raw plasma LBP protein (light cyan shading marks the 5-control raw protein range); RNA shows no group difference ($q = 0.990$), whereas raw plasma LBP falls below the control range in all 13 cases ($q = 0.0047$).}
\label{fig:si-anzhen-seven}
\end{figure}

\begin{figure}[!htbp]
\centering
\includegraphics[width=\linewidth]{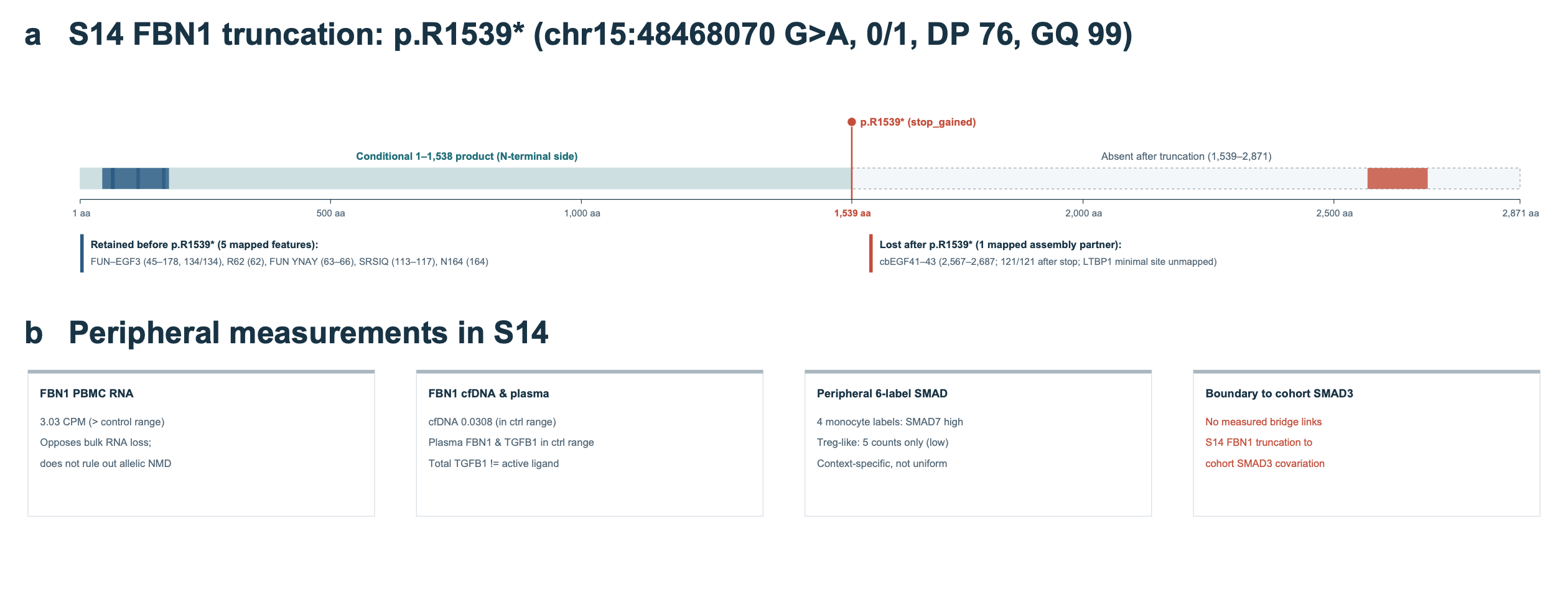}
\caption{\textbf{Conditional protein fragment boundaries and peripheral multi-omics response for \textit{FBN1} \texttt{p.R1539*} in case S14.} Maps retention of N-terminal recombinant domains (FUN--EGF3) and loss of C-terminal cbEGF41--43 if a stable 1--1538 truncated protein were formed, alongside \textit{SMAD7}, \textit{SMAD3}, and \textit{TGFB1} directions across six peripheral cell labels (Treg-like counts total only 5, representing a low-count readout; this single-case structural lead is independent of the 13-case \textit{SMAD3} regional association).}
\label{fig:si-anzhen-fbn1}
\end{figure}

This section consolidates fixed-window regional effects and five-layer audits for the seven prioritized candidates (SI Table~\ref{tab:si-anzhen-seven} and SI Figure~\ref{fig:si-anzhen-seven}), per-carrier values and edge-level evidence for CP6B--\textit{GFI1}--\textit{MMP8} (SI Table~\ref{tab:si-anzhen-cp6b}), \textit{SMAD3} sensitivity diagnostics (SI Table~\ref{tab:si-anzhen-smad3}), the 208 across-case process--protein summary (SI Table~\ref{tab:si-anzhen-208}), the S14 \textit{FBN1} structural lead (SI Figure~\ref{fig:si-anzhen-fbn1} and SI Table~\ref{tab:si-anzhen-fbn1}), and the prioritized follow-up validation plan (SI Table~\ref{tab:si-anzhen-followup}).

\begin{table}[!htbp]
\centering
\captionsetup{singlelinecheck=false,justification=raggedright}
\caption{\textbf{Dual-denominator ATAC and cfDNA adjusted effects and cfDNA coverage medians across fixed windows for the seven candidates ($n = 18$; 13 cases / 5 controls).}}\label{tab:si-anzhen-seven}
\footnotesize
\setlength{\tabcolsep}{3pt}
\renewcommand{\arraystretch}{1.15}
\setlength{\anzhenTableWidth}{\dimexpr\linewidth-10\tabcolsep\relax}
\begin{tabular}{@{}>{\raggedright\arraybackslash}p{0.22\anzhenTableWidth}>{\raggedleft\arraybackslash}p{0.13\anzhenTableWidth}>{\raggedleft\arraybackslash}p{0.13\anzhenTableWidth}>{\raggedleft\arraybackslash}p{0.13\anzhenTableWidth}>{\raggedleft\arraybackslash}p{0.13\anzhenTableWidth}>{\raggedright\arraybackslash}p{0.26\anzhenTableWidth}@{}}
\toprule
\textbf{Candidate and hg38 window} & \textbf{ATAC whole / BH} & \textbf{ATAC prom.\ / BH} & \textbf{cfDNA / BH} & \textbf{cfDNA med.\ (rec./depth)} & \textbf{Key gap or counterexample} \\
\midrule
\textit{MMP8} \texttt{chr11:\allowbreak{}[102726049,\allowbreak{}102728050)} & $-0.1342$ / 0.3030 & $+0.8212$ / 0.0397 & $-0.0626$ / 0.2277 & 30 / 407 & S5 low RNA / high protein; \texttt{rs11225395} lies 301 bp outside window \\
\textit{FOLR3} \texttt{chr11:\allowbreak{}[72113868,\allowbreak{}72115869)} & $+0.0620$ / 0.1710 & $+0.6543$ / 0.0054 & $-0.0532$ / 0.3704 & 16 / 212 & Opposite RNA/protein directions in \texttt{rs60054591} \texttt{0/1} carriers S2 and S14 \\
\textit{LBP} \texttt{chr20:\allowbreak{}[38345481,\allowbreak{}38347482)} & $+0.3329$ / 0.0002 & $+1.1377$ / 0.0032 & $-0.0062$ / 0.8586 & 52 / 670 & Raw 13/13 low vs.\ centered 10 low / 3 high; conflicts with external AAD upregulation~\cite{Cheng2020AADProteomics,Yesitayi2025LBPDissection} \\
\textit{CSTB} \texttt{chr21:\allowbreak{}[43775578,\allowbreak{}43777579)} & $-1.0938$ / 0.0008 & $+0.0401$ / 0.8369 & $-0.0010$ / 0.9730 & 297 / 1,457.5 & RNA low in 11/13 while protein high in 13/13; complex indel unnormalized \\
\textit{PTX3} \texttt{chr3:\allowbreak{}[157435849,\allowbreak{}157437850)} & $-0.4400$ / 0.0126 & $+0.5636$ / 0.0943 & $-0.0027$ / 0.8751 & 149.5 / 1,503 & RNA high in only 4/13 while protein high in 13/13; source unresolved \\
\textit{IL1R2} \texttt{chr2:\allowbreak{}[101990823,\allowbreak{}101992824)} & $-0.4961$ / 0.0107 & $+0.5177$ / 0.1321 & $-0.0055$ / 0.9240 & 52 / 578.5 & Strongest RNA support; no matching plasma protein assay (four layers) \\
\textit{S100A12} \texttt{chr1:\allowbreak{}[153374620,\allowbreak{}153376621)} & $-0.5242$ / 0.0222 & $+0.4822$ / 0.2186 & $-0.0275$ / 0.4514 & 50 / 395.5 & No candidate VCF call inside window (four layers; does not establish homozygous reference) \\
\bottomrule
\end{tabular}
\end{table}

Detailed reconciliation of the \textit{LBP} directional conflict with external literature: in this cohort, local PBMC ATAC exceeded the entire control range in all 13 cases under both denominators while all-cell \textit{LBP} RNA showed no group difference ($q = 0.990$). At promoter variant \texttt{rs2232571}, cases comprised 4 CT, 7 TT, and 2 unknown genotypes versus 1 CT and 4 TT among controls, with both TT and CT cases showing low raw plasma LBP. Guinan et al.\ reported higher circulating LBP in CT versus TT carriers among hematopoietic cell transplant recipients~\cite{Guinan2014LBP}, which does not generalize as allelic causality here; Cheng et al.\ reported a $2.99$-fold pooled iTRAQ LBP elevation in discovery profiling ($p = 0.0055$) but did not list LBP among the three significant proteins in expanded ELISA validation~\cite{Cheng2020AADProteomics}; and Yesitayi et al.\ reported elevated plasma and aortic-tissue LBP in acute dissection~\cite{Yesitayi2025LBPDissection}, with sampling windows that may differ from this cohort.

\begin{table}[!htbp]
\centering\footnotesize
\setlength{\tabcolsep}{3pt}
\caption{\textbf{Five-layer measurements for the three CP6B--\textit{GFI1}--\textit{MMP8} heterozygous (\texttt{0/1}) carriers and edge-level evidence assessment.} Across all 13 cases, promoter-normalized CP6B ATAC, regional cfDNA methylation, and \textit{GFI1} RNA each showed 9 cases above and 4 below the control median, with different cases accounting for the lower readouts. NPX is a relative log scale (negative values do not denote negative concentrations), and ``high/low'' refers to the corresponding control median. Reported $\sim 34$-fold activation by Gfi-1/C/EBP$\varepsilon$ co-transfection in 32Dwt18 cells ($p = 0.04$) tested the \textit{MMP8} promoter rather than the human CP6B variant~\cite{KhannaGupta2007Gfi1}; two short \textit{CEBPE} regions lacked coverage.}
\label{tab:si-anzhen-cp6b}
\begin{tabularx}{\linewidth}{@{}>{\raggedright\arraybackslash}Xrrr@{}}
\toprule
\textbf{Measurement} & \textbf{S7} & \textbf{S12} & \textbf{S14} \\
\midrule
Variant read depth (DP) & 57 & 44 & 34 \\
Genotype quality (GQ) & 99 & 99 & 99 \\
ATAC, whole-fragment denominator (CPM) & 1.1638 & 1.1517 & 1.4186 \\
ATAC, promoter denominator (CPM) & 5.8052 & \textbf{4.5637} (low) & 6.2339 \\
cfDNA mean & \textbf{0.1300} (low) & 0.1390 & 0.1812 \\
\textit{GFI1} RNA & 1,222.48 & 751.43 & \textbf{531.73} (low) \\
\textit{MMP8} RNA (CPM) & 2.1962 & 2.9642 & 2.0856 \\
Plasma MMP8, raw NPX & $+2.0703$ & $-0.6223$ & $-0.9012$ \\
Plasma MMP8, centered NPX & $+3.0491$ & $+0.0663$ & $-0.0487$ \\
Discordant readout & Low cfDNA & Low prom.\ ATAC & Low \textit{GFI1} RNA \\
\bottomrule
\end{tabularx}

\vspace{0.6em}
\begin{tabularx}{\linewidth}{@{}>{\raggedright\arraybackslash}p{0.20\linewidth}>{\raggedright\arraybackslash}p{0.25\linewidth}>{\raggedright\arraybackslash}p{0.33\linewidth}>{\raggedright\arraybackslash}X@{}}
\toprule
\textbf{Candidate edge} & \textbf{Established basis} & \textbf{Counterevidence, alternatives, and gaps} & \textbf{Current assessment} \\
\midrule
\texttt{92486968 G>A}--CP6B state & Variant falls inside CP6B; all three carriers have ATAC and cfDNA data & Genotypes unknown in 15 participants; discordant states across carriers; no allelic assay & Weak positional and donor alignment \\
CP6B ATAC--\textit{GFI1} RNA & Promoter annotation and observed readouts across 18 participants & Denominators flip direction; annotation is not causal proof; composition unseparated & Low-confidence inferred edge \\
CP6B cfDNA--upstream branch & Regional circulating DNA covering the variant interval & Shared CpG sites unaudited; cell-source mixing unseparated from within-source change & Retained regional-state branch \\
GFI1/CEBPE--\textit{MMP8} & Reporter site-dependence, probe binding, and differentiation precedent~\cite{KhannaGupta2007Gfi1} & Does not test human CP6B locus; \textit{GFI1} RNA is not transcription-factor activity & Bounded external functional precedent \\
\textit{MMP8} RNA--plasma MMP8 & Gene-matched, same-donor elevation in 12/13 cases & S5 counterexample; protein not BH-significant; non-PBMC sources and clearance unresolved & Exploratory cross-material alignment \\
\bottomrule
\end{tabularx}
\end{table}

\begin{table}[!htbp]
\centering\footnotesize
\setlength{\tabcolsep}{4pt}
\caption{\textbf{Sensitivity analysis of the \textit{SMAD3} across-case negative correlation across measurement definitions and S14 exclusion.} The first two rows belong to the prespecified 48-pair family; the remaining three rows are within-cohort post-discovery sensitivity checks.}
\label{tab:si-anzhen-smad3}
\begin{tabularx}{\linewidth}{@{}>{\raggedright\arraybackslash}p{0.24\linewidth}>{\raggedright\arraybackslash}p{0.28\linewidth}>{\raggedright\arraybackslash}p{0.26\linewidth}>{\raggedright\arraybackslash}X@{}}
\toprule
\textbf{Specification} & \textbf{Measurement definition} & \textbf{Result} & \textbf{Evidential scope} \\
\midrule
\textit{SMAD3} 48-pair (regional mean) & Regional mean vs.\ whole / promoter ATAC ($n = 13$) & $\rho = -0.8407$ ($q = 0.0091$) / $-0.7582$ ($q = 0.0320$) & Prespecified 48-pair cross-material family \\
\textit{SMAD3} 48-pair (matched offset) & Matched offset vs.\ whole / promoter ATAC ($n = 13$) & $\rho = -0.8352$ ($q = 0.0091$) / $-0.7747$ ($q = 0.0299$) & Alternative cfDNA estimate within same family \\
\textit{SMAD3} shared CpG records & 54 shared strand-specific CpG records (depth $\ge 5$) vs.\ whole ATAC & $\rho = -0.8516$ ($9{,}999$ perm.\ $p = 0.0009$) & Within-cohort site-coverage sensitivity check \\
Leave-S14-out & Regional mean vs.\ whole ATAC ($n = 12$ cases) & $\rho = -0.818$ & Parallel within-cohort sensitivity check \\
Shared CpG records + leave-S14-out & 54 shared CpG records vs.\ whole ATAC ($n = 12$ cases) & $\rho = -0.832$ & Combined within-cohort sensitivity check \\
\bottomrule
\end{tabularx}
\end{table}

\begin{table}[!htbp]
\centering\footnotesize
\setlength{\tabcolsep}{4pt}
\caption{\textbf{Summary of the 208 across-case correlations between immune-process RNA scores and plasma MMP8 or LBP.} Each test includes only cases with observed values in both layers; the 8 prespecified all-cell comparisons appear in SI Figure~\ref{fig:si-anzhen-mofa-208}d.}
\label{tab:si-anzhen-208}
\begin{tabularx}{\linewidth}{@{}>{\raggedright\arraybackslash}p{0.22\linewidth}>{\raggedright\arraybackslash}p{0.36\linewidth}>{\raggedright\arraybackslash}X@{}}
\toprule
\textbf{Comparison scope} & \textbf{Tested relationship} & \textbf{Summary result} \\
\midrule
All 208 tests & Across-case rank correlation between innate immunity or degranulation RNA scores and plasma MMP8 or LBP across all RNA scopes & All 208 estimable; 0 passed BH correction; lowest $q = 0.4368$ (bright-like NK vs.\ centered LBP). \\
Prespecified 8 all-cell subset & 2 processes $\times$ 2 proteins (MMP8, LBP) $\times$ 2 protein scales (nested within the 208-test family) & $\rho$ ranged from $-0.3571$ to $+0.0879$; all 8 had 208-test BH-adjusted $q = 1.0$. \\
\bottomrule
\end{tabularx}
\end{table}

\begin{table}[!htbp]
\centering\footnotesize
\setlength{\tabcolsep}{4pt}
\caption{\textbf{Protein fragment boundaries and peripheral multi-omics readouts for \textit{FBN1} \texttt{p.R1539*} in case S14.} Genotypes at \texttt{p.R1539*} were unknown in the remaining 17 participants, and this single-case structural lead is independent of the 13-case \textit{SMAD3} regional association (SI Figure~\ref{fig:si-anzhen-fbn1}).}
\label{tab:si-anzhen-fbn1}
\begin{tabularx}{\linewidth}{@{}>{\raggedright\arraybackslash}p{0.20\linewidth}>{\raggedright\arraybackslash}p{0.28\linewidth}>{\raggedright\arraybackslash}p{0.24\linewidth}>{\raggedright\arraybackslash}X@{}}
\toprule
\textbf{Target} & \textbf{Coordinates or readouts} & \textbf{Relation to truncation / controls} & \textbf{Evidence boundary} \\
\midrule
S14 \textit{FBN1} N-terminal domain & FUN--EGF3 (45--178); YNAY (63--66); SRSIQ (113--117); R62; N164 & 100\% upstream of \texttt{p.R1539*} & Recombinant binding/mutagenesis~\cite{Yadin2013Fibrillin1} (N164 cited therein); conditional on stable 1--1538 product \\
S14 \textit{FBN1} C-terminal domain & cbEGF41--43 (2567--2687) & 121/121 residues downstream of \texttt{p.R1539*} & Human recombinant fragment binding~\cite{Yadin2013Fibrillin1}; nonsense-mediated decay (NMD) unexcluded \\
S14 peripheral readouts & PBMC \textit{FBN1} RNA; regional cfDNA, plasma FBN1, total TGFB1; monocyte and Treg \textit{SMAD7}, \textit{SMAD3} RNA & \textit{FBN1} RNA above control range; cfDNA and both proteins within range; downstream RNA discordant & Current assays do not distinguish intact from truncated FBN1 or confirm TGF-$\beta$ activation \\
\bottomrule
\end{tabularx}
\end{table}

\subsection{Follow-Up Validation Plan}\label{sec:si-anzhen-s5-1}

Two independent priorities govern follow-up validation. For \textit{SMAD3}, replication requires an independent acute aortic dissection cohort profiling cfDNA methylation and PBMC ATAC across the same $\sim 2\text{ kb}$ window, paired with a non-dissection acute-illness comparator to evaluate disease specificity. For CP6B, the locus must first be reassessed under uniform genotype-calling and quality criteria, retaining unresolved genotypes as unknown, before assessing whether cohort allelic comparisons or cellular perturbation assays are warranted.

\begingroup
\footnotesize
\setlength{\tabcolsep}{3pt}
\renewcommand{\arraystretch}{1.15}
\setlength{\LTleft}{0pt}
\setlength{\LTright}{\fill}
\setlength{\LTcapwidth}{\linewidth}
\captionsetup{singlelinecheck=false,justification=raggedright}
\setlength{\anzhenTableWidth}{\dimexpr\linewidth-6\tabcolsep\relax}
\begin{longtable}{@{}>{\raggedright\arraybackslash}p{0.20\anzhenTableWidth}>{\raggedright\arraybackslash}p{0.24\anzhenTableWidth}>{\raggedright\arraybackslash}p{0.30\anzhenTableWidth}>{\raggedright\arraybackslash}p{0.26\anzhenTableWidth}@{}}
\caption{\textbf{Prioritized follow-up validation plan.} Entries are ordered by the scientific question to resolve, integrating the follow-up directions generated by CytoCoScientist for CP6B, MMP8, LBP, F3, and FBN1 with the independent-cohort and acute-comparator replication design formulated by the authors for \textit{SMAD3}.}\label{tab:si-anzhen-followup}\\
\toprule
\textbf{Lead and open question} & \textbf{Current cohort evidence} & \textbf{Priority next step} & \textbf{Discriminative criterion} \\
\midrule
\endfirsthead
\multicolumn{4}{@{}l}{\footnotesize\textbf{Table~\thetable\ (continued)}}\\
\toprule
\textbf{Lead and open question} & \textbf{Current cohort evidence} & \textbf{Priority next step} & \textbf{Discriminative criterion} \\
\midrule
\endhead
\midrule
\multicolumn{4}{r@{}}{\footnotesize\emph{Continued on next page}}\\
\endfoot
\bottomrule
\endlastfoot
\textbf{\textit{SMAD3}:} Does the regional association replicate in new patients? & Across the 13 cases, higher regional cfDNA methylation tracks with lower PBMC ATAC; direction persists across 54 shared strand-specific CpG records and leave-S14-out checks without adding new patients. & Prespecify the same $\sim 2\text{ kb}$ window, biospecimen types, and readouts in an independent acute dissection cohort, alongside non-dissection acute-illness controls with identical assays, comparing effect sizes and uncertainties. & Replication confirms robustness across cohorts; a comparable effect in acute-illness controls indicates a general acute-injury response rather than dissection specificity (significance in one group alone does not prove a group difference). \\
\textbf{CP6B:} Can carrier status be resolved before testing allelic effects? & 3 cases carry confirmed \texttt{0/1} calls while 15 participants have unknown genotypes; all 3 carriers have elevated \textit{MMP8} RNA and protein but discordant upstream readouts. & Reassess the CP6B locus (\texttt{chr1:92486968}) under uniform genotype-calling and quality criteria, retaining unresolved genotypes as unknown; if comparable groups exist, evaluate allelic contrast and test allele-specific nascent RNA/protein output in matched myeloid backgrounds. & Completing genotypes establishes valid comparators rather than proving regulation; if comparable groups remain unavailable, retain carrier descriptions without estimating allelic effects from this cohort. \\
\textbf{Four-omics F3:} Does monocyte RNA track specific proteins or overall panel level? & Three-layer models showed 0/1,200 conditional associations; in F3 ($n = 16$), partial correlation dropped from $0.652$ to $-0.039$ after adjusting for individual plasma protein panel median. & Prespecify panel-level background proteins versus target proteins before new assays, and verify associations with monocyte RNA using an independent quantitative platform. & Persistence only for global panel shifts argues against monocyte-specific protein coupling; retention of target-specific associations prioritizes mechanistic follow-up. \\
\textbf{MMP8:} Does elevated plasma protein reflect concurrent PBMC transcription? & Case S5 shows low PBMC \textit{MMP8} RNA with high plasma MMP8; a single draw cannot separate de novo synthesis, granule release, or non-PBMC sources. & Measure nascent RNA, intracellular granule stores, intact circulating MMP8, and enzymatic activity or cleavage products from the same blood draw. & Separates concurrent transcriptional synthesis from preformed store release; existing NPX readouts establish neither tissue origin nor enzymatic activity. \\
\textbf{LBP:} Does low raw NPX reflect lower intact plasma protein? & Raw NPX fell below the control range in 13/13 cases, became non-significant after panel centering, and conflicted with external AAD reports~\cite{Cheng2020AADProteomics,Yesitayi2025LBPDissection}. & Align onset-to-sampling windows, re-assay using an antibody recognizing a distinct epitope, and quantify intact LBP independently (distinguishing complexes and fragments if needed). & Persistent platform discordance supports assay-epitope or sampling-window differences rather than a generalizable directional biomarker. \\
\textbf{S14 \textit{FBN1}:} Is the predicted truncated protein actually present? & \texttt{p.R1539*} lies between known N- and C-terminal interaction fragments, but existing plasma assays cannot distinguish intact from truncated FBN1. & Assay allele-specific RNA and resolve intact versus truncated FBN1 protein; subsequently test microfibril assembly, LTBP deposition, and active TGF-$\beta$ signaling. & Absence of a stable truncated protein rules out a C-terminal-loss mechanism in S14; this single-case lead remains independent of the cohort-wide \textit{SMAD3} association. \\
\end{longtable}
\endgroup

\FloatBarrier

\FloatBarrier
\phantomsection
\addcontentsline{toc}{section}{Supplementary References}
\begingroup
\renewcommand{\refname}{Supplementary References}
\putbib[references]
\endgroup
\end{bibunit}
\end{document}